\pdfoutput=1

\documentclass[11pt]{article}

\usepackage[final]{coling}

\usepackage{times}
\usepackage{latexsym}
\usepackage{subcaption}

\usepackage[T1]{fontenc}

\usepackage[utf8]{inputenc}

\usepackage{microtype}

\usepackage{inconsolata}
\usepackage{booktabs}
\usepackage{subcaption}
\usepackage{multirow}
\usepackage{soul}
\usepackage{array}
\usepackage{tabularx}
\usepackage{longtable}
\usepackage{amsmath}
\usepackage{tcolorbox}   
\usepackage{xcolor}      
\usepackage{graphicx}    

\usepackage{graphicx}

\definecolor{darkred}{RGB}{200, 0, 0}
\definecolor{darkyellow}{RGB}{250, 153, 0}
\definecolor{darkgreen}{RGB}{0, 180, 0}
\title{Chain-of-Thought in Large Language Models: Decoding, Projection, and Activation}

\author{Hao Yang \\
   Nanjing University \\
  \texttt{howyoung9901@gmail.com} \\\And
 Qinghua Zhao \\
  SAIBD, Hefei University \\
  SKLSDE, Beihang University \\
  \texttt{zhaoqh@buaa.edu.cn} \\\And
  Lei Li \\
  University of Washington \\
  University of Copenhagen \\
  \texttt{lilei@di.ku.dk} \\}

\begin{document}
\maketitle
\begin{abstract}
Chain-of-Thought prompting has significantly enhanced the reasoning capabilities of large language models, with numerous studies exploring factors influencing its performance. However, the underlying mechanisms  remain poorly understood. To further demystify the operational principles, this work examines three key aspects: decoding, projection, and activation, aiming to elucidate the changes that occur within models when employing Chain-of-Thought.  Our findings reveal that LLMs effectively imitate exemplar formats while integrating them with their understanding of the question, exhibiting fluctuations in token logits during generation but ultimately producing a more concentrated logits distribution, and activating a broader set of neurons in the final layers, indicating more extensive knowledge retrieval compared to standard prompts. Our code and data will be publicly avialable when the paper is accepted.
\end{abstract}

\section{Introduction}
Large language models (LLMs) have demonstrated remarkable performance across various tasks, yet their efficacy remains suboptimal in certain reasoning tasks \citep{wei2022chain}. This underscores the urgent need to understand and optimize LLMs in complex reasoning scenarios. The Chain-of-Thought (CoT) prompting has emerged as a promising solution, significantly enhancing LLMs performance in these challenging tasks \citep{wei2022chain,zhang2024cocot}. Recent research has extended CoT to more complex structures such as trees and graphs \citep{yao2024tree,besta2024graph}, while other studies have explored various factors affecting CoT performance, including the number of reasoning steps, difficulty, length, and correctness of reasoning processes \citep{webson2021prompt,chia2023contrastive,wang2023towards,schaeffer2023invalid,jin2024impact,prabhakar2024deciphering}.

Several hypotheses are proposed regarding the underlying principles of CoT's functionality. \citet{schaeffer2023invalid} suggest that CoT enhances the reasoning capabilities of LLMs, while \citet{madaan2023makes} argue that CoT reduces task difficulty, making tasks more comprehensible to models. \citet{saparovlanguage} propose that models  mimic its format, filling in answers based on the provided exemplar. \citet{merrillexpressive} offer a theoretical analysis of how CoT improves the computational power of models. Although these hypotheses await further experimental validation, they provide valuable insights for exploring the fundamental mechanisms of CoT.

Despite these advancements, the specific changes that occur both internally and externally within the model when employing CoT remain unclear. This paper aims to address this knowledge gap by examining CoT mechanisms from three perspectives: the decoding phase, the projection phase, and neuron activation. We seek to answer the following questions:
\begin{enumerate}
\item Does the large model merely mimic the patterns found in CoT exemplars?
\vspace{-6pt}
\item What changes occur in the model's projection space (e.g., logits) when using CoT prompt as opposed to the standard prompt?
\vspace{-6pt}
\item Does CoT enable models' deeper and broader utilization of knowledge acquired during the pre-training phase?
\end{enumerate}
To answer these questions, we employ a multi-faceted approach:
\begin{itemize}
\item In the decoding phase, we analyze CoT-generated text, focusing on four test points: time, action, location, and number.  We also conduct a transfer test to evaluate CoT's performance across different datasets.
\vspace{-6pt}
\item In the projection phase, we examine horizontal and vertical changes in projected logits, analyzing logits values and probability distributions at each time step.
\vspace{-6pt}
\item We compare the range, intensity, and dynamic evolution of neuron activations under standard and CoT-style prompts. 
\end{itemize}

Our experiments yield insightful results that contribute to a deeper understanding of CoT mechanisms in large language models. 

\section{Related Work}
Recent years have witnessed remarkable progress in LLMs, with models such as GPT-4 \citep{achiam2023gpt} and LLaMA \citep{meta2024introducing} demonstrating exceptional capabilities across a wide range of tasks. The introduction of CoT has further elevated their performance, particularly in complex reasoning tasks including arithmetic, commonsense, and symbolic reasoning \citep{wei2022chain}.

The success of CoT has spurred a wave of research aimed at extending and refining this approach. Notable contributions include the development of self-consistency \citep{wangself}, Boosting of Thoughts \citep{chenboosting}, and least-to-most prompting \citep{zhouleast}. These methods have been complemented by zero-shot CoT \citep{kojima2022large}, which eliminates the need for task-specific examples, and more specialized techniques such as Program-of-Thought \citep{bi2024program}, Mixture-of-Thoughts  \citep{li2023mot}, and Contrastive Chain-of-Thought  \citep{zhang2024cocot}. The versatility of CoT has been further demonstrated through its application in enhancing performance across diverse tasks \citep{tanneru2024hardness}.

Research on CoT has uncovered key factors influencing its performance. Studies show that maintaining a constant number of reasoning steps, even with errors, does not hinder performance as long as logical coherence is maintained \citep{jin2024impact,chia2023contrastive,wang2023towards,schaeffer2023invalid,webson2021prompt,prabhakar2024deciphering}. While intermediate steps in CoT do not directly improve problem-solving, they enhance task comprehension \citep{madaan2023makes}. CoT also stabilizes significance scores \citep{wu2023analyzing}, and diverse examples complement each other to improve reasoning \citep{ye2023complementary}.

Understanding why CoT is effective remains limited, but recent research has proposed several hypotheses based on experimental findings. 
\citet{madaan2023makes} use counterfactual methods to modify exemplar content, observing that while intermediate steps in CoT may not enhance task-solving abilities, they improve task comprehension. 
\citet{saparovlanguage,chia2023contrastive,wang2023towards,wang2024chain} suggest reasoning abilities are acquired during pre-training, with exemplars guiding response generation. 
\citet{madaan2022text} posit that intermediate steps act as templates for the model to fill in answers rather than aiding task-solving. 
\citet{nowak2024representational,merrillexpressive} argue CoT boosts computational power. 
\citet{gudibande2023false} find that while imitation improves style and adherence to instructions, it does not enhance factuality, coding, or problem-solving. 
\citet{rai2024investigation} use GPT-4 to study neuronal activation patterns, explaining the importance of equations, textual explanations, and exemplar diversity, as well as why incorrect reasoning does not necessarily impair performance.

Our paper contributes to the growing body of research investigating the underlying mechanisms of CoT prompting. Unlike existing work, we approach this question from three distinct perspectives: decoding, pre-decoding steps, and neuronal activation, specifically exploring whether the model mimics exemplars during decoding, how the model's projection space changes with CoT, and whether CoT facilitates deeper knowledge exploration, aiming to provide a more comprehensive understanding of how CoT enhances the reasoning capabilities of language models.

\section{Methodology}
\subsection{Evaluation Tasks} \label{sec:datasets}
Following the methodologies outlined by  \citet{chia2023contrastive,madaan2023makes,jin2024impact}, we conducted experiments on reasoning tasks, specifically focusing on arithmetic reasoning, commonsense reasoning, and symbolic reasoning.
\begin{figure*}[!htbp]
\centering
    \begin{subfigure}{0.30\linewidth}
        \includegraphics[width=\linewidth]{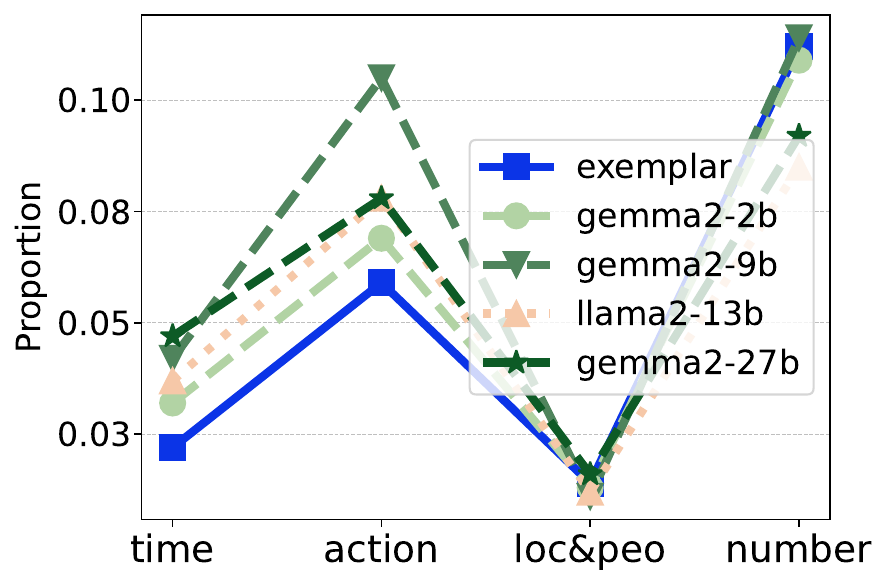}
        \caption{\textit{AQuA}}
    \end{subfigure}
    \begin{subfigure}{0.30\linewidth}
        \includegraphics[width=\linewidth]{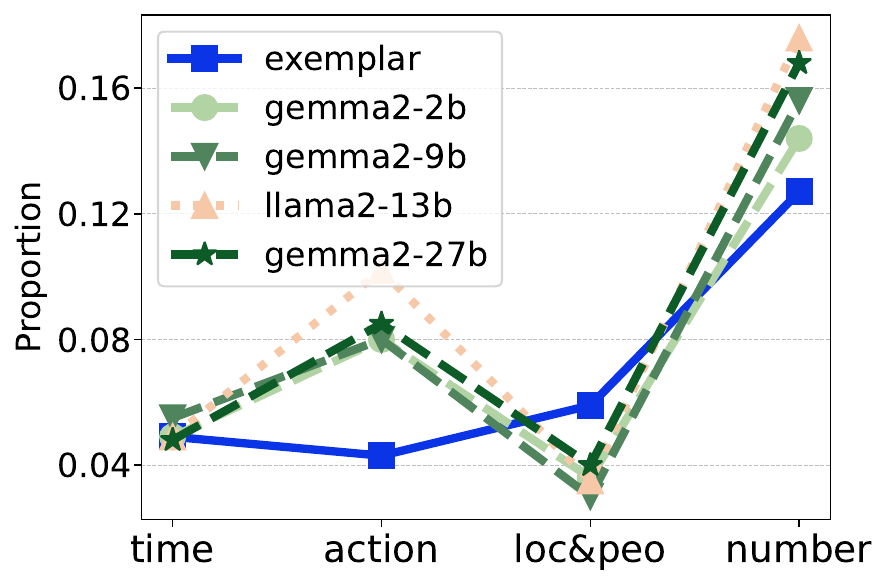}
        \caption{\textit{GSM8K}}
    \end{subfigure}
    \begin{subfigure}{0.3\linewidth}
        \includegraphics[width=\linewidth]{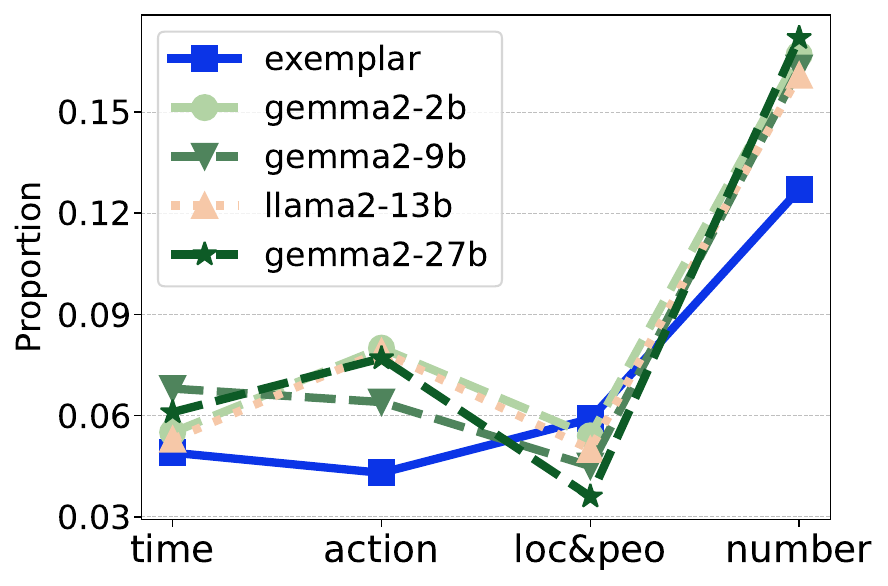}
        \caption{\textit{SVAMP}}
    \end{subfigure}
    \begin{subfigure}{0.3\linewidth}
        \includegraphics[width=\linewidth]{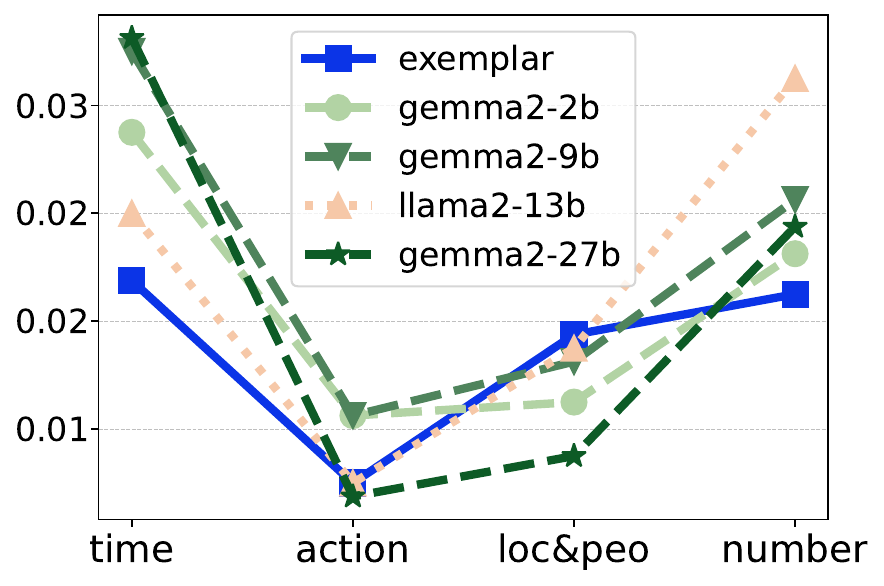}
        \caption{\textit{Bamboogle}}
    \end{subfigure}
    \begin{subfigure}{0.3\linewidth}
        \includegraphics[width=\linewidth]{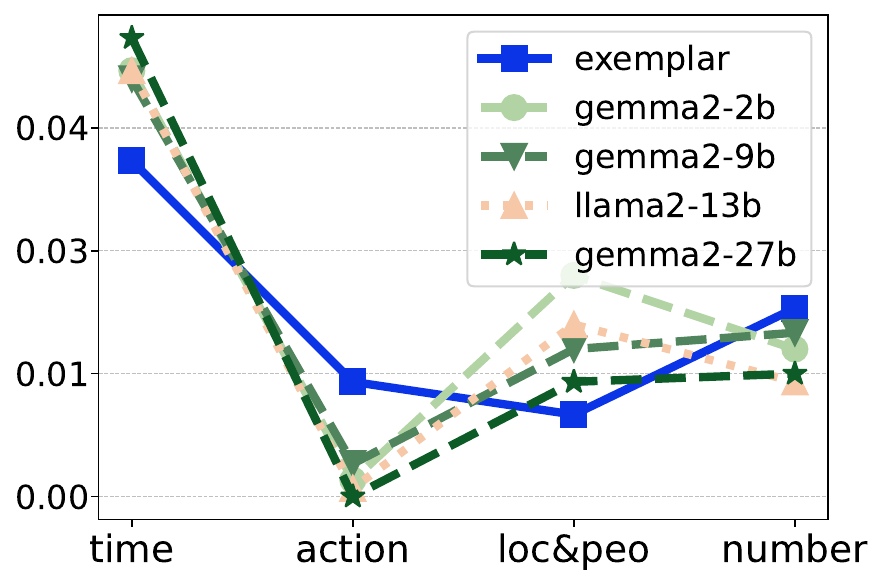}
        \caption{\textit{StrategyQA}}
    \end{subfigure}
    \begin{subfigure}{0.3\linewidth}
        \includegraphics[width=\linewidth]{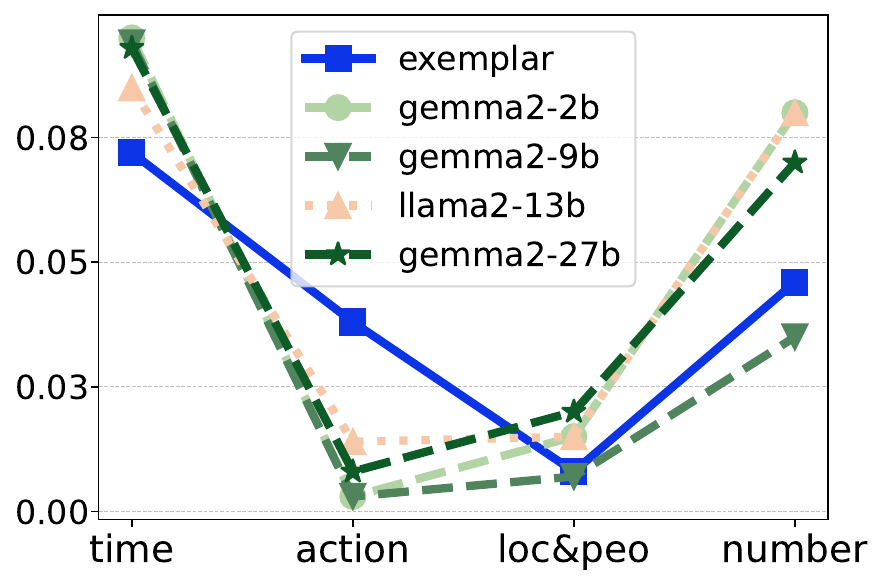}
        \caption{Date}
    \end{subfigure}
    \begin{subfigure}{0.3\linewidth}
        \includegraphics[width=\linewidth]{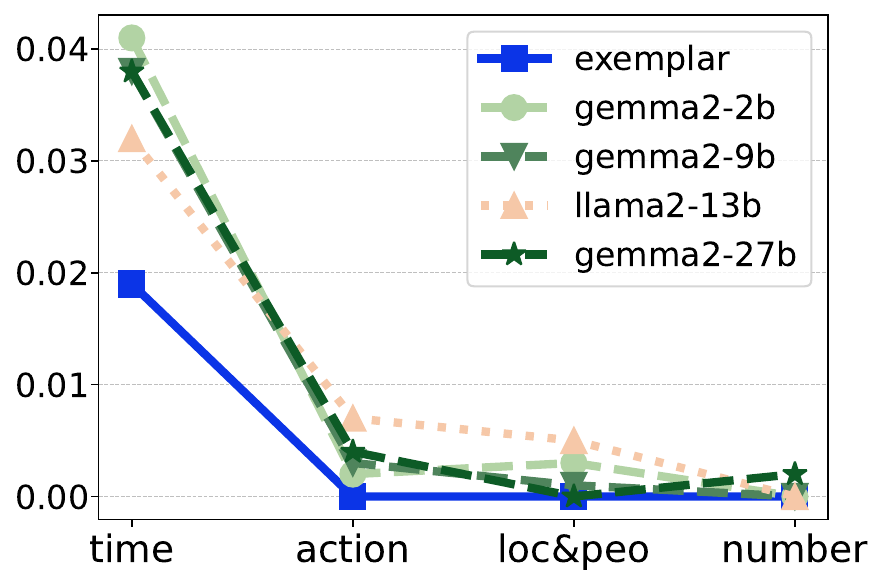}
        \caption{\textit{Sports}}
    \end{subfigure}
    \begin{subfigure}{0.3\linewidth}
        \includegraphics[width=\linewidth]{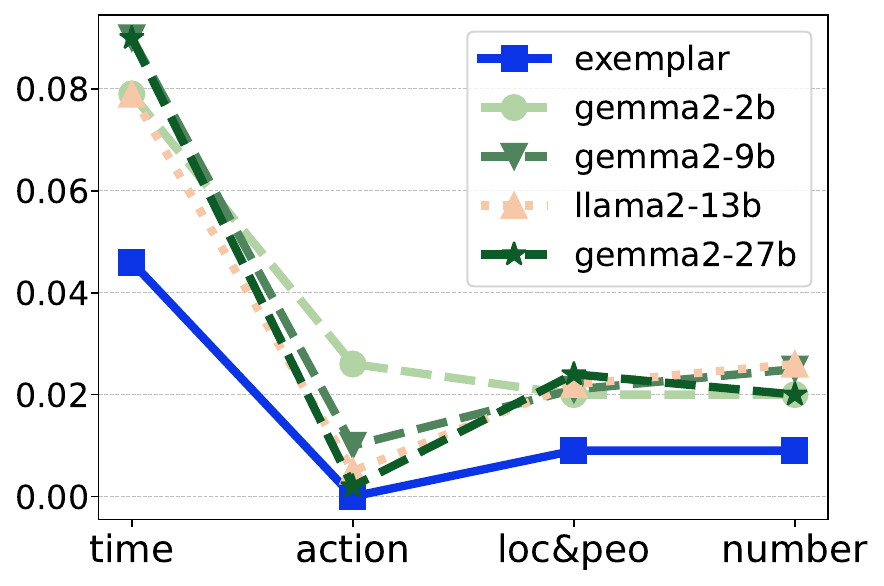}
        \caption{Coin Flip}
    \end{subfigure}
    \begin{subfigure}{0.3\linewidth}
        \includegraphics[width=\linewidth]{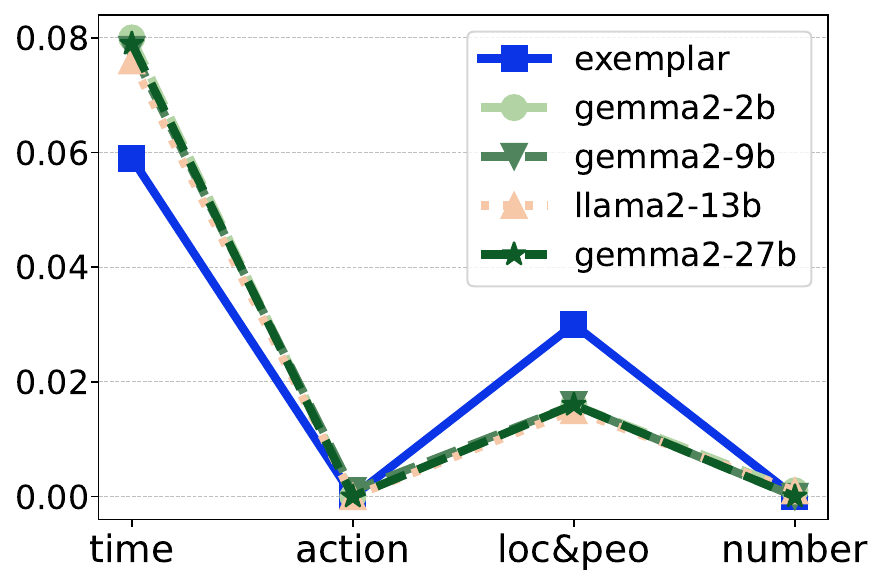}
        \caption{\textit{Last Letter}}
    \end{subfigure}
\vspace{-2pt}
\caption{Statistical analysis of \textit{\textcolor{darkyellow}{\textbf{test points}}} matches in model-generated content when using \textcolor{darkgreen}{\textbf{CoT}}.}
\label{fig:test_points}
\end{figure*}

\vspace{-3pt}
\paragraph{Arithmetic reasoning.} To evaluate arithmetic reasoning, we used three well-established datasets: GSM8K \citep{cobbe2021training} with grade-school math problems, SVAMP \citep{patel-etal-2021-nlp} featuring diverse mathematical structures, and AQuA \citep{ling2017program} with multiple-choice problems. These challenging benchmarks are widely used in mathematical reasoning research.
\vspace{-3pt}
\paragraph{Commonsense reasoning.} For commonsense reasoning tasks, we employed the Bamboogle \citep{press2023measuring}, StrategyQA \cite{geva2021strategyqa}, Date, and Sports \citep{srivastava2023beyond} datasets. Bamboogle is a small, handcrafted dataset containing 2-hop questions designed to test commonsense reasoning abilities. StrategyQA is a question answering benchmark where the required reasoning steps are implicit in the question, and should be inferred using a strategy. The Date and Sports datasets comprise compositional questions specifically designed to assess common sense.

\vspace{-3pt}
\paragraph{Symbolic reasoning.} To evaluate symbolic reasoning, we considered two tasks: the Coin Flip task \citep{wei2022chain} and the Last Letters Concatenation dataset \citep{NEURIPS2022_8bb0d291}. The Coin Flip task assesses the ability to reason about probabilistic outcomes, while the Last Letters Concatenation dataset evaluates the capacity to manipulate and reason about symbolic representations.

It is important to note that due to budget constraints, we randomly sampled 50 test examples from each dataset for evaluation purposes. 


\subsection{Models \& parameters setting}  \label{sec:models}
In our experiments, we used publicly available pre-trained models, including Gemma2 (2B, 9B, 27B) \citep{gemmateam2024gemma2improvingopen} and LLaMA2-13B \citep{touvron2023llama2openfoundation}, to cover different model series and sizes. Generation employed greedy decoding, capped at 300 tokens, following \citet{wei2022chain}.

\vspace{-3pt}
\paragraph{Prompts}
For each dataset, we employed a 4-shot prompt, with the CoT exemplars manually constructed based on \citet{wei2022chain, wang2023towards,NEURIPS2022_8bb0d291}. The full prompts are included in Table \ref{standard:gsm8k} to  \ref{cot:letter}.


\subsection{RQ-1: Imitation or Understanding?}

Current views suggest that LLMs develop core problem-solving and analogical skills during pre-training, with exemplars guiding their output by providing style and format cues. Interestingly, even if the exemplars in the prompt contain multiple errors, they do not seem to affect the model's output \citep{wang2023towards,wang2024chain,madaan2022text}.

However, recent studies have shown that LLMs can learn the mapping relationships between inputs and outputs from exemplars, even when these relationships are inconsistent with the knowledge acquired during pre-training \citep{liu2023towards}. This raises the question of whether LLMs are merely imitating the exemplars or truly understanding the underlying concepts.

To investigate this question, we analyzed the behavior of LLMs from both fine-grained and coarse-grained perspectives. These experiments aim to shed light on the extent to which LLMs rely on pre-training versus exemplars for problem-solving and understanding.

\begin{figure*}[!ht]
\centering
    \begin{subfigure}{0.3\linewidth}
        \includegraphics[width=\linewidth]{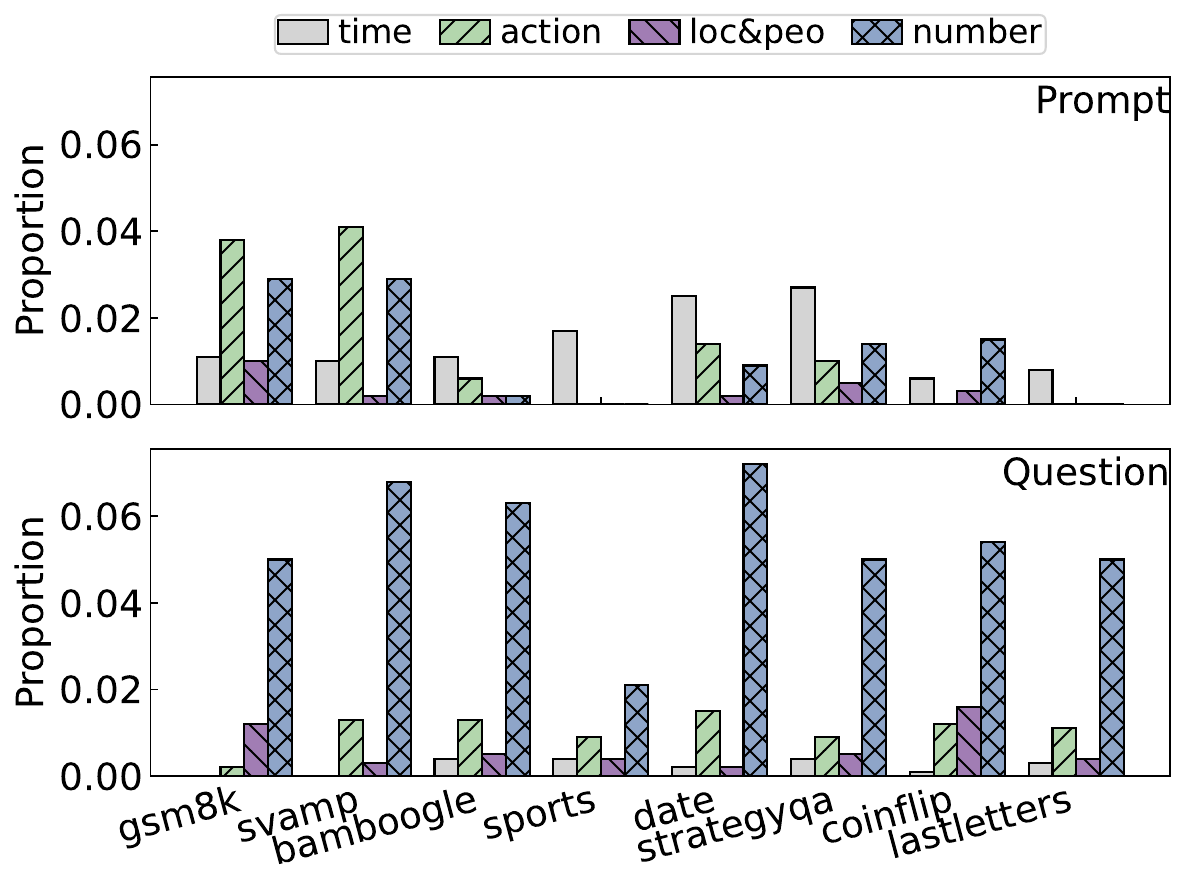}
        \caption{\textit{AQuA}}
        \label{fig:transfer_test_27b_a}
    \end{subfigure}
    \begin{subfigure}{0.3\linewidth}
        \includegraphics[width=\linewidth]{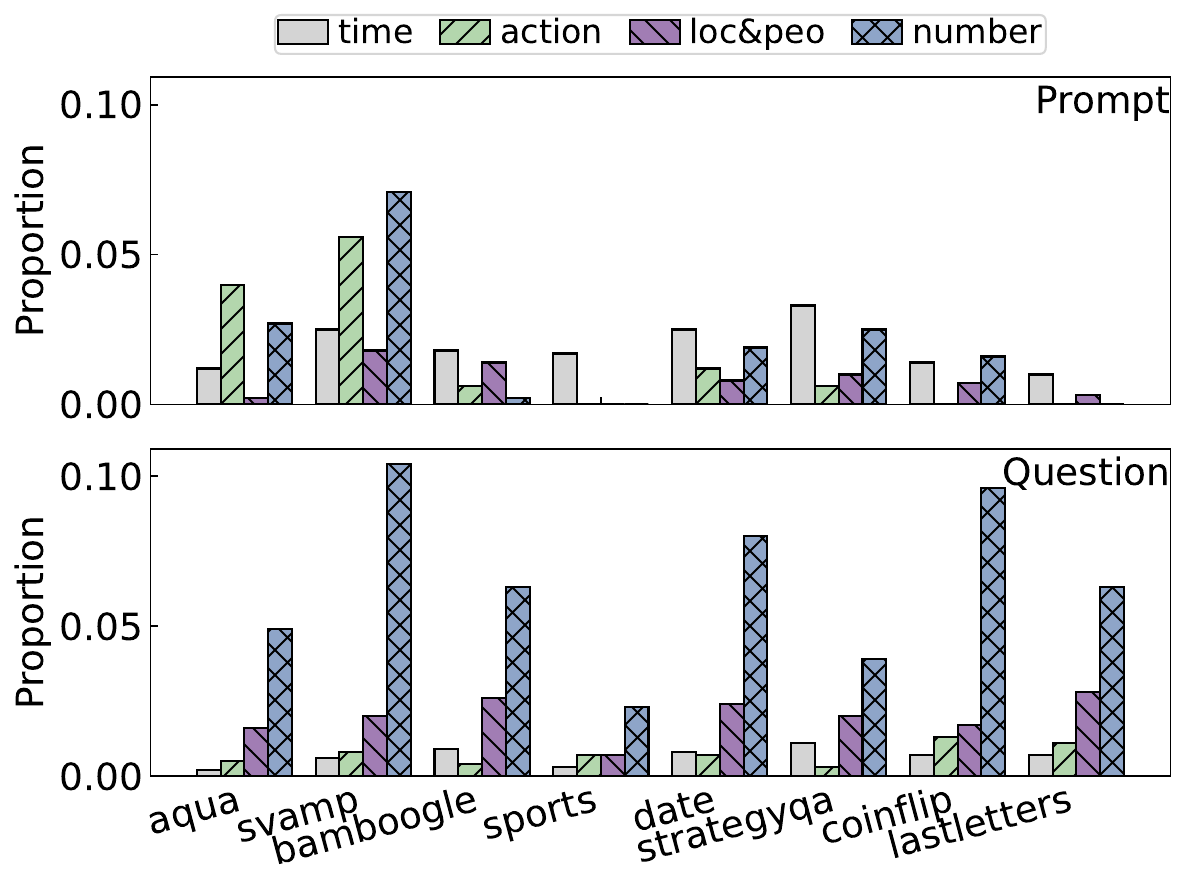}
        \caption{\textit{GSM8K}}
        \label{fig:transfer_test_27b_b}
    \end{subfigure}
    \begin{subfigure}{0.3\linewidth}
        \includegraphics[width=\linewidth]{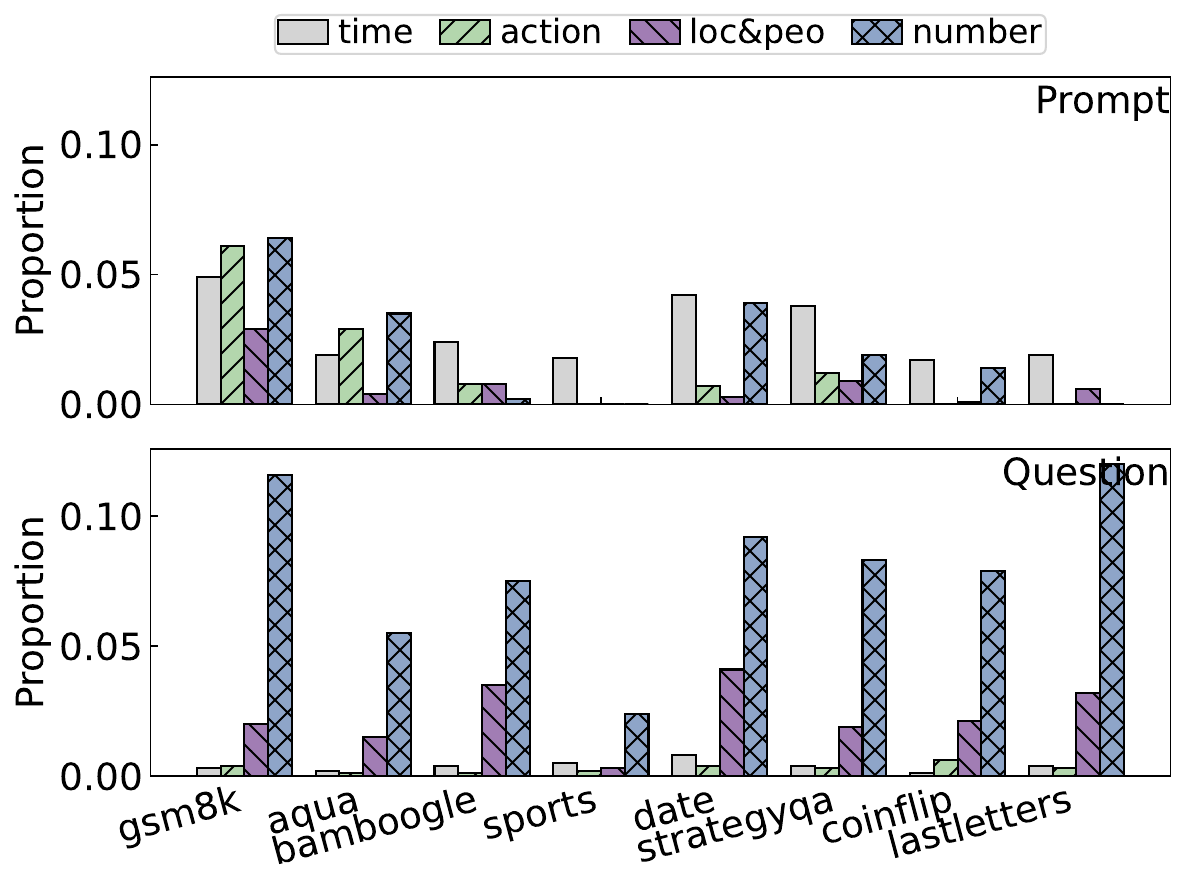}
        \caption{\textit{SVAMP}}
        \label{fig:transfer_test_27b_c}
    \end{subfigure}
    \begin{subfigure}{0.3\linewidth}
        \includegraphics[width=\linewidth]{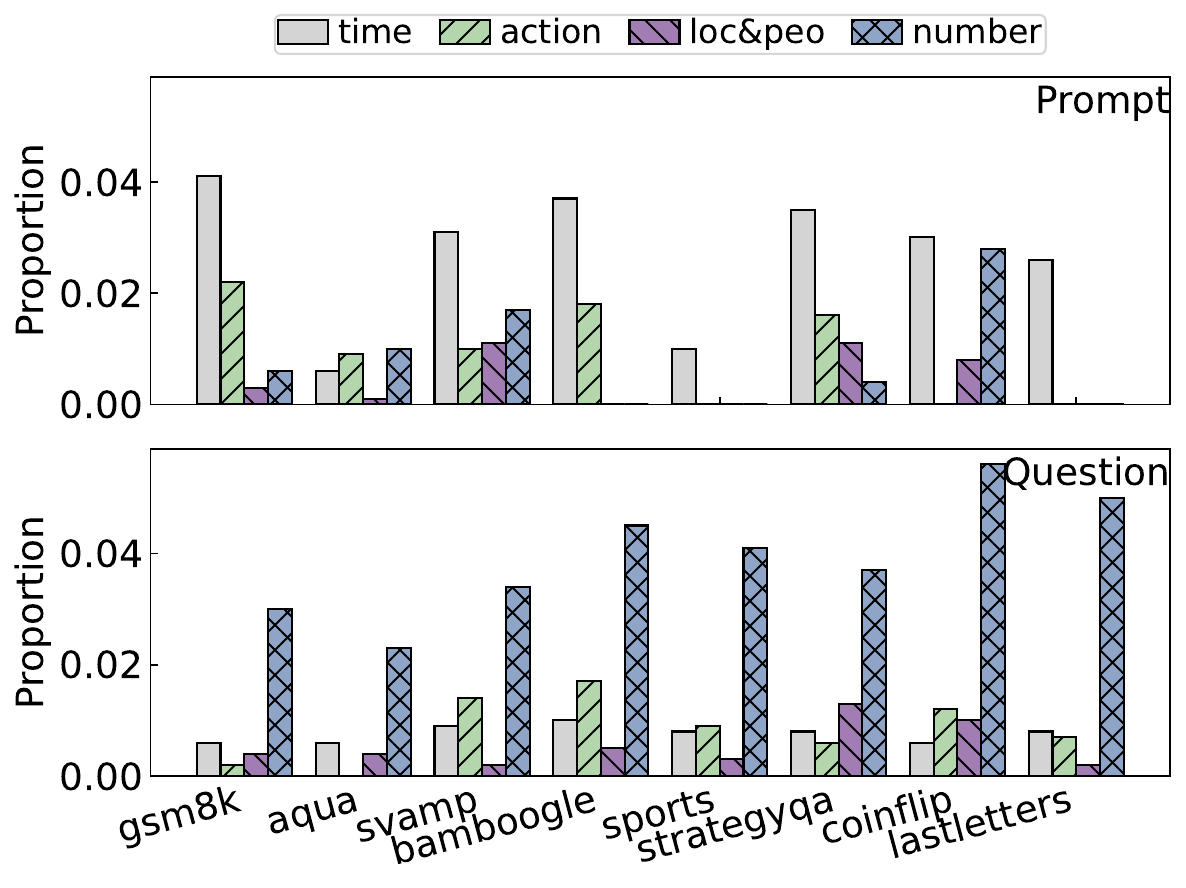}
        \caption{\textit{Date}}
        \label{fig:transfer_test_27b_d}
    \end{subfigure}
  \begin{subfigure}{0.3\linewidth}
        \includegraphics[width=\linewidth]{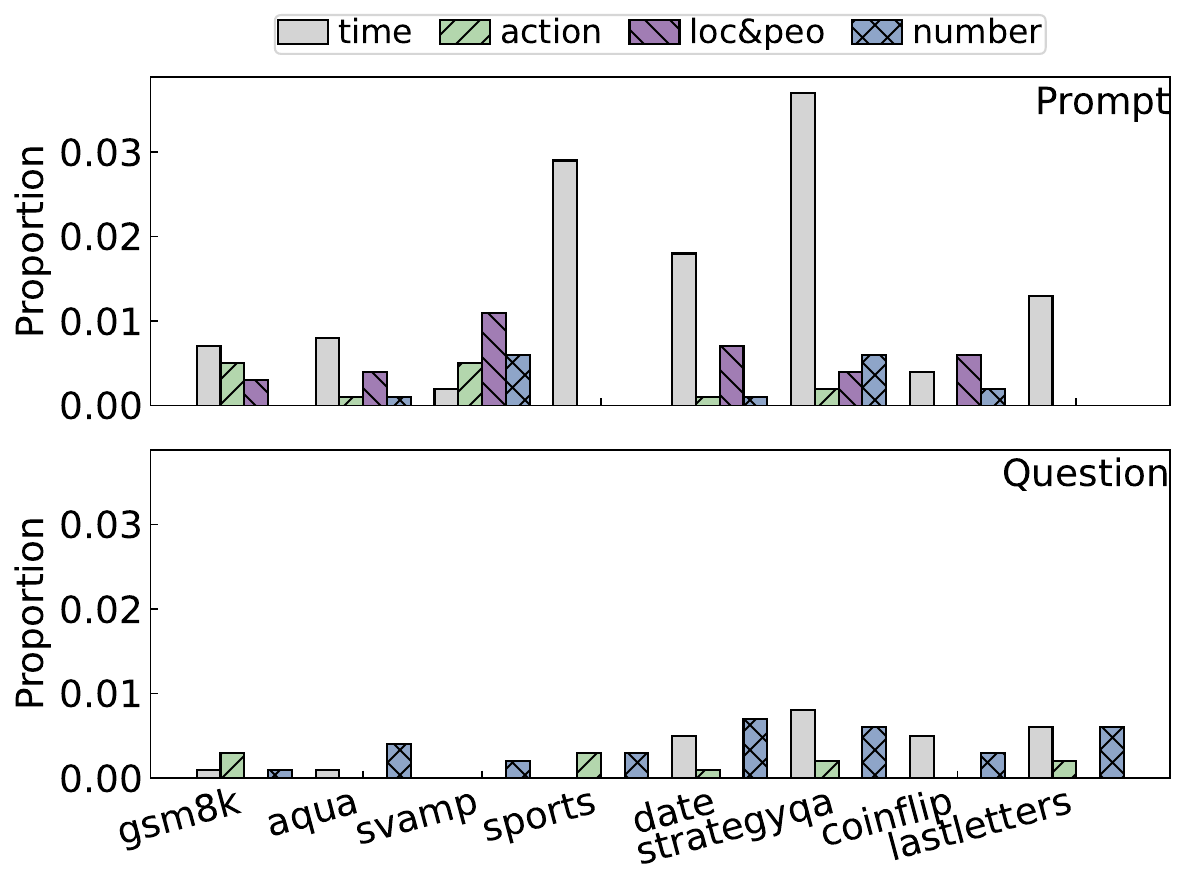}
        \caption{\textit{Bamboogle}}
        \label{fig:transfer_test_27b_e}
    \end{subfigure}
    \begin{subfigure}{0.3\linewidth}
        \includegraphics[width=\linewidth]{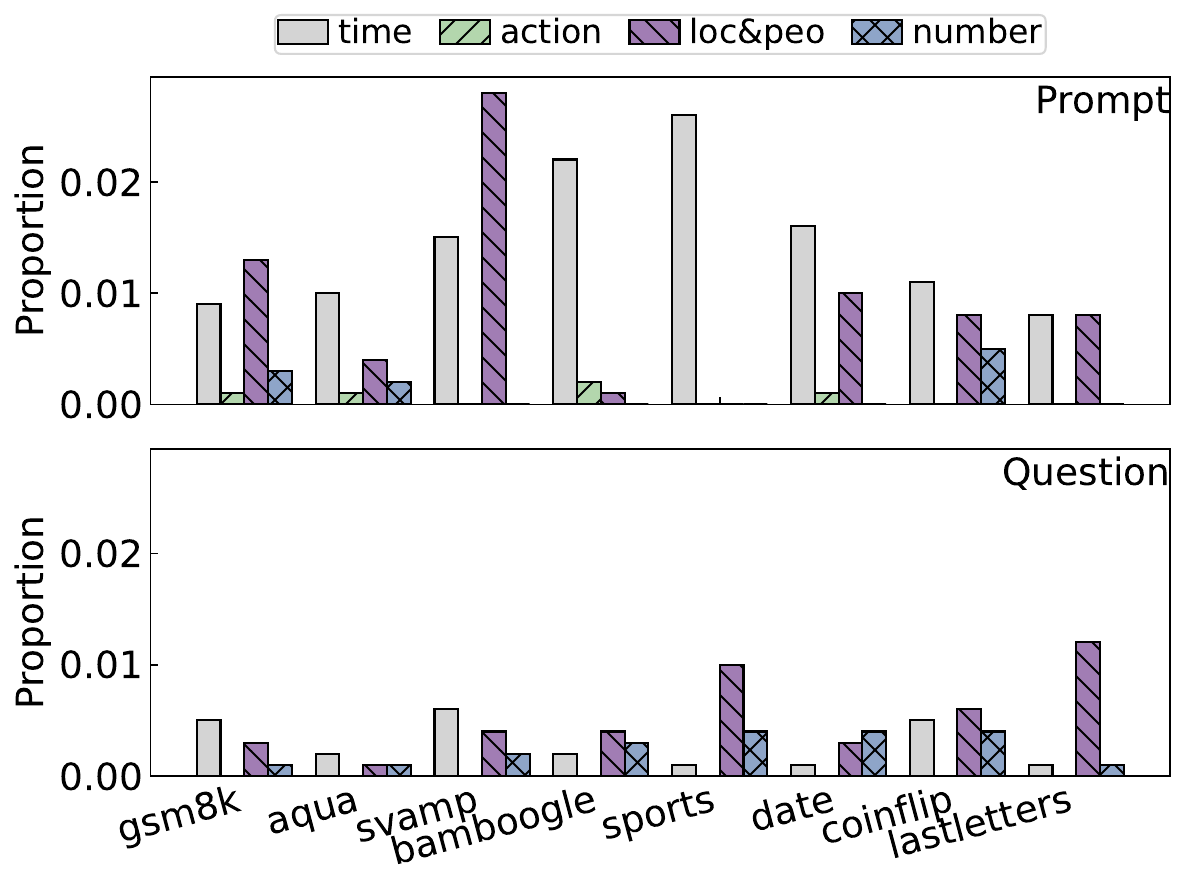}
        \caption{\textit{StrategyQA}}
    \end{subfigure}
    
     \begin{subfigure}{0.3\linewidth}
        \includegraphics[width=\linewidth]{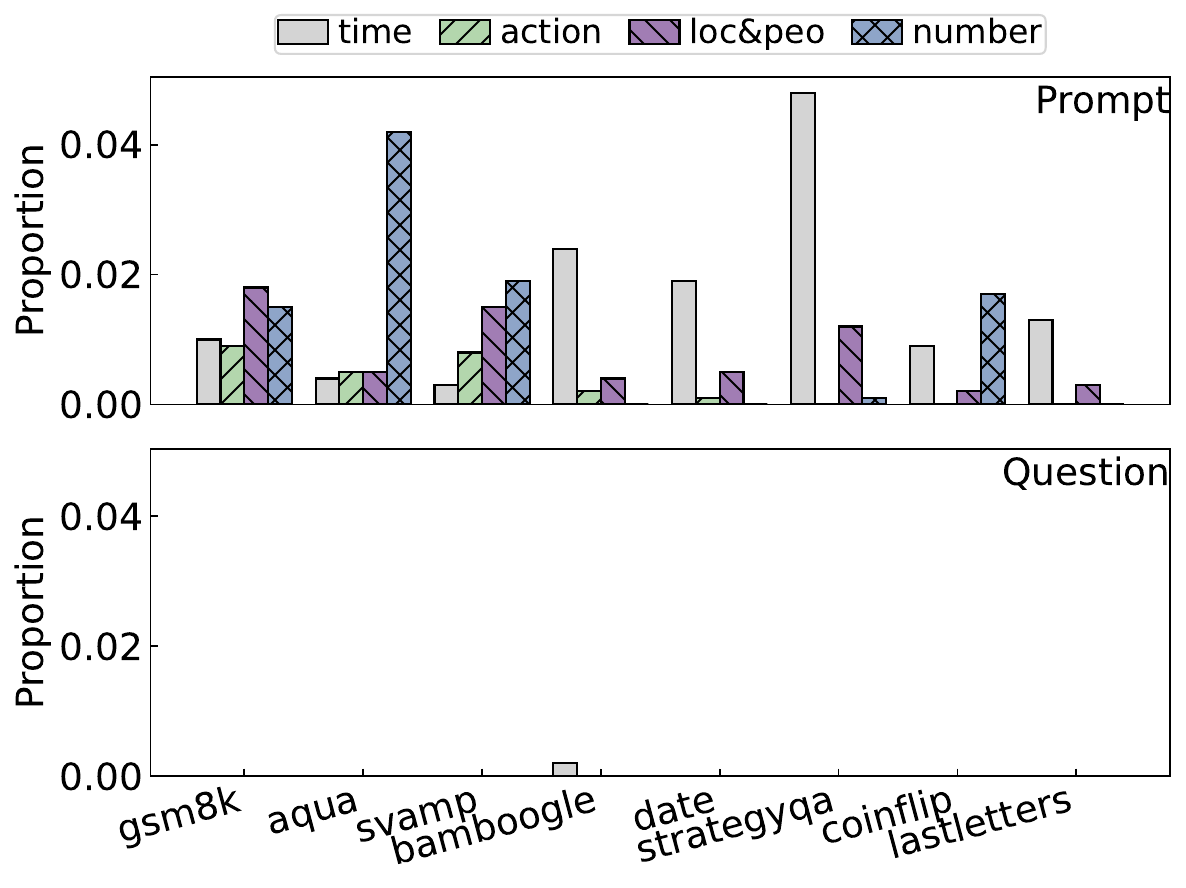}
        \caption{\textit{Sports}}
    \end{subfigure}
    \begin{subfigure}{0.3\linewidth}
        \includegraphics[width=\linewidth]{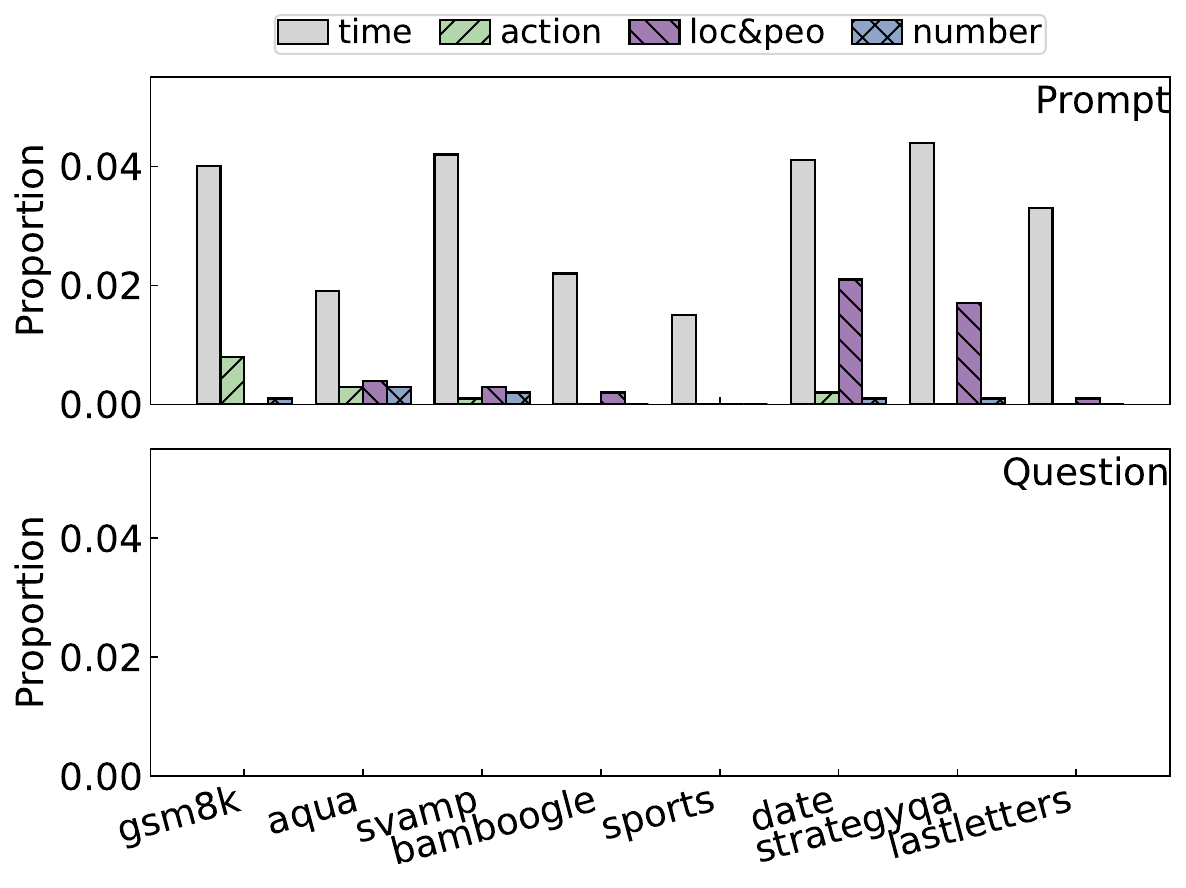}
        \caption{\textit{Coin Flip}}
    \end{subfigure}
    \begin{subfigure}{0.3\linewidth}
        \includegraphics[width=\linewidth]{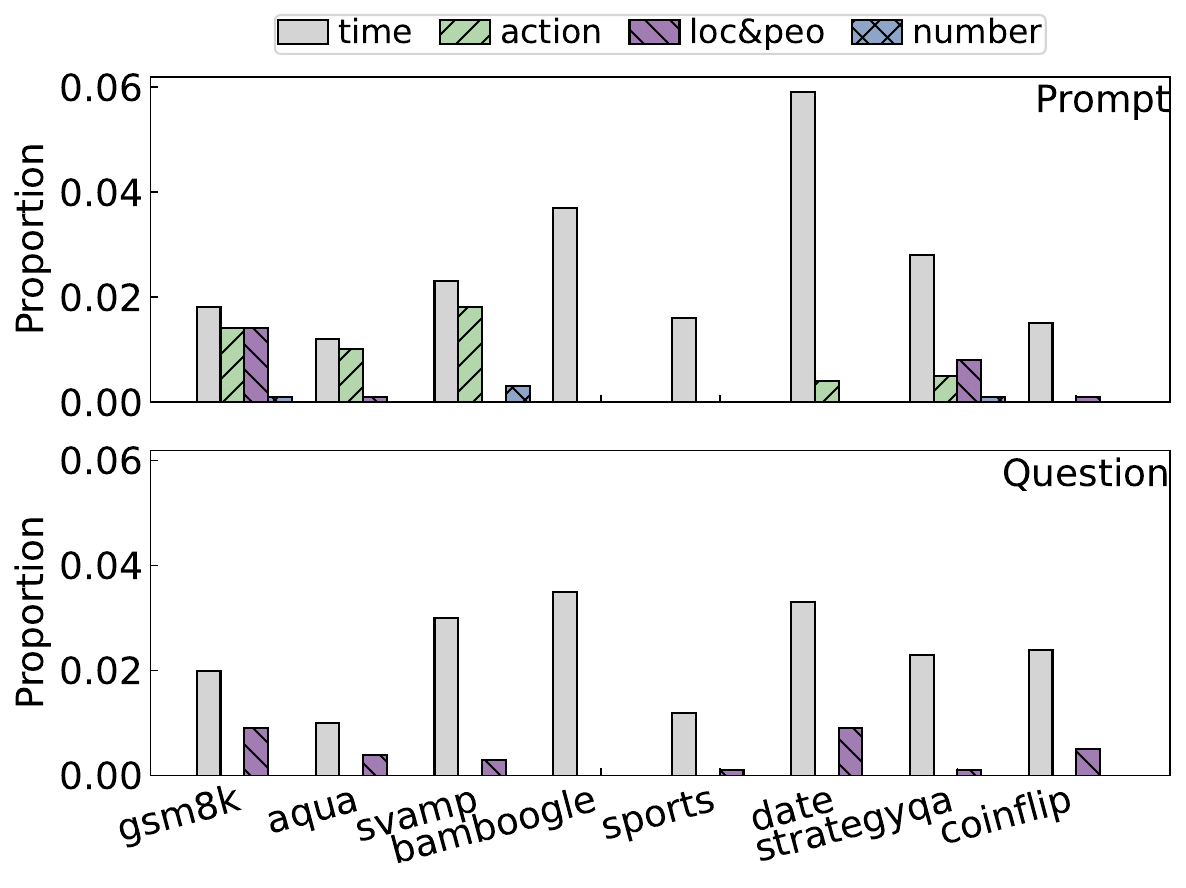}
        \caption{\textit{Last Letter}}
        \label{fig:transfer_test_27b_i}
    \end{subfigure}
\vspace{-2pt}
\caption{Results of \textit{\textbf{\textcolor{darkgreen}{transfer}}} test for Gemma2-27b, comparing test point \textbf{\textcolor{darkred}{overlap}} between model-generated content and exemplars (upper) or input questions (lower). See Figures \ref{fig:transfer_test_2b}, \ref{fig:transfer_test_9b}, and \ref{fig:transfer_test_13b} for complete results.}
\label{fig:transfer_test_27b}
\end{figure*}

\vspace{-3pt}
\paragraph{Fine-grained Analysis (Test Points Match)} 
Inspired by computer vision methods for comparing actions using key points \citep{wan2018survey,zou2024cross}, we applied a similar approach with ``test points'' to assess whether the model's generated content aligns more with the exemplar or pre-trained knowledge. We predefined four types of test points:
\begin{enumerate}
\vspace{-3pt}
    \item \textit{time} test points: including conjunctions like ``first, therefore, after'' that indicate sequence.
    \vspace{-6pt}
    \item \textit{action} test points: including verbs like ``add, divide'' that denote mathematical operations.
    \vspace{-6pt}
    \item \textit{loc\&peo} test points: including location and person words like ``table, outside, we, he'' that denote specific people or places.
    \vspace{-6pt}
    \item \textit{number} test points: including ``two, 53'', etc. 
\end{enumerate}
For the details of test points, please refer to Table \ref{tab:test_points} in Appendix.  We first manually extracted words belonging to each test point category from each dataset, and then we used exact match to determine the statistical presence of these test points. It is worth noting that,  test point templates are not universally applicable. Therefore, we took the union of the test points from all datasets.

\noindent \textbf{Analysis.} Figure \ref{fig:test_points} presents the statistical analysis of test point matches in the content generated by various models when using CoT across multiple datasets. The experimental results reveal several notable observations.

First, regarding the datasets, the arithmetic reasoning datasets (i.e., AQuA, GSM8K, and SVAMP) have yielded consistent results. Specifically, the proportion of \textit{action} and \textit{number} test points included in the generated content is relatively high (compared to the proportion of test points in the exemplars), while \textit{time} and \textit{loc\&peo} are relatively low. The other datasets have also shown another consistent results: the proportion of \textit{time} test points is relatively high, whereas \textit{action} test points are relatively low. Since \textit{time} test points (e.g., ``first'', ``then'', ``so'', etc.) in the exemplars serve to indicate the sequence of reasoning steps, their presence in the generated content suggests that the model is imitating the format presented in the exemplars.

Second, from the perspective of models, the proportions of the four types of test points in the generated content do not differ significantly among different models. This indicates that these exemplar formats are easy for models to \textit{imitate}, even smaller ones like Gemma2-2b.

Finally, the overall proportion of test points in the generated content is relatively small, which indicates that the majority of the content generated by the model consists of personalized responses based on its \textit{understanding} of the question.

\vspace{-3pt}
\paragraph{Coarse-grained Analysis (Transfer Test)} The aforementioned experiment illustrated the model’s mimicry of specific words in the generated content. To evaluate the mimicry at a higher scale, we drew inspiration from the concept of transfer learning \cite{pan2009survey}, transferring exemplars from one task to another. Specifically, we migrated exemplars within and between three types of reasoning tasks: arithmetic reasoning, commonsense reasoning, and symbolic reasoning. 

\noindent \textbf{Analysis.}  Figure \ref{fig:transfer_test_27b} shows the test point match between model-generated content and the input question (lower part) and the exemplar (upper part).

It can be observed that in the transfer use of CoT prompts across different tasks, the test point most frequently shared between the model-generated content and the question content is ``number'' (e.g., Figure \ref{fig:transfer_test_27b_a} to Figure \ref{fig:transfer_test_27b_d}), while the test point most frequently shared with the tested prompt (i.e., exemplar) is ``time'' (e.g., Figure \ref{fig:transfer_test_27b_a} to \ref{fig:transfer_test_27b_i}). This experiment further supports the above empirical hypothesis that the model understands the purpose of using CoT prompts, as it can both leverage its pre-trained knowledge to handle tasks and comprehend the template embodied in the prompt, thereby outputting the response in the given template format.

In addition, we also observed some interesting experimental results. For example, in the Sports and Coin Flip datasets, the generated content has no overlap of test points with the input question. However, since this is far from the main topic of this paper, we will not discuss it here.

\begin{figure}[!ht]
\centering
    \includegraphics[width=0.523\columnwidth]{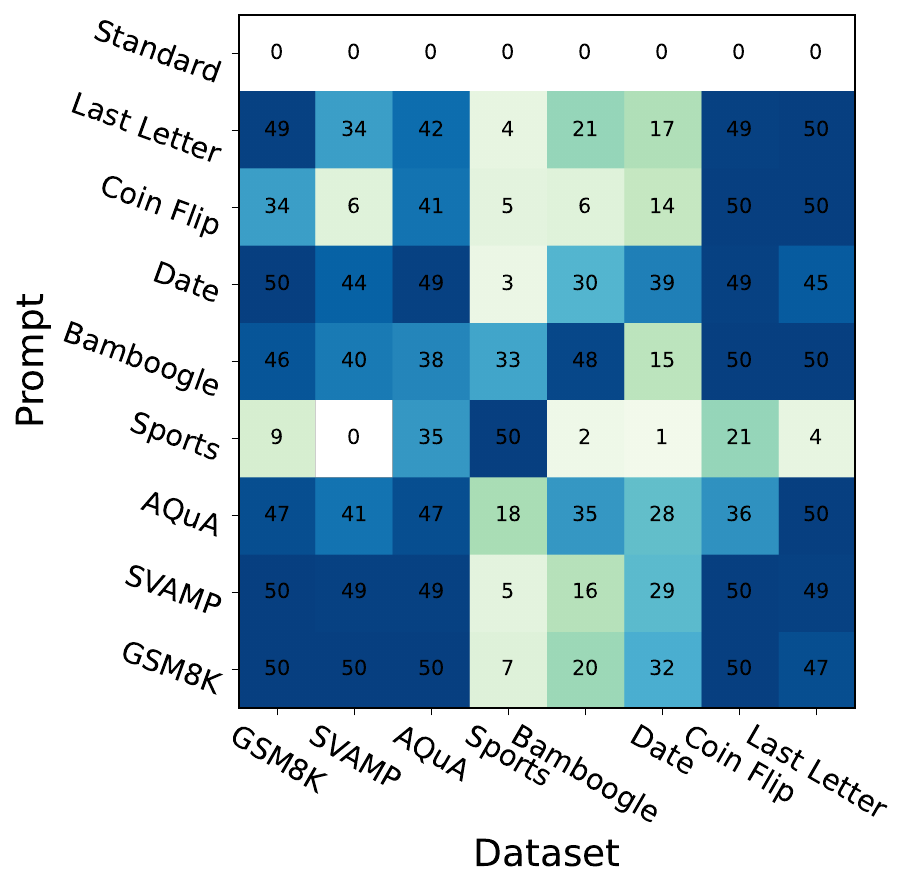}
    \includegraphics[width=0.4\columnwidth]{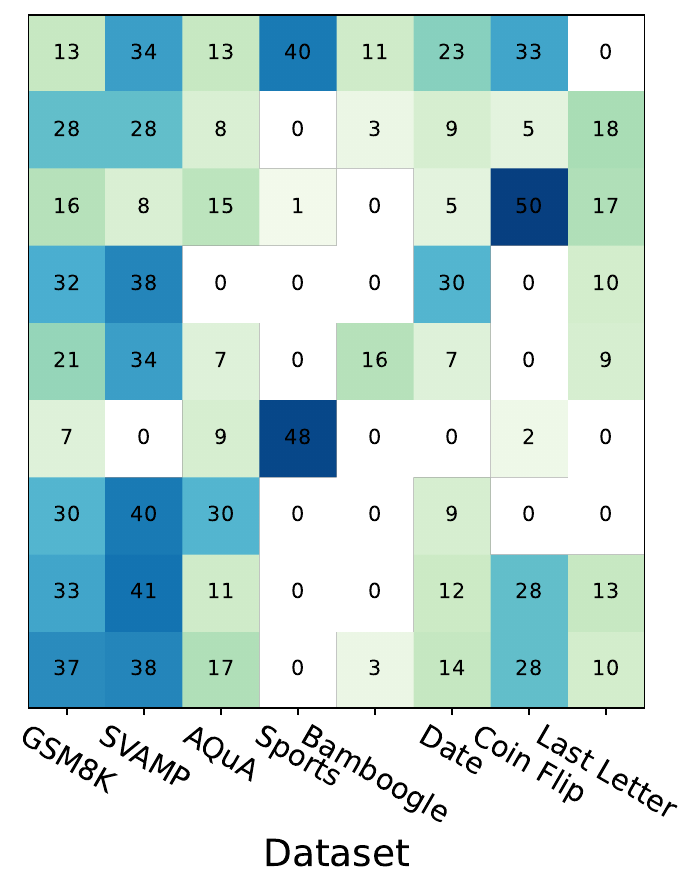}
\vspace{-2pt}
\caption{Number of samples \textbf{\textcolor{darkred}{imitating}} exemplars (left) and also  answering \textbf{\textcolor{darkgreen}{ correctly}} (right) for Gemma2-9b. See Figure \ref{fig:hotmap2b}, \ref{fig:hotmap13b}, and \ref{fig:hotmap27b} for complete results.}
\vspace{-2pt}
\label{fig:heatmap}
\end{figure}

The two above experiments show LLMs' imitation of the CoT exemplar format and their understanding of the input question. To explore the correlation between imitation and understanding, Figure \ref{fig:heatmap} displays the number of samples imitating the exemplar (left) and those also generating correct answers (right) during the transfer test. Details on quantifying imitation are in Appendix \ref{sec:quantify_imitation}.


\textbf{Analysis.}  We observe from the diagonal results in Figure \ref{fig:heatmap} (left) that the model consistently follows the CoT format. For example, in the Coin Flip task, across 50 test questions, whether using its own CoT exemplar or transferring prompts from other tasks (such as Last Letter Concatenation or GSM8K), the model's responses consistently adhere to the CoT format.  
Comparing the left and right figures reveals an unexpected finding: better imitation with transferred prompts leads to higher accuracy. This suggests that successful imitation requires the model to understand the exemplars, resulting in more accurate responses. Without this understanding, the model fails to answer correctly.


\vspace{-7pt}
\begin{tcolorbox}[colback=gray!30, colframe=gray!30, boxrule=1pt, rounded  corners]
  \textbf{\textit{Takeaway:}} LLMs  imitate exemplar formats while integrating them with their understanding of the question for answers.
\end{tcolorbox}
\vspace{-2pt}


\begin{figure*}[htbp]
\centering
    \begin{subfigure}{0.32\linewidth}
        \includegraphics[width=\linewidth]{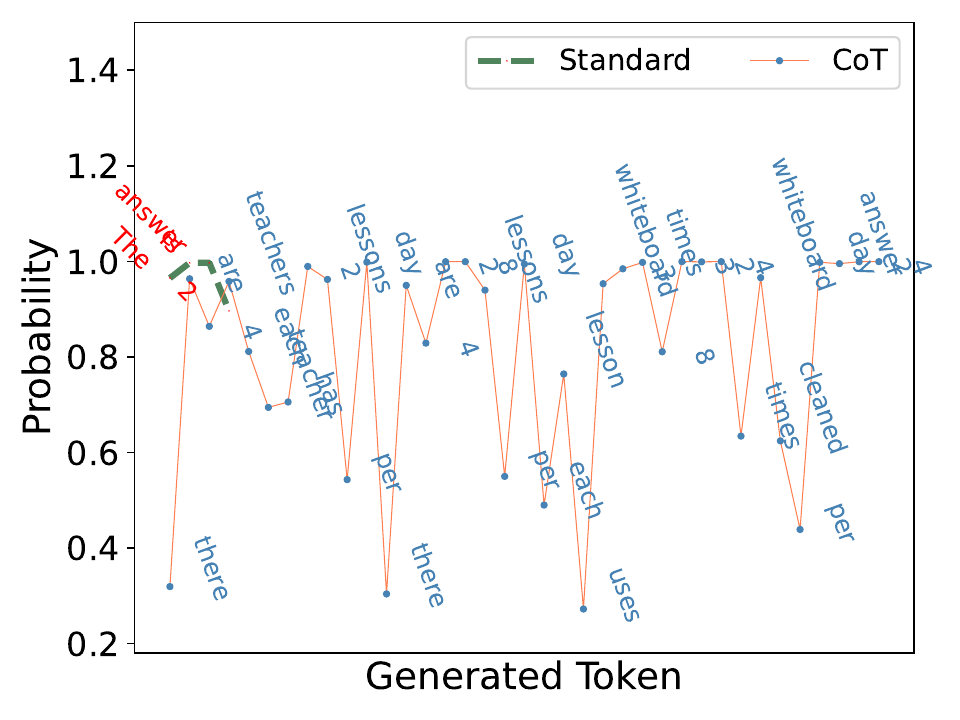}
        \caption{\textit{GSM8K}}
    \end{subfigure}
    \begin{subfigure}{0.32\linewidth}
        \includegraphics[width=\linewidth]{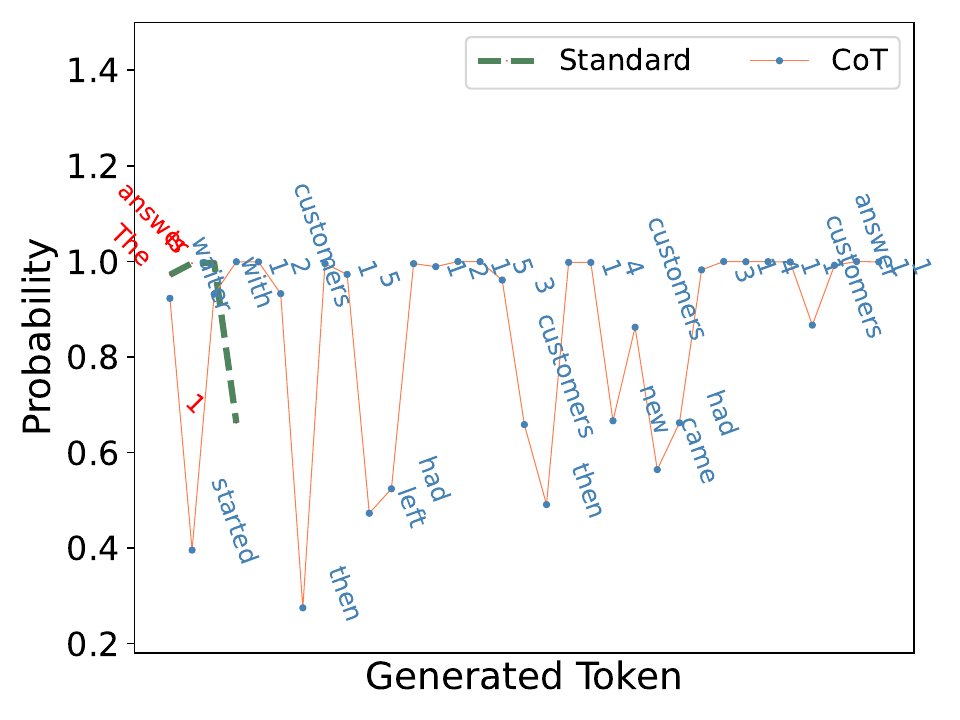}
        \caption{\textit{SVAMP}}
    \end{subfigure}
    \begin{subfigure}{0.32\linewidth}
        \includegraphics[width=\linewidth]{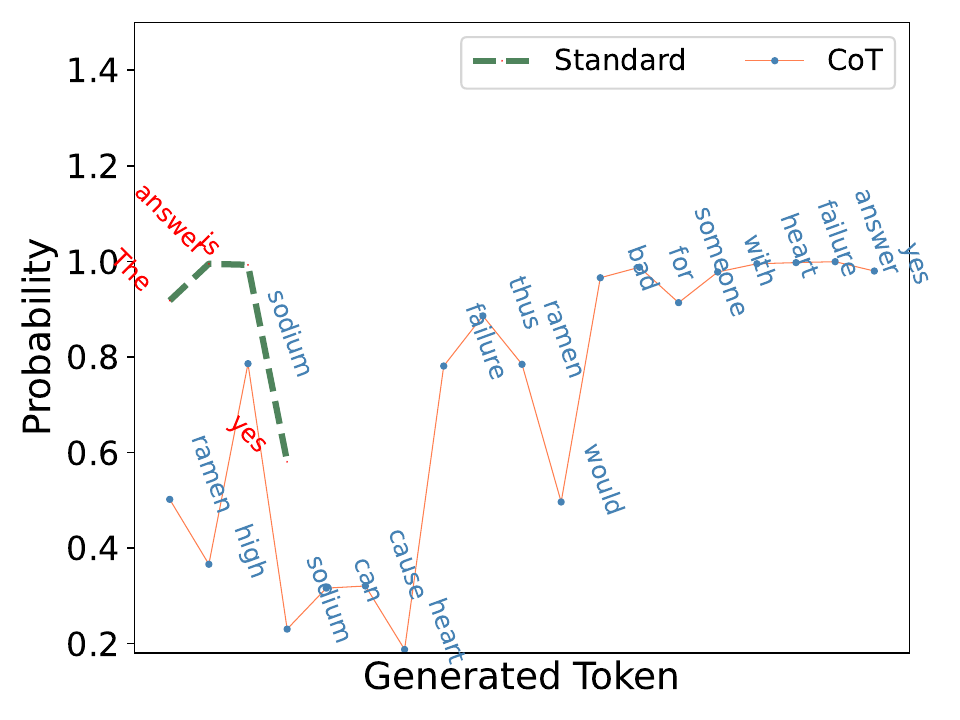}
        \caption{\textit{StrategyQA}}
    \end{subfigure}
    \begin{subfigure}{0.32\linewidth}
        \includegraphics[width=\linewidth]{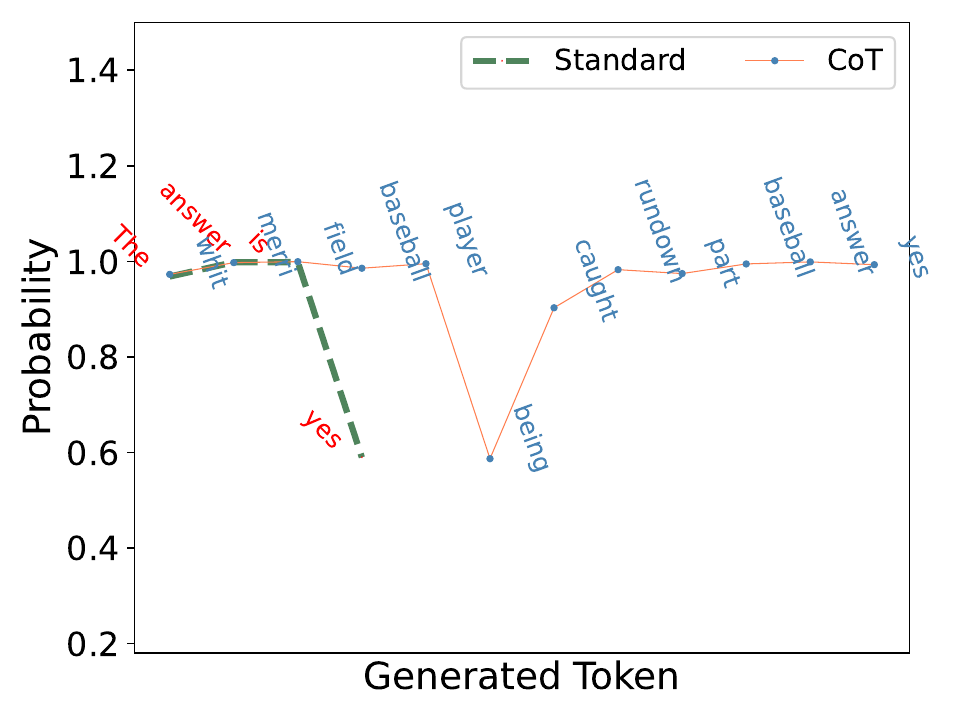}
        \caption{\textit{Sports}}
    \end{subfigure}
    \begin{subfigure}{0.32\linewidth}
        \includegraphics[width=\linewidth]{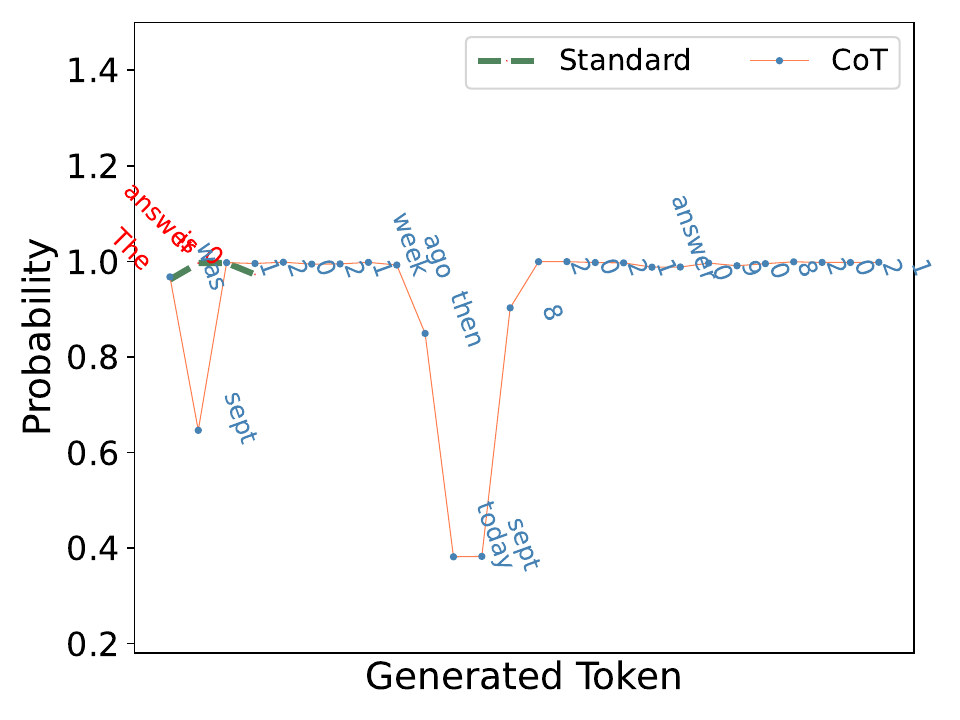}
        \caption{\textit{Date}}
    \end{subfigure}
    \begin{subfigure}{0.32\linewidth}
        \includegraphics[width=\linewidth]{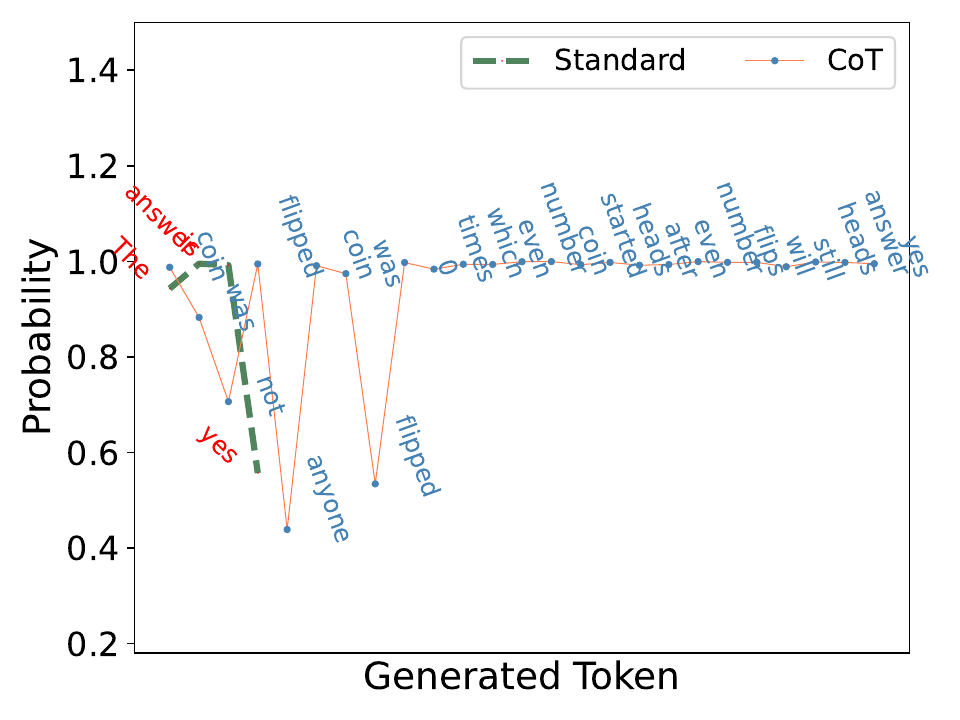}
        \caption{\textit{Coin Flip}}
    \end{subfigure}
\vspace{-2pt}
\caption{The normalized \textbf{\textcolor{darkred}{logits value}} of each generated token (Gemma2-9b is reported. See Figures \ref{fig:question2_logits_value_2b}, \ref{fig:question2_logits_value_13b}, and \ref{fig:question2_logits_value_27b} for other
models and other datasets)}
\label{fig:question2_horizontal_1}
\end{figure*}

\subsection{RQ-2: Clustering or Spread?}
In the previous step, we investigated the source of knowledge the model relies on to answer questions. In this section, we delve deeper into the internal workings of the model to examine the changes that occur in the projection space.

The generated text is derived from logits through a decoding strategy, meaning any change in the text reflects a change in logits. So, what behavioral shifts does a CoT-style prompt cause? \textit{To our knowledge, no prior work has explored the internal changes in LLMs when using CoT prompts.} We are the first to investigate CoT's logits patterns during the decoding phase.

To address these questions, we conducted experiments along two dimensions: the horizontal dimension tracks the evolution of logits over the generated token sequence, while the vertical dimension examines logits distribution across the vocabulary at each time step.
\vspace{-3pt}
\paragraph{Horizontal Dimension (Logits Value)}
During the decoding phase (with standard or CoT prompt), we obtain each generated token and its corresponding normalized logits (i.e., probability). Note that we removed spaces and newlines, and skipped spaces, punctuations, articles, and some prepositions for better visualization.

\noindent \textbf{Analysis.} Figure \ref{fig:question2_horizontal_1} shows the generated tokens and their probabilities. With a standard prompt, token probabilities remain consistently high and stable. In contrast, CoT prompts cause sharp drops and oscillations in probability values. However, when generating the final answer (e.g., ``the answer is...''), the CoT pattern stabilizes, resembling the standard prompt. For instance, on the GSM8K dataset, the average token probability is around 0.95 with a standard prompt, compared to 0.85 with CoT.

Additionally, with a standard prompt, while ``the answer is'' generates high probabilities, the probabilities for the final answer drop sharply. This sharp drop is absent with CoT prompts.

\begin{table}[!t]
\centering
\resizebox{0.9\columnwidth}{!}{%
\begin{tabular}{@{}l|l@{}}
\toprule
\textbf{Datasets} & \textbf{Top 10 tokens} \\
\midrule
GSM8K & ., 0, So, 2, the, \$, 5, 1, 4 \\
SVAMP & ,, ., 1, 2, So, the, 7, 0, 5 \\
AQuA & ., the, 0, 1, So, is, 2, (, of \\
\midrule
Bamboogle & :, the, is, of, ., was, person, What, The, in \\
Sports & ., is, the, a, So, part, of, answer, player \\
Date & 1, /, ., is, 0, 2, 9, the, So \\
\midrule
Coin Flip & ., coin, pped, the, number, an, is, f, odd, So \\
Last Letter Concatenation & ``, ``., is, last, The, ., So, the, aten \\
\bottomrule
\end{tabular}%
}
\vspace{-5pt}
\caption{The top 10 tokens with the \textbf{\textcolor{darkyellow}{lowest}} probability (spaces and line breaks are omitted).}
\label{tab:top10_tokens}
\end{table}

\begin{figure*}[!t]
\centering
    \includegraphics[width=0.24\linewidth]{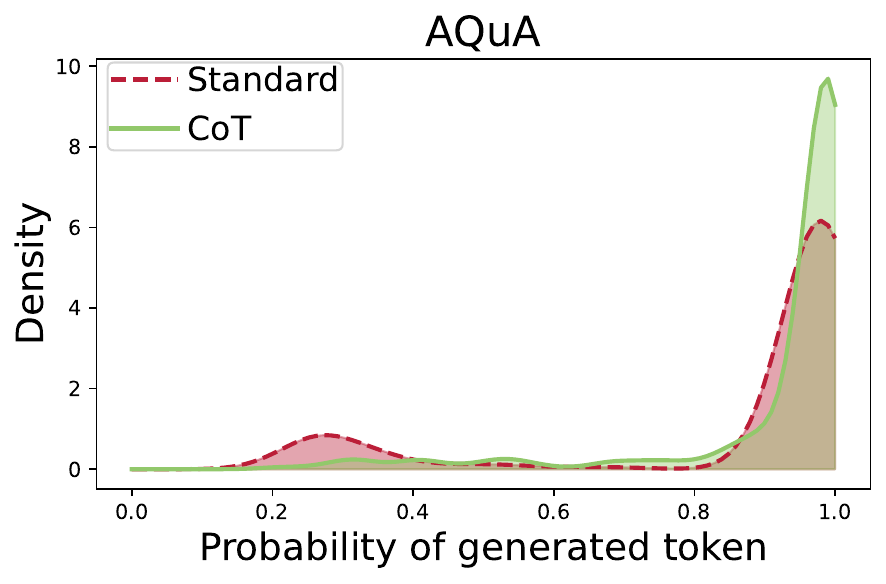}
    \includegraphics[width=0.24\linewidth]{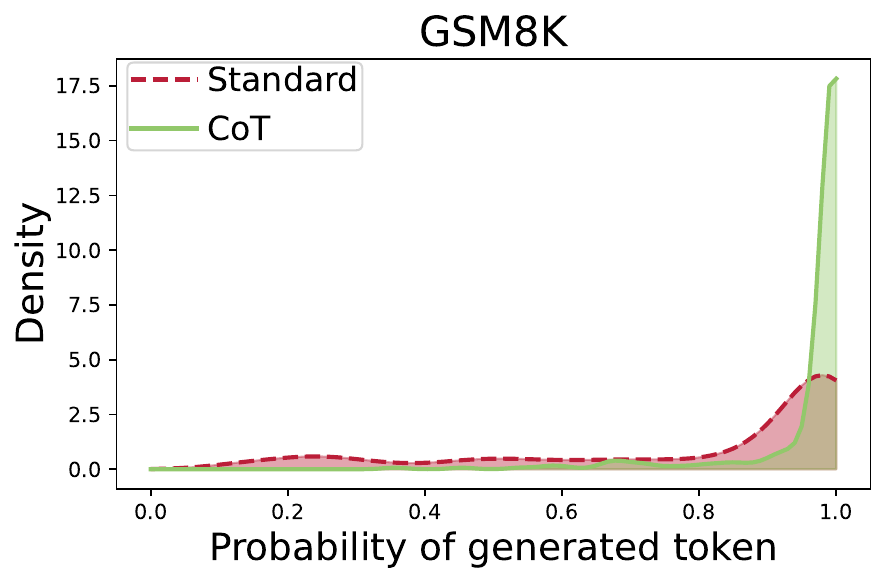}
    \includegraphics[width=0.24\linewidth]{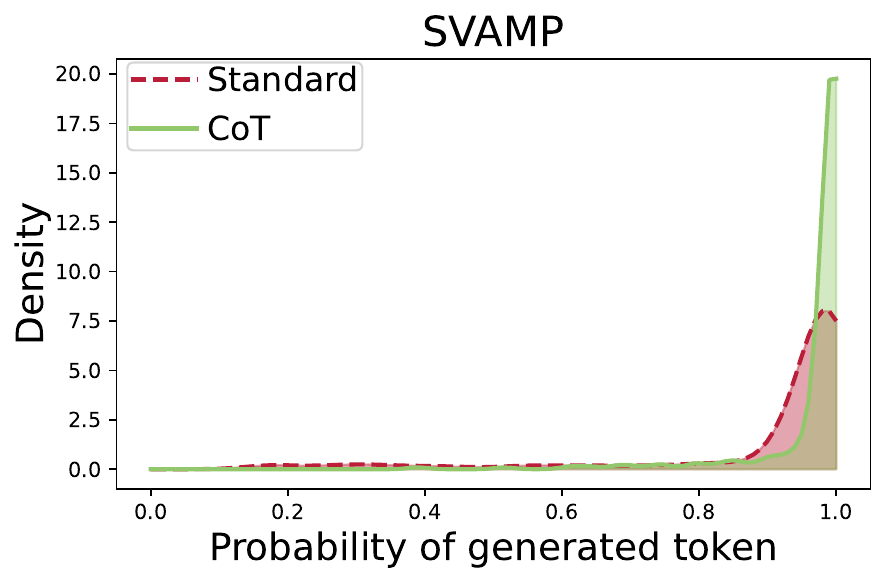}
    \includegraphics[width=0.24\linewidth]{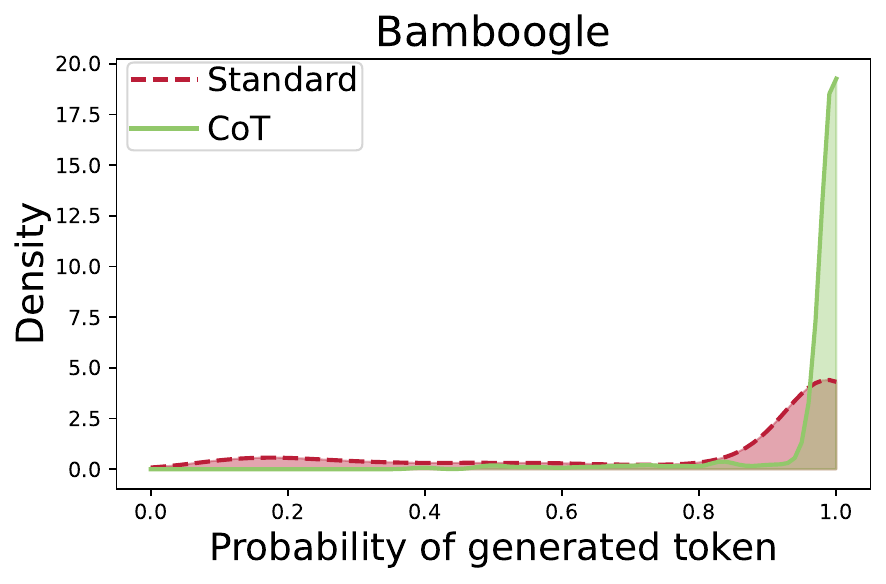}
    \includegraphics[width=0.24\linewidth]{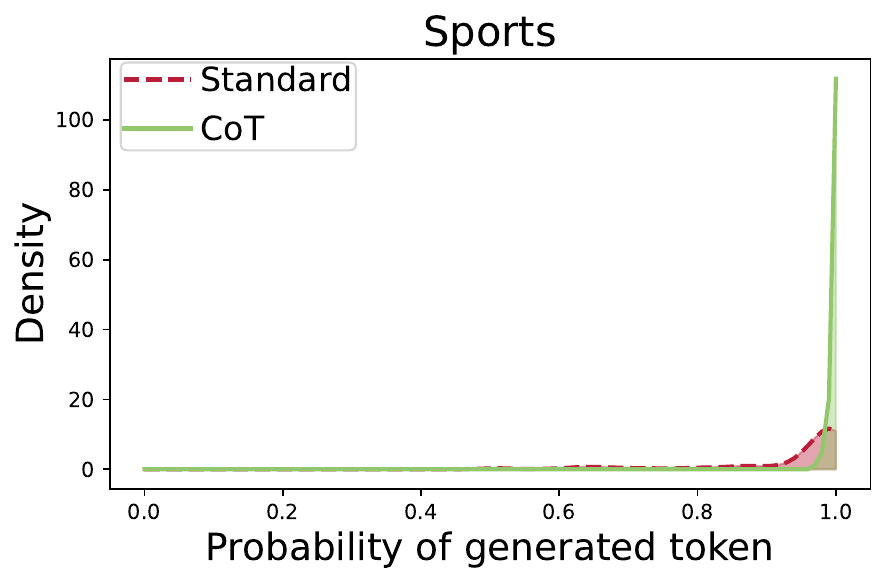}
    \includegraphics[width=0.24\linewidth]{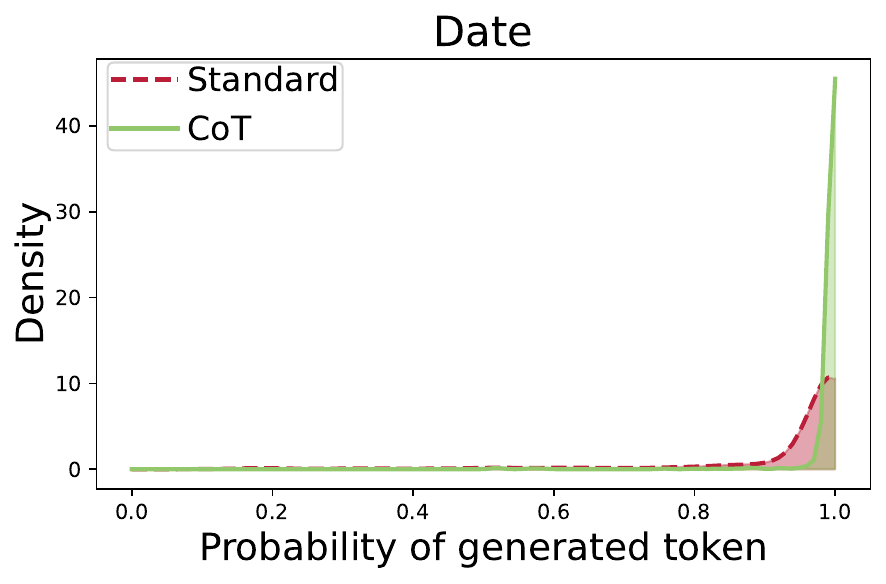}
     \includegraphics[width=0.24\linewidth]{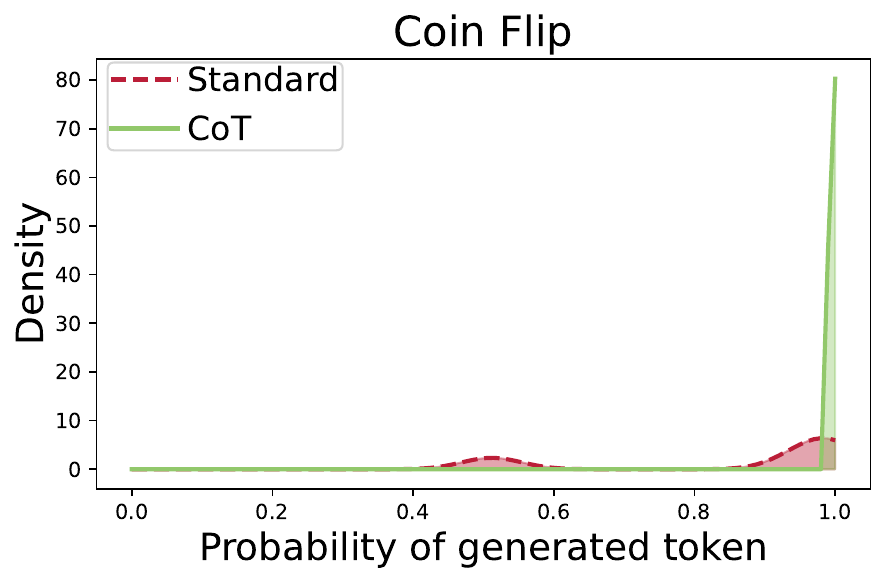}
    \includegraphics[width=0.24\linewidth]{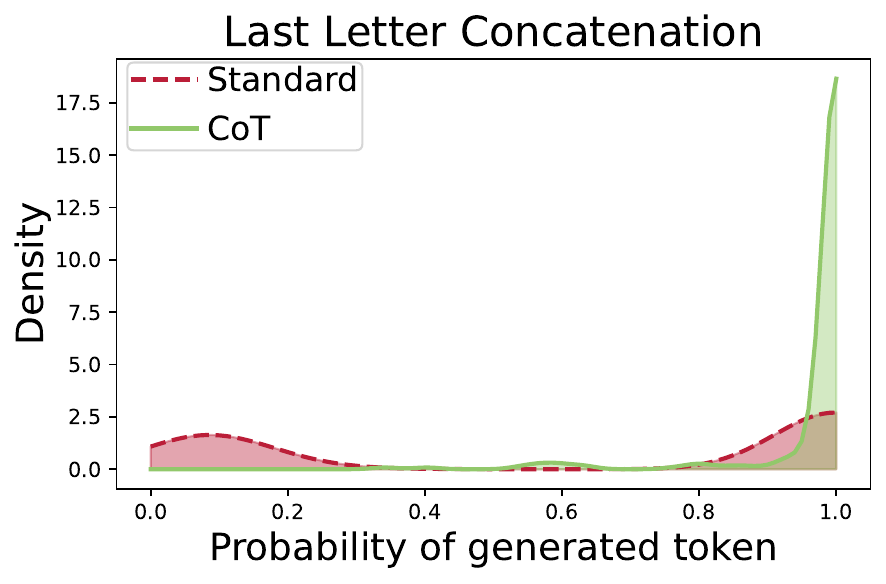}
    \vspace{-2pt}
    \caption{\textbf{\textcolor{darkred}{Kernel density}} estimation of normalized logits for `` \textbf{the answer is ...}'' generated by Gemma2-9b (see Figures \ref{fig:question2_logits_value_kernel_2b}-\ref{fig:question2_logits_value_kernel_27b} for more results).}
    \label{fig:question2_horizontal_2}
\end{figure*}
\begin{figure*}
\centering
    \begin{subfigure}{0.32\linewidth}
        \includegraphics[width=\linewidth]{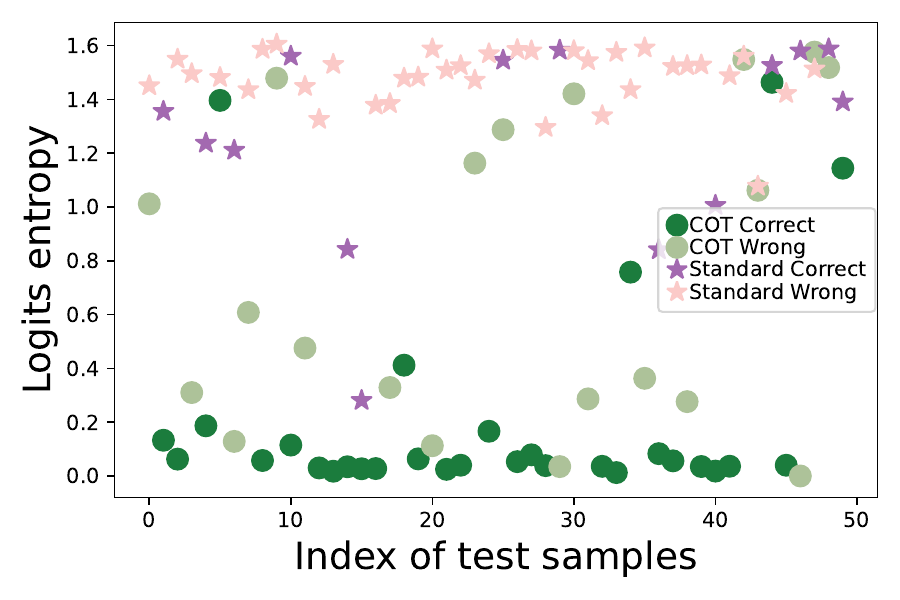}
        \caption{\textit{AQuA}}
    \end{subfigure}
    \begin{subfigure}{0.32\linewidth}
        \includegraphics[width=\linewidth]{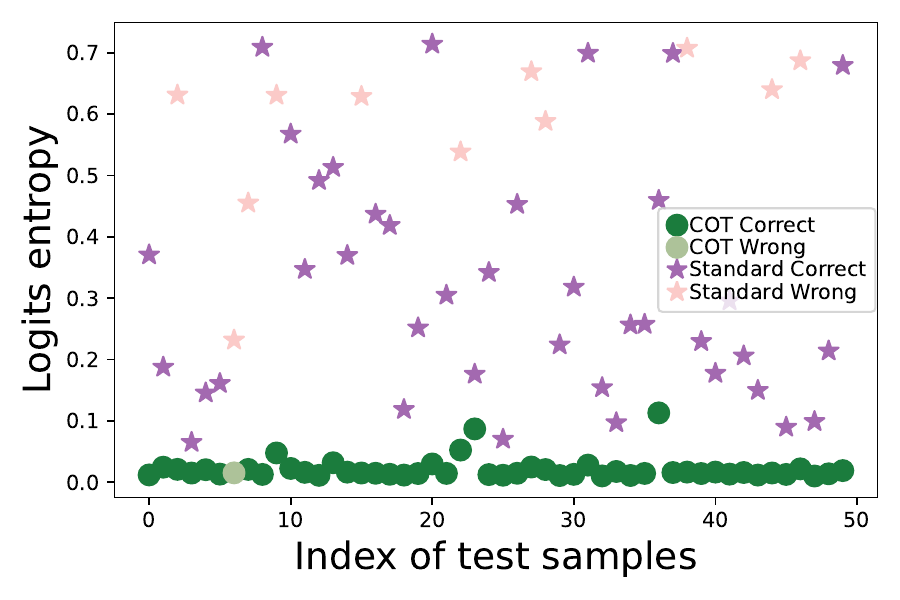}
        \caption{\textit{Sports}}
    \end{subfigure}
    \begin{subfigure}{0.32\linewidth}
        \includegraphics[width=\linewidth]{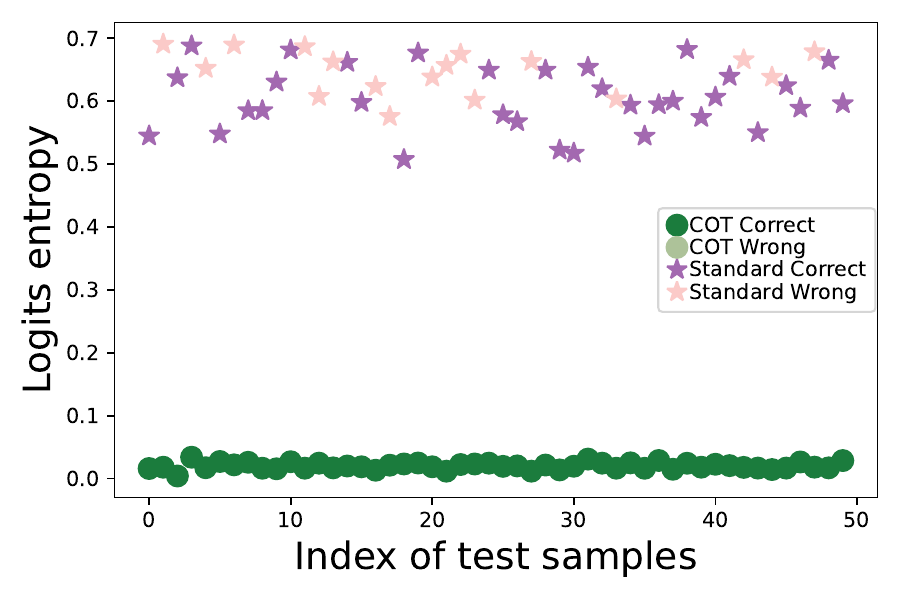}
        \caption{\textit{Coin Flip}}
    \end{subfigure}
\vspace{-2pt}
\caption{\textbf{\textcolor{darkred}{Entropy}} of normalized logits (Gemma2-27b; see Figures \ref{fig:question2_vertical_entropy_2b} to \ref{fig:question2_vertical_entropy_13b} for complete results).}
\label{fig:question2_vertical_entropy_27b}
\end{figure*}

To identify tokens with lower probability values, we calculated the difference between each token's generation probability and its adjacent tokens' probabilities, selected the top 1/3 points with larger differences, and compiled statistics across the entire test dataset.

\noindent \textbf{Analysis.}   Table \ref{tab:top10_tokens} shows the top 10 tokens with larger differences, which are mainly punctuation marks, articles, conjunctions, etc. This may be due to these words having stronger substitutability, providing the model with multiple candidate words, thus lowering individual token probabilities.

To further analyze this phenomenon, we conducted a statistical analysis of probabilities when the model is about to generate the answer, specifically when outputting  ``the answer is...''. For the CoT prompt, we obtain a probability sequence $P_{\text{CoT}} = (p_{\text{the}}, p_{\text{answer}}, p_{\text{is}}, \dots)$, and similarly, for the standard prompt, we obtain a probability sequence $P_{\text{Std}} = (p_{\text{the}}, p_{\text{answer}}, p_{\text{is}}, \dots)$. We then compute their kernel density \citep{parzen1962estimation,davis2011remarks} (refer to Appendix \ref{appendix:density} for more details) and present the results in Figure \ref{fig:question2_horizontal_2}. The figure demonstrates that the probability values in the $P_{\text{CoT}}$ sequence are generally higher and more concentrated, while the $P_{\text{Std}}$ probability sequence exhibits a more dispersed distribution. This observation potentially indicates that CoT helps the model more clearly understand what content should be generated next and in which direction to generate it.

\begin{figure*}[!t]
\centering
     \includegraphics[width=0.18\linewidth]{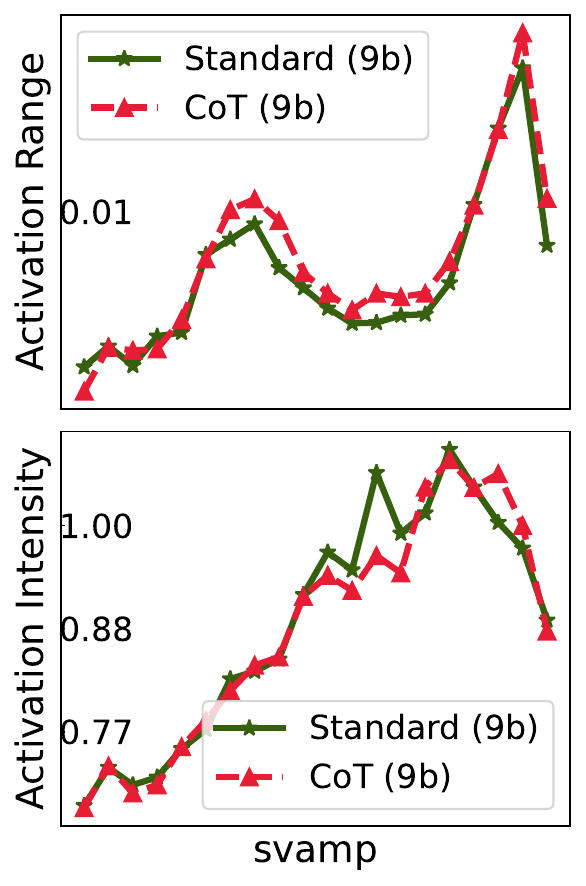}
    \includegraphics[width=0.18\linewidth]{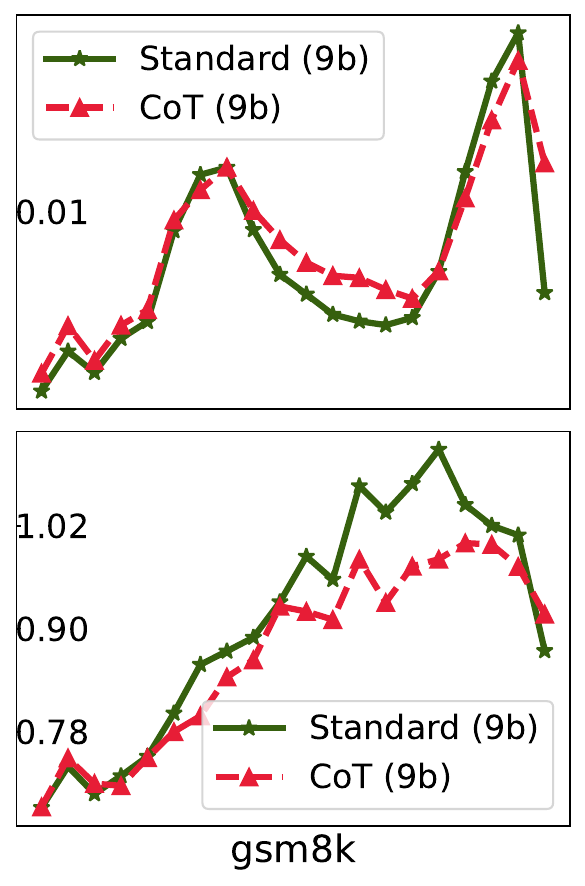}
    \includegraphics[width=0.18\linewidth]{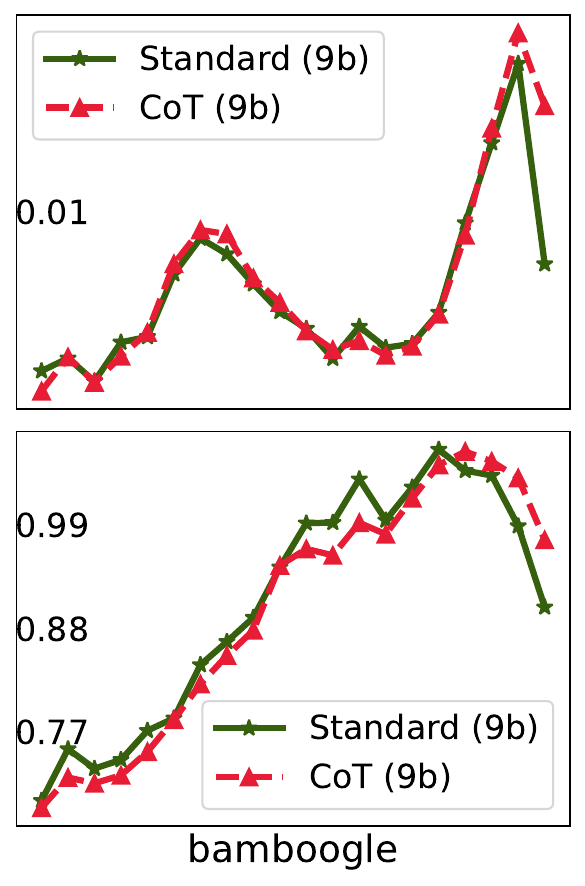}
    \includegraphics[width=0.18\linewidth]{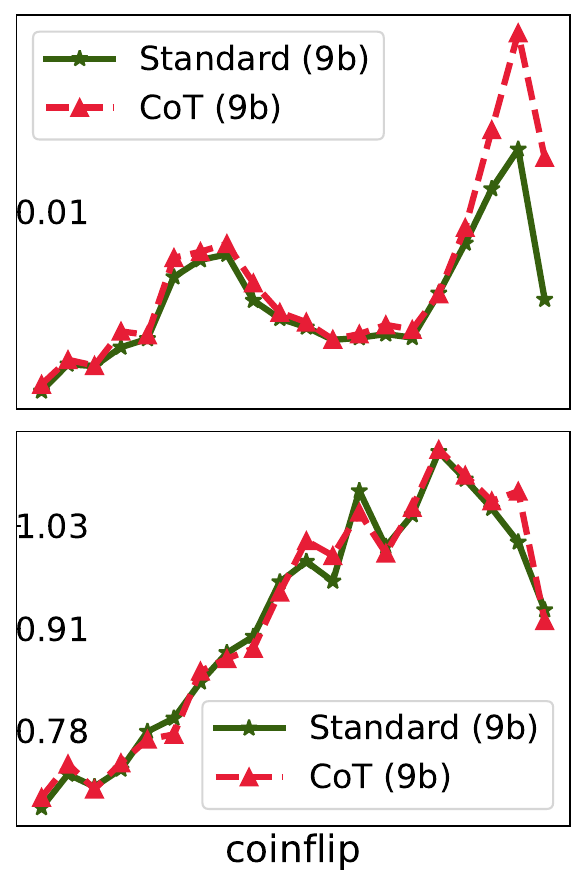}
    \includegraphics[width=0.18\linewidth]{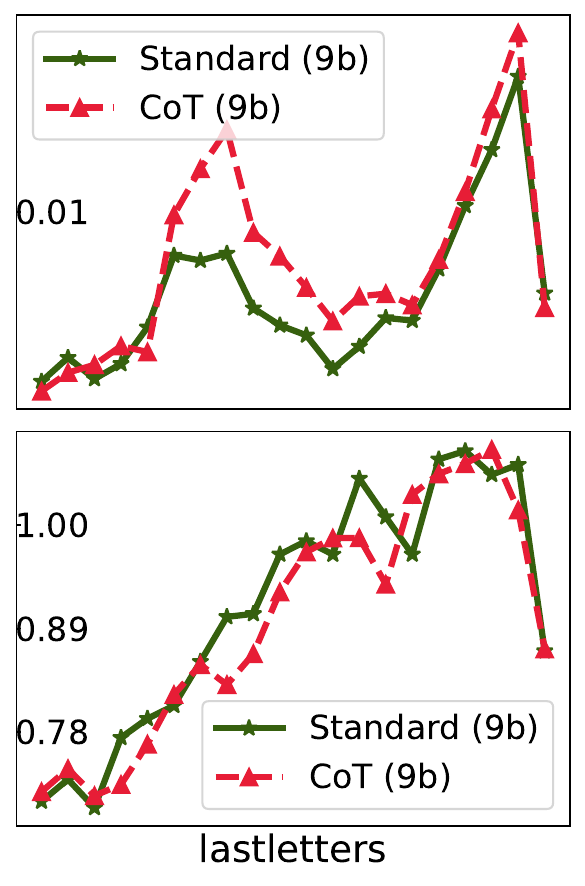}
    \vspace{-2pt}
    \caption{Neuron activation \textcolor{darkred}{\textbf{range}} and \textbf{\textcolor{darkgreen}{intensity}} across the final 20 layers of Gemma2-9b (See Figure \ref{fig:question3_activation_2b_27b} for complete results.)}
    \label{fig:question3_activation_9b}
\end{figure*}

\vspace{-3pt}
\paragraph{Vertical Dimension (Logits Distribution)}
We investigated the impact of standard and CoT-style prompts on the probability distribution across the vocabulary at each decoding step. To quantify the concentration of logits distribution, we employed entropy, a measure of uncertainty or information content in probability theory and information theory. Entropy increases with distribution dispersion and decreases with concentration \citep{pereira-etal-1993-distributional}. For a probability distribution $P = \{p_1, p_2, \ldots, p_n\}$, the entropy is calculated as $H(P) = -\sum_{i=1}^{n} p_i \log(p_i)$.

We focused on datasets with finite answer options, such as AQuA (answer space: ``a, b, c, d, e''), Sports, and Coin Flip (answer space for both: ``yes, no''). Interestingly, at the answer generation step, the top $k$ probabilities corresponded exactly to the $k$ candidate answer options. For instance, in AQuA, where $k = 5$, the top 5 highest probability values matched ``a, b, c, d, e'', while in Sports, where $k = 2$, the top 2 highest probability values corresponded to ``yes, no''.

Therefore, for each dataset, we selected the top $k$ probabilities, normalized them, and calculated the entropy values. Figure \ref{fig:question2_vertical_entropy_27b} illustrates that standard prompts result in higher entropy values, indicating that the model assigns higher probabilities to multiple candidates. In contrast, CoT prompts lead to the model focusing more on a target token, suggesting that CoT enables the model to understand more clearly the answer it should provide to the question.

\vspace{-6pt}
\begin{tcolorbox}[colback=gray!30, colframe=gray!30, boxrule=1pt, rounded  corners]
  \textbf{\textit{Takeaway:}} With CoT, token logits fluctuate during generation, but the final output shows a more concentrated logits distribution.
\end{tcolorbox}

\vspace{-2pt}
\subsection{RQ-3: Deepen and Broaden?}

\vspace{-3pt}
\paragraph{Hypothesis} We hypothesize that the effectiveness of Chain-of-Thought (CoT) prompts can be attributed to their ability to encompass a broader array of content, providing the model with additional cues that prompt a more profound and extensive exploration of the knowledge acquired during the pre-training phase. This process enables the integration of more extensive knowledge for content generation. It is important to note that the knowledge referenced here represents all content learned during the model's pre-training phase, not just factual or commonsense knowledge.

\paragraph{Quantifying Neuronal Activation} To validate our hypothesis, we explore the premise that more cues lead to more activation points, where activation points refer to activated neurons. 
GPT-style models comprise multiple layers, each consisting of an attention layer and a feedforward neural network (FFN) layer. Each FFN layer includes downsampling, nonlinear activation, upsampling, and residual connections. The FFN layers encapsulate rich semantic information and have been identified to contain neurons associated with knowledge, skills, and concepts \cite{dai-etal-2022-knowledge,dai2022knowledge,lee2017gene}. Following the approach of \cite{geva2021transformer,rai2024investigation}, we analyze the neuronal conditions in the FFN layers, defining values greater than zero as active and values less than or equal to zero as inactive.
We define two metrics to quantify neuronal activation:
\vspace{-2pt}
\begin{itemize}
\item \textbf{Activation Range:} The proportion of activated neurons relative to the total number of neurons in the FFN layer.
\vspace{-2pt}
\item \textbf{Activation Intensity:} The averaged activation value of the activated neurons.
\end{itemize}

\noindent \textbf{Analysis.}  We investigate \textit{activation evolution}, focusing on neuronal activation  across all FFN layers. As model sizes vary, we report on the last 20 layers.    Figure \ref{fig:question3_activation_9b} shows that both Standard and CoT prompts exhibit similar dynamic trends, but the CoT prompt leads to a broader activation range and lower intensity. Notably, in the final layer, which directly impacts the output, the activation range significantly expands with the CoT prompt.

These findings partially support our hypothesis that CoT encourages the model to
perform deeper and broader knowledge retrieval, as evidenced by the increased activation range and lower activation intensity in the final layers of the model.

\vspace{-6pt}
\begin{tcolorbox}[colback=gray!30, colframe=gray!30, boxrule=1pt, rounded  corners]
  \textbf{\textit{Takeaway:}} CoT activates a wider range of neurons in the final layers, suggesting more extensive knowledge retrieval.
\end{tcolorbox}
\vspace{-2pt}

\section{Conclusion}
This paper attempts to explore how the model’s behavior patterns differ when using CoT in reasoning tasks compared to standard prompts, in order to investigate the mechanism by which CoT functions. Specifically, we approach this from three perspectives: decoding, projection, and activation. We designed statistical analyses of test point matches, prompt transfer tests, logits value and distribution statistics, as well as FFN layer neuron activation statistics. We found that when using CoT, the model primarily mimics the sequential test point information from the exemplar while also understanding its numerical significance, particularly the number test points. Additionally, CoT leads to more focused token predictions and broader neuron activations.
\section*{Limitations}
Although the experiments yielded very exciting results, there are still some shortcomings in the experimental process. First, we only tested on multiple datasets within three types of reasoning tasks, so the results may not generalize to other datasets. Second, we used common prompts and did not study the number or types of CoT prompts, which means that the existing results may not generalize under other prompt conditions. Additionally, we only analyzed and observed neurons in the FFN layers, while large models contain numerous neurons, making further and broader exploration necessary. Lastly, and most importantly, we provided only a few analyses on the changes in model behavior patterns induced by CoT, without conducting a comprehensive exploration. It would be meaningful to explore more perspectives and undertake more diverse investigations.
\bibliography{custom}
\clearpage
\appendix
\onecolumn
\section{Prompts in this Paper}\label{sec:prompts}
This paper utilizes two major types of prompts: Standard prompt, CoT prompt. We adopt the Standard and CoT prompts from previous work \cite{wei2022chain, wang2023towards,kojima2022large}. The prompts for the AQuA, GSM8K, SVAMP, Sports, Date, and Coin Flip datasets are derived from \citet{wei2022chain}, the prompts for the Bamboogle dataset are from \citet{wang2023towards}, and the prompts for the Last Letter Concatenation dataset are from \citet{kojima2022large}. We made slight modifications to the above prompts; for example, when returning the answer, the CoT Prompt consistently uses ``\textit{\textbf{So the answer is...}}'' while the Standard Prompt uses  ``\textbf{\textit{The answer is...}}''.
In this paper, we provide 4 exemplars per prompt for each test dataset.The following sections will provide a detailed introduction to the two types of prompts for each dataset.

\subsection{Standard Prompt}
The Standard Prompt provides the model with several question-answer pairs, where the model directly outputs the answer without generating intermediate steps. Table \ref{standard:gsm8k} shows the Standard Prompt for the GSM8K dataset, Table \ref{standard:aqua} shows the Standard Prompt for the AQuA dataset, Table \ref{standard:svamp} shows the Standard Prompt for the SVAMP dataset, Table \ref{standard:bamboogle} shows the Standard Prompt for the Bamboogle dataset, Table \ref{standard:sports} shows the Standard Prompt for the Sports dataset, Table \ref{standard:date} shows the Standard Prompt for the Date dataset, Table \ref{standard:coin} shows the Standard Prompt for the Coin Flip dataset, and Table \ref{standard:letter} shows the Standard Prompt for the Last Letter Concatenation dataset.

\begin{table*}[ht]
    \centering
    \begin{tabularx}{0.96\textwidth}{X c r}
        \hline
         Standard-Prompt for GSM8K & & \\
        \hline
        Q: There are 15 trees in the grove. Grove workers will plant trees in the grove today. After they are done, there will be 21 trees. How many trees did the grove workers plant today?
        && \\
A: The answer is 6.& & \\
\\
Q: Leah had 32 chocolates and her sister had 42. If they ate 35, how many pieces do they have left in total?.& & \\
A: The answer is 39.& & \\
\\
Q: There were nine computers in the server room. Five more computers were installed each day, from Monday to Thursday. How many computers are now in the server room?&&\\
A: The answer is 29.&&\\
\\
Q: Michael had 58 golf balls. On Tuesday, he lost 23 golf balls. On Wednesday, he lost 2 more. How many golf balls did he have at the end of Wednesday?&&\\
A: The answer is 33.&&\\
        \hline
    \end{tabularx}
    \caption{Standard-Prompt for GSM8K dataset (used in \citet{wei2022chain})}
    \label{standard:gsm8k}
\end{table*}
\clearpage

\begin{table}
    \centering
    \begin{tabularx}{0.9\textwidth}{X c r}
        \hline
         Standard-Prompt for SVAMP & & \\
        \hline
Q: If there are 3 cars in the parking lot and 2 more cars arrive, how many cars are in the parking lot?&&\\
A: The answer is 5. &&\\
\\
Q: Jason had 20 lollipops. He gave Denny some lollipops. Now Jason has 12 lollipops. How many lollipops did Jason give to Denny? &&\\
A: The answer is 8. &&\\
\\
Q: Shawn has five toys. For Christmas, he got two toys each from his mom and dad. How many toys does he have now?&&\\
A: The answer is 9.&&\\
\\
Q: Olivia has \$23. She bought five bagels for \$3 each. How much money does she have left? &&\\
A: The answer is 8.&&\\
        \hline
    \end{tabularx}
    \caption{Standard-Prompt for SVAMP dataset(used in \citet{wei2022chain})}
    \label{standard:svamp}
\end{table}


\begin{table}[ht]
    \centering
    \begin{tabularx}{0.9\textwidth}{X c r}
        \hline
         Standard-Prompt for AQuA & & \\
        \hline
       Q: John found that the average of 15 numbers is 40. If 10 is added to each number then the mean of the numbers is?&&\\
Answer Choices: (a) 50 (b) 45 (c) 65 (d) 78 (e) 64&&\\
A: The answer is (a).&&\\
\\
Q: If a / b = 3/4 and 8a + 5b = 22, then find the value of a.&&\\
Answer Choices: (a) 1/2 (b) 3/2 (c) 5/2 (d) 4/2 (e) 7/2&&\\
A: The answer is (b).&&\\
\\
Q: A person is traveling at 20 km/hr and reached his destiny in 2.5 hr then find the distance?&&\\
Answer Choices: (a) 53 km (b) 55 km (c) 52 km (d) 60 km (e) 50 km&&\\
A: The answer is (e).&&\\
\\
Q: How many keystrokes are needed to type the numbers from 1 to 500?&&\\
Answer Choices: (a) 1156 (b) 1392 (c) 1480 (d) 1562 (e) 1788&&\\
A: The answer is (b).&&\\
        \hline
    \end{tabularx}
    \caption{Standard-Prompt for AQuA dataset(used in \citet{wei2022chain})}
    \label{standard:aqua}
\end{table}


\begin{table}[ht]
    \centering
    \begin{tabularx}{0.9\textwidth}{X c r}
        \hline
         Standard-Prompt for Bamboogle & & \\
        \hline
      Q: Who lived longer, Theodor Haecker or Harry Vaughan Watkins?&&\\
A: The answer is Harry Vaughan Watkins.&&\\
\\
Q: Why did the founder of Versus die?&&\\
A: The answer is Shot.&&\\
\\
Q: Who is the grandchild of Dambar Shah?&&\\
A: The answer is Rudra Shah.&&\\
\\
Q: Are both director of film FAQ: Frequently Asked Questions and director of film The Big Money from the same country?&&\\
A: The answer is No.
&&\\
        \hline
    \end{tabularx}
    \caption{Standard-Prompt for Bamboogle dataset(used in \citet{wang2023towards})}
    \label{standard:bamboogle}
\end{table}

\begin{table}[ht]
    \centering
    \begin{tabularx}{0.9\textwidth}{X c r}
        \hline
         Standard-Prompt for Sports & & \\
        \hline
Q: Is the following sentence plausible? ``Kyle Palmieri was called for slashing ''. & & \\
A: The answer is yes.& & \\
\\
Q: Is the following sentence plausible? ``Joao Moutinho caught the screen pass in the NFC championship ''. & & \\
A: The answer is no.& & \\
\\
Q: Is the following sentence plausible? ``Carson Wentz set the pick and roll ''.& & \\ 
A: The answer is no.& & \\
\\
Q: Is the following sentence plausible? ``Malcolm Brogdon banked the shot in ''.& & \\ 
A: The answer is yes.& & \\
        \hline
    \end{tabularx}
    \caption{Standard-Prompt for Sports dataset(used in \citet{wei2022chain})}
    \label{standard:sports}
\end{table}
\begin{table}[h]
    \centering
    \begin{tabularx}{0.9\textwidth}{X c r}
        \hline
         Standard-Prompt for Date & & \\
        \hline
Q: 2015 is coming in 36 hours. What is the date one week from today in MM/DD/YYYY? & & \\
A: The answer is 01/05/2015. & & \\
\\
Q: The first day of 2019 is a Tuesday, and today is the first Monday of 2019. What is the date today in MM/DD/YYYY? & & \\
A: The answer is 01/07/2019. & & \\
\\
Q: The concert was scheduled to be on 06/01/1943, but was delayed by one day to today. What is the date 10 days ago in MM/DD/YYYY? & & \\
A: The answer is 05/23/1943. & & \\
\\
Q: Jane was born on the last day of Feburary in 2001. Today is her 16-year-old birthday. What is the date yesterday in MM/DD/YYYY? & & \\
A: The answer is 02/27/2017.& & \\
        \hline
    \end{tabularx}
    \caption{Standard-Prompt for Date dataset(used in \citet{wei2022chain})}
    \label{standard:date}
\end{table}
\begin{table}[ht]
    \centering
    \begin{tabularx}{0.9\textwidth}{X c r}
        \hline
         Standard-Prompt for Coin Flip & & \\
        \hline
Q: A coin is heads up. Ka flips the coin. Sherrie flips the coin. Is the coin still heads up? & & \\
A: The answer is yes.& & \\
\\
Q: A coin is heads up. Maybelle flips the coin. Shalonda does not flip the coin. Is the coin still heads up? & & \\
A: The answer is no.& & \\
\\
Q: A coin is heads up. Millicent does not flip the coin. Conception flips the coin. Is the coin still heads up? & & \\
A: The answer is no.& & \\
\\
Q: A coin is heads up. Ryan flips the coin. Shaunda flips the coin. Is the coin still heads up?& & \\
A: The answer is yes.
& & \\
        \hline
    \end{tabularx}
    \caption{Standard-Prompt for Coin Flip dataset(used in \citet{wei2022chain})}
    \label{standard:coin}
\end{table}


\begin{table*}[ht]
    \centering
    \begin{tabularx}{0.9\textwidth}{X c r}
        \hline
         Standard-Prompt for Last Letter Concatenation & & \\
        \hline
Q: Take the last letters of each words in ``Tim Candace Cecil Misael'' and concatenate them.& & \\
A: The answer is mell.& & \\
\\
Q: Take the last letters of each words in ``Alina Alessandra Amina Bianca'' and concatenate them.& & \\
A: The answer is aaaa.& & \\
\\
Q: Take the last letters of each words in ``Felipe Heidi Nino Bradley'' and concatenate them.& & \\
A: The answer is eioy.& & \\
\\
Q: Take the last letters of each words in ``Lacey Nora Debra Ashleigh'' and concatenate them.& & \\
A: The answer is yaah.& & \\
        \hline
    \end{tabularx}
    \caption{Standard-Prompt for Last Letter Concatenation dataset(used in \citet{kojima2022large})}
    \label{standard:letter}
\end{table*}
\clearpage
\subsection{Cot Prompt}
The CoT Prompt adds intermediate reasoning steps to the exemplars provided to the model, guiding it to answer step by step, eventually arriving at the final answer. Table \ref{cot:gsm8k} shows the CoT Prompt for the GSM8K dataset, Table \ref{cot:aqua} shows the CoT Prompt for the AQuA dataset, Table \ref{cot:svamp} shows the CoT Prompt for the SVAMP dataset, Table \ref{cot:bamboogle} shows the CoT Prompt for the Bamboogle dataset, Table \ref{cot:sports} shows the CoT Prompt for the Sports dataset, Table \ref{cot:date} shows the CoT Prompt for the Date dataset, Table \ref{cot:coin} shows the CoT Prompt for the Coin Flip dataset, and Table \ref{cot:letter} shows the CoT Prompt for the Last Letter Concatenation dataset.
\begin{table}[ht]
    \centering
    \begin{tabularx}{0.96\textwidth}{X c r}
        \hline
         CoT-Prompt for GSM8K & & \\
        \hline
        Q: There are 15 trees in the grove. Grove workers will plant trees in the grove today. After they are done, there will be 21 trees. How many trees did the grove workers plant today?
        && \\
A: There are 15 trees originally. Then there were 21 trees after some more were planted. So there must have been 21 - 15 = 6. So the answer is 6.& & \\
\\
Q: Leah had 32 chocolates and her sister had 42. If they ate 35, how many pieces do they have left in total?& & \\
A: Originally, Leah had 32 chocolates. Her sister had 42. So in total they had 32 + 42 = 74. After eating 35, they had 74 - 35 = 39. So the answer is 39.& & \\
\\
Q: There were nine computers in the server room. Five more computers were installed each day, from Monday to Thursday. How many computers are now in the server room?&&\\
A: There were originally 9 computers. For each of 4 days, 5 more computers were added. So 5 * 4 = 20 computers were added. 9 + 20 is 29. So the answer is 29.&&\\
\\
Q: Michael had 58 golf balls. On Tuesday, he lost 23 golf balls. On Wednesday, he lost 2 more. How many golf balls did he have at the end of Wednesday?&&\\
A: Michael started with 58 golf balls. After losing 23 on Tuesday, he had 58 - 23 = 35. After losing 2 more, he had 35 - 2 = 33 golf balls. So the answer is 33.&&\\
        \hline
    \end{tabularx}
    \caption{CoT-Prompt for GSM8K dataset(used in \citet{wei2022chain})}
    \label{cot:gsm8k}
\end{table}
\begin{table}[ht]
    \centering
    \begin{tabularx}{0.9\textwidth}{X c r}
        \hline
         CoT-Prompt for SVAMP & & \\
        \hline
Q: If there are 3 cars in the parking lot and 2 more cars arrive, how many cars are in the parking lot?&&\\
A: There are originally 3 cars. 2 more cars arrive. 3 + 2 = 5. So the answer is 5. &&\\
\\
Q: Jason had 20 lollipops. He gave Denny some lollipops. Now Jason has 12 lollipops. How many lollipops did Jason give to Denny? &&\\
A: Jason started with 20 lollipops. Then he had 12 after giving some to Denny. So he gave Denny 20 - 12 = 8. So the answer is 8. &&\\
\\
Q: Shawn has five toys. For Christmas, he got two toys each from his mom and dad. How many toys does he have now?&&\\
A: Shawn started with 5 toys. If he got 2 toys each from his mom and dad, then that is 4 more toys. 5 + 4 = 9. So the answer is 9.&&\\
\\
Q: Olivia has \$23. She bought five bagels for \$3 each. How much money does she have left? &&\\
A: Olivia had 23 dollars. 5 bagels for 3 dollars each will be 5 x 3 = 15 dollars. So she has 23 - 15 dollars left. 23 - 15 is 8. So the answer is 8.&&\\
        \hline
    \end{tabularx}
    \caption{CoT-Prompt for SVAMP dataset(used in \citet{wei2022chain})}
    \label{cot:svamp}
\end{table}
\begin{table}[ht]
    \centering
    \begin{tabularx}{0.9\textwidth}{X c r}
        \hline
         CoT-Prompt for AQuA & & \\
        \hline
       Q: John found that the average of 15 numbers is 40. If 10 is added to each number then the mean of the numbers is?&&\\
Answer Choices: (a) 50 (b) 45 (c) 65 (d) 78 (e) 64&&\\
A: If 10 is added to each number, then the mean of the numbers also increases by 10. So the new mean would be 50. So the answer is (a).&&\\
\\
Q: If a / b = 3/4 and 8a + 5b = 22,then find the value of a.&&\\
Answer Choices: (a) 1/2 (b) 3/2 (c) 5/2 (d) 4/2 (e) 7/2&&\\
A: If a / b = 3/4, then b = 4a / 3. So 8a + 5(4a / 3) = 22. This simplifies to 8a + 20a / 3 = 22, which means 44a / 3 = 22. So a is equal to 3/2. So the answer is (b).&&\\
\\
Q: A person is traveling at 20 km/hr and reached his destiny in 2.5 hr then find the distance?&&\\
Answer Choices: (a) 53 km (b) 55 km (c) 52 km (d) 60 km (e) 50 km&&\\
A: The distance that the person traveled would have been 20 km/hr * 2.5 hrs = 50 km. So the answer is (e).&&\\
\\
Q: How many keystrokes are needed to type the numbers from 1 to 500?&&\\
Answer Choices: (a) 1156 (b) 1392 (c) 1480 (d) 1562 (e) 1788&&\\
A: There are 9 one-digit numbers from 1 to 9. There are 90 two-digit numbers from 10 to 99. There are 401 three-digit numbers from 100 to 500. 9 + 90(2) + 401(3) = 1392. So the answer is (b).&&\\
        \hline
    \end{tabularx}
    \caption{CoT-Prompt for AQuA dataset(used in \citet{wei2022chain})}
    \label{cot:aqua}
\end{table}
\begin{table}[ht]
    \centering
    \begin{tabularx}{0.9\textwidth}{X c r}
        \hline
         CoT-Prompt for Bamboogle & & \\
        \hline
      Q: Who lived longer, Theodor Haecker or Harry Vaughan Watkins?&&\\
A: Theodor Haecker was 65 years old when he died. Harry Vaughan Watkins was 69 years old when he died. So the answer is Harry Vaughan Watkins.&&\\
\\
Q: Why did the founder of Versus die?&&\\
A: The founder of Versus was Gianni Versace. Gianni Versace was shot and killed on the steps of his Miami Beach mansion on July 15, 1997. So the answer is Shot.&&\\
\\
Q: Who is the grandchild of Dambar Shah?&&\\
A: Dambar Shah (? - 1645) was the king of the Gorkha Kingdom. He was the father of Krishna Shah. Krishna Shah (? - 1661) was the king of the Gorkha Kingdom. He was the father of Rudra Shah. So the answer is Rudra Shah.&&\\
\\
Q: Are both director of film FAQ: Frequently Asked Questions and director of film The Big Money from the same country?&&\\
A: The director of the film FAQ: Frequently Asked Questions is Carlos Atanes. The director of the film The Big Money is John Paddy Carstairs. The nationality of Carlos Atanes is Spanish. The nationality of John Paddy Carstairs is British. So the answer is No.
&&\\
        \hline
    \end{tabularx}
    \caption{CoT-Prompt for Bamboogle dataset(used in \citet{wang2023towards})}
    \label{cot:bamboogle}
\end{table}

\begin{table}[h]
    \centering
    \begin{tabularx}{0.9\textwidth}{X c r}
        \hline
         CoT-Prompt for Sports & & \\
        \hline
Q: Is the following sentence plausible? ``Kyle Palmieri was called for slashing ''. & & \\
A: Kyle Palmieri is a hockey player. Being called for slashing is part of hockey. So the answer is yes.& & \\
\\
Q: Is the following sentence plausible? ``Joao Moutinho caught the screen pass in the NFC championship''. & & \\
A: Joao Moutinho is a soccer player. The NFC championship is part of American football, not soccer. So the answer is no.& & \\
\\
Q: Is the following sentence plausible? ``Carson Wentz set the pick and roll''.& & \\ 
A: Carson Wentz is an American football player. Pick and roll is part of basketball, not football. So the answer is no.& & \\
\\
Q: Is the following sentence plausible? ``Malcolm Brogdon banked the shot in''.& & \\ 
A: Malcolm Brogdon is a basketball player. Banking the shot in is part of basketball. So the answer is yes.& & \\
        \hline
    \end{tabularx}
    \caption{CoT-Prompt for Sports dataset(used in \citet{wei2022chain})}
    \label{cot:sports}
\end{table}
\begin{table}[ht]
    \centering
    \begin{tabularx}{0.9\textwidth}{X c r}
        \hline
         CoT-Prompt for Date & & \\
        \hline
Q: 2015 is coming in 36 hours. What is the date one week from today in MM/DD/YYYY? & & \\
A: If 2015 is coming in 36 hours, then it is coming in 2 days. 2 days before 01/01/2015 is 12/30/2014, so today is 12/30/2014. So one week from today will be 01/05/2015. So the answer is 01/05/2015. & & \\
\\
Q: The first day of 2019 is a Tuesday, and today is the first Monday of 2019. What is the date today in MM/DD/YYYY? & & \\
A: If the first day of 2019 was Tuesday, then 01/01/2019 was a Tuesday. Today is the first monday, would be six days later. So today is 01/07/2019. So the answer is 01/07/2019. & & \\
\\
Q: The concert was scheduled to be on 06/01/1943, but was delayed by one day to today. What is the date 10 days ago in MM/DD/YYYY? & & \\
A: One day after 06/01/1943 is 06/02/1943, so today is 06/02/1943. 10 days before today is 05/23/1943. So the answer is 05/23/1943. & & \\
\\
Q: Jane was born on the last day of Feburary in 2001. Today is her 16-year-old birthday. What is the date yesterday in MM/DD/YYYY? & & \\
A: The last day of February is the 28th, so Jane was born on 02/28/2001. Today is her 16-year old birthday, so today is 02/28/2017. So yesterday was 02/27/2017. So the answer is 02/27/2017.& & \\
        \hline
    \end{tabularx}
    \caption{CoT-Prompt for Date dataset(used in \citet{wei2022chain})}
    \label{cot:date}
\end{table}
\begin{table}[ht]
    \centering
    \begin{tabularx}{0.9\textwidth}{X c r}
        \hline
         CoT-Prompt for Coin Flip & & \\
        \hline
Q: A coin is heads up. Ka flips the coin. Sherrie flips the coin. Is the coin still heads up? & & \\
A: The coin was flipped by Ka and Sherrie. So the coin was flipped 2 times, which is an even number. The coin started heads up, so after an even number of flips, it will still be heads up. So the answer is yes.& & \\
\\
Q: A coin is heads up. Maybelle flips the coin. Shalonda does not flip the coin. Is the coin still heads up? & & \\
A: The coin was flipped by Maybelle. So the coin was flipped 1 time, which is an odd number. The coin started heads up, so after an odd number of flips, it will be tails up. So the answer is no.& & \\
\\
Q: A coin is heads up. Millicent does not flip the coin. Conception flips the coin. Is the coin still heads up? & & \\
A: The coin was flipped by Conception. So the coin was flipped 1 time, which is an odd number. The coin started heads up, so after an odd number of flips, it will be tails up. So the answer is no.& & \\
\\
Q: A coin is heads up. Ryan flips the coin. Shaunda flips the coin. Is the coin still heads up?& & \\
A: The coin was flipped by Ryan and Shaunda. So the coin was flipped 2 times, which is an even number. The coin started heads up, so after an even number of flips, it will still be heads up. So the answer is yes.
& & \\
        \hline
    \end{tabularx}
    \caption{CoT-Prompt for Coin Flip dataset(used in \citet{wei2022chain})}
    \label{cot:coin}
\end{table}
\begin{table}[ht]
    \centering
    \begin{tabularx}{0.9\textwidth}{X c r}
        \hline
         CoT-Prompt for Last Letter Concatenation & & \\
        \hline
Q: Take the last letters of each words in ``Tim Candace Cecil Misael'' and concatenate them.& & \\
A: The last letter of ``Tim'' is ``m''. The last letter of ``Candace'' is ``e''. The last letter of ``Cecil'' is ``l''. The last letter of ``Misael'' is ``l''. Concatenating them is ``mell''. So the answer is mell.& & \\
\\
Q: Take the last letters of each words in ``Alina Alessandra Amina Bianca'' and concatenate them.& & \\
A: The last letter of ``Alina'' is ``a''. The last letter of ``Alessandra'' is ``a''. The last letter of ``Amina'' is ``a''. The last letter of ``Bianca'' is ``a''. Concatenating them is ``aaaa''. So the answer is aaaa.& & \\
\\
Q: Take the last letters of each words in ``Felipe Heidi Nino Bradley'' and concatenate them.& & \\
A: The last letter of ``Felipe'' is ``e''. The last letter of ``Heidi'' is ``i''. The last letter of ``Nino'' is ``o''. The last letter of ``Bradley'' is ``y''. Concatenating them is ``eioy''. So the answer is eioy.& & \\
\\
Q: Take the last letters of each words in ``Lacey Nora Debra Ashleigh'' and concatenate them.& & \\
A: The last letter of ``Lacey'' is ``y''. The last letter of ``Nora'' is ``a''. The last letter of ``Debra'' is ``a''. The last letter of ``Ashleigh'' is ``h''. Concatenating them is ``yaah''. So the answer is yaah.& & \\
        \hline
    \end{tabularx}
    \caption{CoT-Prompt for Last Letter Concatenation dataset(used in \citet{kojima2022large})}
    \label{cot:letter}
\end{table}
\clearpage
\section{Test points}

\begin{longtable}{@{}p{2cm}p{12cm}@{}}
\caption{Test points and their associated words.}\label{tab:test_points}\\
\toprule
\textbf{Test points} & \textbf{Contained words} \\
\midrule
\endfirsthead
\multicolumn{2}{l}{\textit{Continued from previous page}}\\
\toprule
\textbf{Test points} & \textbf{Contained words} \\
\midrule
\endhead
\midrule
\multicolumn{2}{r}{\textit{Continued on next page}}\\
\endfoot
\bottomrule
\endlastfoot
time & originally, then, after, so, start, first, next, last, finally, 
      before, later, afterwards, subsequently, meanwhile, during, while, 
      when, once, as, since, because, due, hence, therefore, thus, 
      consequently, accordingly, result, resulting, resulted, initially,
      earlier, until, at the same time \\
\midrule
action & +, -, *, /, =, >, <, add, subtract, multiply, divide, average, 
        increase, decrease, equal, calculate, total, square, root, cube, prime \\
\midrule
loc\&peo & there, location, site, area, spot, venue, someone, somebody, anyone, 
          nobody, everyone, person, individual, participant, operator, handler, 
          's, his, her, their, its, he, she, they, it \\
\midrule
number & ``it was obtained by regular expression''  \\
\end{longtable}
\section{Resutls of transfer test}
\begin{figure*}[!h]
\centering
    \begin{subfigure}{0.3\linewidth}
        \includegraphics[width=\linewidth]{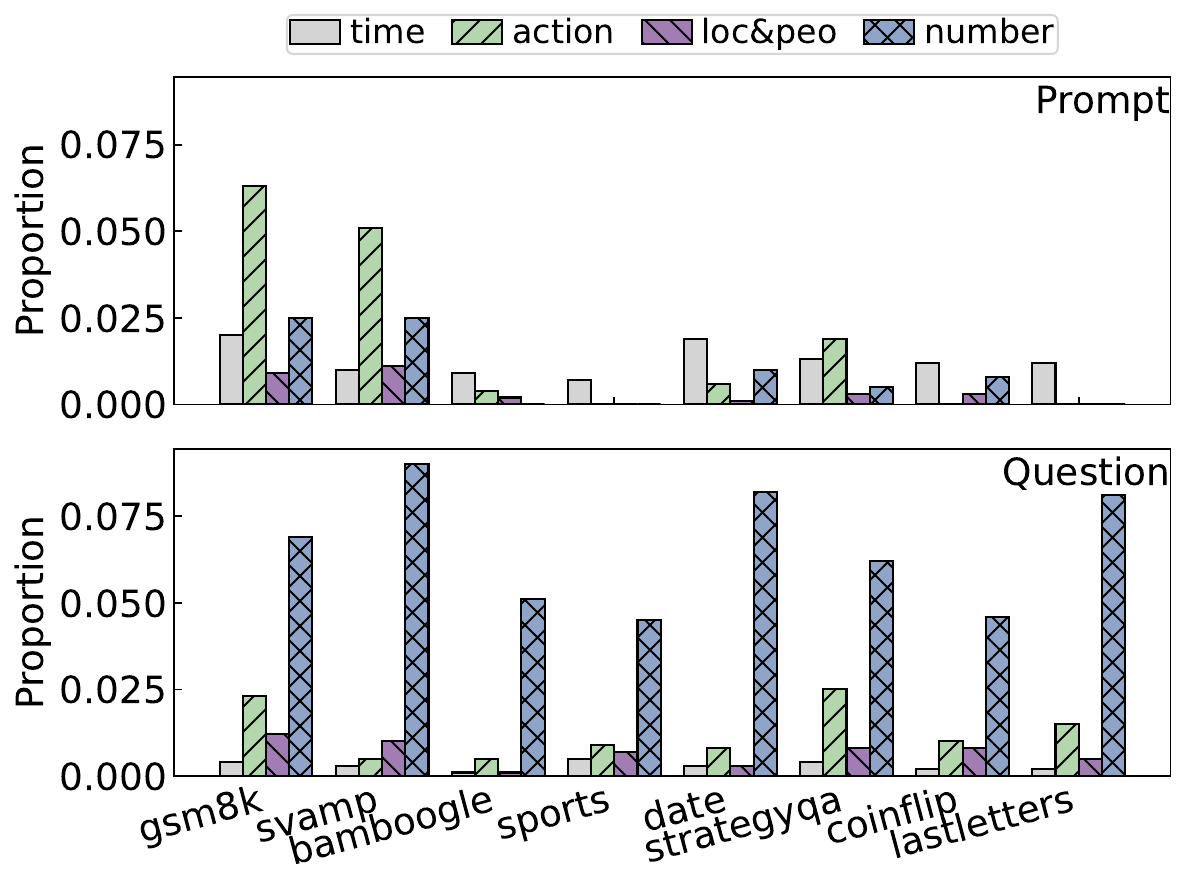}
        \caption{\textit{AQuA}}
        \label{fig:question1_a2_a}
    \end{subfigure}
    \begin{subfigure}{0.3\linewidth}
        \includegraphics[width=\linewidth]{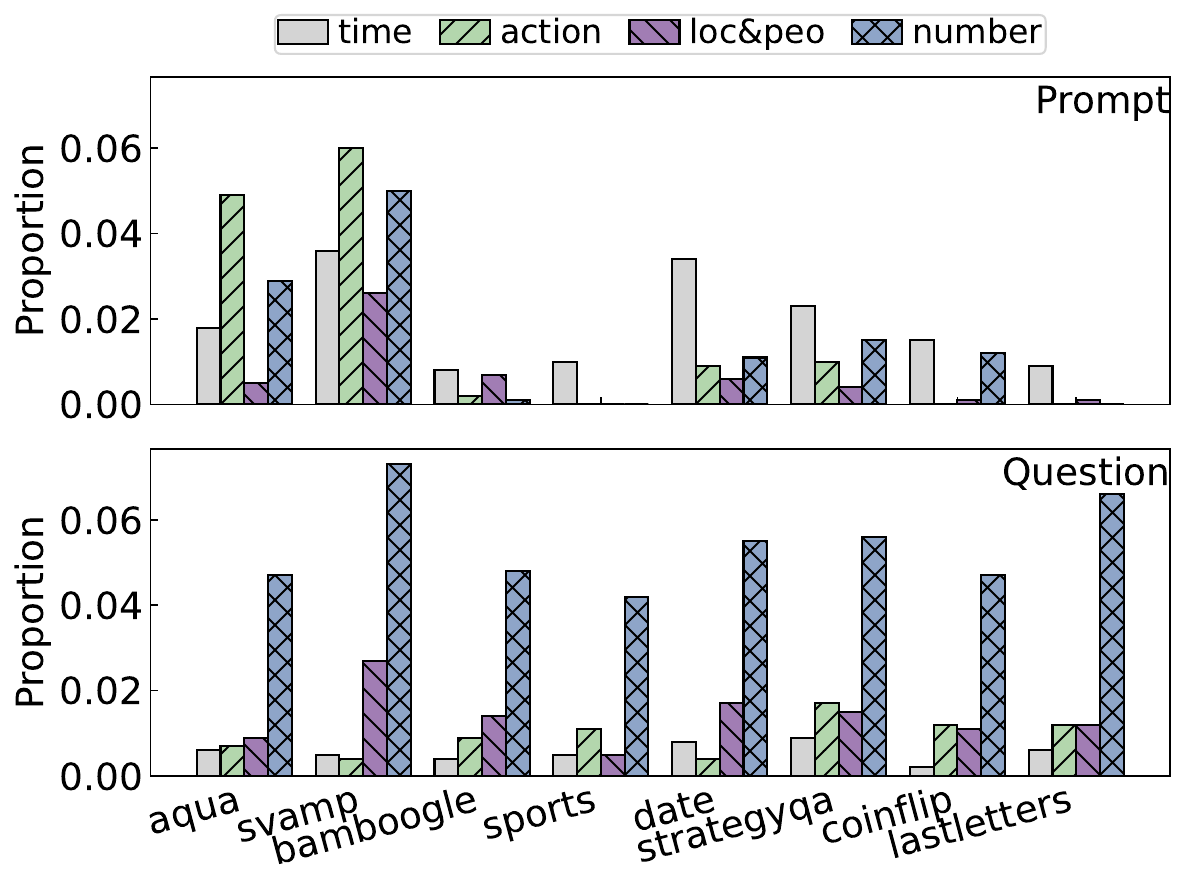}
        \caption{\textit{GSM8K}}
        \label{fig:question1_a2_b}
    \end{subfigure}
    \begin{subfigure}{0.3\linewidth}
        \includegraphics[width=\linewidth]{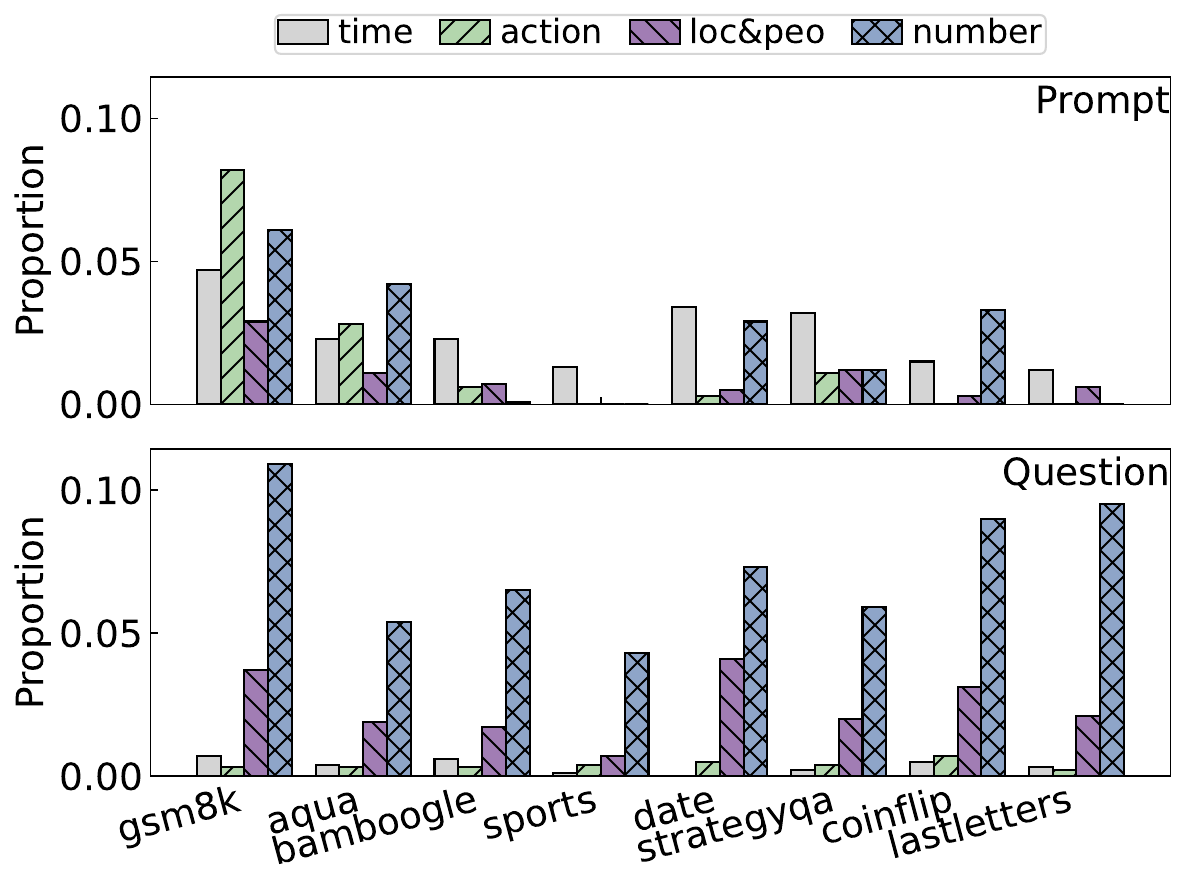}
        \caption{\textit{SVAMP}}
        \label{fig:question1_a2_c}
    \end{subfigure}

  \begin{subfigure}{0.3\linewidth}
        \includegraphics[width=\linewidth]{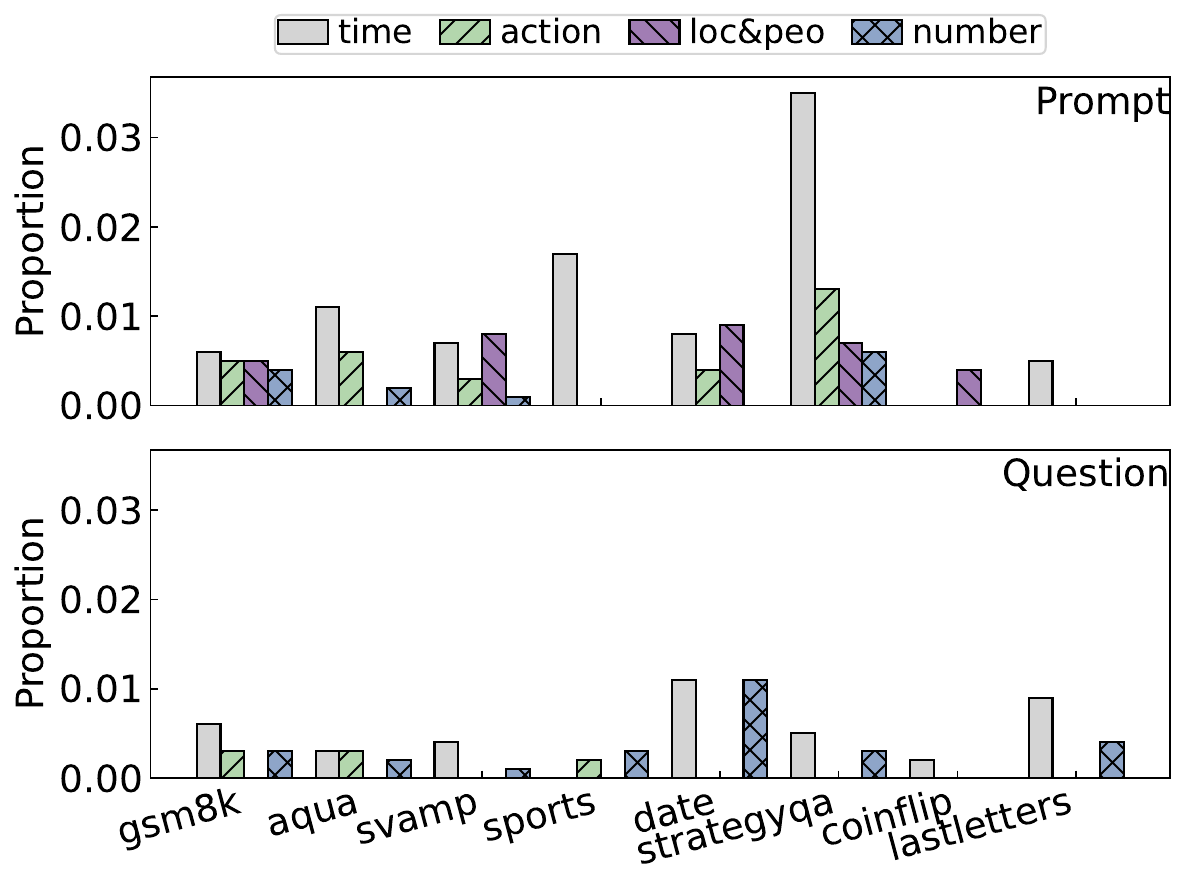}
        \caption{\textit{Bamboogle}}
        \label{fig:question1_a2_e}
    \end{subfigure}
  \begin{subfigure}{0.3\linewidth}
        \includegraphics[width=\linewidth]{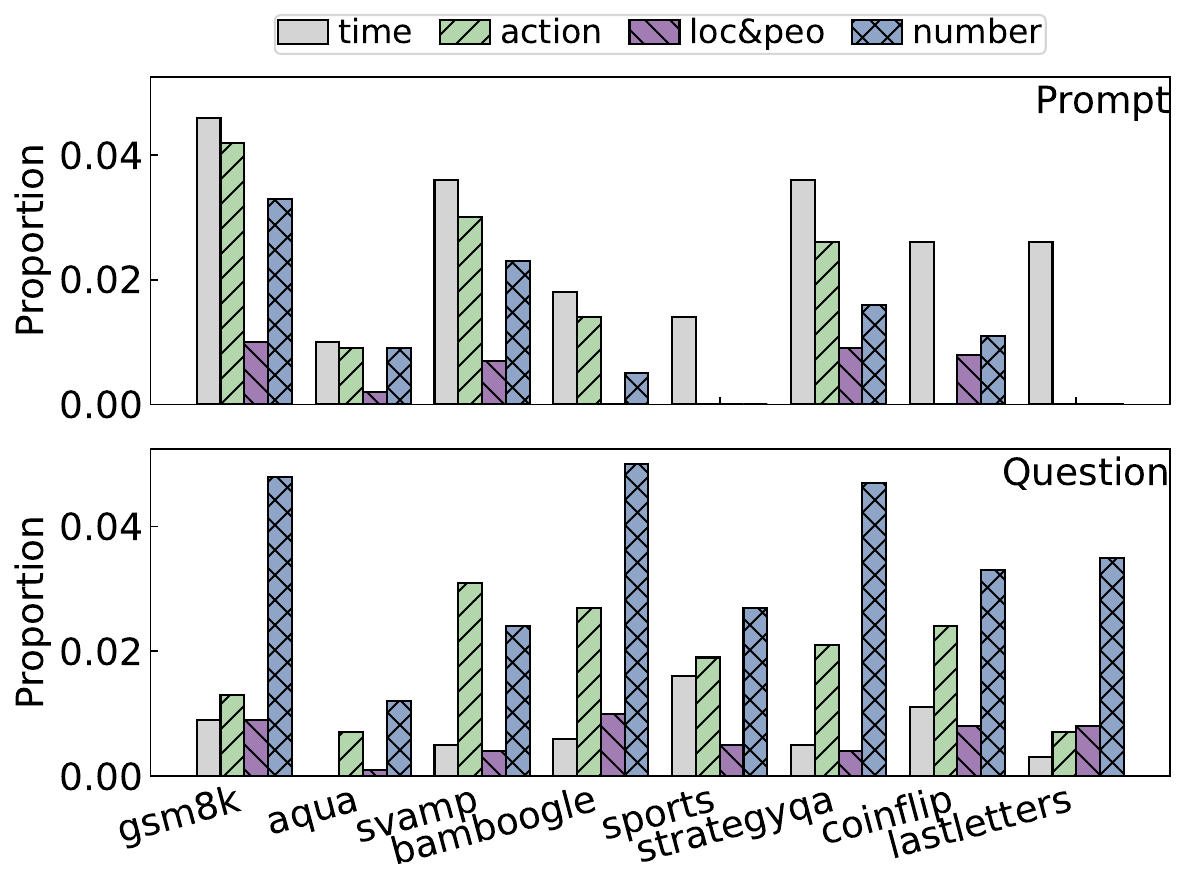}
        \caption{\textit{Date}}
        \label{fig:question1_a2_d}
    \end{subfigure}
    \begin{subfigure}{0.3\linewidth}
        \includegraphics[width=\linewidth]{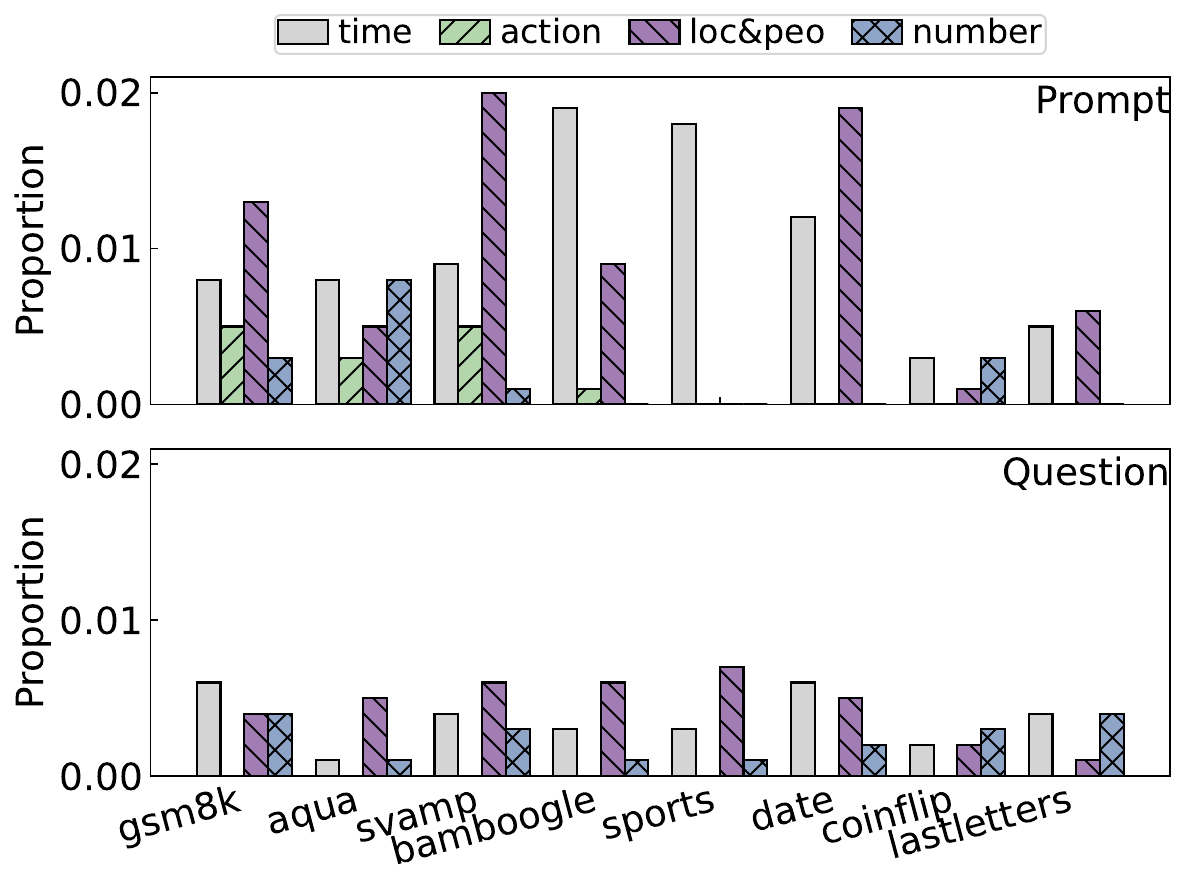}
        \caption{\textit{StrategyQA}}
    \end{subfigure}
    
     \begin{subfigure}{0.3\linewidth}
        \includegraphics[width=\linewidth]{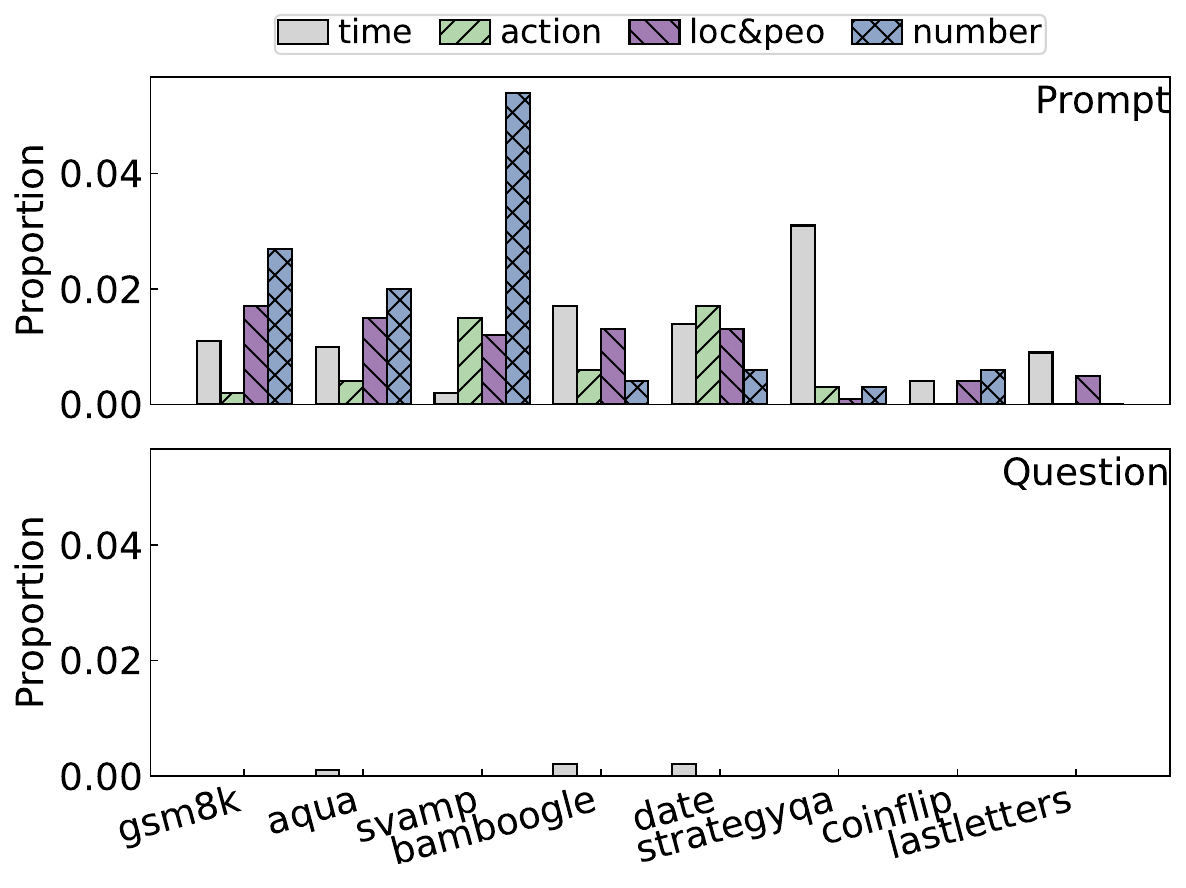}
        \caption{\textit{Sports}}
    \end{subfigure}
    \begin{subfigure}{0.3\linewidth}
        \includegraphics[width=\linewidth]{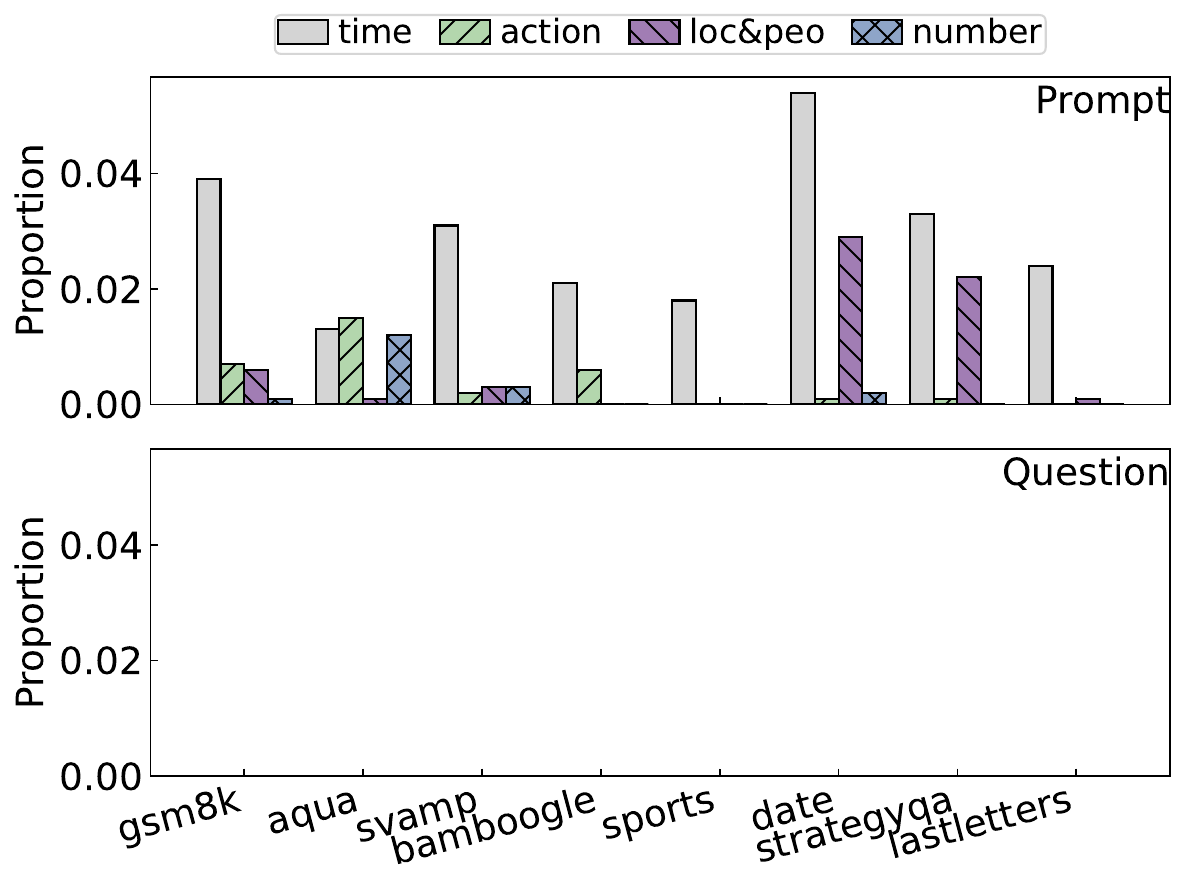}
        \caption{\textit{Coin Flip}}
    \end{subfigure}
    \begin{subfigure}{0.3\linewidth}
        \includegraphics[width=\linewidth]{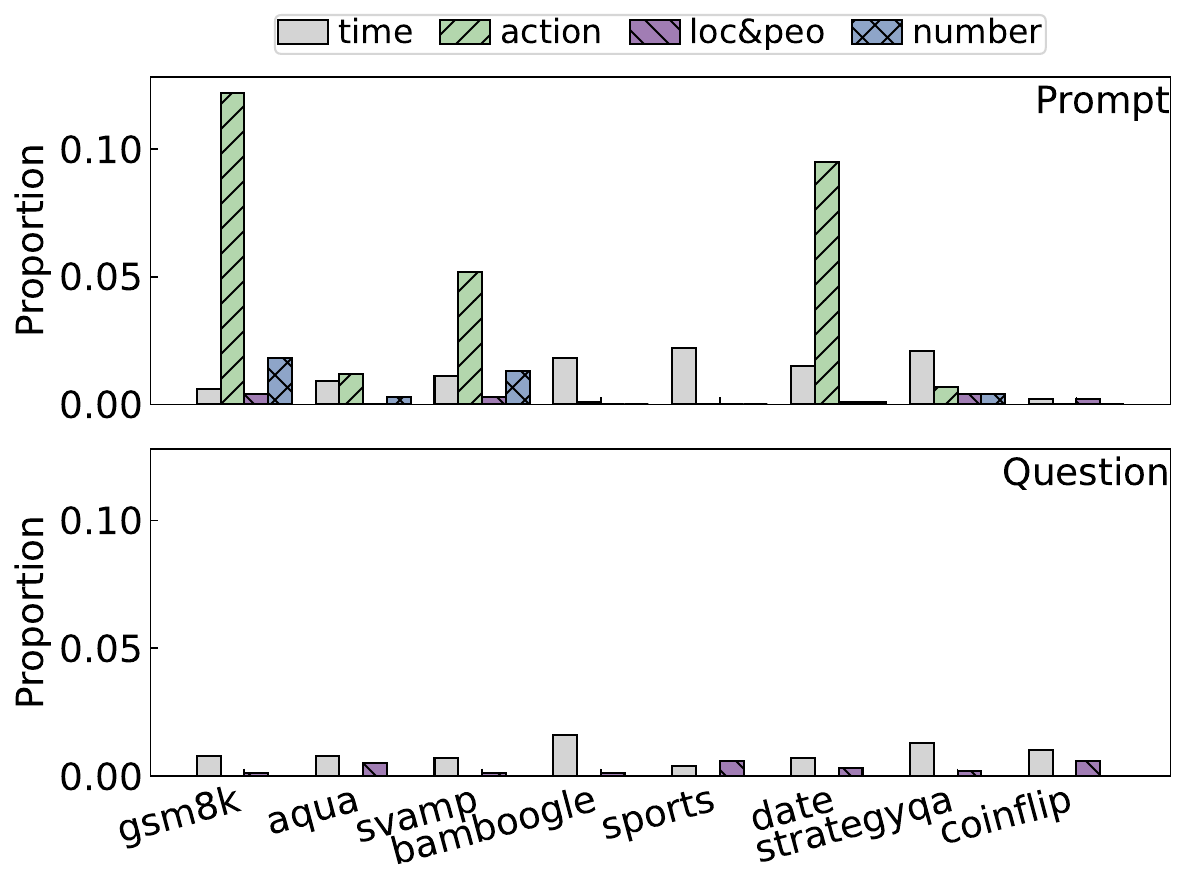}
        \caption{\textit{Last Letter}}
    \end{subfigure}
\caption{Results of \textit{transfer test} for different tasks under other prompts, results of Gemma2-2b is reported.}
\label{fig:transfer_test_2b}
\end{figure*}
\begin{figure*}[!h]
\centering
    \begin{subfigure}{0.3\linewidth}
        \includegraphics[width=\linewidth]{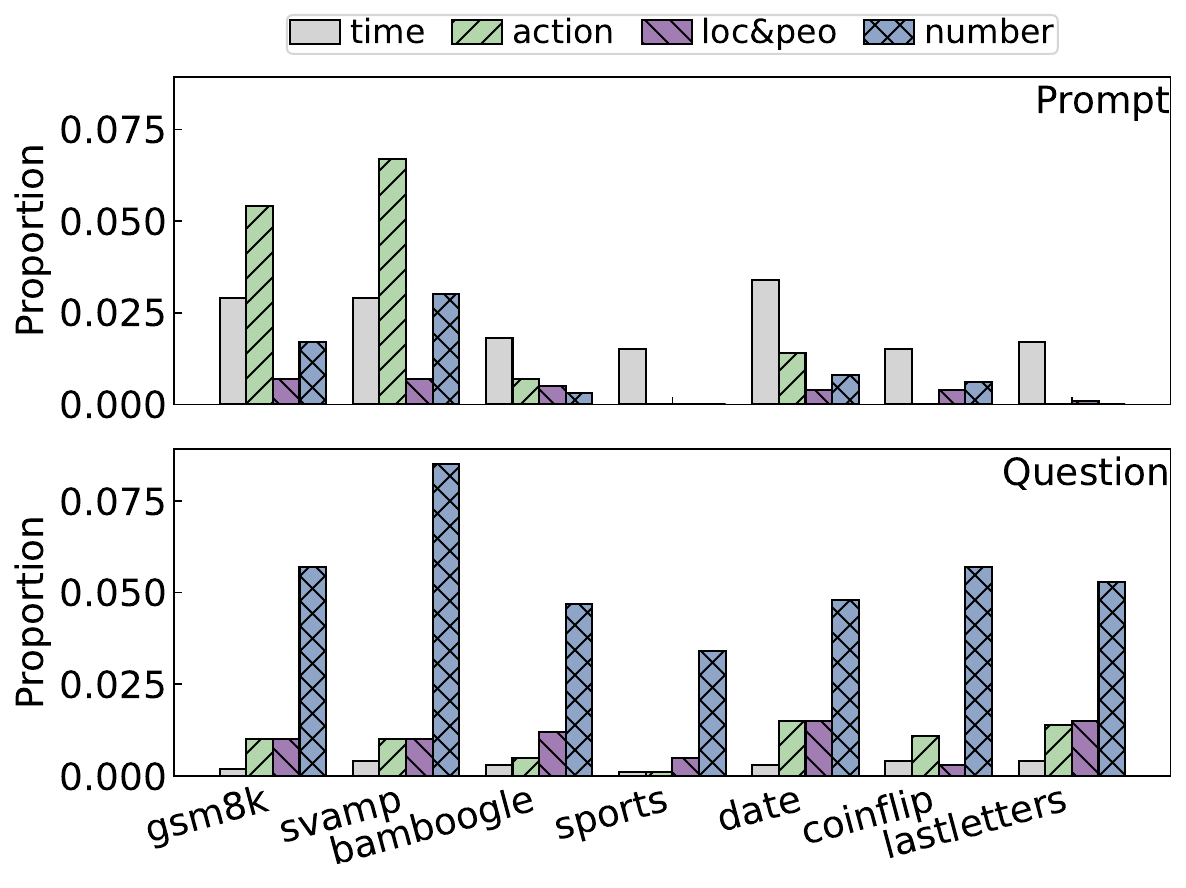}
        \caption{\textit{AQuA}}
        \label{fig:question1_a2_a}
    \end{subfigure}
    \begin{subfigure}{0.3\linewidth}
        \includegraphics[width=\linewidth]{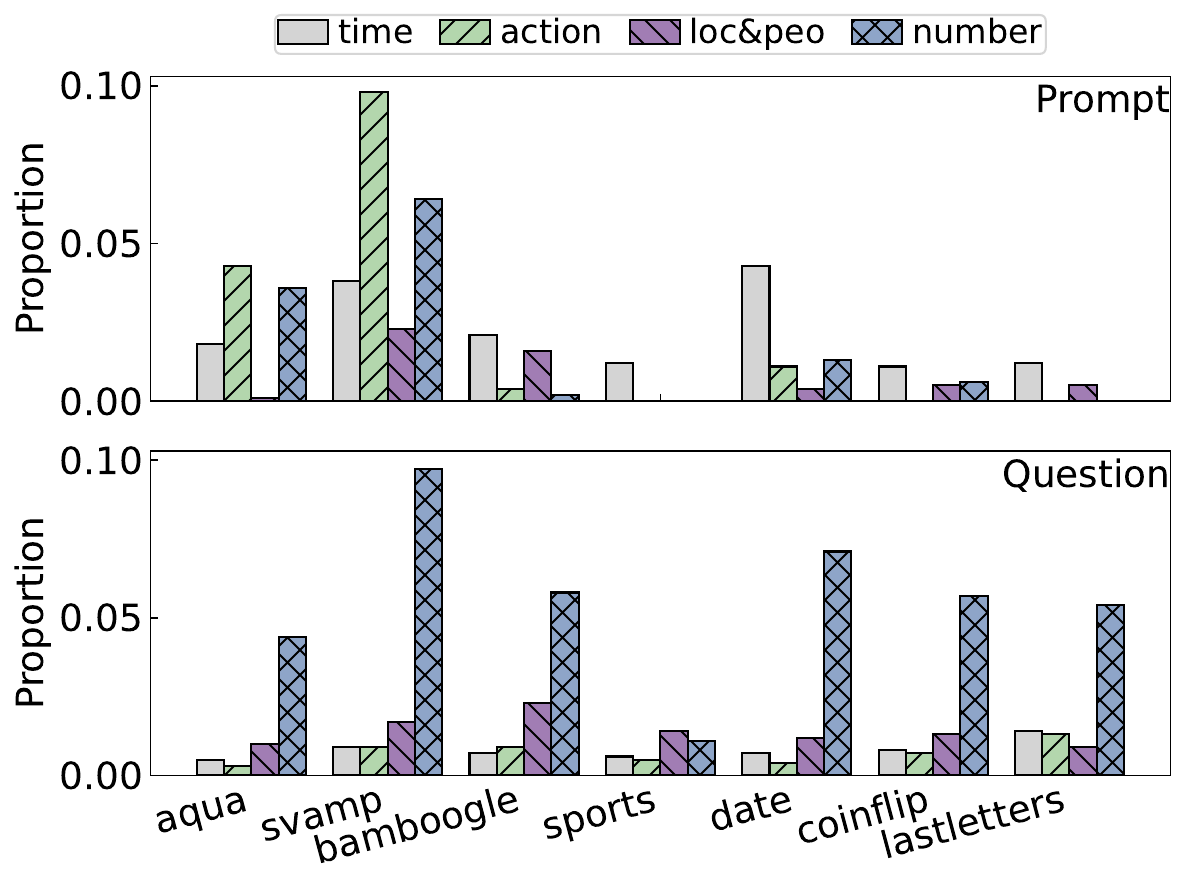}
        \caption{\textit{GSM8K}}
        \label{fig:question1_a2_b}
    \end{subfigure}
    \begin{subfigure}{0.3\linewidth}
        \includegraphics[width=\linewidth]{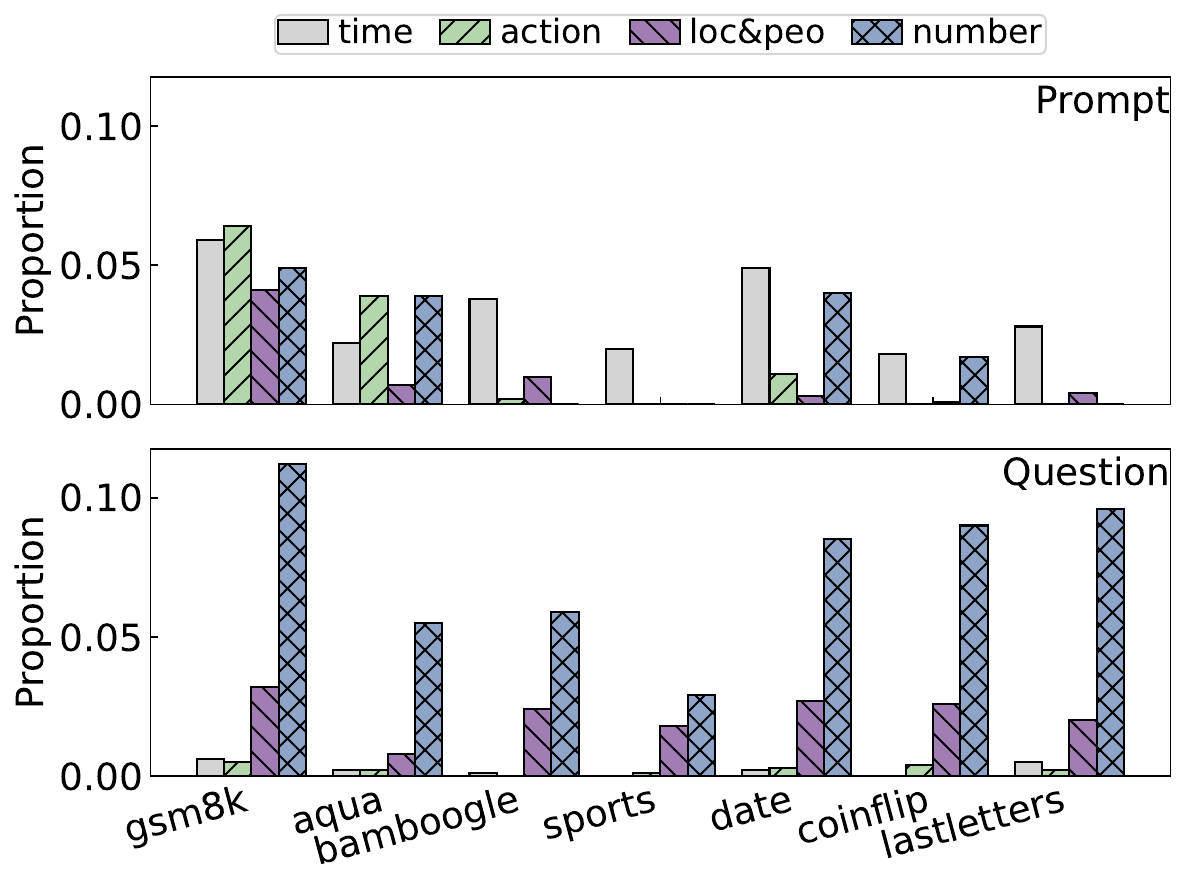}
        \caption{\textit{SVAMP}}
        \label{fig:question1_a2_c}
    \end{subfigure}

  \begin{subfigure}{0.3\linewidth}
        \includegraphics[width=\linewidth]{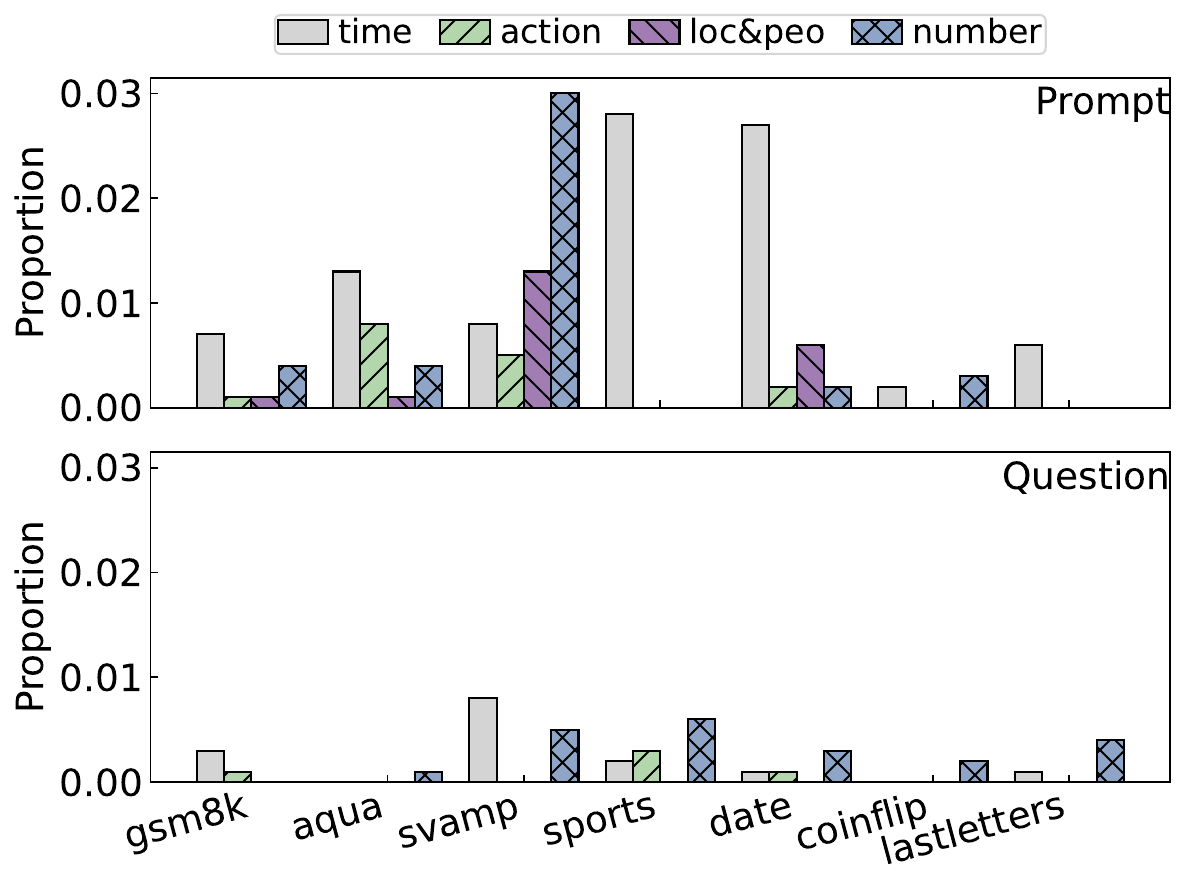}
        \caption{\textit{Bamboogle}}
        \label{fig:question1_a2_e}
    \end{subfigure}
  \begin{subfigure}{0.3\linewidth}
        \includegraphics[width=\linewidth]{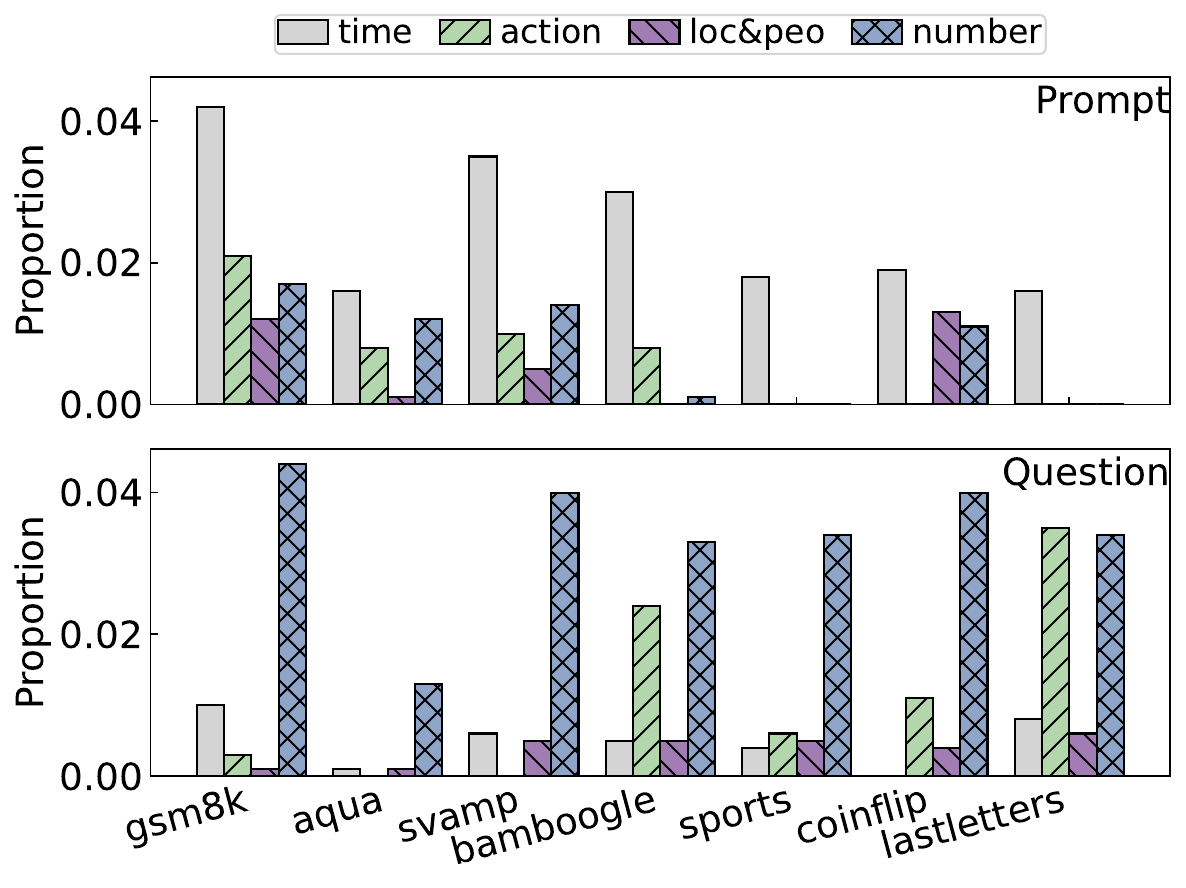}
        \caption{\textit{Date}}
        \label{fig:question1_a2_d}
    \end{subfigure}
    \begin{subfigure}{0.3\linewidth}
        \includegraphics[width=\linewidth]{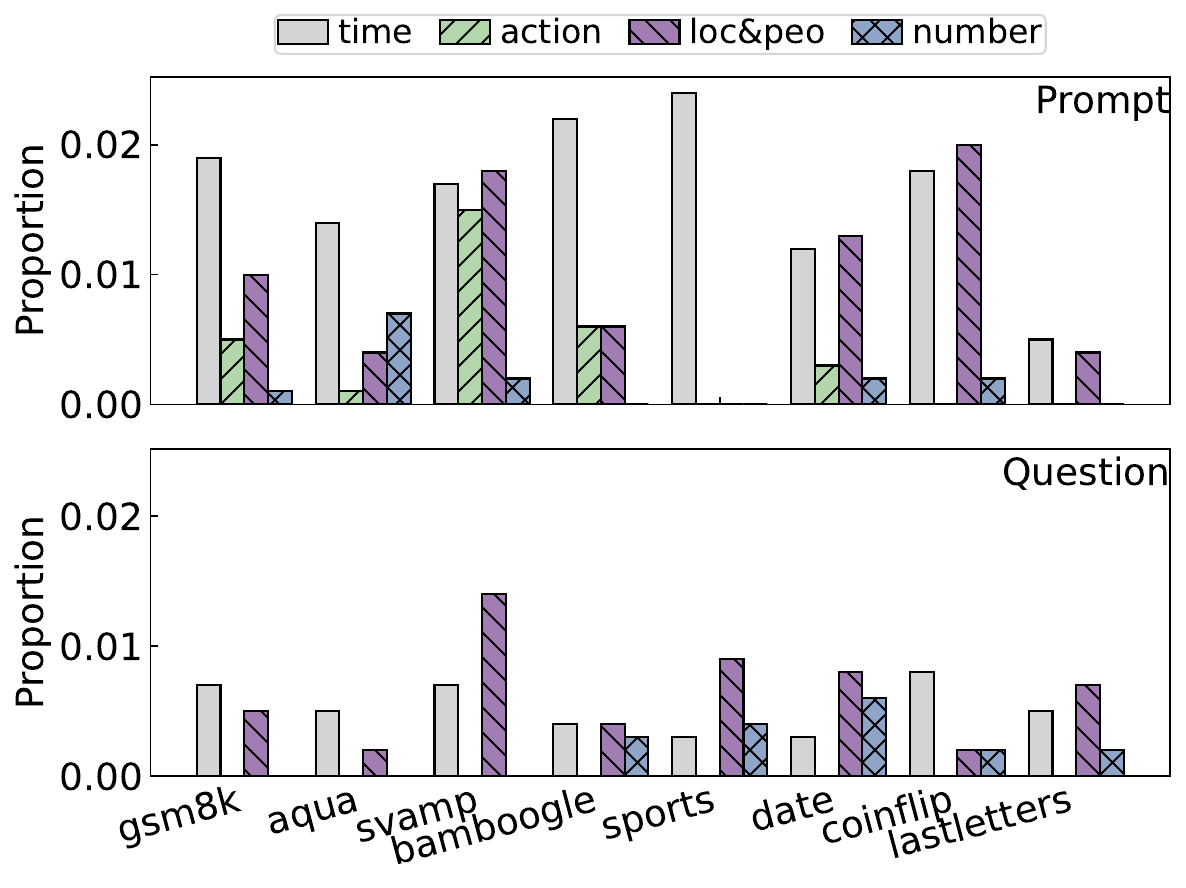}
        \caption{\textit{StrategyQA}}
    \end{subfigure}
    
     \begin{subfigure}{0.3\linewidth}
        \includegraphics[width=\linewidth]{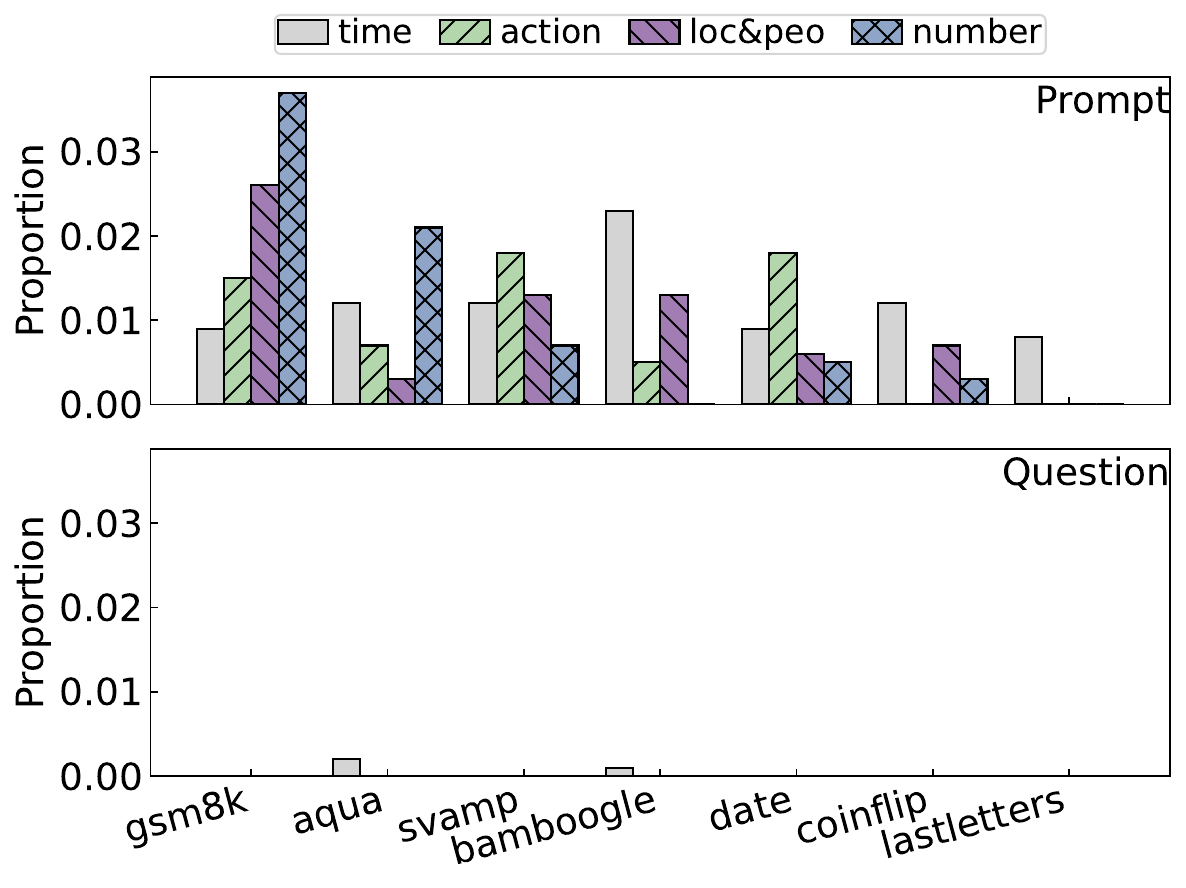}
        \caption{\textit{Sports}}
    \end{subfigure}
    \begin{subfigure}{0.3\linewidth}
        \includegraphics[width=\linewidth]{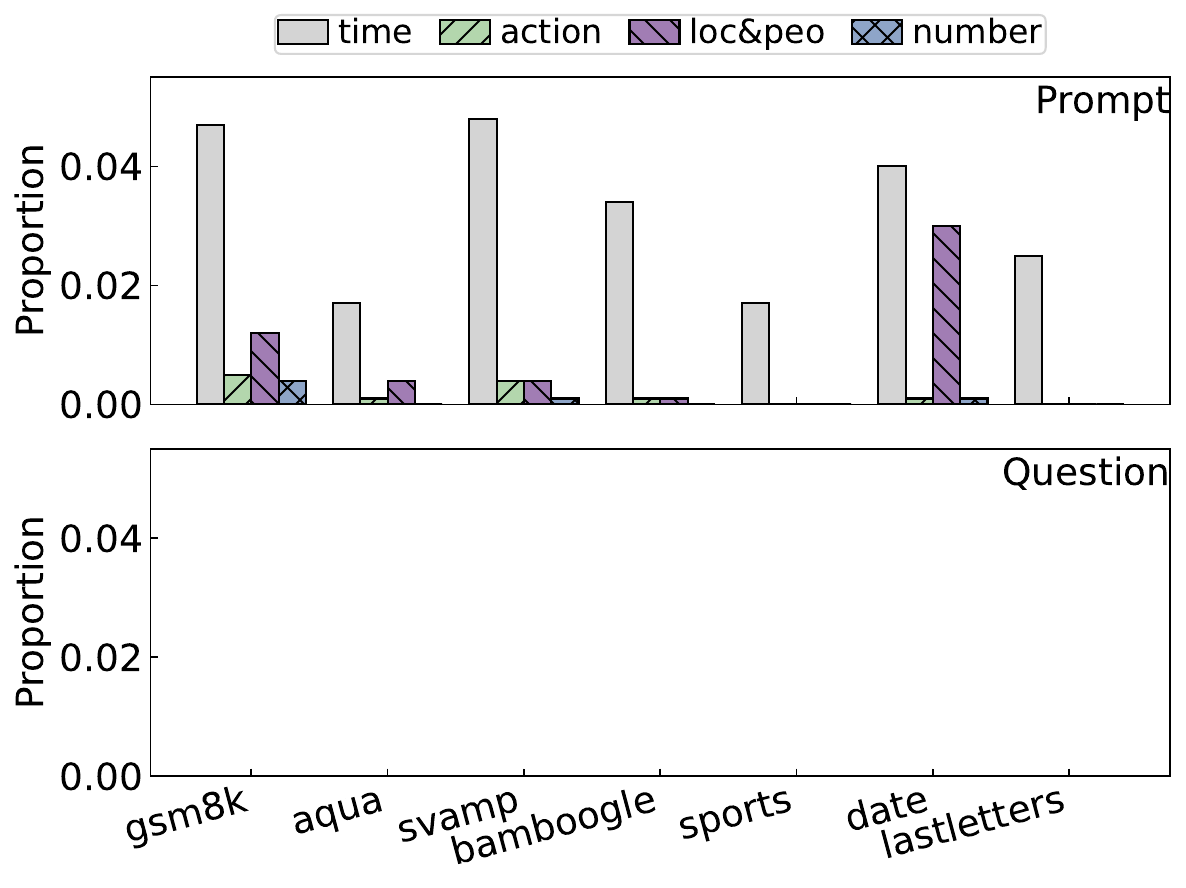}
        \caption{\textit{Coin Flip}}
    \end{subfigure}
    \begin{subfigure}{0.3\linewidth}
        \includegraphics[width=\linewidth]{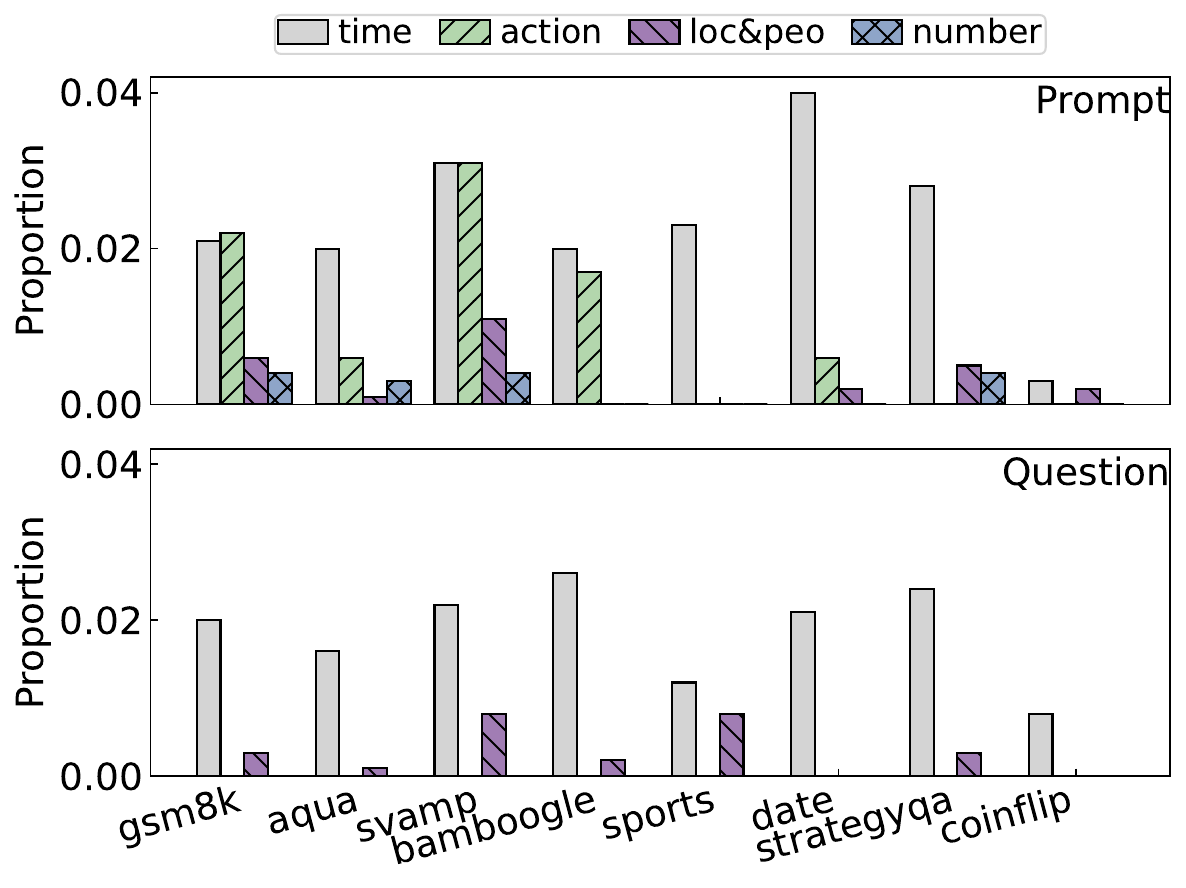}
        \caption{\textit{Last Letter}}
    \end{subfigure}
\caption{Results of \textit{transfer test} for different tasks under other prompts, results of Gemma2-9b is reported.}
\label{fig:transfer_test_9b}
\end{figure*}
\begin{figure*}[!h]
\centering
    \begin{subfigure}{0.3\linewidth}
        \includegraphics[width=\linewidth]{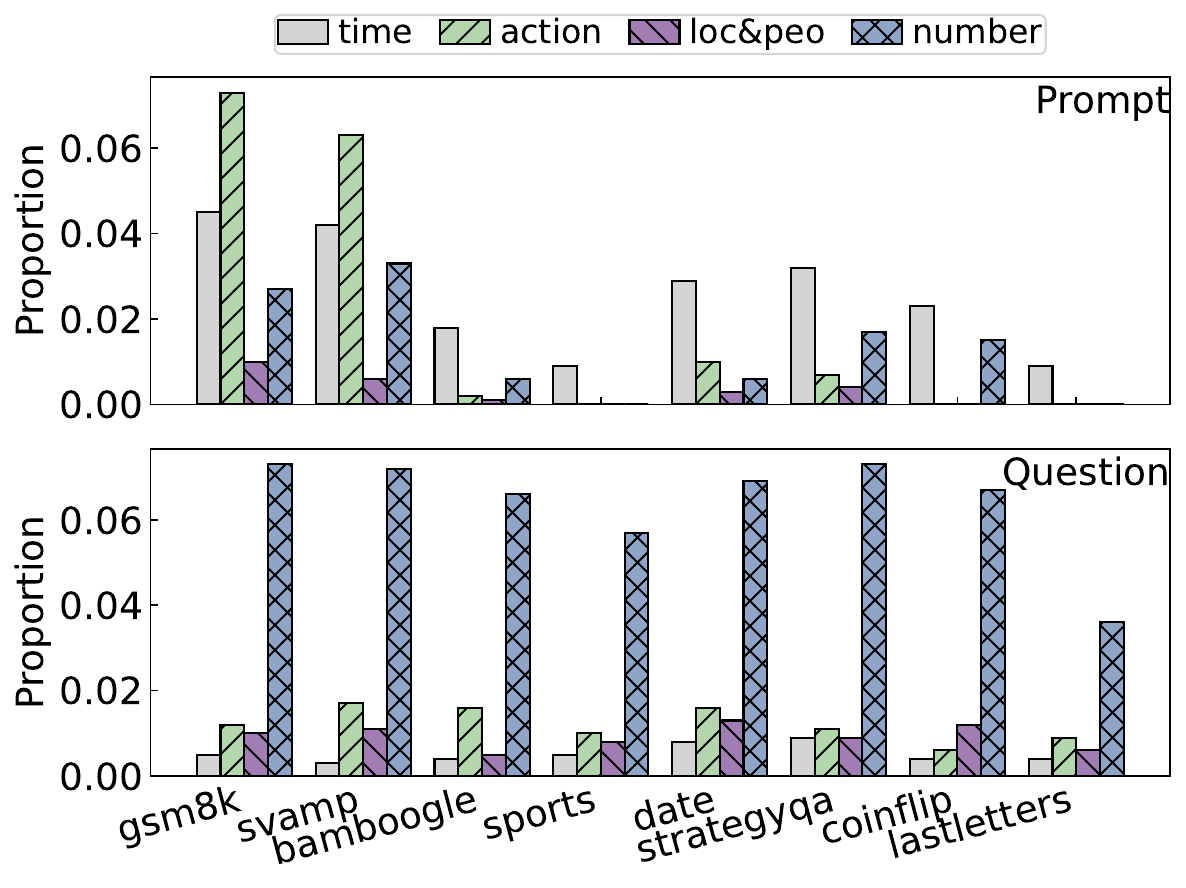}
        \caption{\textit{AQuA}}
        \label{fig:question1_a2_a}
    \end{subfigure}
    \begin{subfigure}{0.3\linewidth}
        \includegraphics[width=\linewidth]{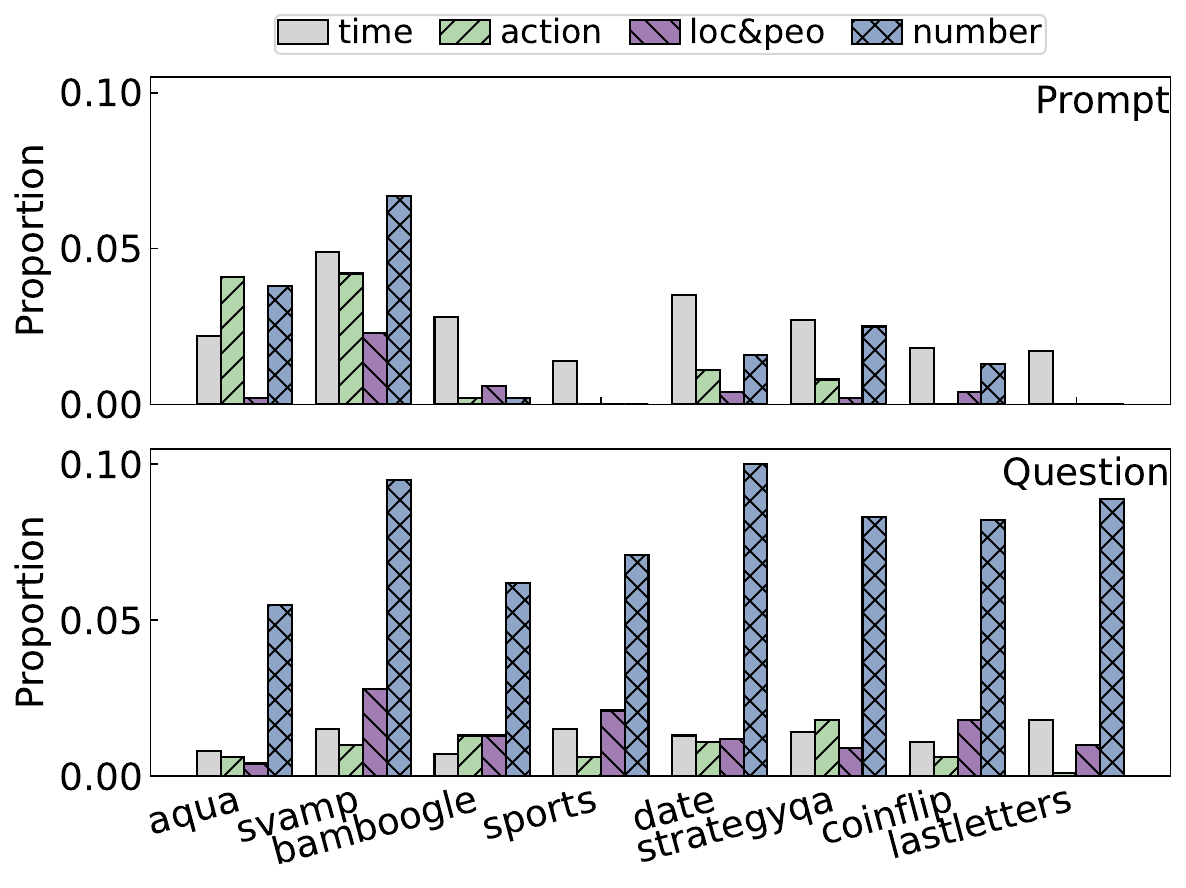}
        \caption{\textit{GSM8K}}
        \label{fig:question1_a2_b}
    \end{subfigure}
    \begin{subfigure}{0.3\linewidth}
        \includegraphics[width=\linewidth]{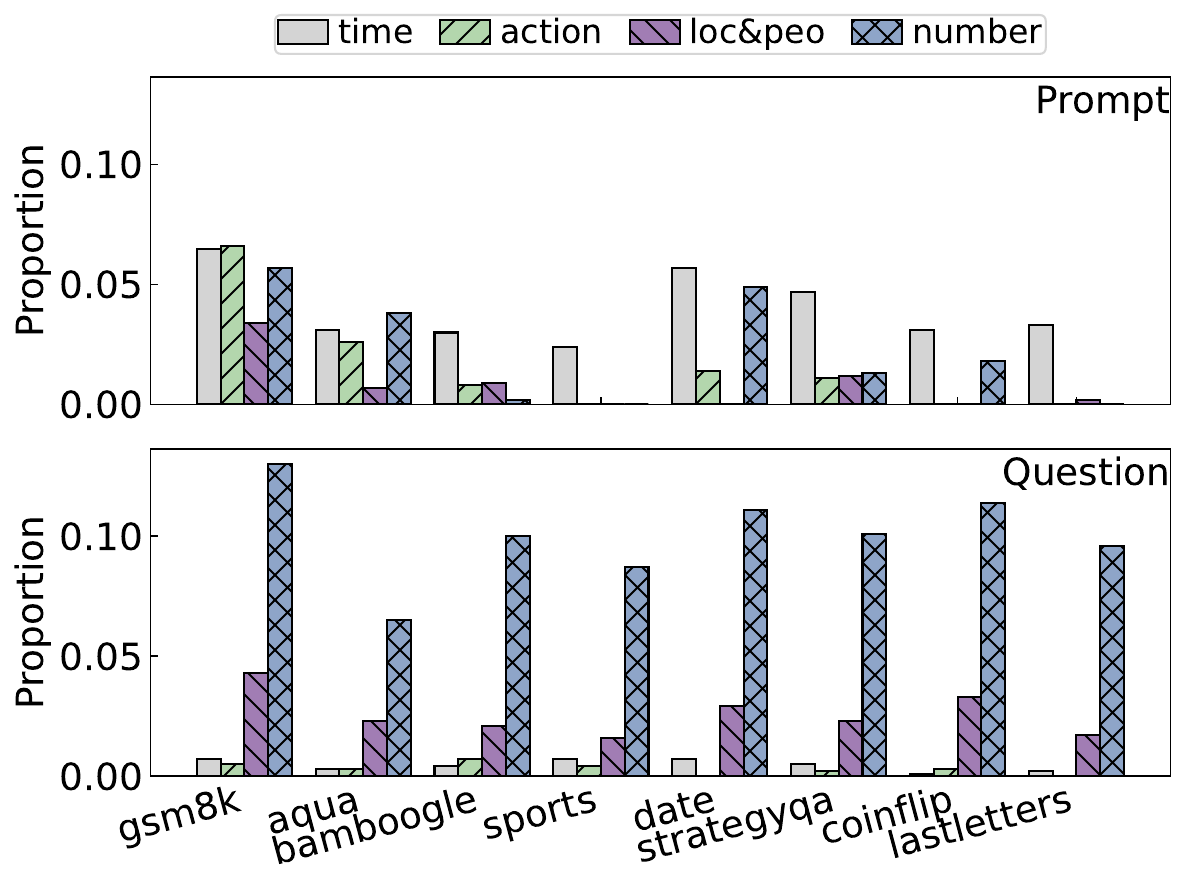}
        \caption{\textit{SVAMP}}
        \label{fig:question1_a2_c}
    \end{subfigure}

  \begin{subfigure}{0.3\linewidth}
        \includegraphics[width=\linewidth]{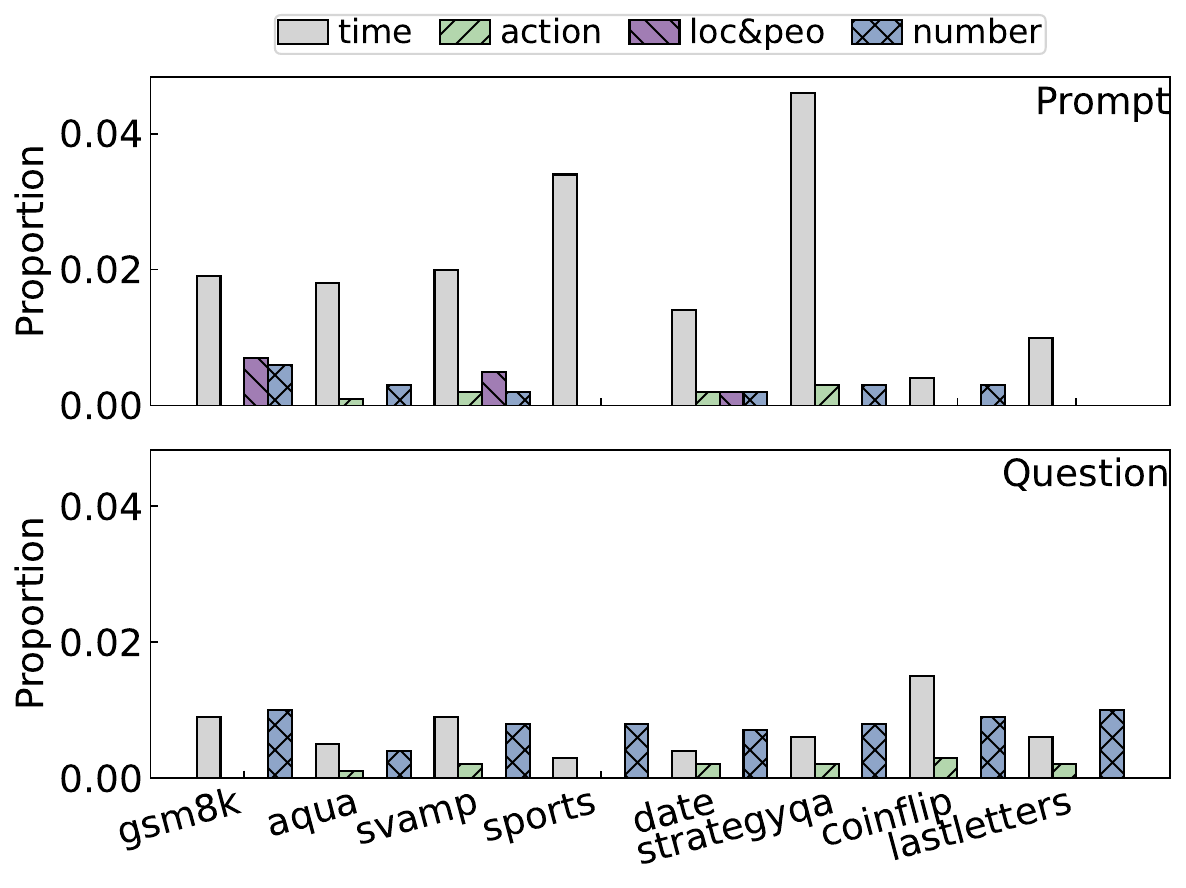}
        \caption{\textit{Bamboogle}}
        \label{fig:question1_a2_e}
    \end{subfigure}
  \begin{subfigure}{0.3\linewidth}
        \includegraphics[width=\linewidth]{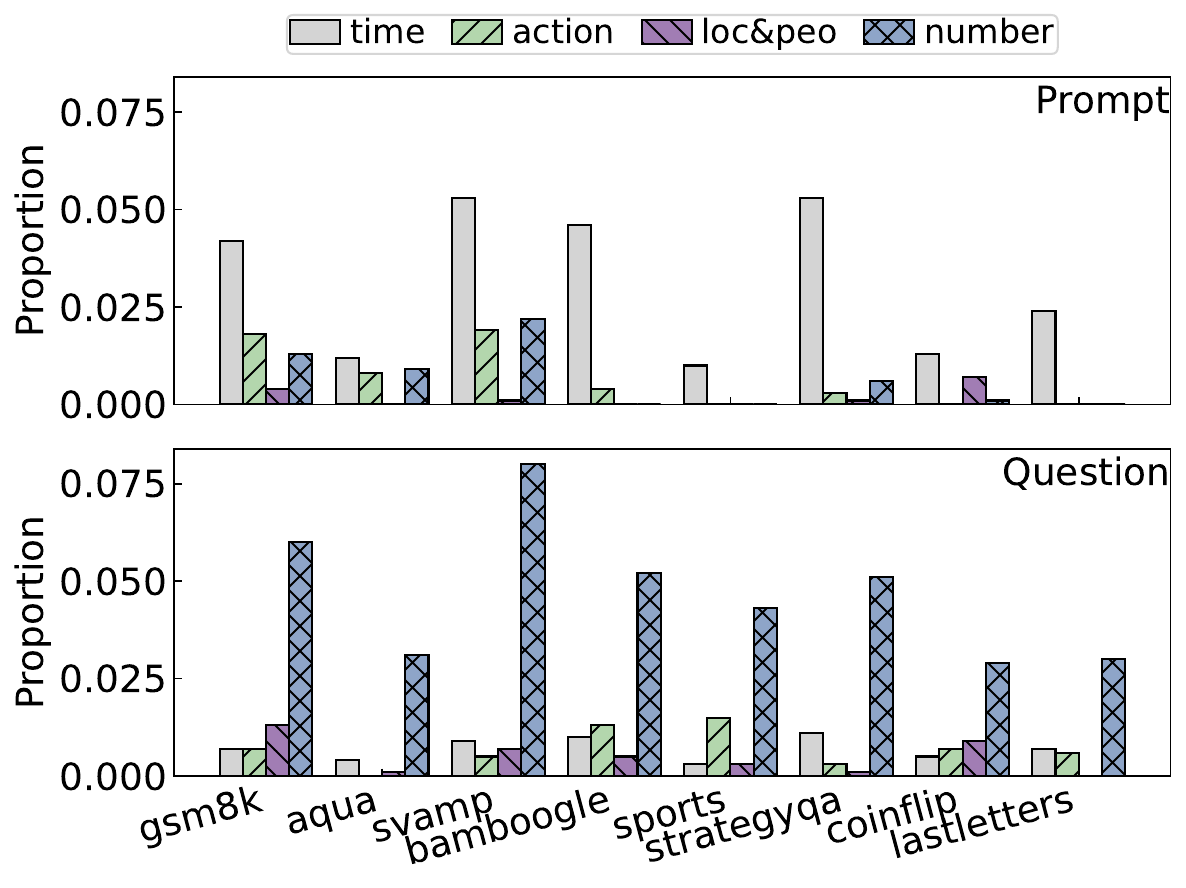}
        \caption{\textit{Date}}
        \label{fig:question1_a2_d}
    \end{subfigure}
    \begin{subfigure}{0.3\linewidth}
        \includegraphics[width=\linewidth]{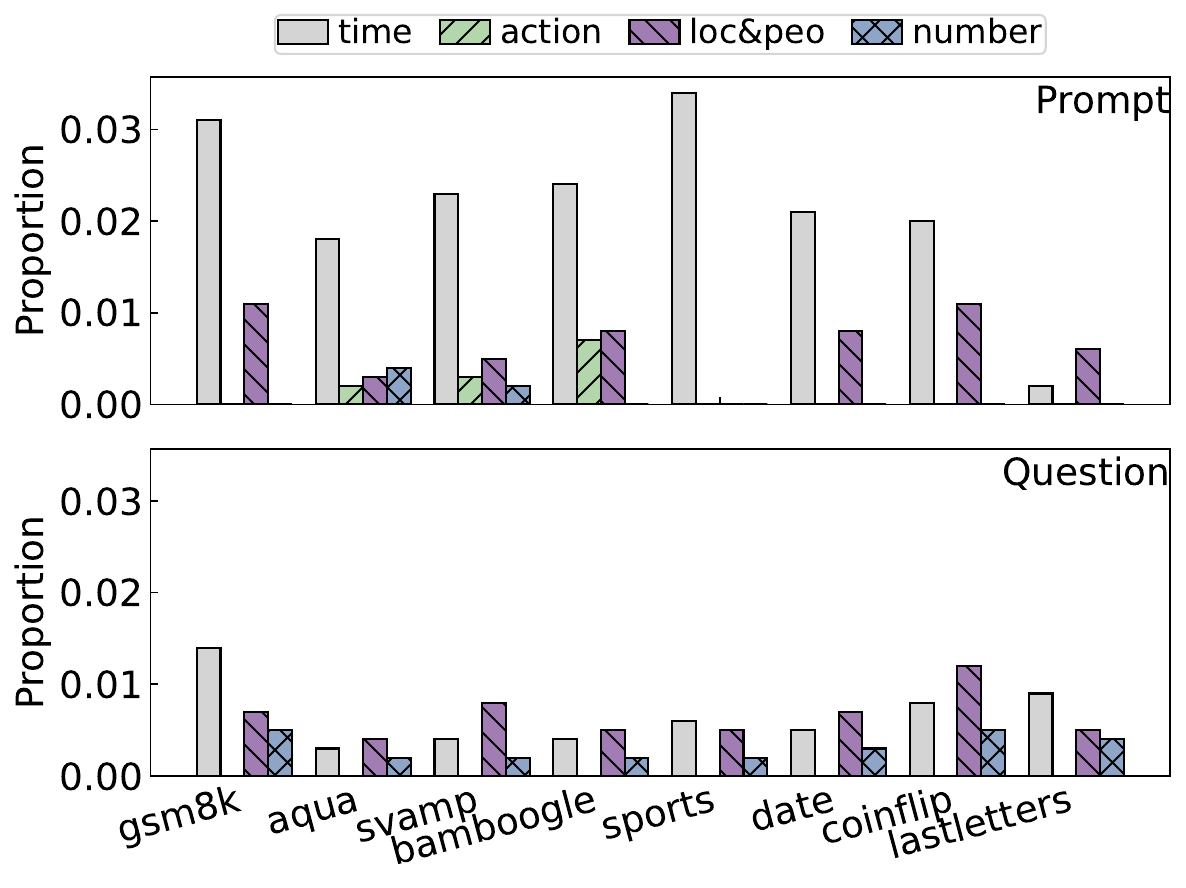}
        \caption{\textit{StrategyQA}}
    \end{subfigure}
    
     \begin{subfigure}{0.3\linewidth}
        \includegraphics[width=\linewidth]{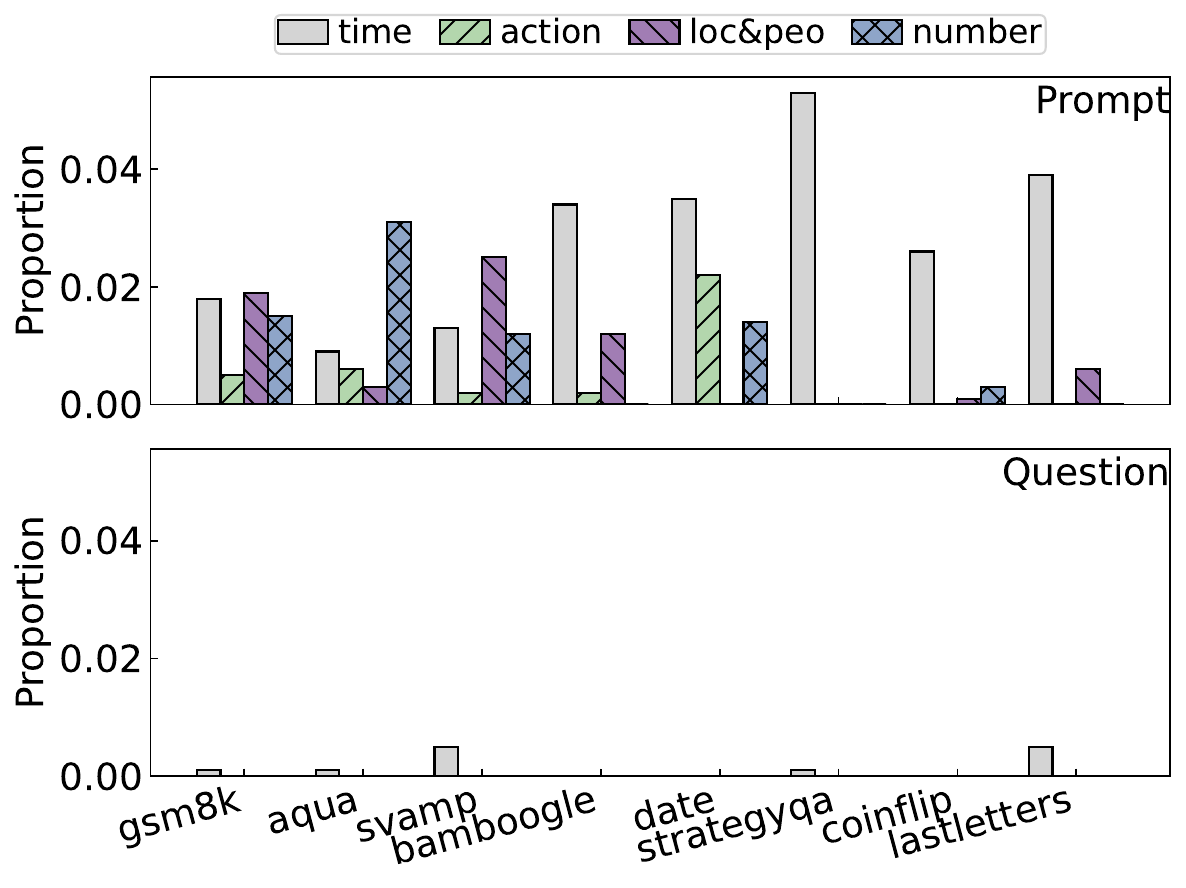}
        \caption{\textit{Sports}}
    \end{subfigure}
    \begin{subfigure}{0.3\linewidth}
        \includegraphics[width=\linewidth]{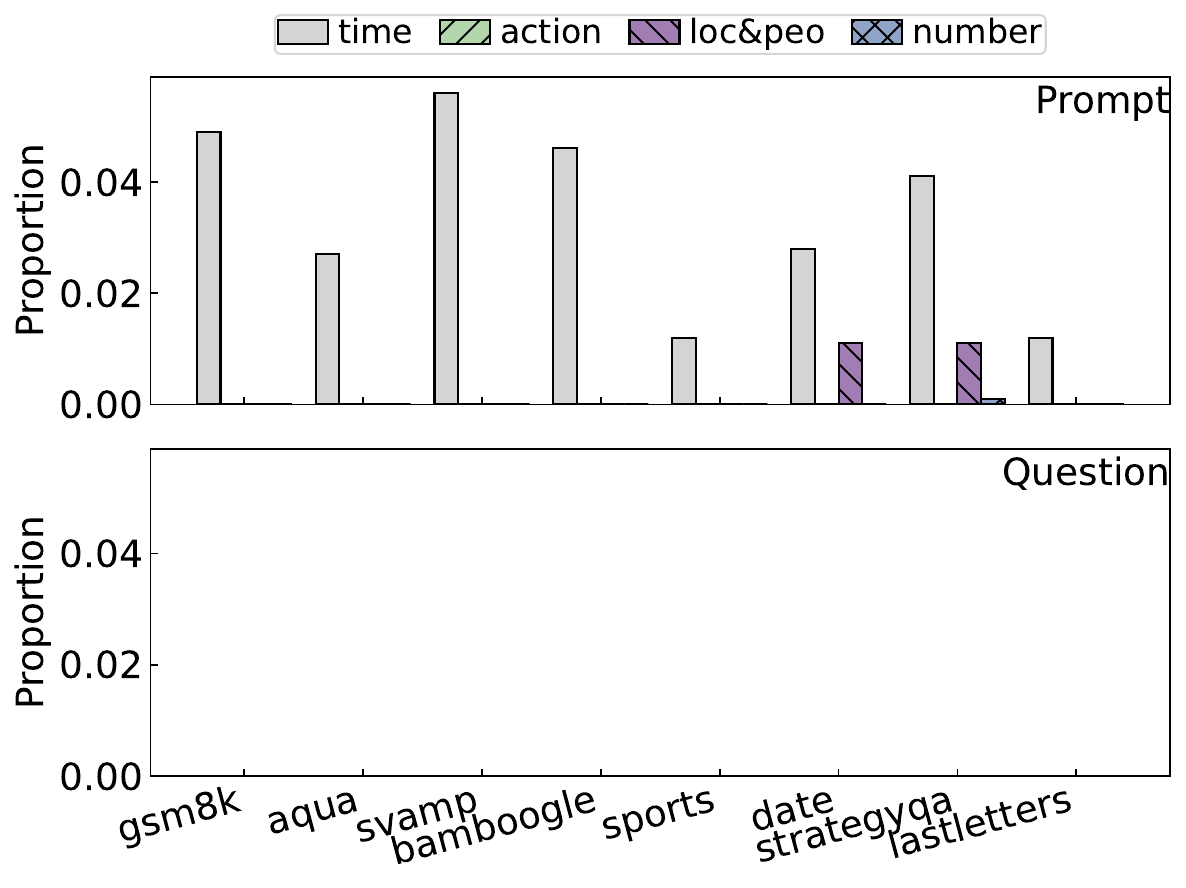}
        \caption{\textit{Coin Flip}}
    \end{subfigure}
    \begin{subfigure}{0.3\linewidth}
        \includegraphics[width=\linewidth]{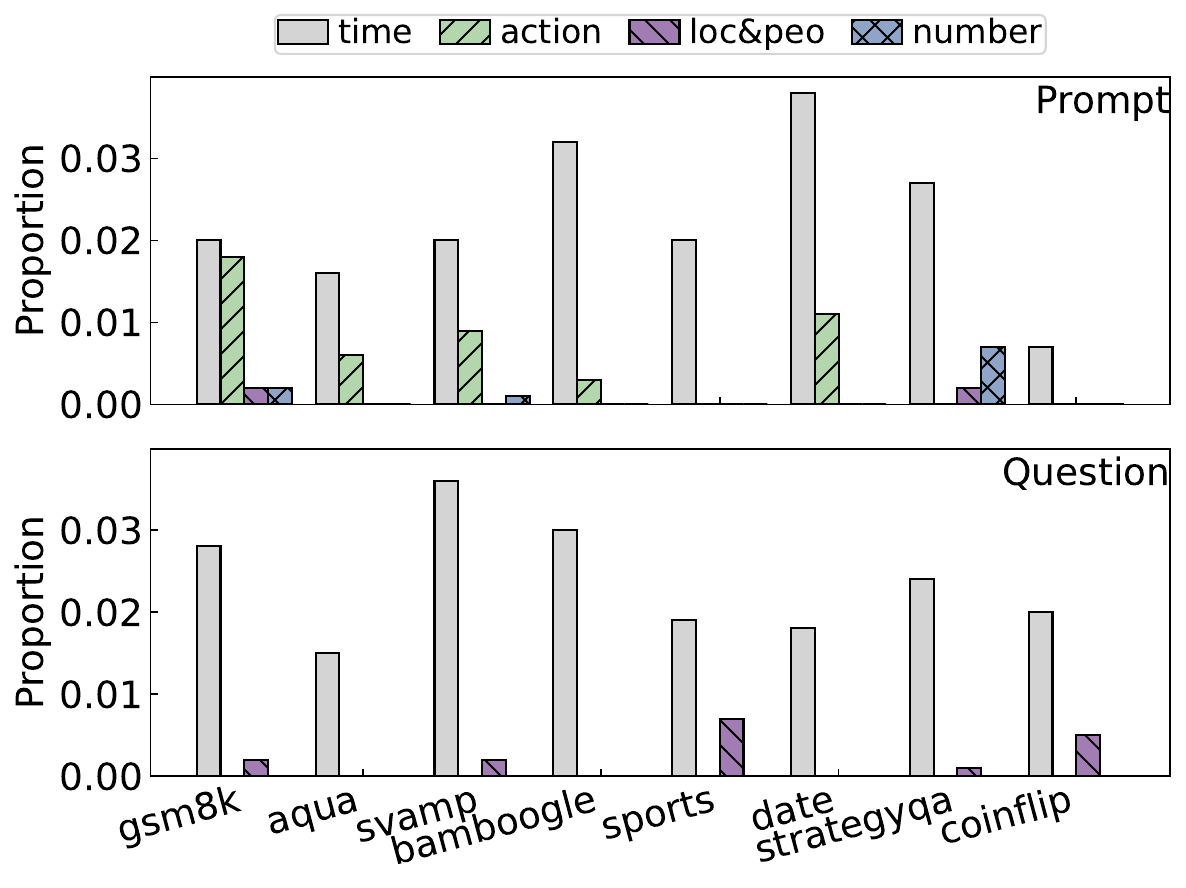}
        \caption{\textit{Last Letter}}
    \end{subfigure}
\caption{Results of \textit{transfer test} for different tasks under other prompts, results of LLaMA2-13b is reported.}
\label{fig:transfer_test_13b}
\end{figure*}
\clearpage
\section{Quantifying imitation}\label{sec:quantify_imitation}
To evaluate whether the model's generated content follows the CoT reasoning format, that is, whether it imitates CoT, we follow these steps:

First, we extracted the entities from the question. For arithmetic reasoning, the entities are numbers; for commonsense reasoning (such as the Bamboogle and Sports datasets), we used the Spacy package to identify entities like names, locations, time, etc. For the Date dataset, we extracted entities using regular expressions; for symbolic reasoning tasks like the Coin Flip dataset, we identified names; and for the Last Letter dataset, we extracted the target words.

Next, we determined whether new entities are generated. Since tasks like Coin Flip and Last Letter do not generate new entities, we evaluated the reasoning based on the number of occurrences of verbs like ``flips, is, was, are, be, were'' in the generated content. If the count exceeds 4, we considered that the model has generated new entities.

Finally, we checked whether an answer has been generated by looking for the phrase ``the answer is'' at the end of the generated content.

If all three steps are satisfied, the model's generated content is considered to follow the CoT reasoning format, meaning it successfully imitates the CoT exemplar.

\clearpage
\section{Result of transfer prompt}
\begin{figure}[ht]
\centering
    \begin{subfigure}{0.45\linewidth}
        \includegraphics[width=\linewidth]{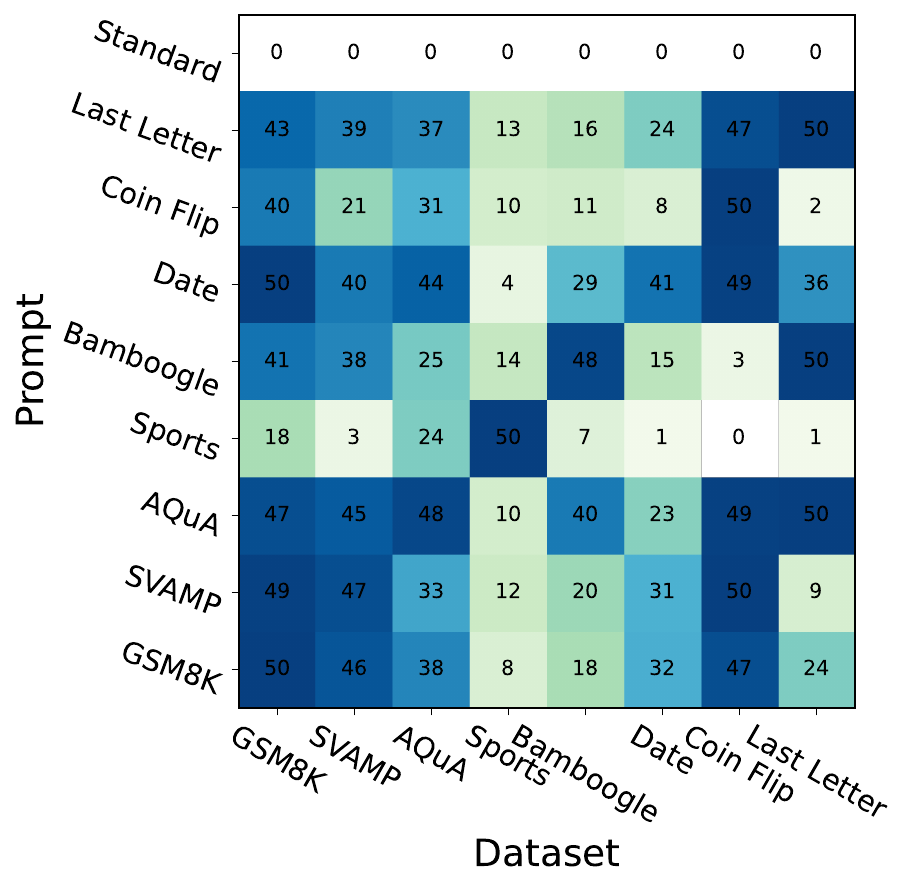}
        \caption{}
        \label{fig:acc_reason_2b}
    \end{subfigure}
    \begin{subfigure}{0.345\linewidth}
        \includegraphics[width=\linewidth]{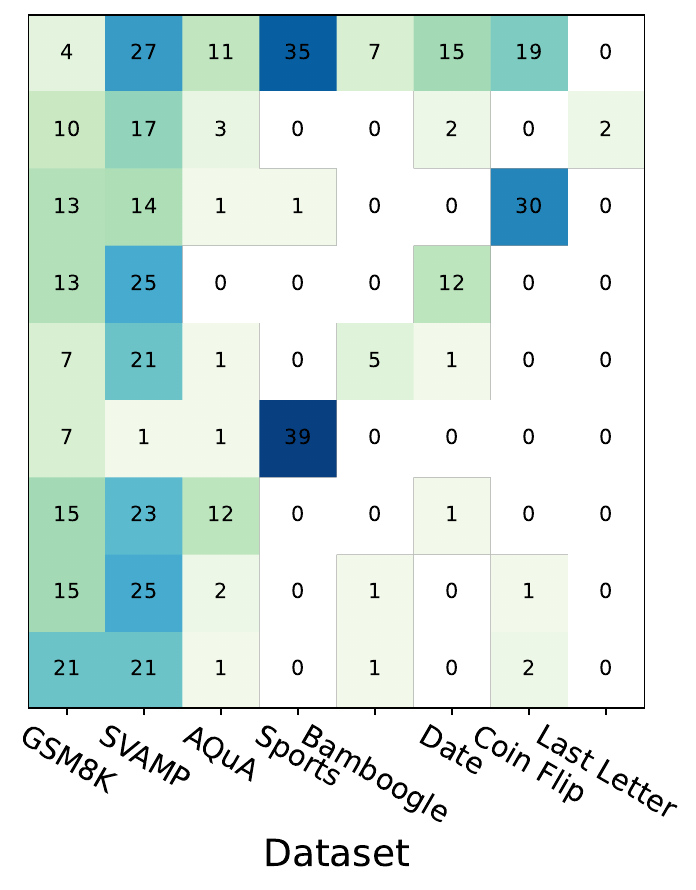}
        \caption{}
        \label{fig:acc_answer_2b}
    \end{subfigure}
\caption{Number of samples imitating exemplars
(left) and also answering correctly (right) for Gemma2-
2b.}
\label{fig:hotmap2b}
\end{figure}

\begin{figure*}[!h]
\centering
    \begin{subfigure}{0.45\linewidth}
        \includegraphics[width=\linewidth]{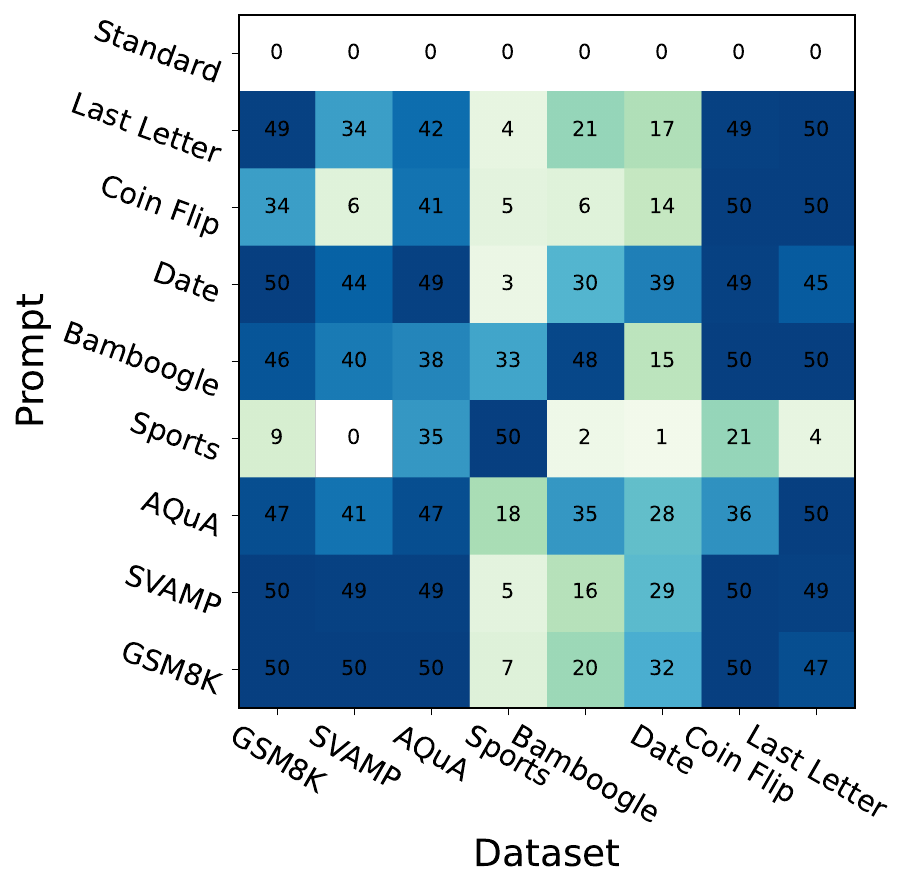}
        \caption{}
        \label{fig:acc_reason_9b}
    \end{subfigure}
    \begin{subfigure}{0.345\linewidth}
        \includegraphics[width=\linewidth]{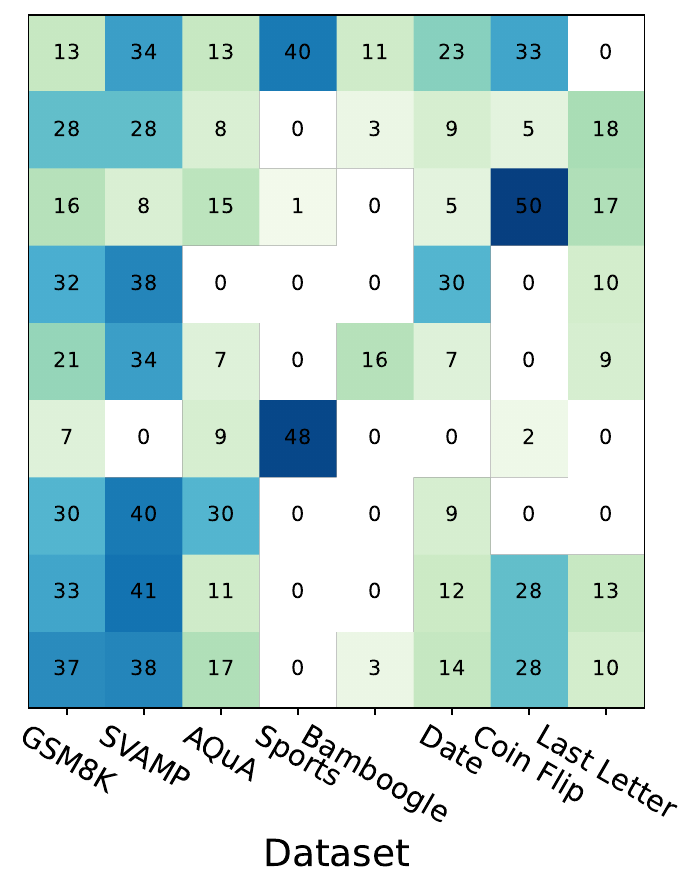}
        \caption{}
        \label{fig:acc_answer_9b}
    \end{subfigure}
\caption{Number of samples imitating exemplars
(left) and also answering correctly (right) for Gemma2-
9b.}
\label{fig:hotmap9b}
\end{figure*}

\begin{figure*}[!ht]
\centering
    \begin{subfigure}{0.45\linewidth}
        \includegraphics[width=\linewidth]{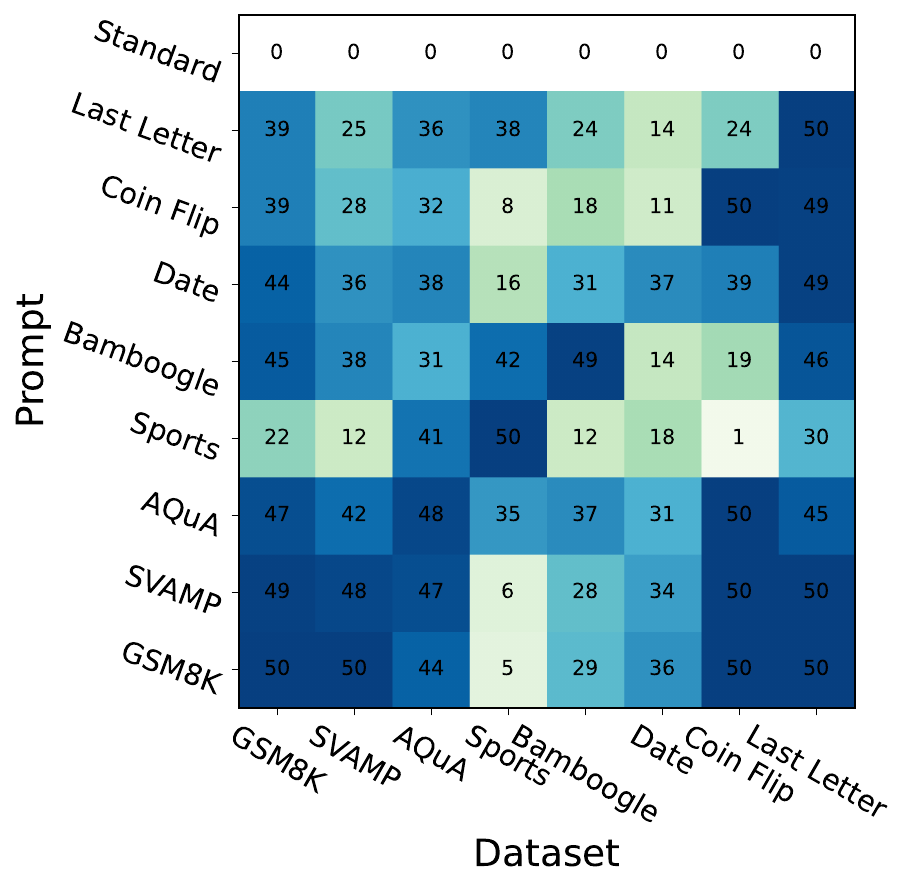}
        \caption{}
        \label{fig:acc_reason_13b}
    \end{subfigure}
    \begin{subfigure}{0.345\linewidth}
        \includegraphics[width=\linewidth]{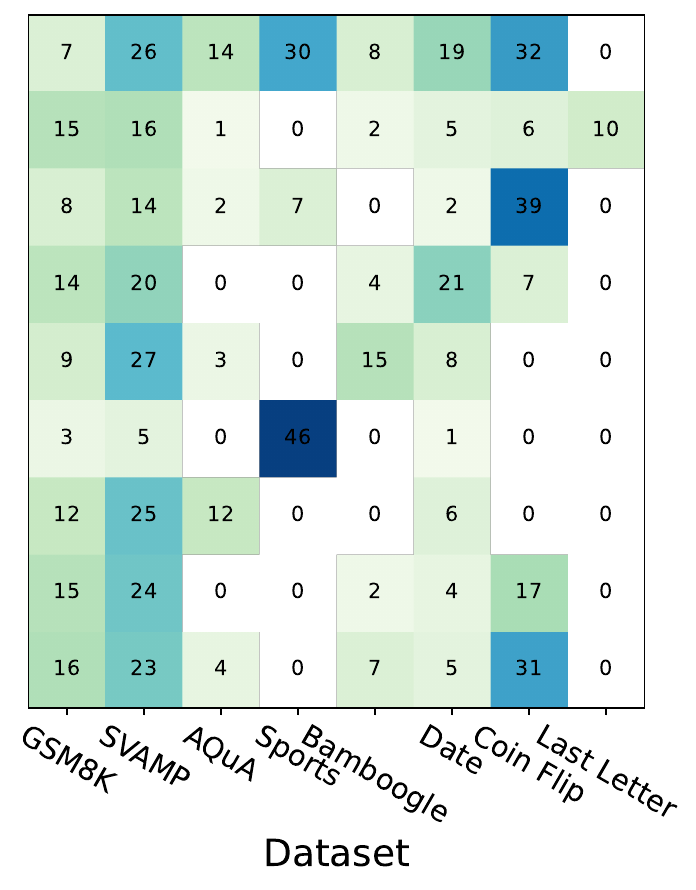}
        \caption{}
        \label{fig:acc_answer_13b}
    \end{subfigure}
\caption{Number of samples imitating exemplars
(left) and also answering correctly (right) for LLaMA2-13b.}
\label{fig:hotmap13b}
\end{figure*}

\begin{figure*}[!ht]
\centering
    \begin{subfigure}{0.45\linewidth}
        \includegraphics[width=\linewidth]{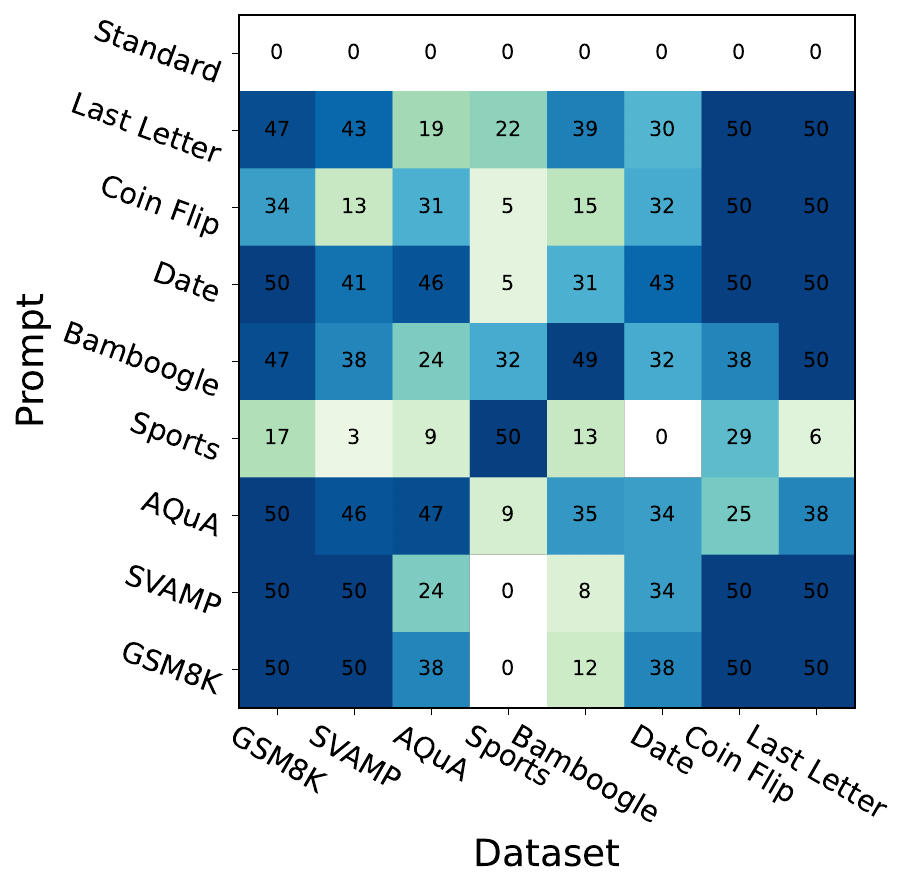}
        \caption{}
        \label{fig:acc_reason_27b}
    \end{subfigure}
    \begin{subfigure}{0.345\linewidth}
        \includegraphics[width=\linewidth]{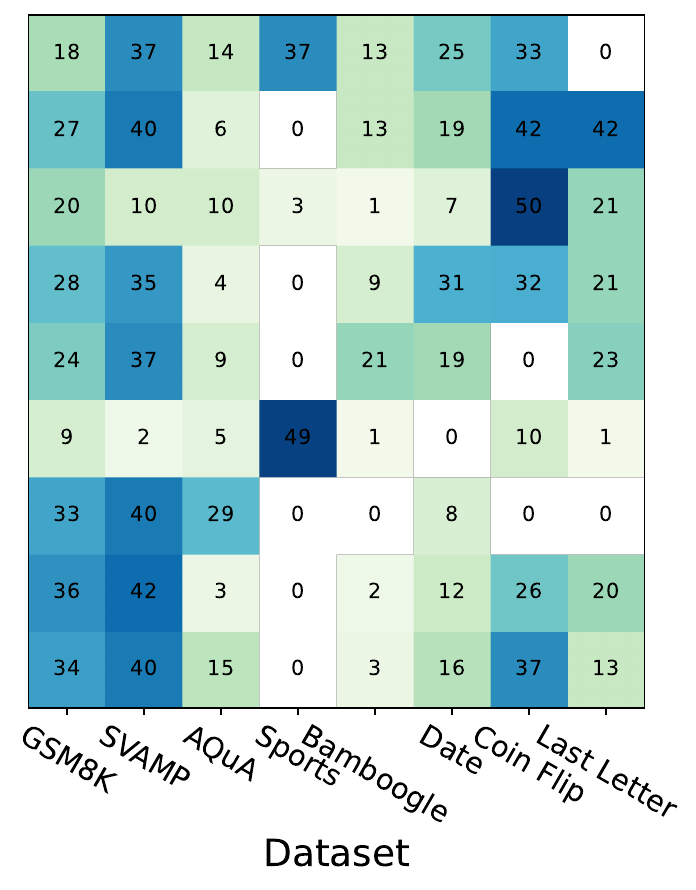}
        \caption{}
        \label{fig:acc_answer_27b}
    \end{subfigure}
\caption{Number of samples imitating exemplars
(left) and also answering correctly (right) for Gemma2-
27b.}
\label{fig:hotmap27b}
\end{figure*}

We applied cross-dataset CoT prompt transfer, testing it on four models: Gemma2-2b, Gemma2-9b, LLaMA2-13b, and Gemma2-27b. We obtained the generated content from each model and counted the number of instances that conformed to the general CoT reasoning format as well as the number of correct answers. Figure \ref{fig:hotmap2b} presents the experimental results for Gemma2-2b, Figure \ref{fig:hotmap9b} presents the experimental results for Gemma2-9b, Figure \ref{fig:hotmap13b} presents the experimental results for LLaMA2-13b, and Figure \ref{fig:hotmap27b} shows the results for Gemma2-27b.

\clearpage
\section{Probabilities of each step for all test models}\label{sec:prob_step}
\begin{figure*}[ht]
\centering
    \begin{subfigure}{0.3\linewidth}
        \includegraphics[width=\linewidth]
        {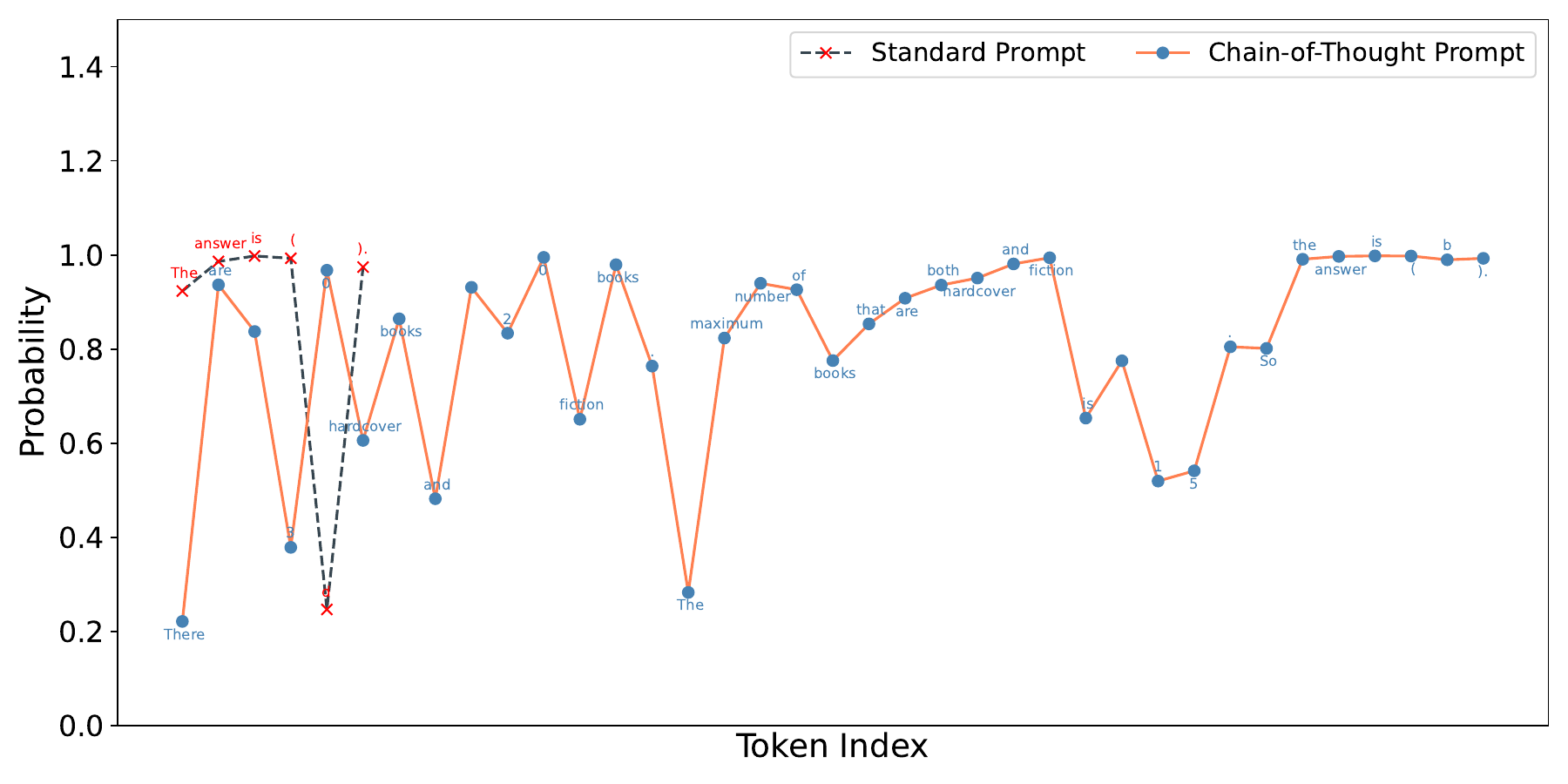}
        \caption{\textit{aqua}}
    \end{subfigure}
    \begin{subfigure}{0.3\linewidth}
        \includegraphics[width=\linewidth]
        {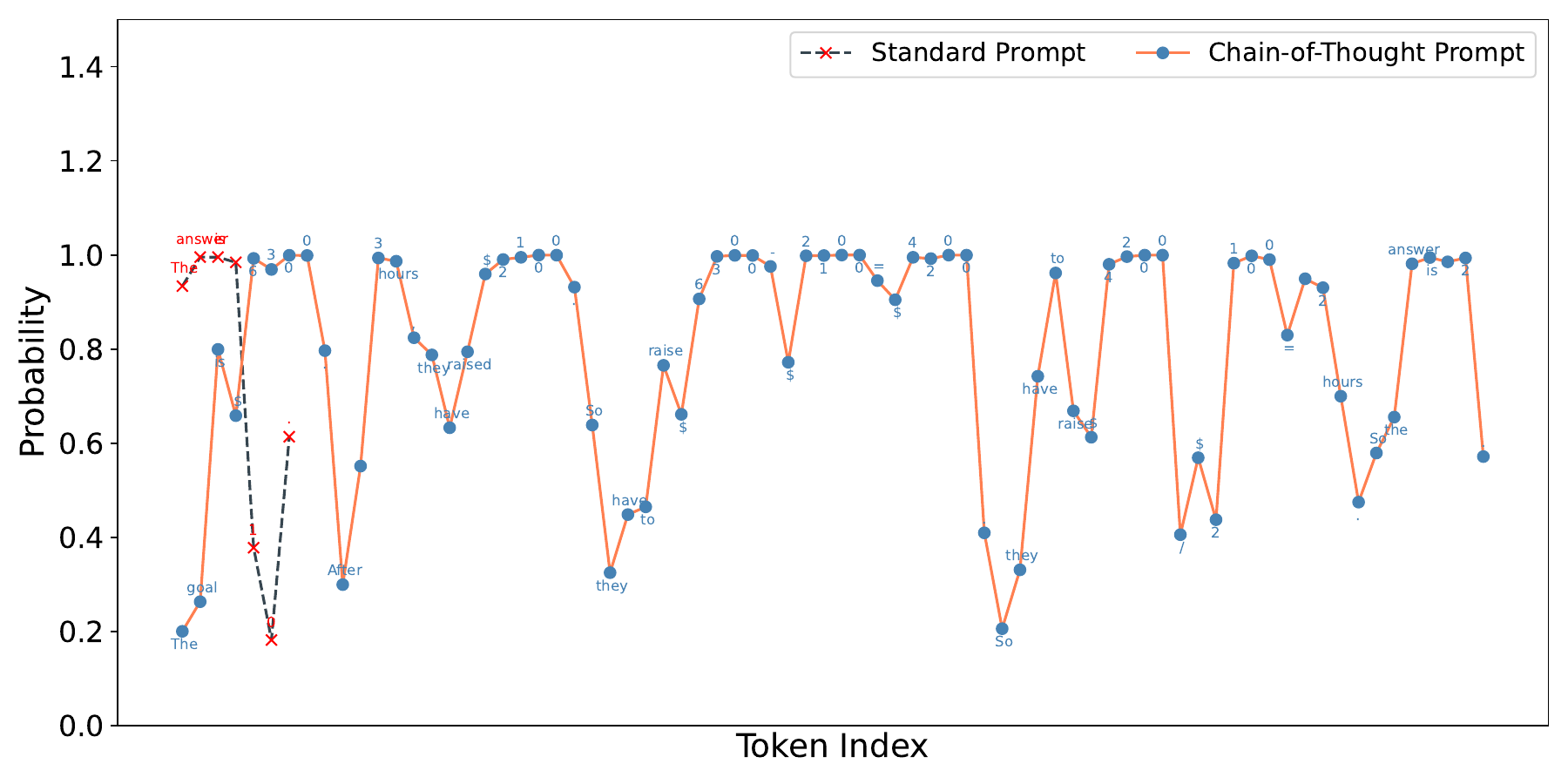}
        \caption{\textit{gsm8k}}
    \end{subfigure}
    \begin{subfigure}{0.3\linewidth}
        \includegraphics[width=\linewidth]{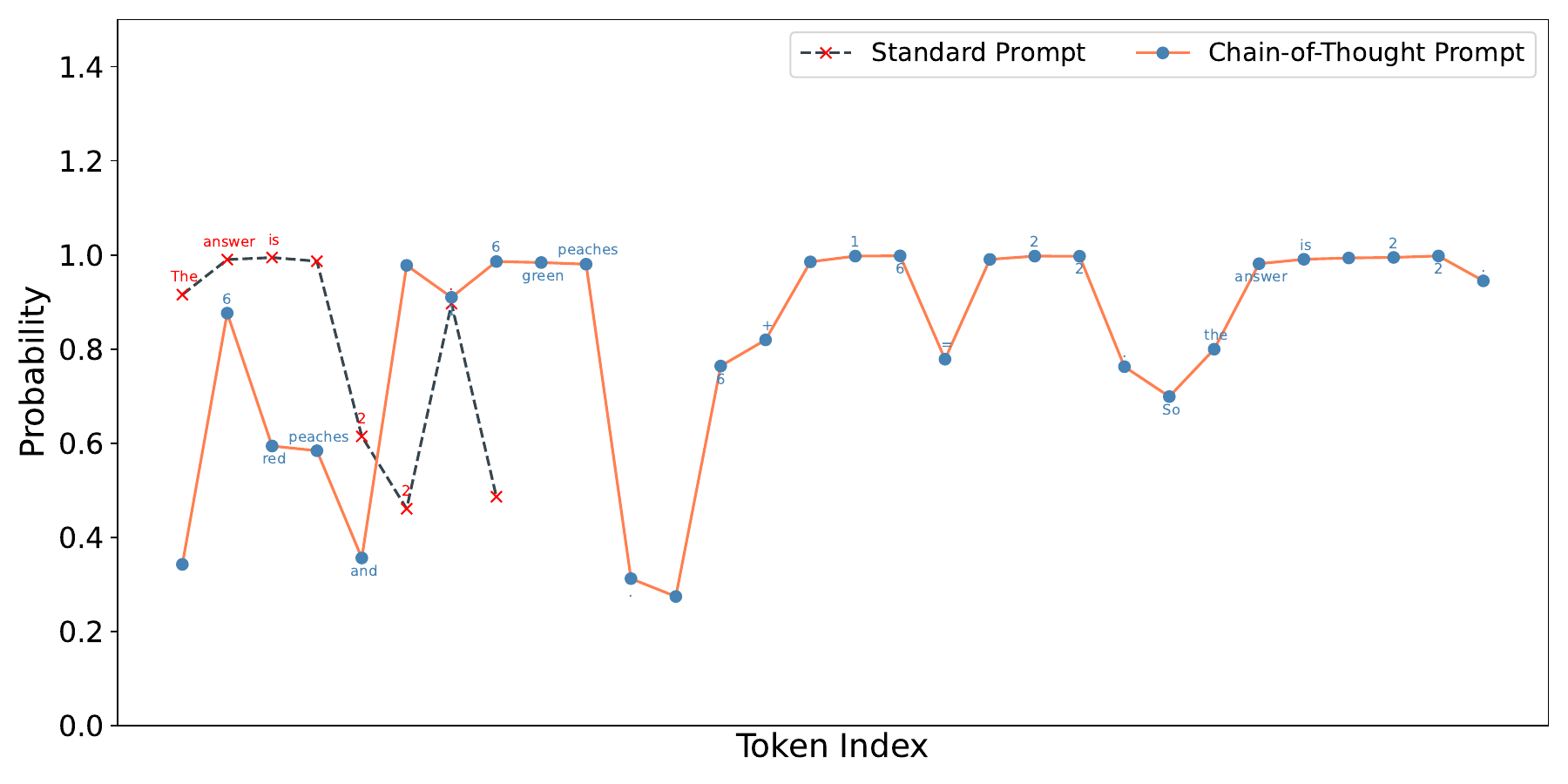}
        \caption{\textit{svamp}}
    \end{subfigure}
    \begin{subfigure}{0.3\linewidth}
        \includegraphics[width=\linewidth]
        {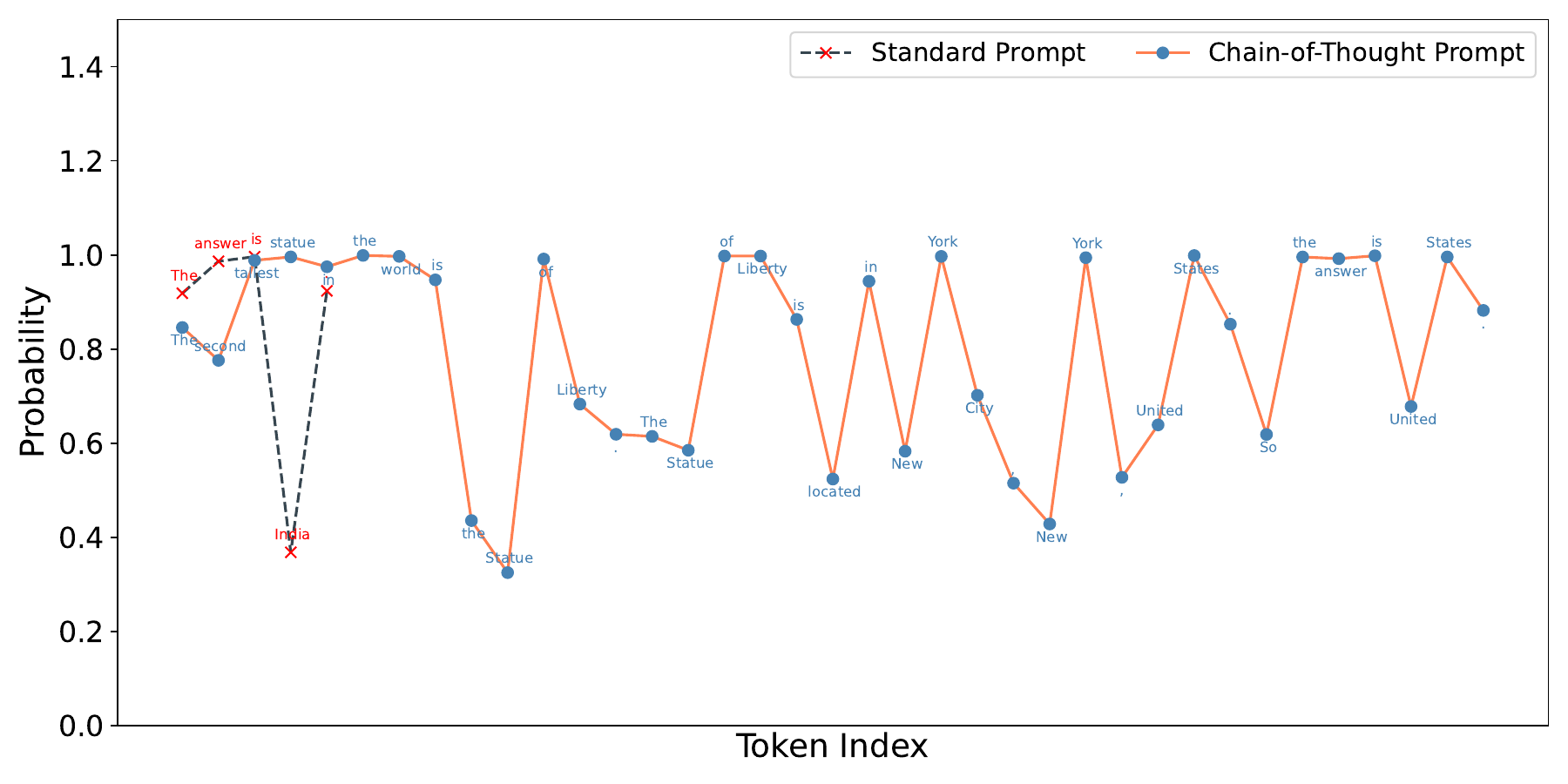}
        \caption{\textit{bamboogle}}
    \end{subfigure}
    \begin{subfigure}{0.3\linewidth}
        \includegraphics[width=\linewidth]
        {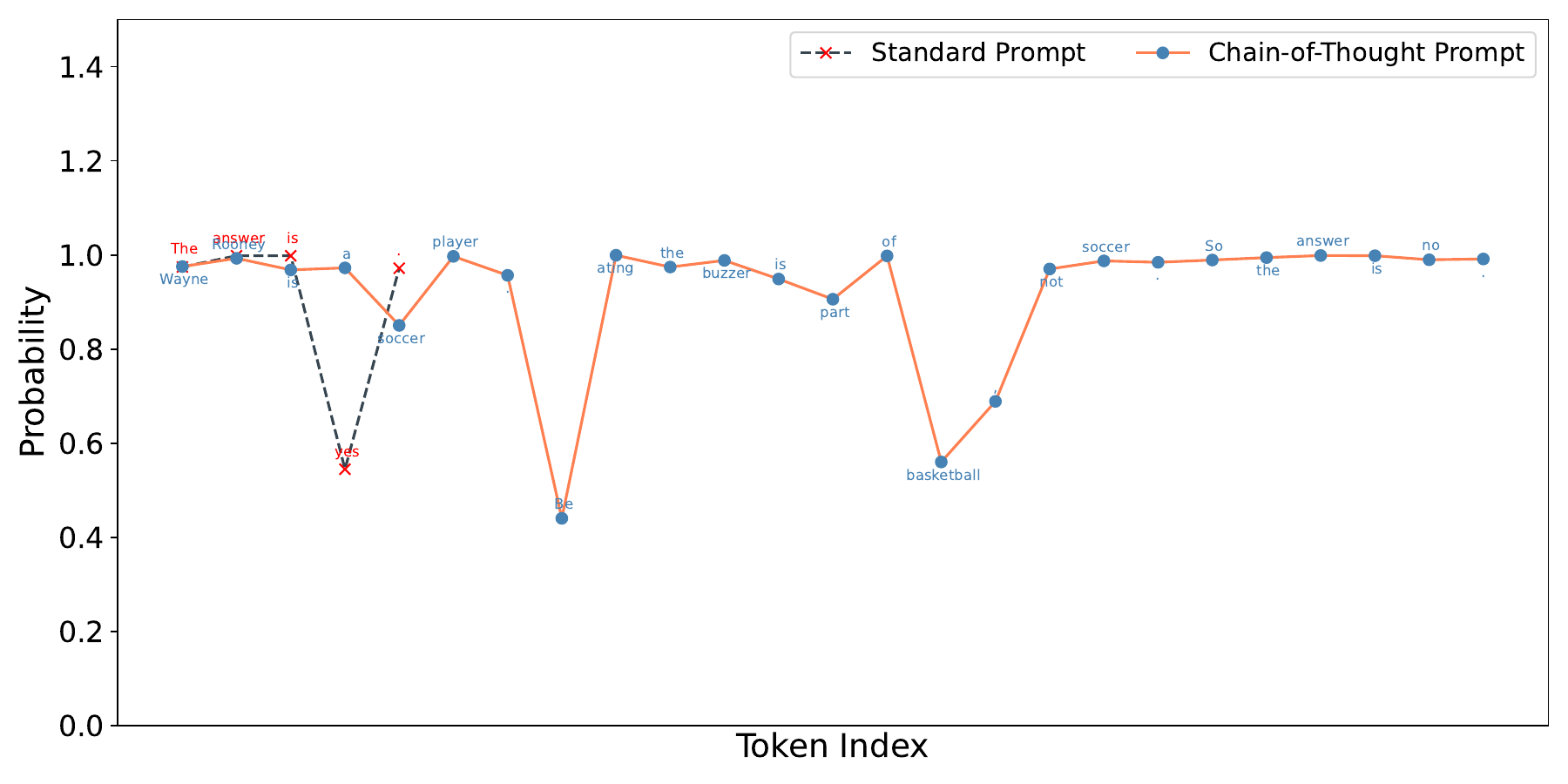}
        \caption{\textit{sports}}
    \end{subfigure}
    \begin{subfigure}{0.3\linewidth}
        \includegraphics[width=\linewidth]
        {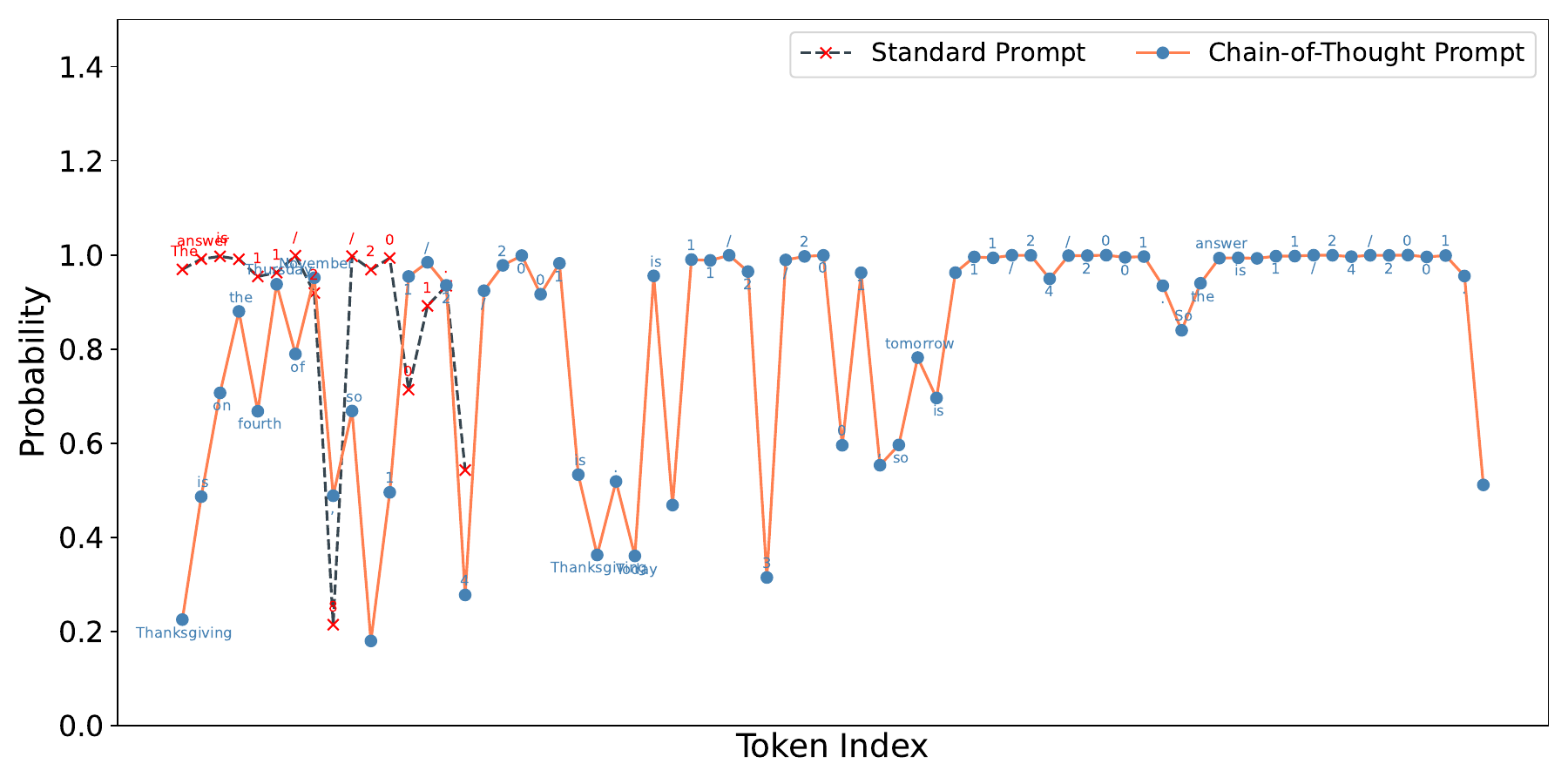}
        \caption{\textit{date}}
    \end{subfigure}
    \begin{subfigure}{0.3\linewidth}
        \includegraphics[width=\linewidth]
        {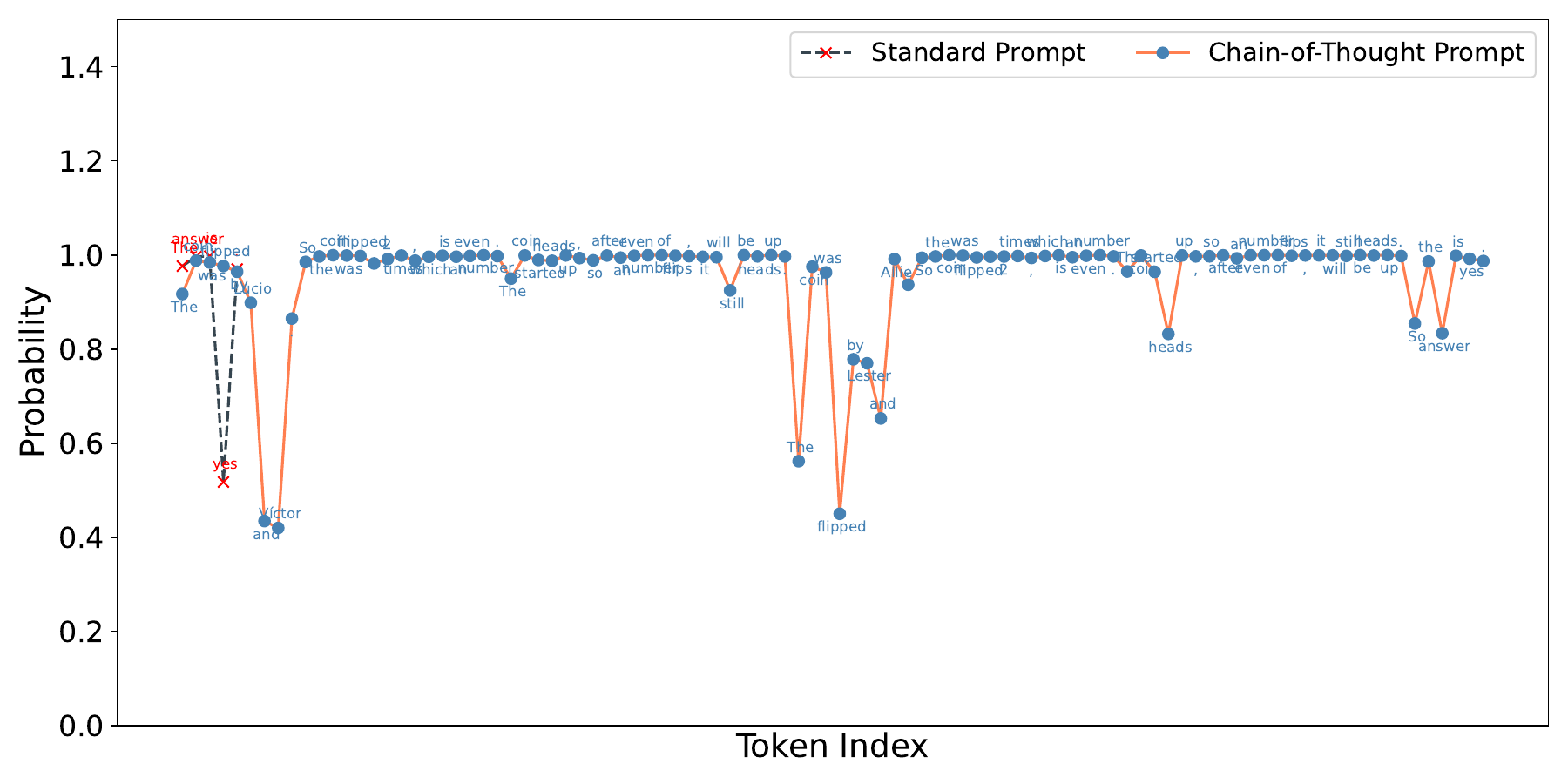}
        \caption{\textit{coin flip}}
    \end{subfigure}
     \begin{subfigure}{0.3\linewidth}
        \includegraphics[width=\linewidth]
        {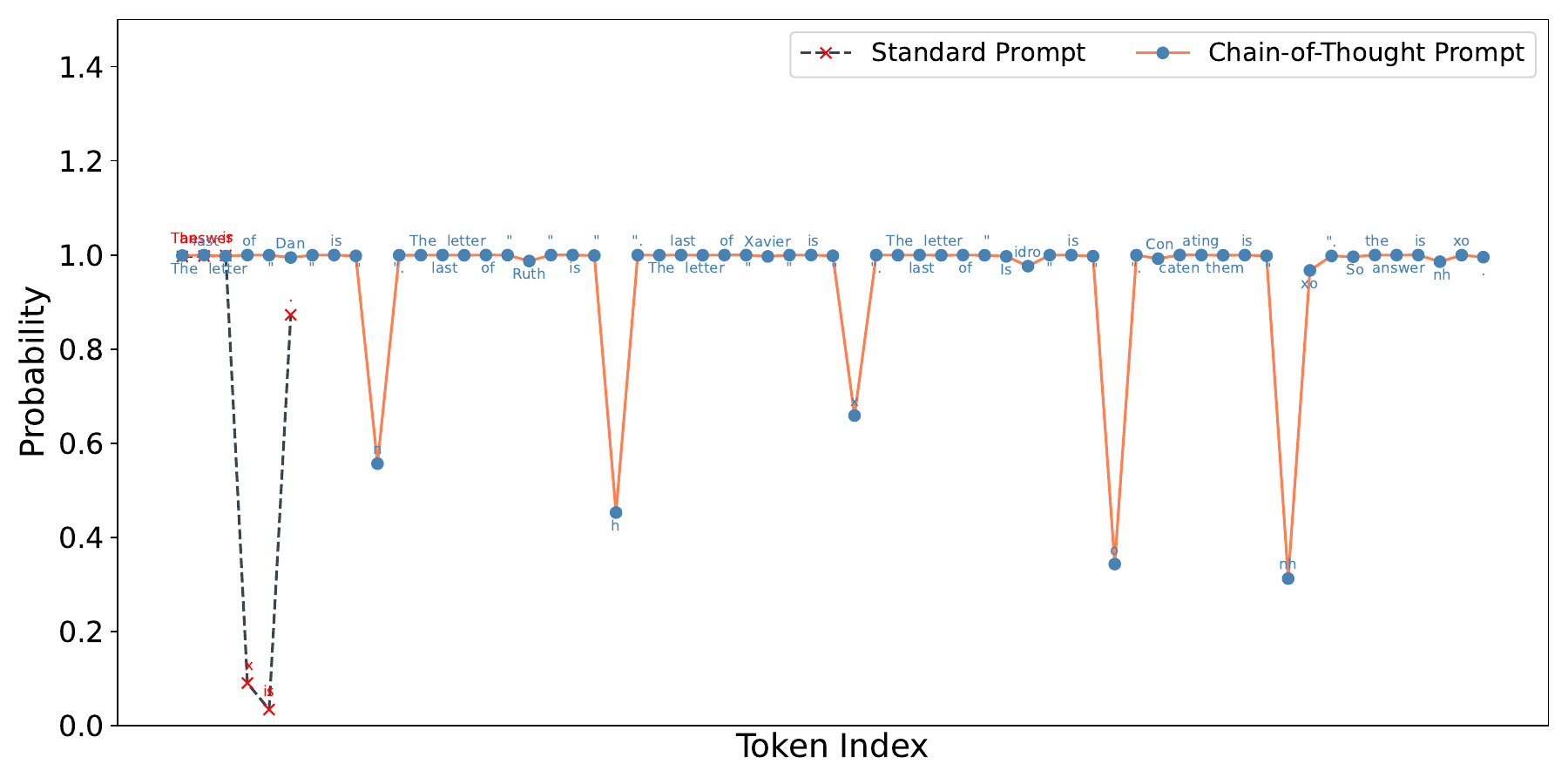}
        \caption{\textit{Last Letter Concatenation}}
    \end{subfigure}
\caption{Probability value of each generated token (results of Gemma2-2b is shown).}
\label{fig:question2_logits_value_2b}
\end{figure*}

\begin{figure*}
\centering
    \begin{subfigure}{0.3\linewidth}
        \includegraphics[width=\linewidth]
        {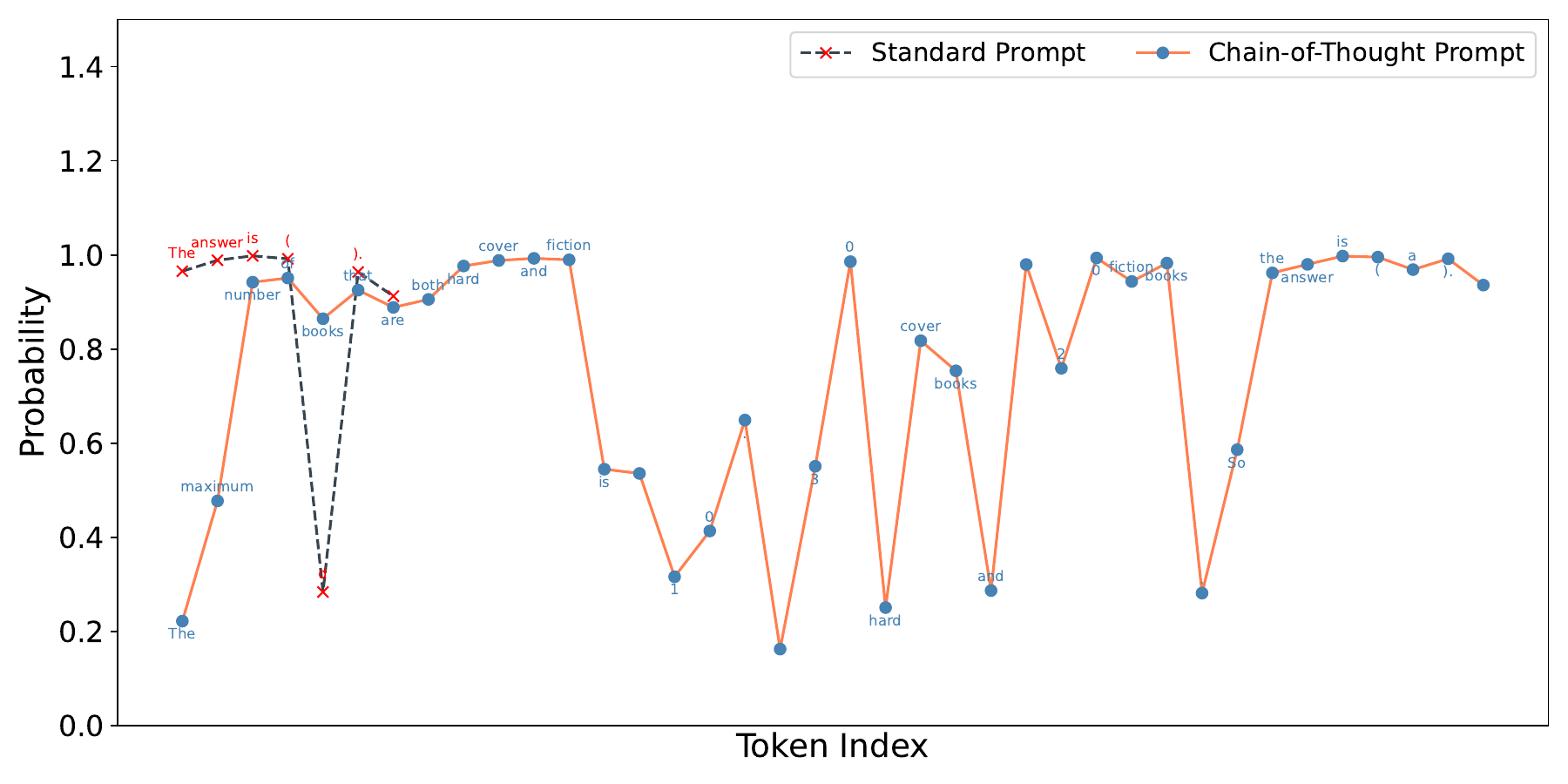}
        \caption{\textit{aqua}}
    \end{subfigure}
    \begin{subfigure}{0.3\linewidth}
        \includegraphics[width=\linewidth]
        {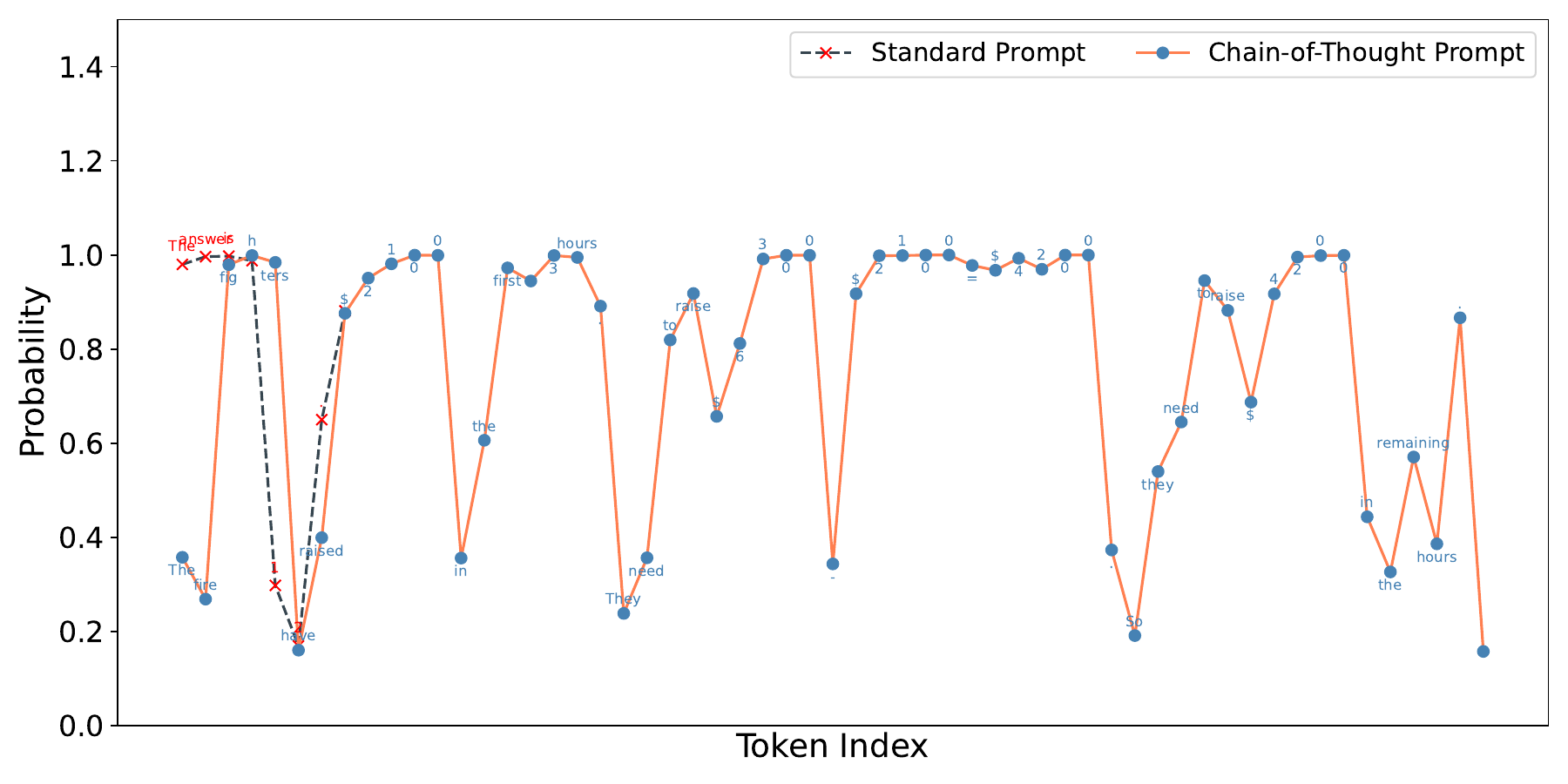}
        \caption{\textit{gsm8k}}
    \end{subfigure}
    \begin{subfigure}{0.3\linewidth}
        \includegraphics[width=\linewidth]{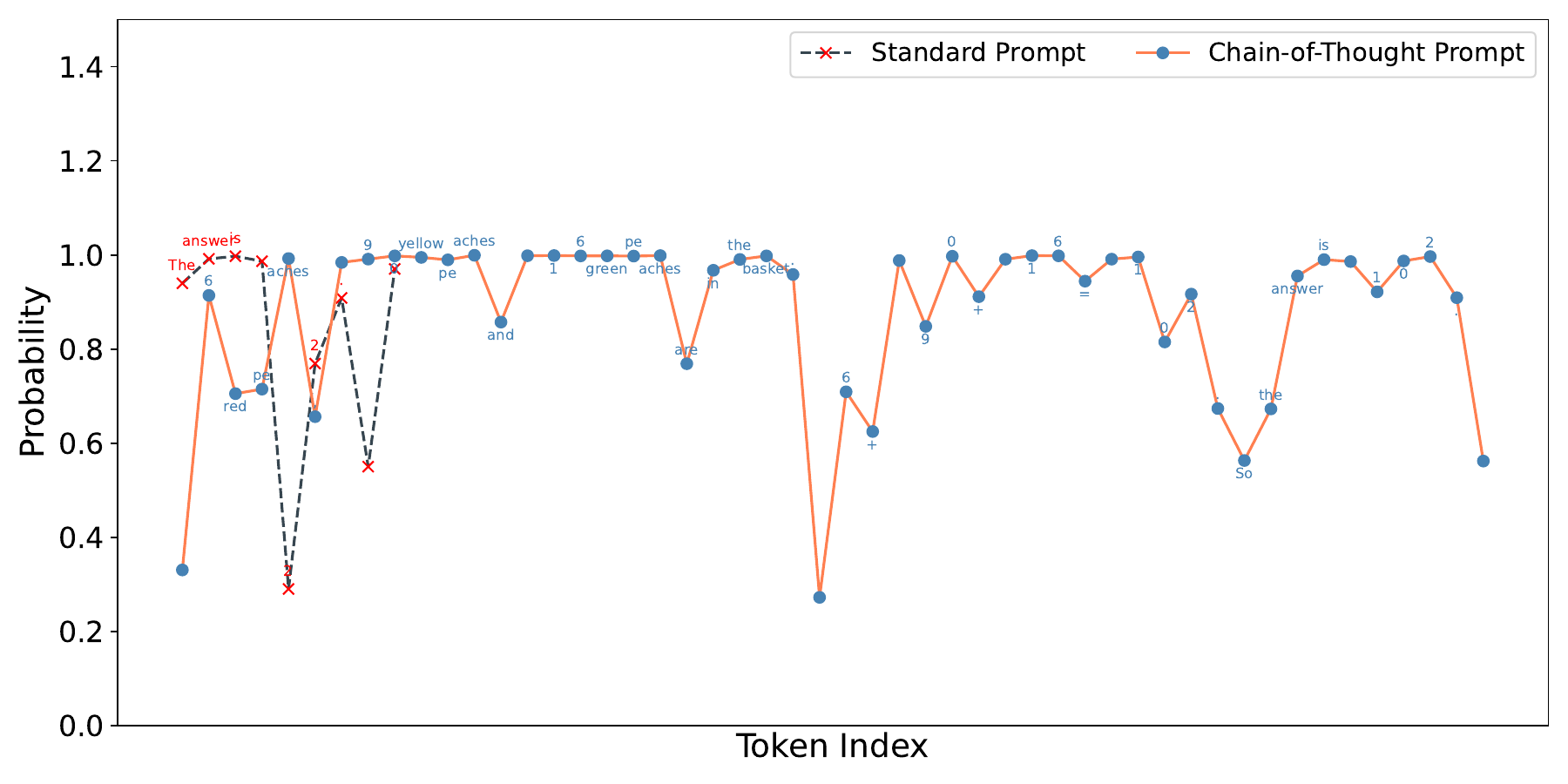}
        \caption{\textit{svamp}}
    \end{subfigure}
    \begin{subfigure}{0.3\linewidth}
        \includegraphics[width=\linewidth]
        {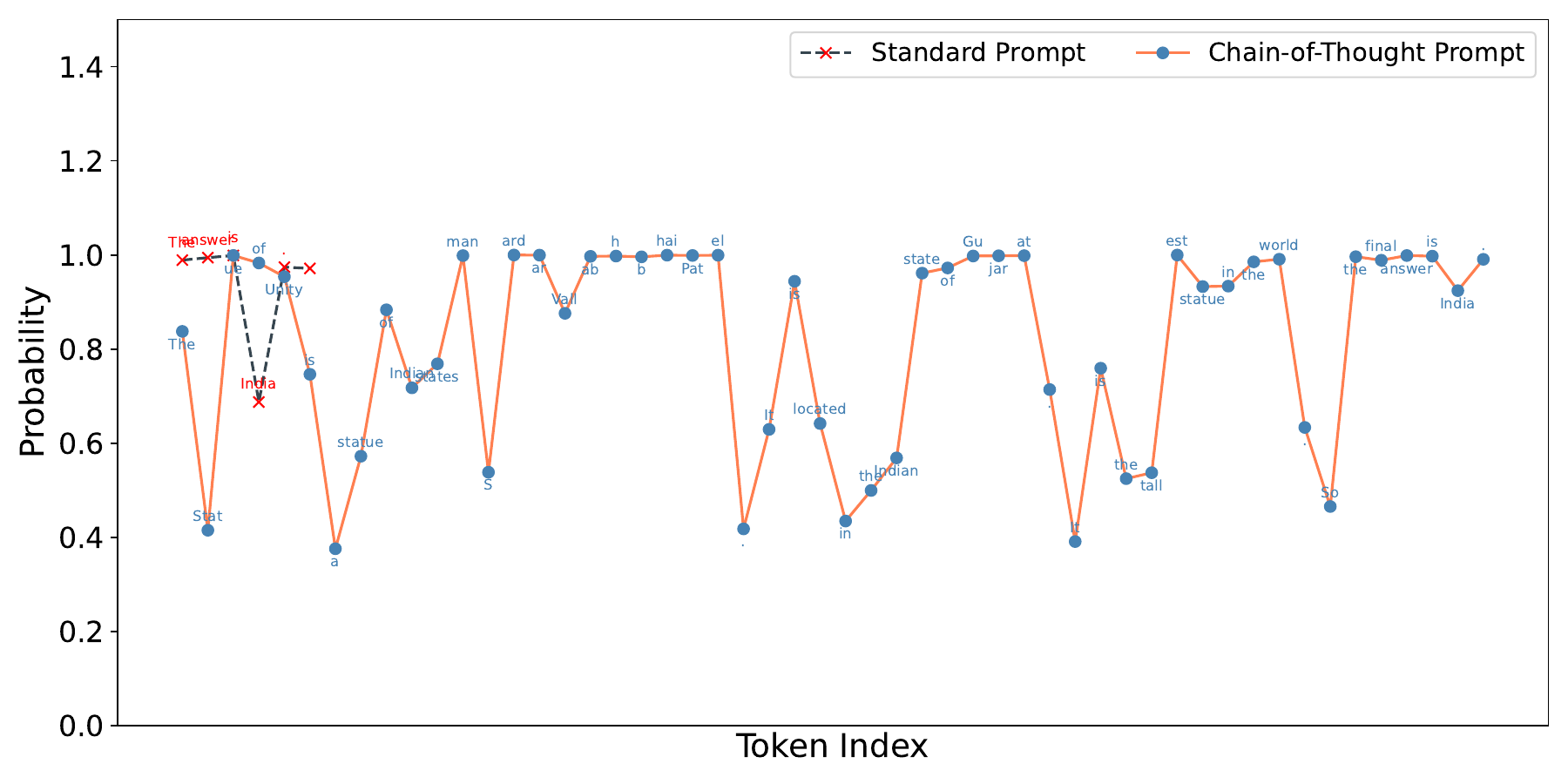}
        \caption{\textit{bamboogle}}
    \end{subfigure}
    \begin{subfigure}{0.3\linewidth}
        \includegraphics[width=\linewidth]
        {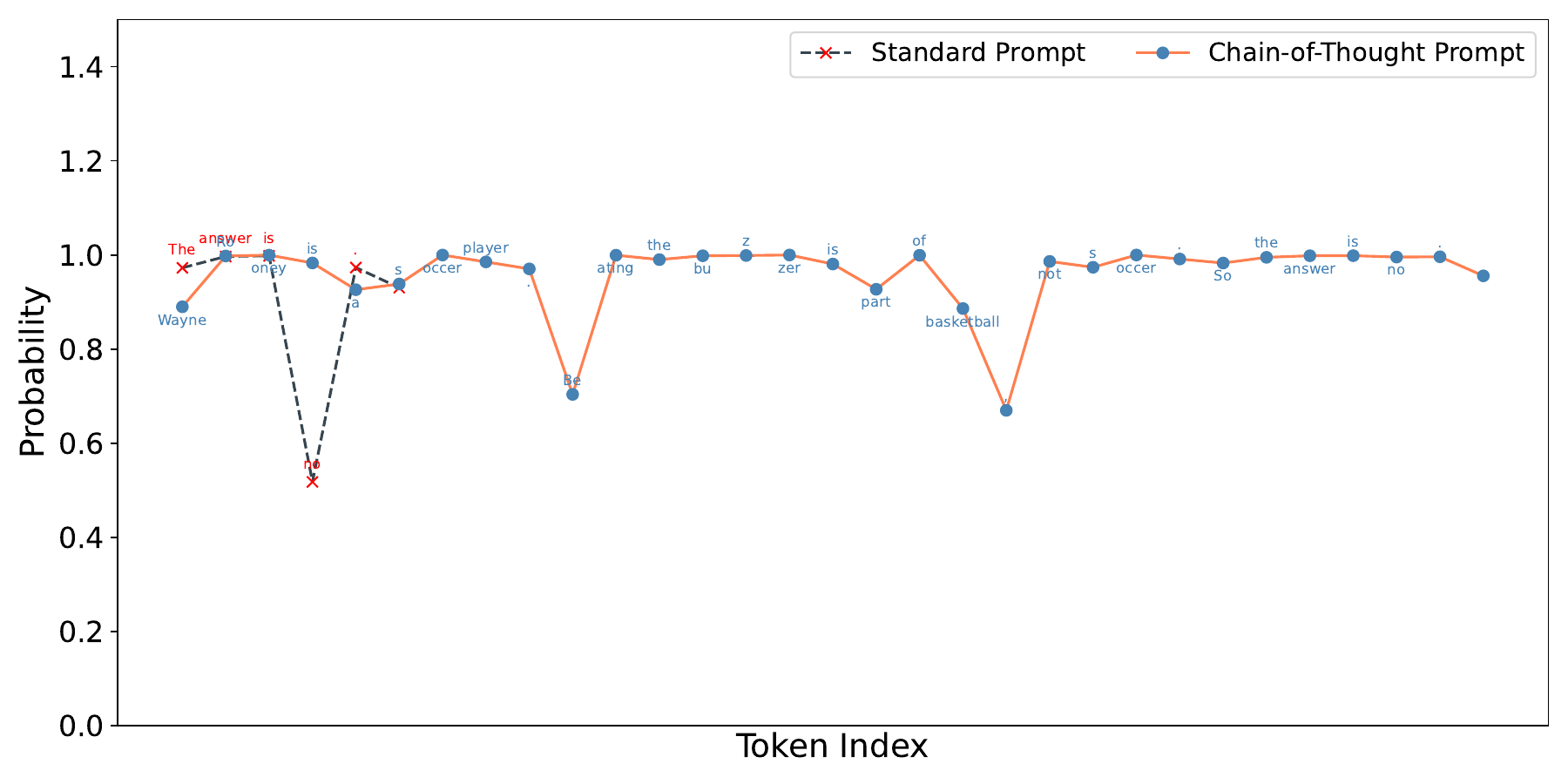}
        \caption{\textit{sports}}
    \end{subfigure}
    \begin{subfigure}{0.3\linewidth}
        \includegraphics[width=\linewidth]
        {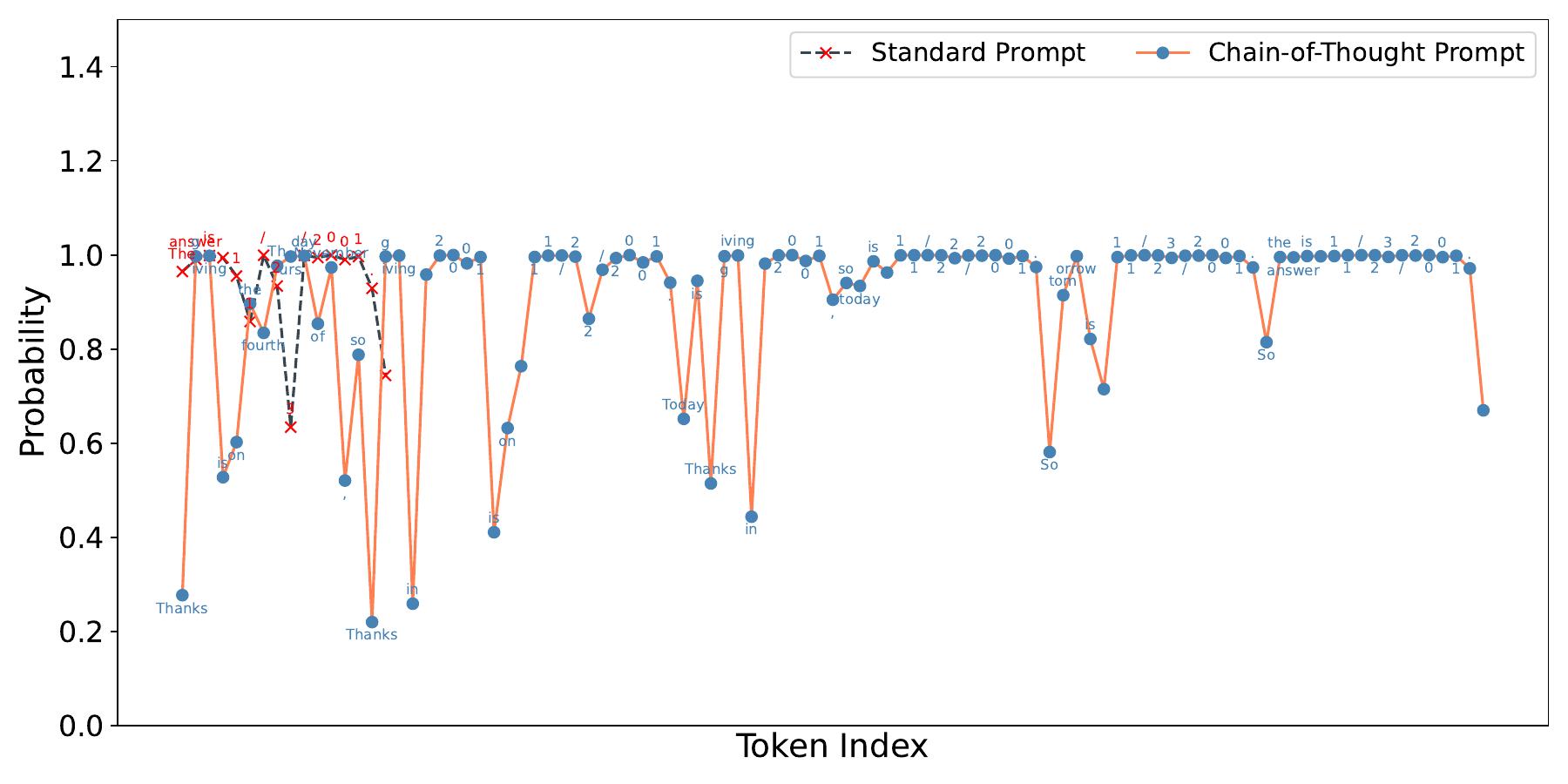}
        \caption{\textit{date}}
    \end{subfigure}
    \begin{subfigure}{0.3\linewidth}
        \includegraphics[width=\linewidth]
        {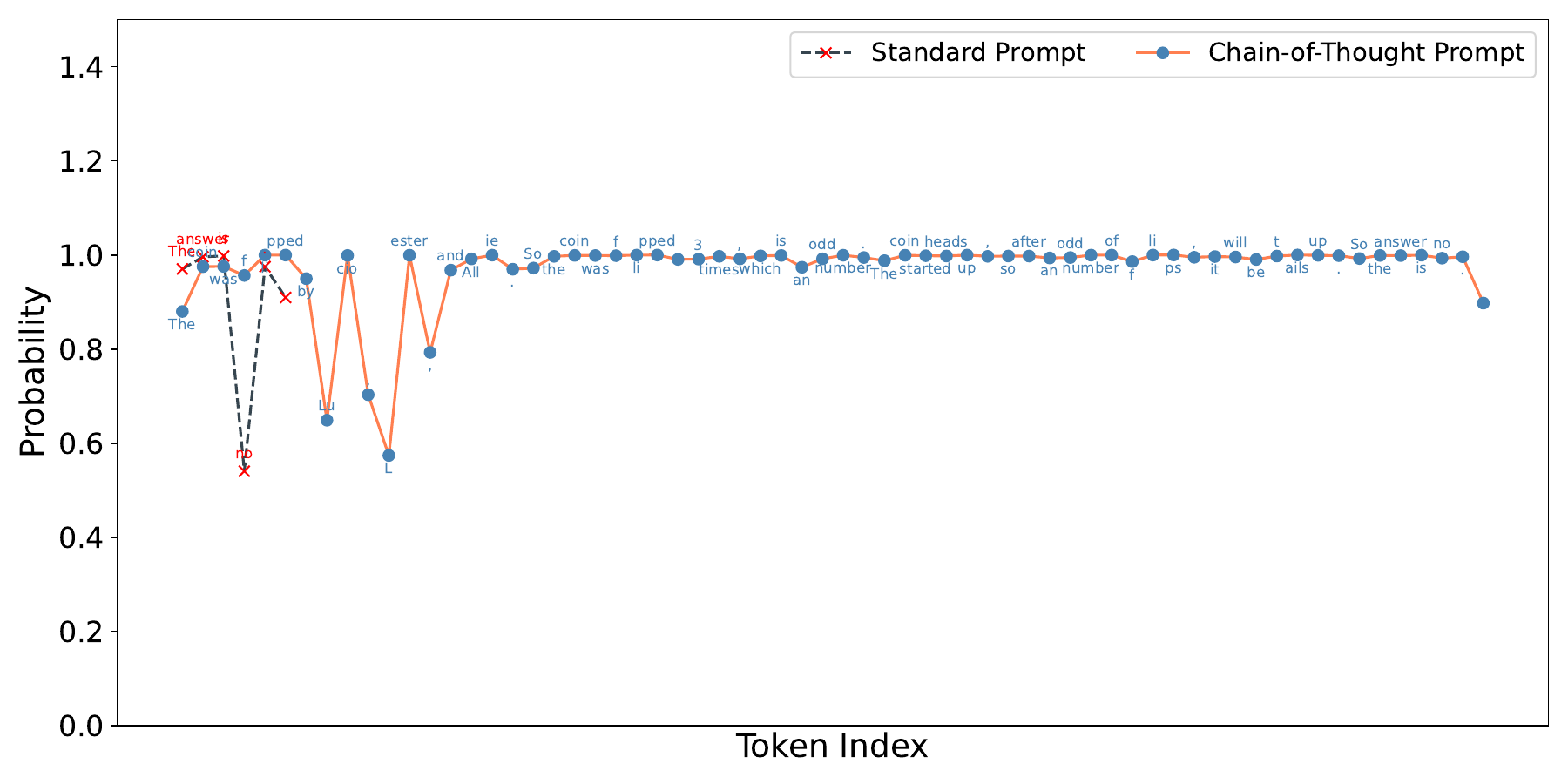}
        \caption{\textit{coin flip}}
    \end{subfigure}
     \begin{subfigure}{0.3\linewidth}
        \includegraphics[width=\linewidth]
        {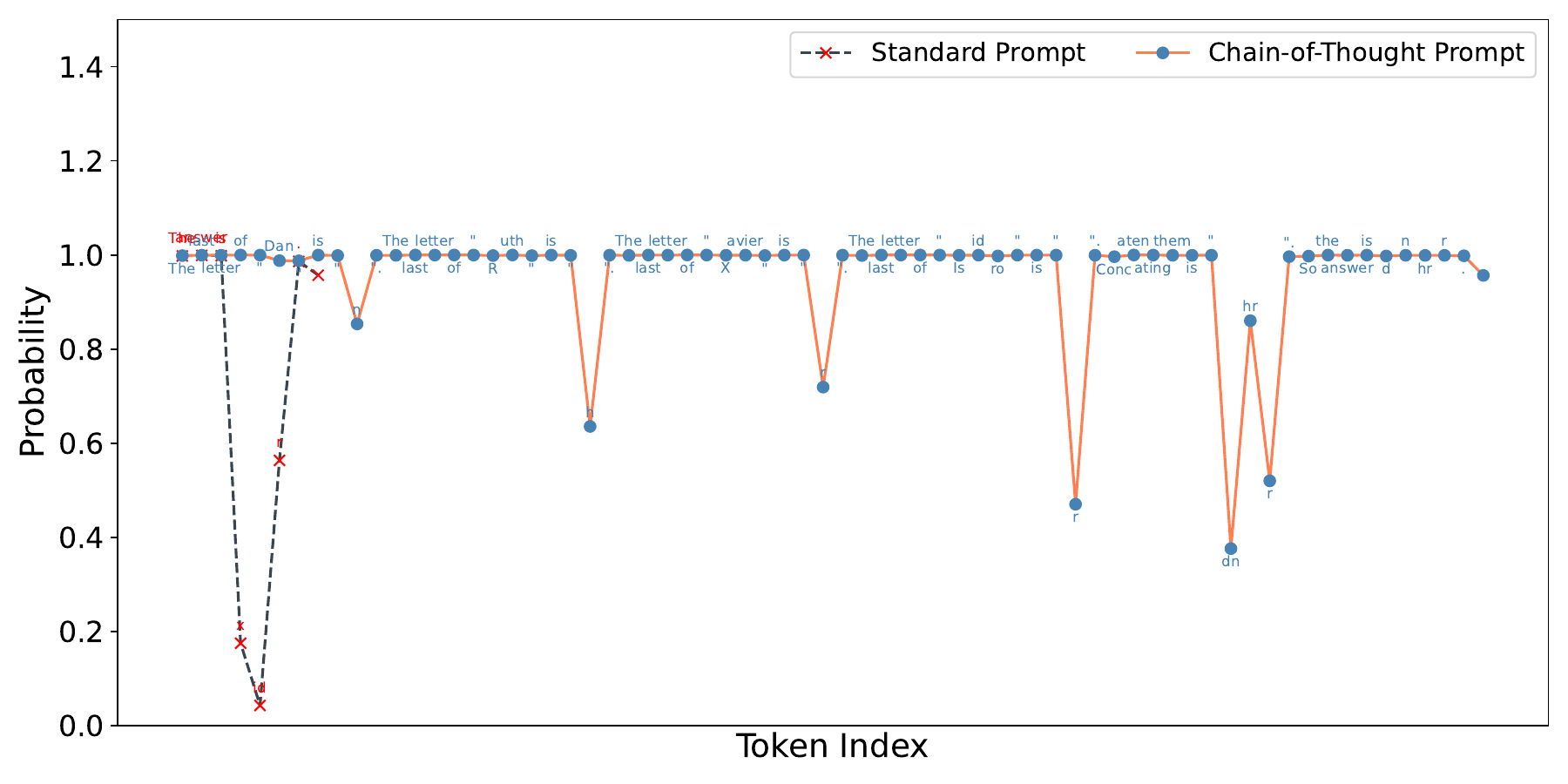}
        \caption{\textit{Last Letter Concatenation}}
    \end{subfigure}
\caption{Probability value of each generated token (the results of LLaMA2-13b is shown).}
\label{fig:question2_logits_value_13b}
\end{figure*}

\begin{figure*}
\centering
    \begin{subfigure}{0.3\linewidth}
        \includegraphics[width=\linewidth]
        {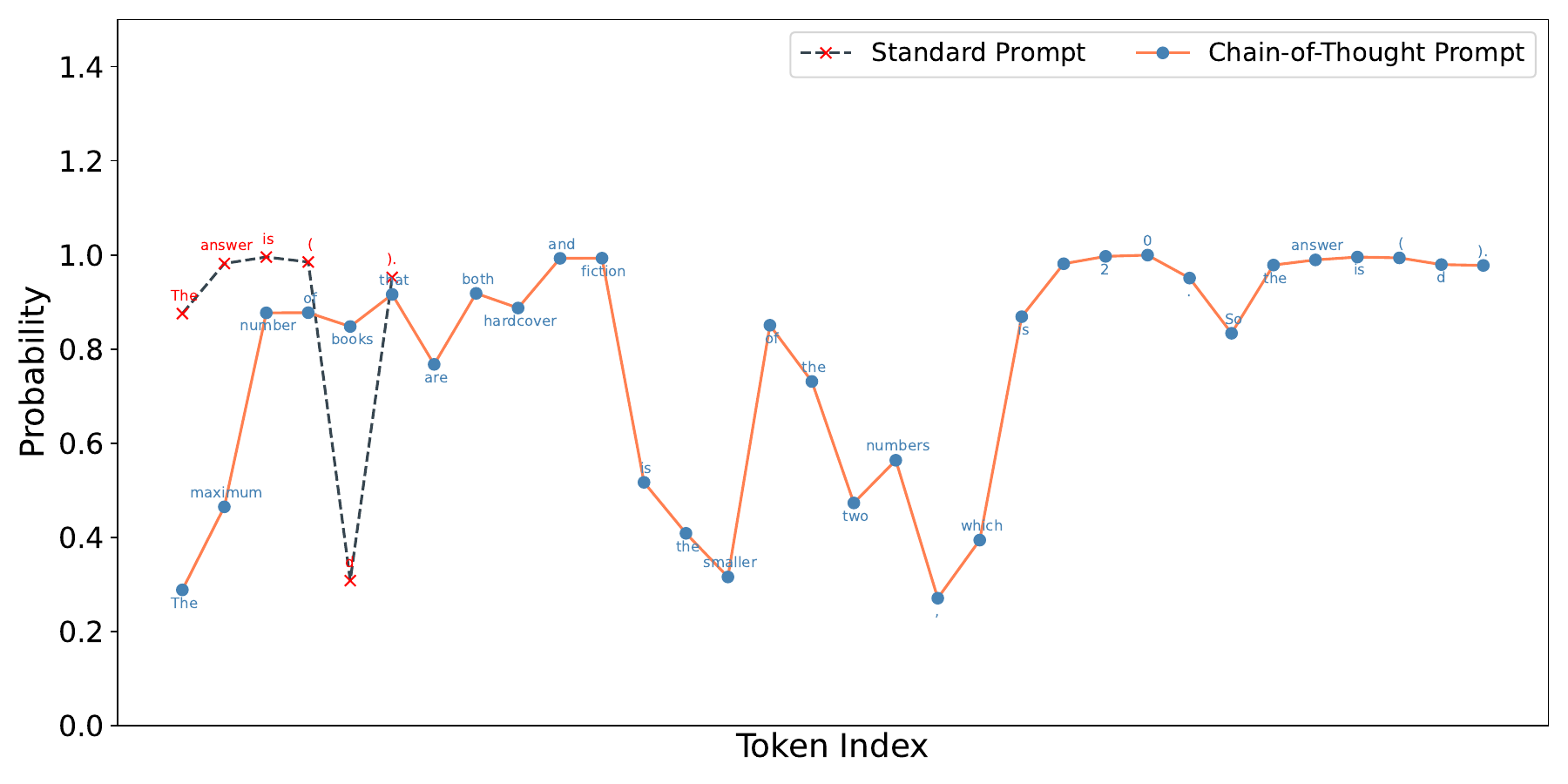}
        \caption{\textit{aqua}}
    \end{subfigure}
    \begin{subfigure}{0.3\linewidth}
        \includegraphics[width=\linewidth]
        {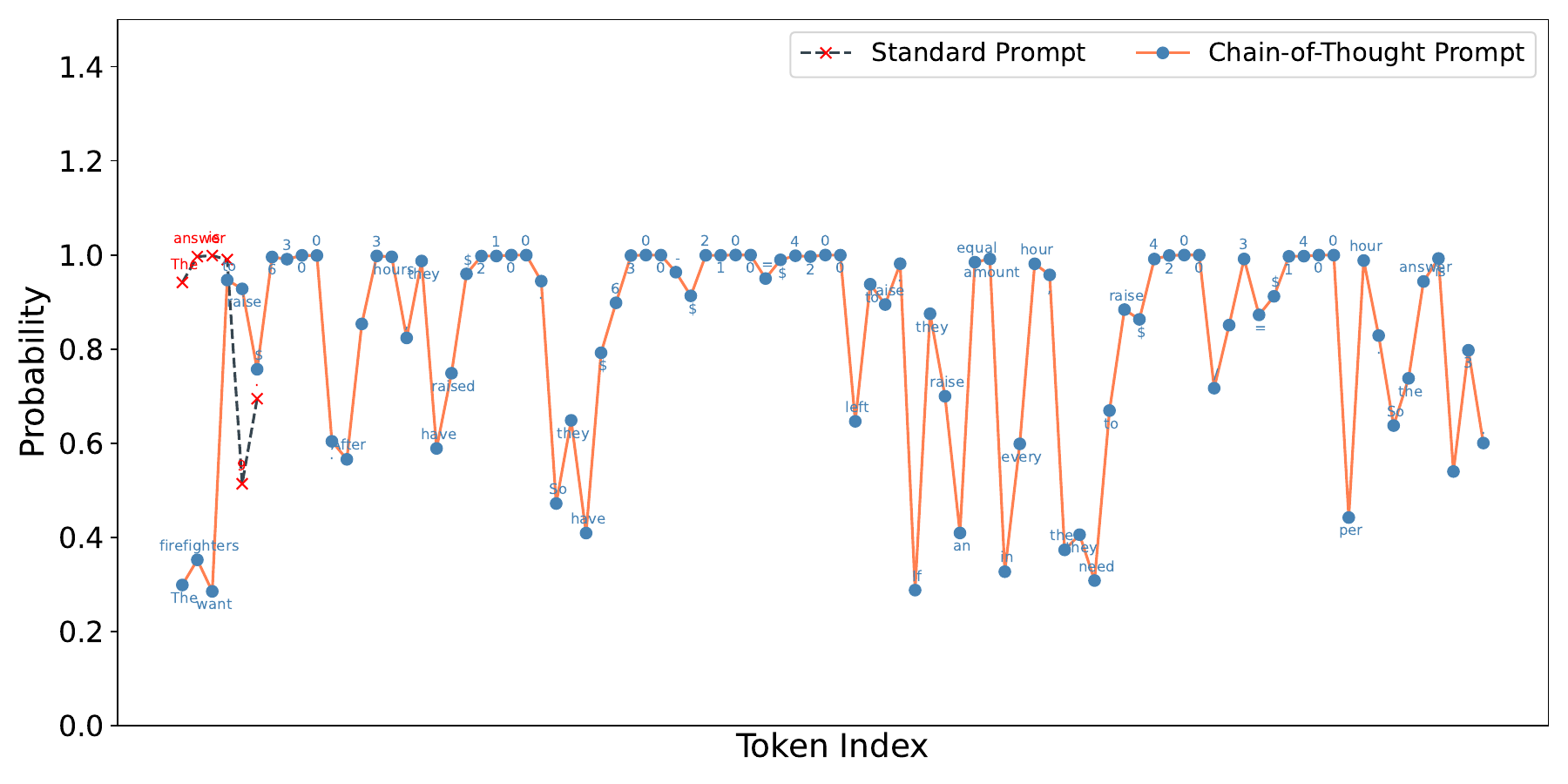}
        \caption{\textit{gsm8k}}
    \end{subfigure}
    \begin{subfigure}{0.3\linewidth}
        \includegraphics[width=\linewidth]{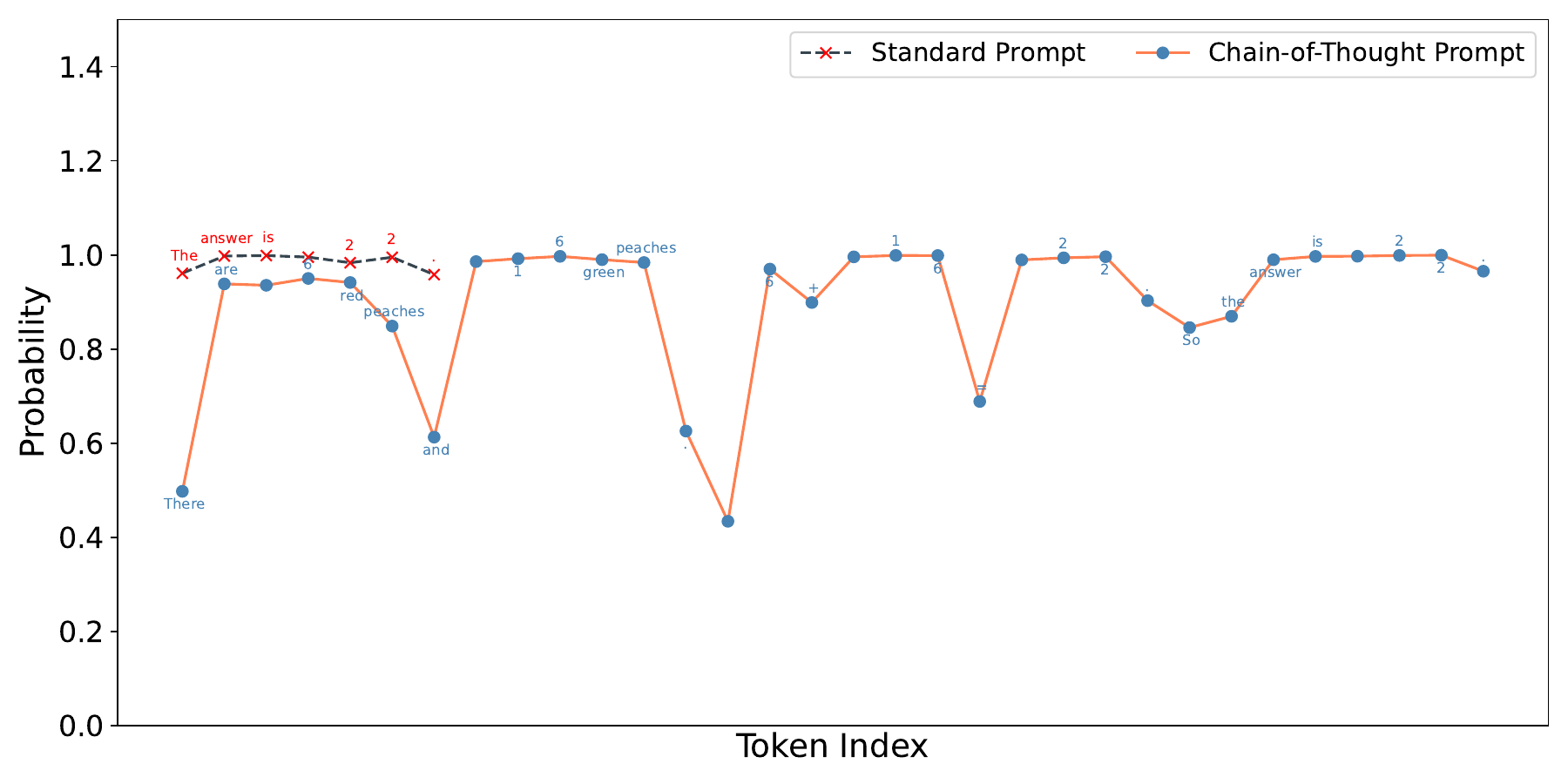}
        \caption{\textit{svamp}}
    \end{subfigure}
    \begin{subfigure}{0.3\linewidth}
        \includegraphics[width=\linewidth]
        {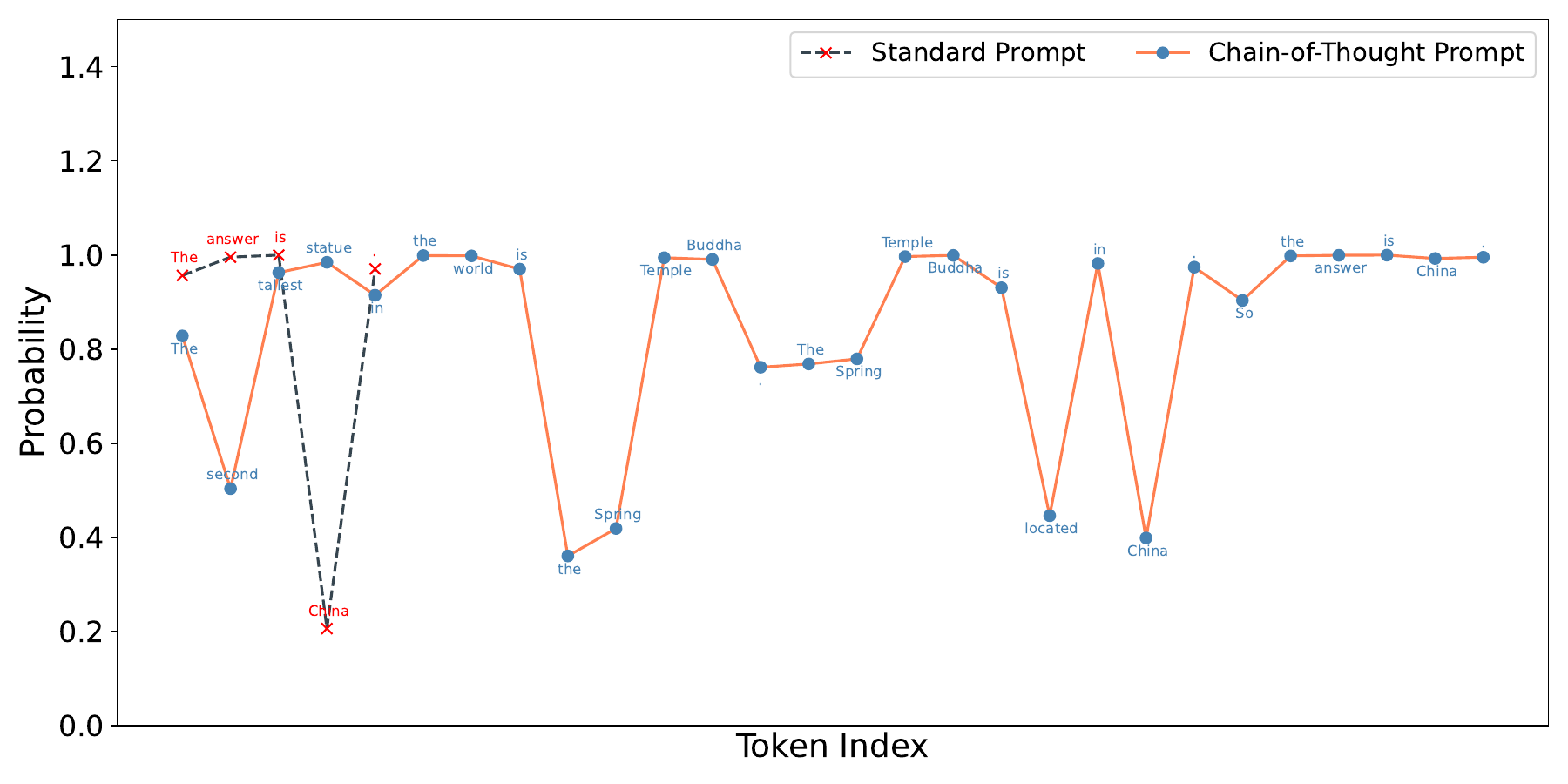}
        \caption{\textit{bamboogle}}
    \end{subfigure}
    \begin{subfigure}{0.3\linewidth}
        \includegraphics[width=\linewidth]
        {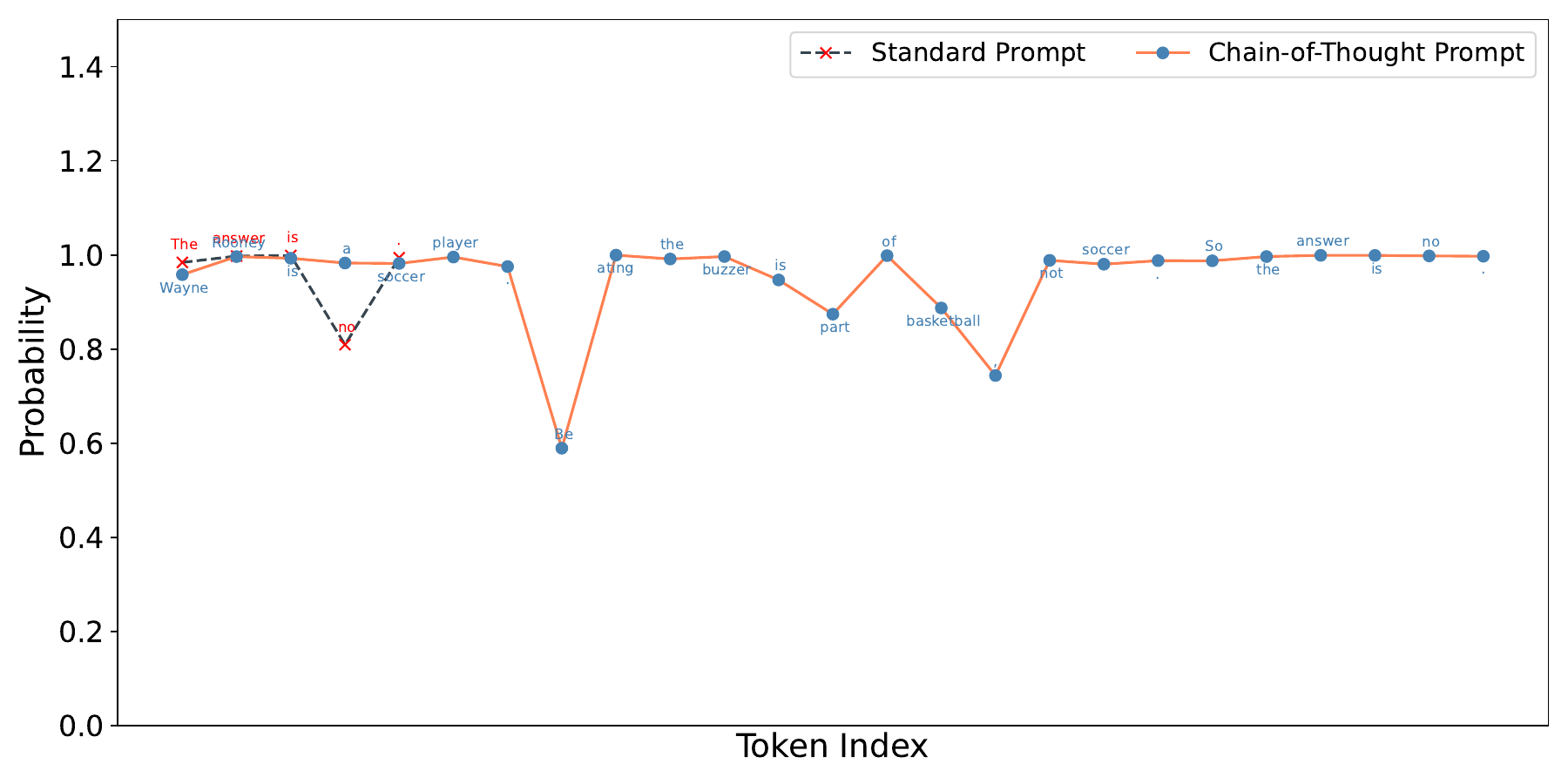}
        \caption{\textit{sports}}
    \end{subfigure}
    \begin{subfigure}{0.3\linewidth}
        \includegraphics[width=\linewidth]
        {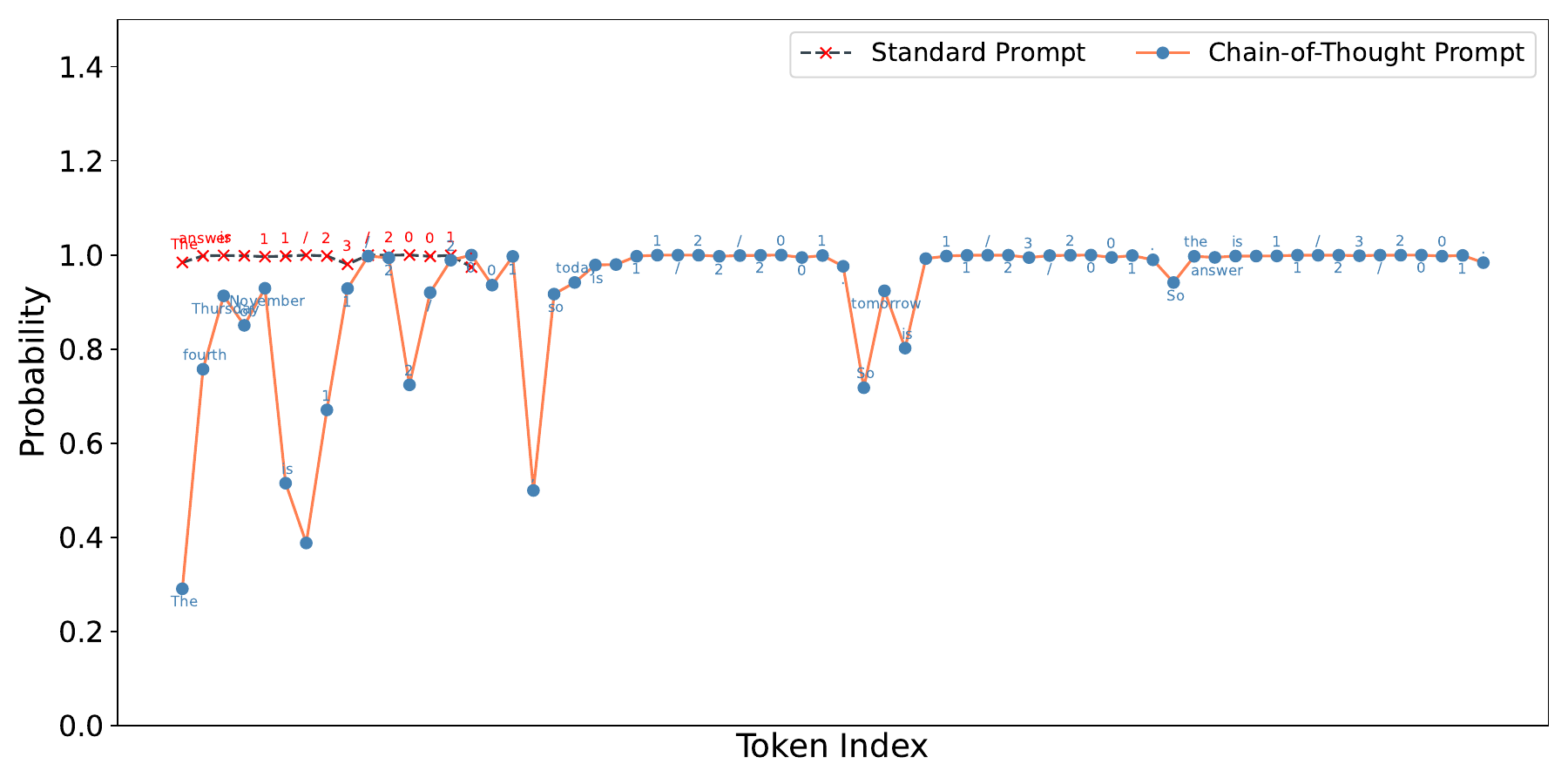}
        \caption{\textit{date}}
    \end{subfigure}
    \begin{subfigure}{0.3\linewidth}
        \includegraphics[width=\linewidth]
        {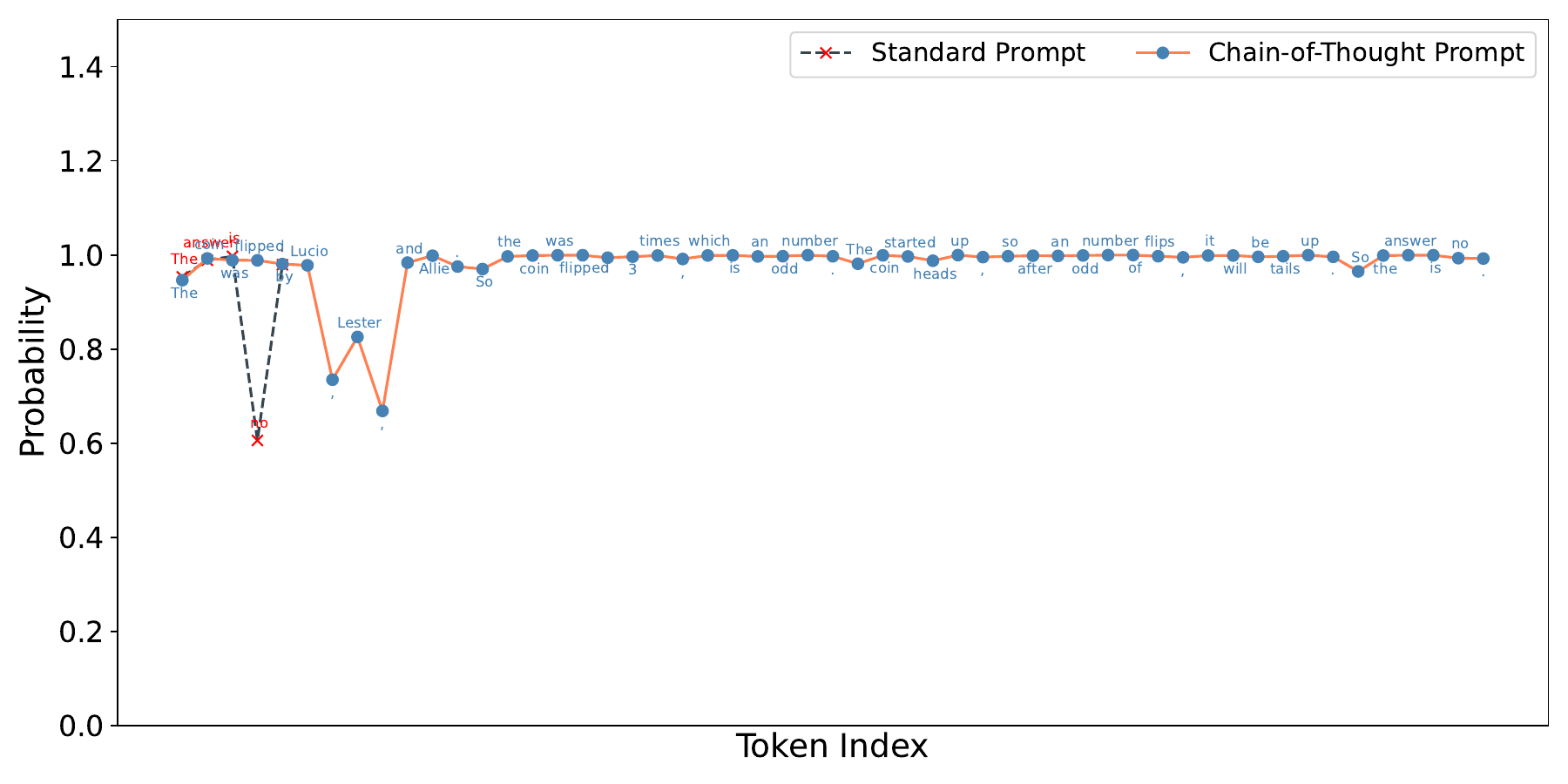}
        \caption{\textit{coin flip}}
    \end{subfigure}
     \begin{subfigure}{0.3\linewidth}
        \includegraphics[width=\linewidth]
        {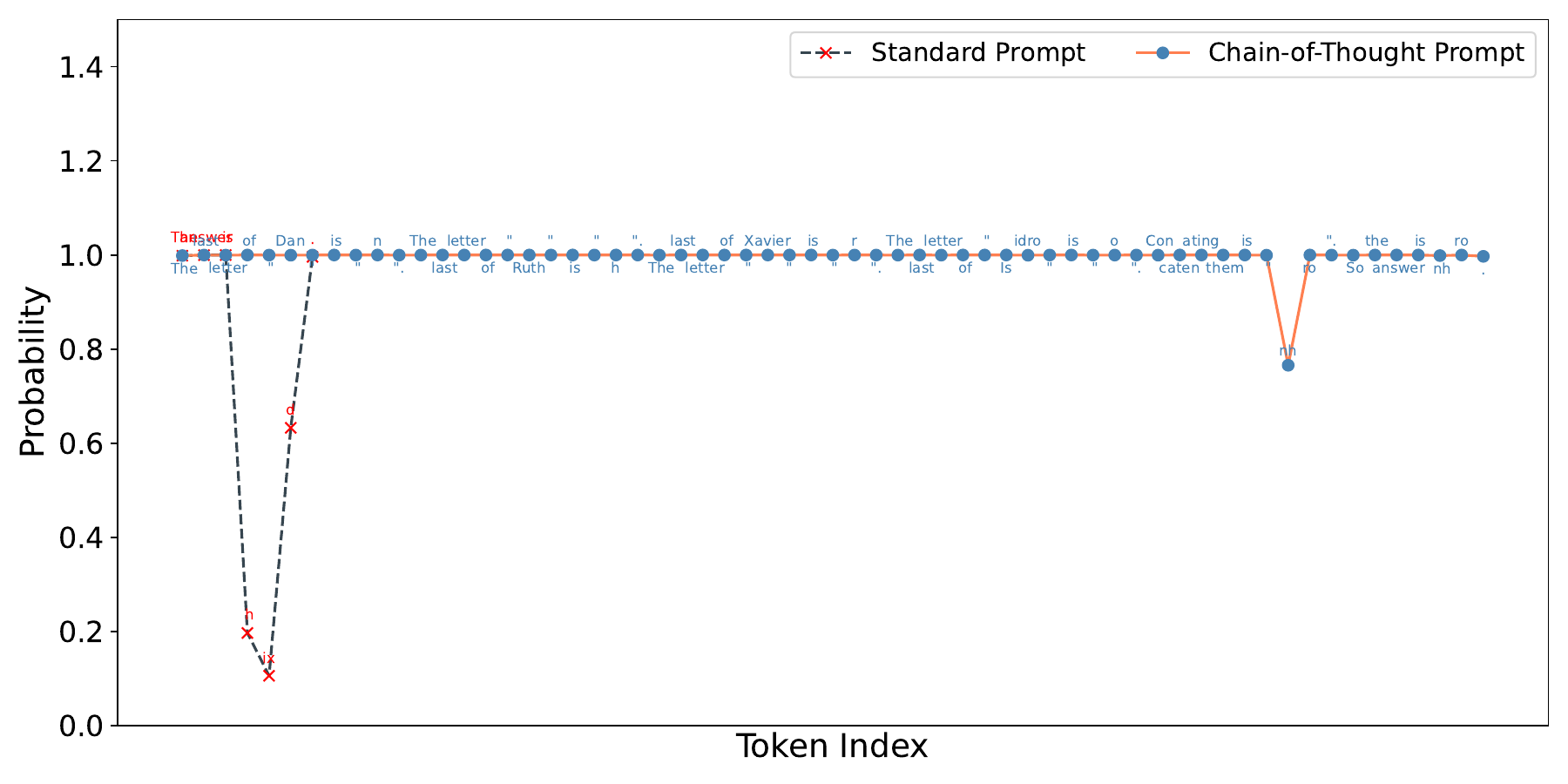}
        \caption{\textit{Last Letter Concatenation}}
    \end{subfigure}
\caption{Probability value of each generated token (the results of Gemma2-27b is shown).}
\label{fig:question2_logits_value_27b}
\end{figure*}

We conducted experiments on all test datasets across various models, obtaining the token sequences and their corresponding generation probability sequences under both CoT and Standard prompting methods. Figure \ref{fig:question2_logits_value_2b} presents the test results for the Gemma2-2b model, Figure \ref{fig:question2_logits_value_13b} shows the results for the LLaMA2-13b model, and Figure \ref{fig:question2_logits_value_27b} illustrates the results for the Gemma2-27b model. The experimental results indicate that the probability trends for token generation under both CoT and Standard prompts are consistent across models of different sizes and series.
\clearpage
\section{Kernel Density Estimate Plots}\label{appendix:density}
\begin{figure*}[ht]
\centering
    \includegraphics[width=0.24\linewidth]{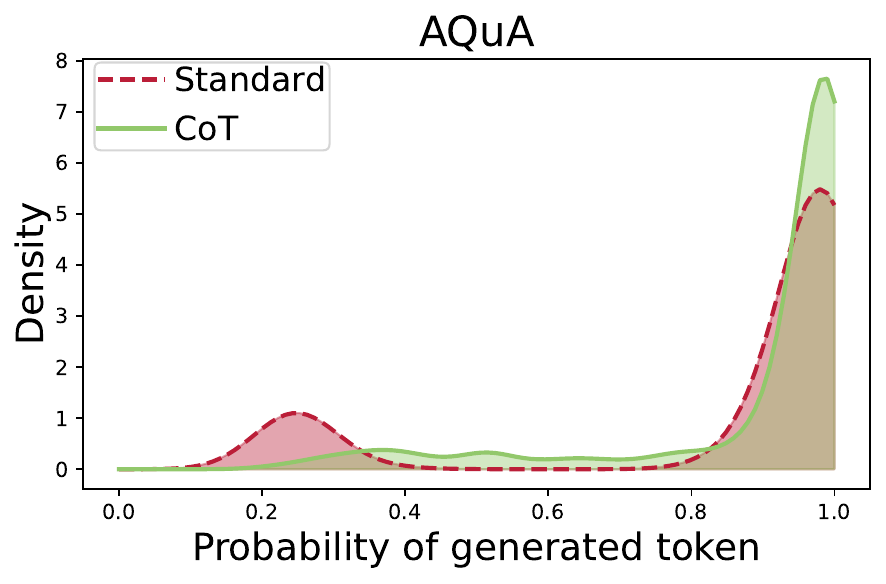}
    \includegraphics[width=0.24\linewidth]{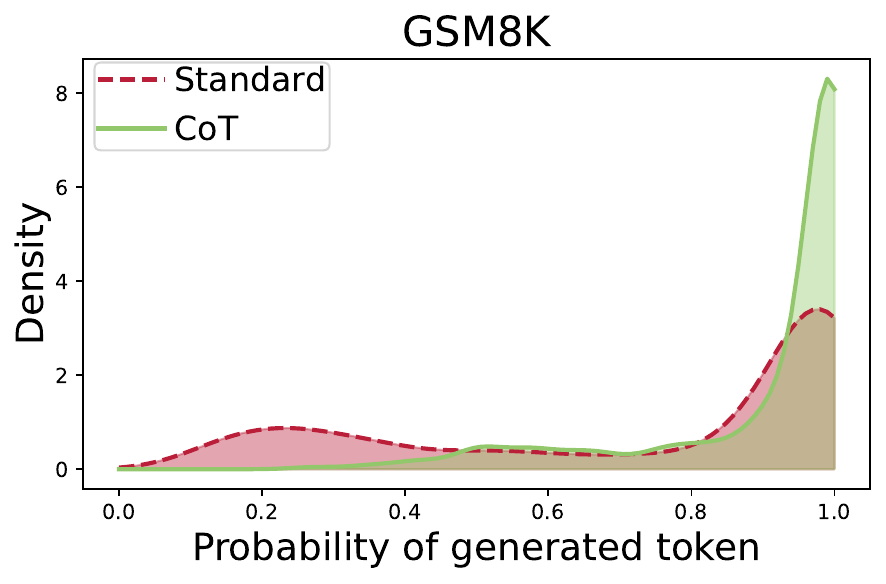}
    \includegraphics[width=0.24\linewidth]{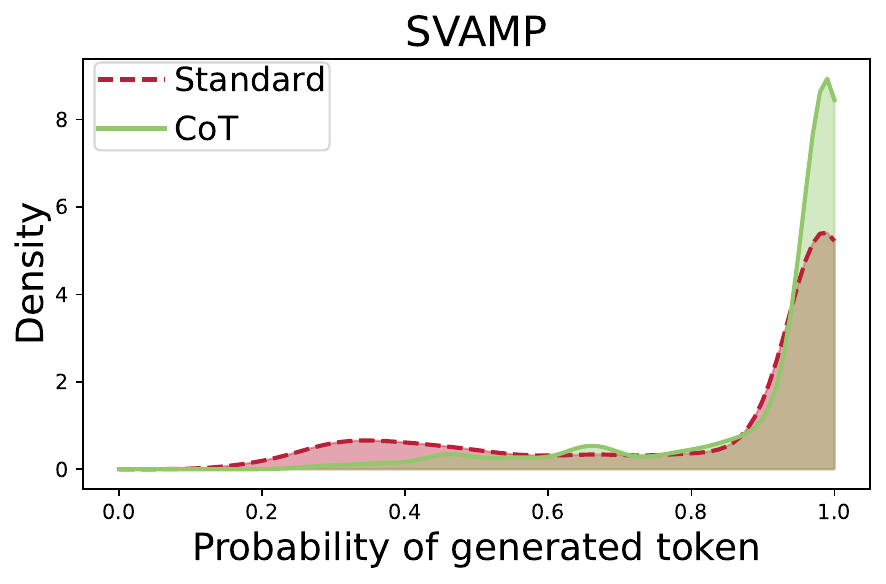}
    \includegraphics[width=0.24\linewidth]{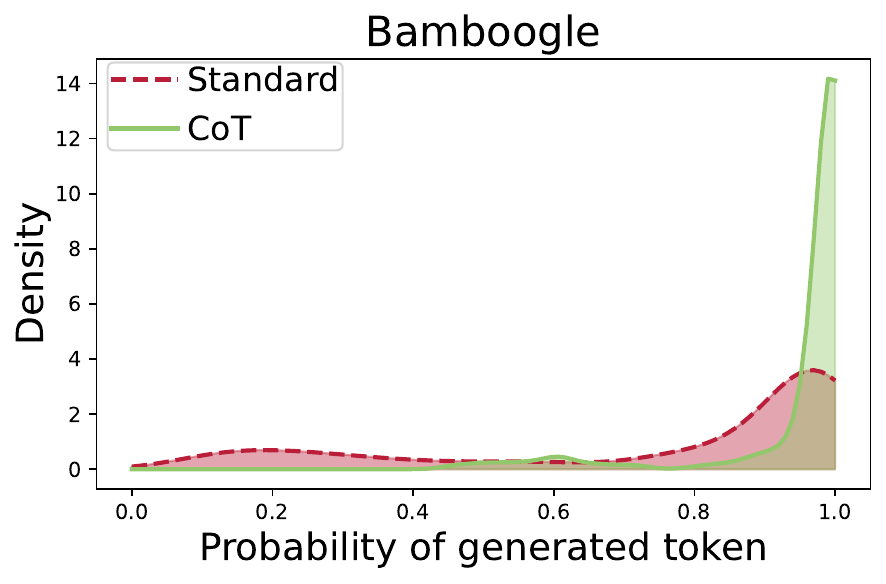}
    \includegraphics[width=0.24\linewidth]{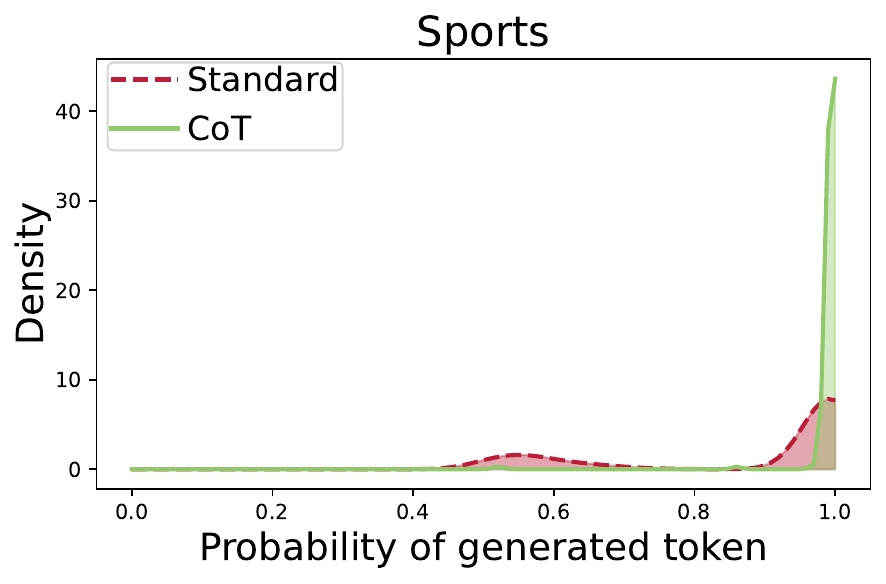}
    \includegraphics[width=0.24\linewidth]{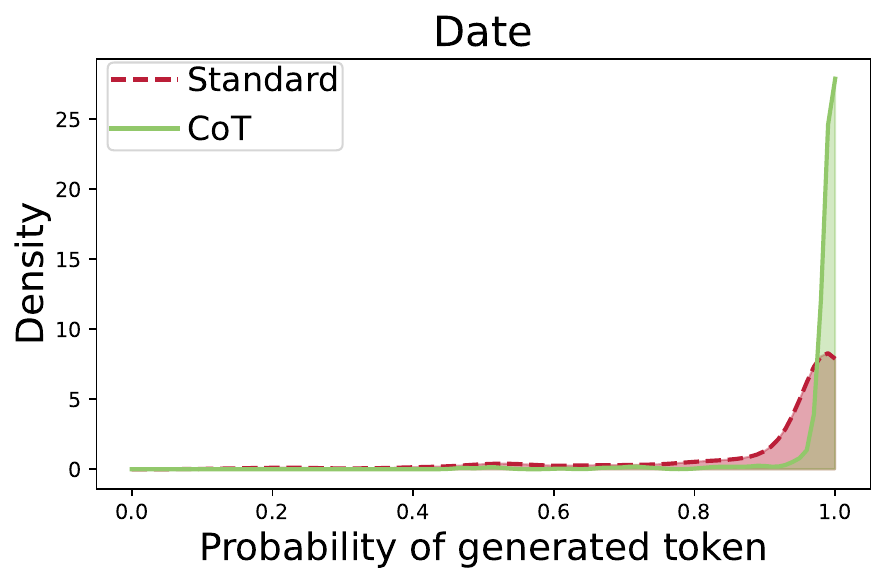}
     \includegraphics[width=0.24\linewidth]{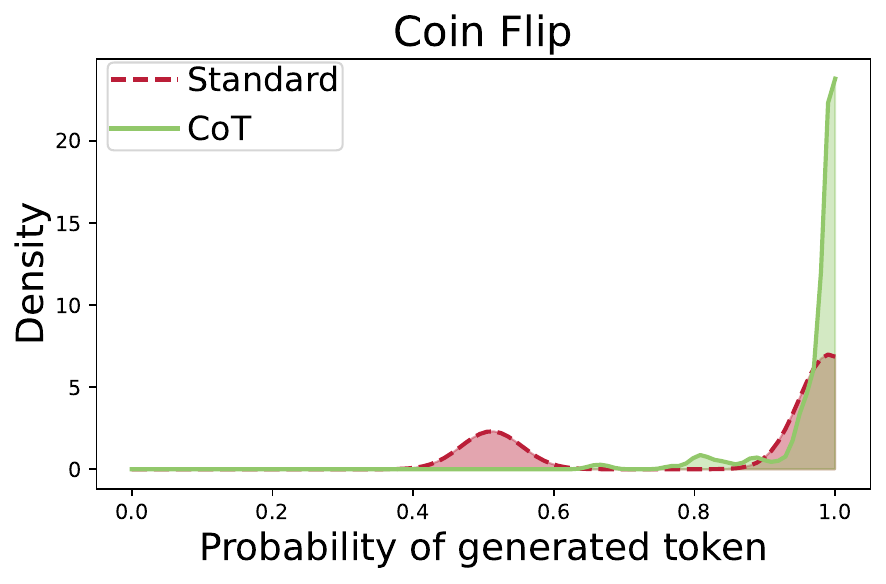}
    \includegraphics[width=0.24\linewidth]{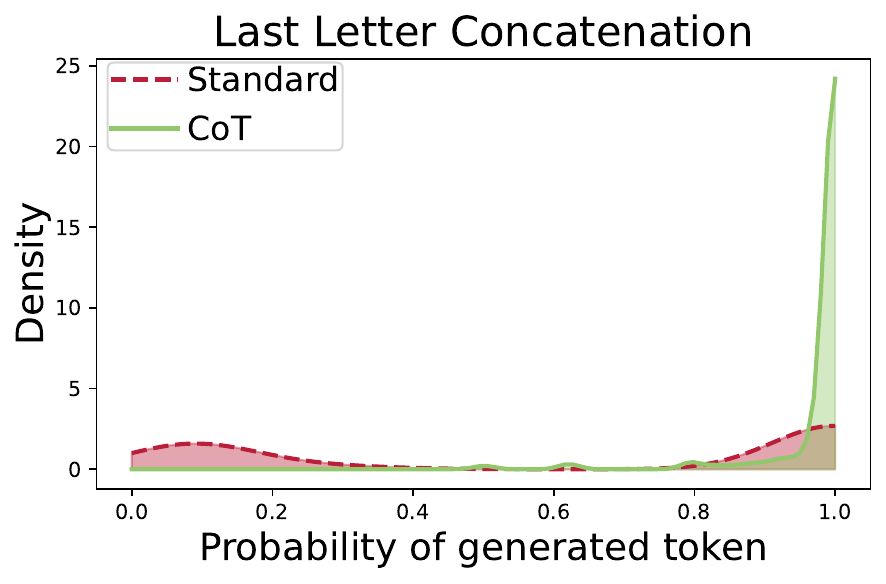}
    \caption{Kernal density for probabilities of sequence ``the answer is...'' (the results of Gemma2-2b is shown).}
    \label{fig:question2_logits_value_kernel_2b}
\end{figure*}

\begin{figure*}[ht]
\centering
    \includegraphics[width=0.24\linewidth]{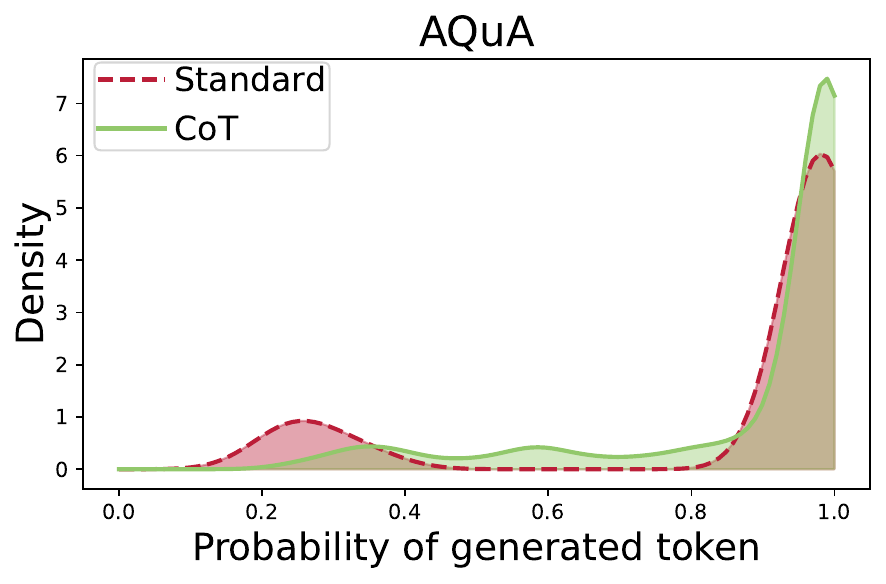}
    \includegraphics[width=0.24\linewidth]{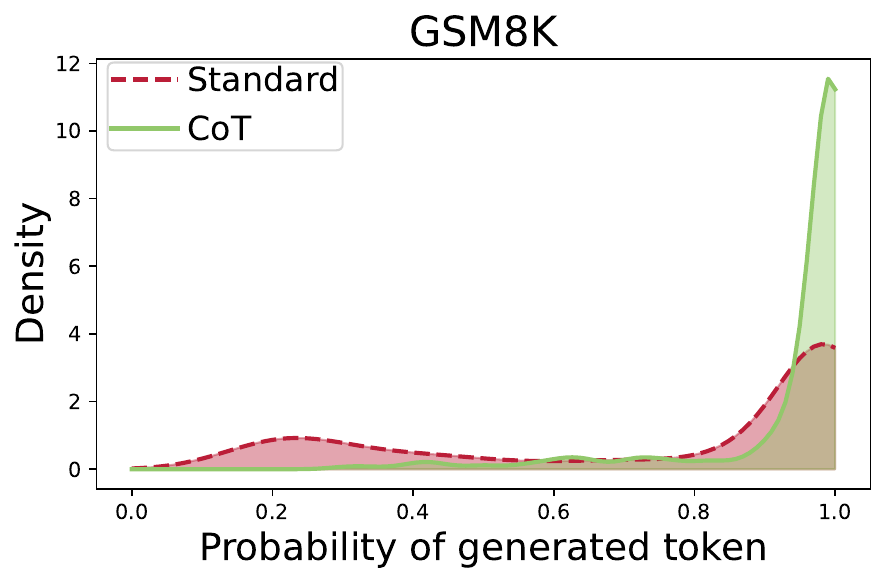}
    \includegraphics[width=0.24\linewidth]{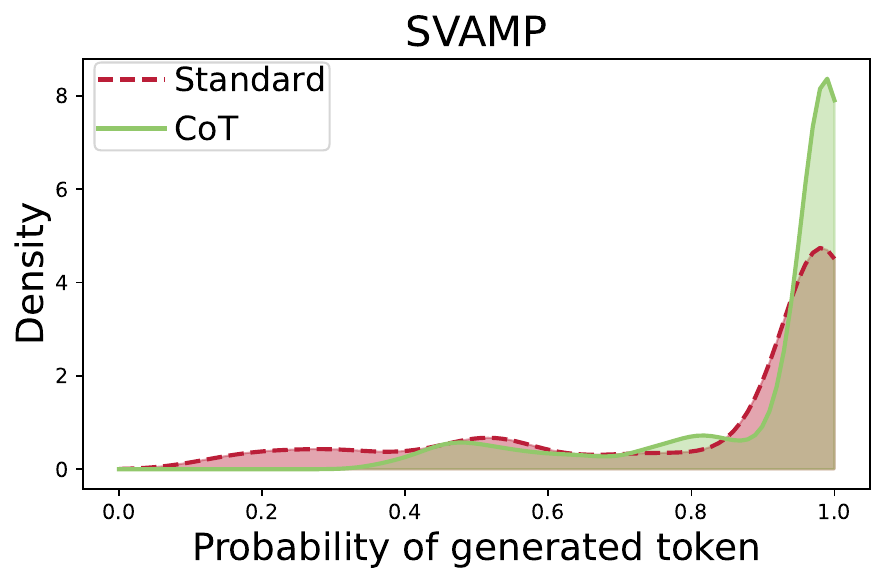}
    \includegraphics[width=0.24\linewidth]{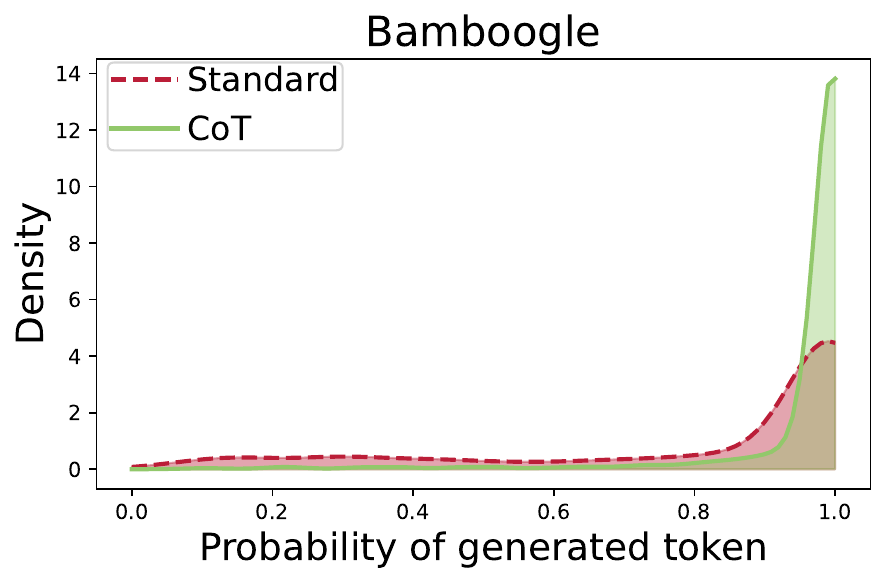}
    \includegraphics[width=0.24\linewidth]{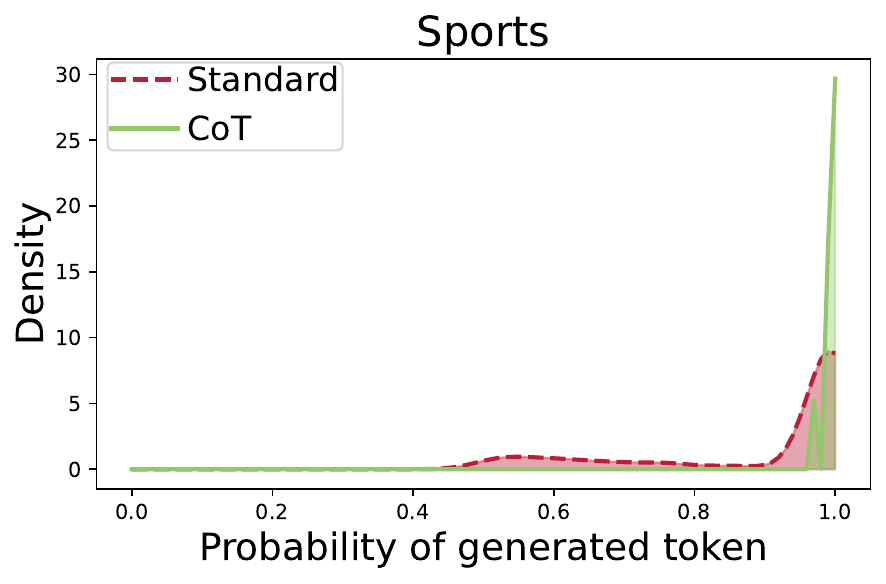}
    \includegraphics[width=0.24\linewidth]{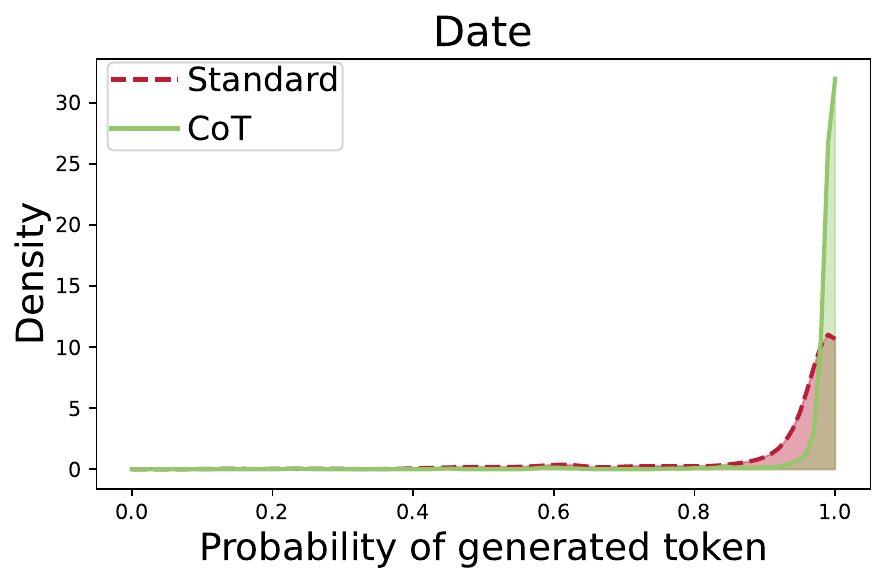}
     \includegraphics[width=0.24\linewidth]{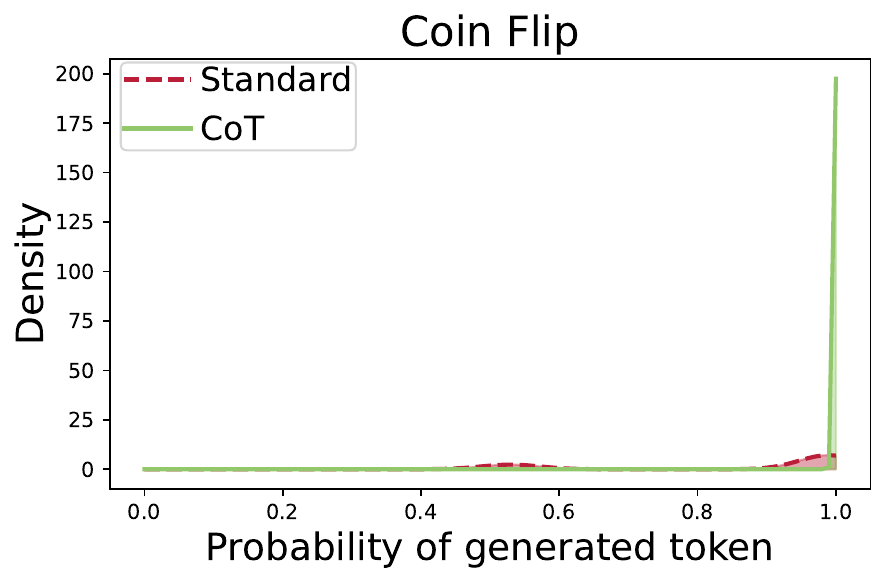}
    \includegraphics[width=0.24\linewidth]{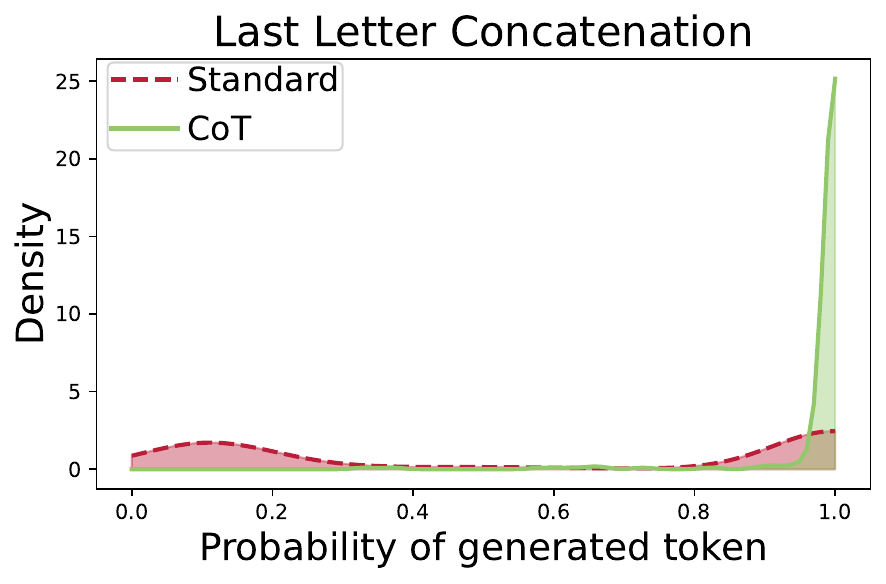}
    \caption{Kernal density for probabilities of sequence ``the answer is...'' (the results of LLaMA2-13b is shown).}
    \label{fig:question2_logits_value_kernel_13b}
\end{figure*}

\begin{figure*}[ht]
\centering
    \includegraphics[width=0.24\linewidth]{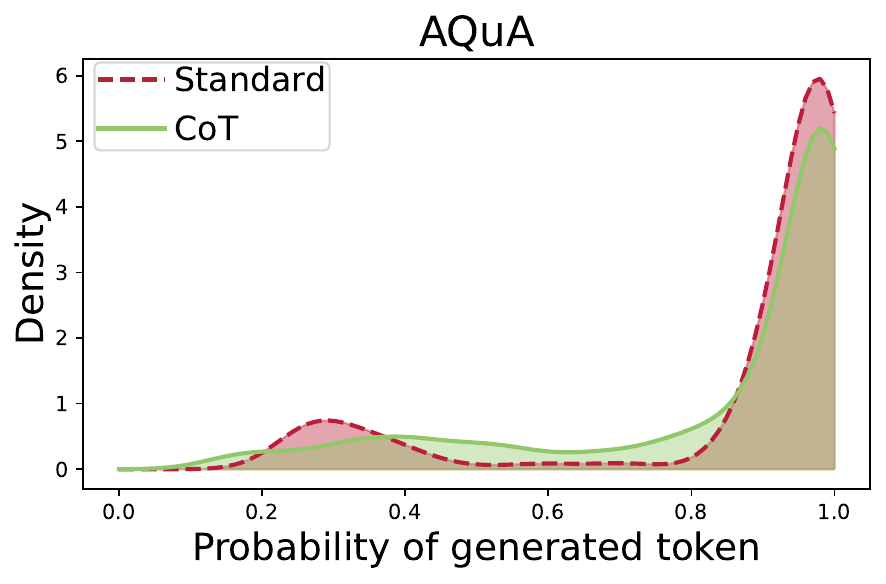}
    \includegraphics[width=0.24\linewidth]{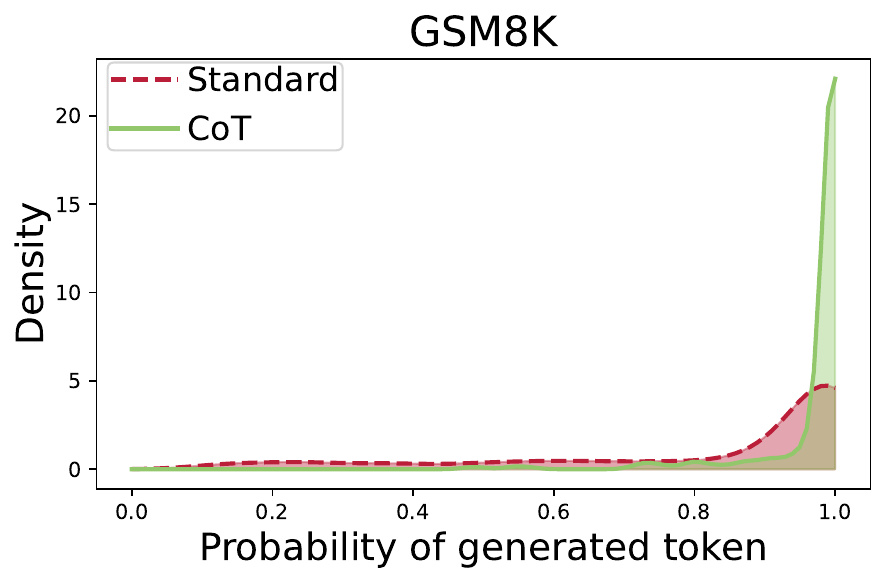}
    \includegraphics[width=0.24\linewidth]{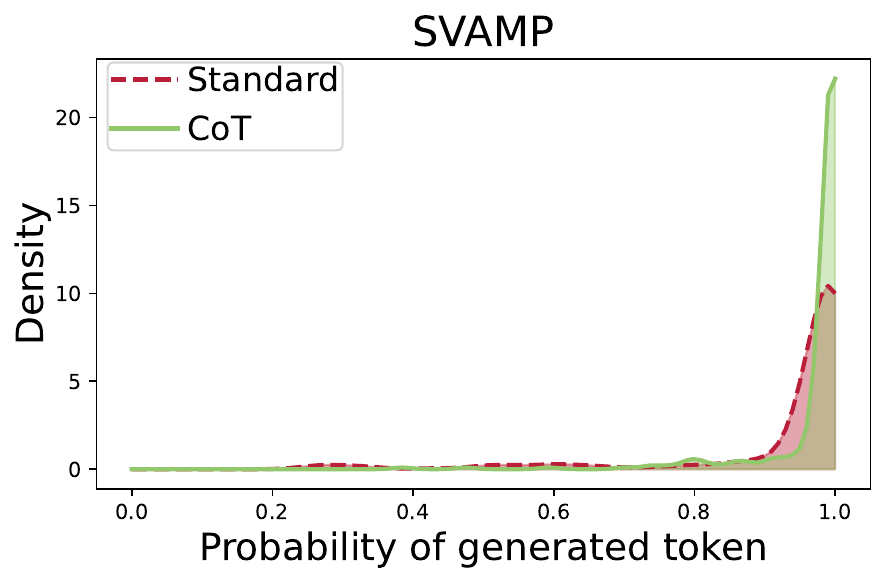}
    \includegraphics[width=0.24\linewidth]{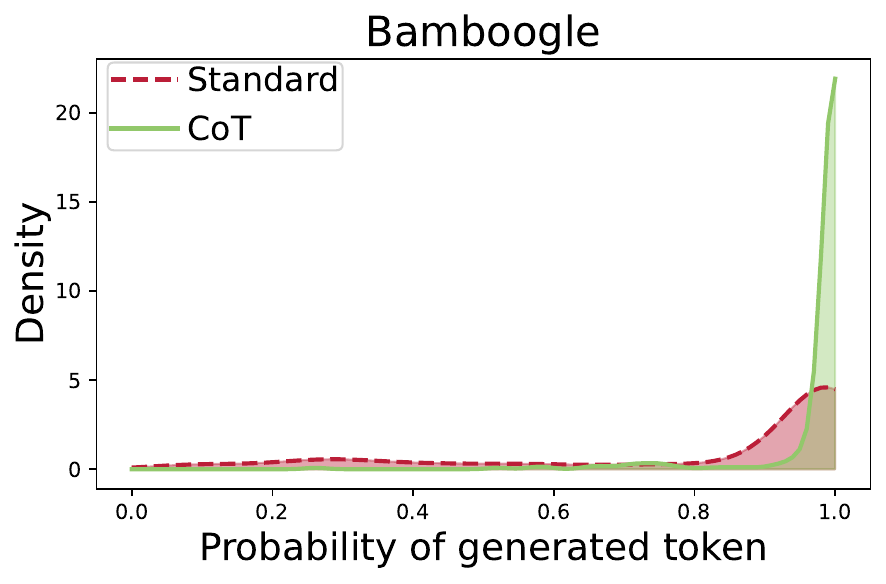}
    \includegraphics[width=0.24\linewidth]{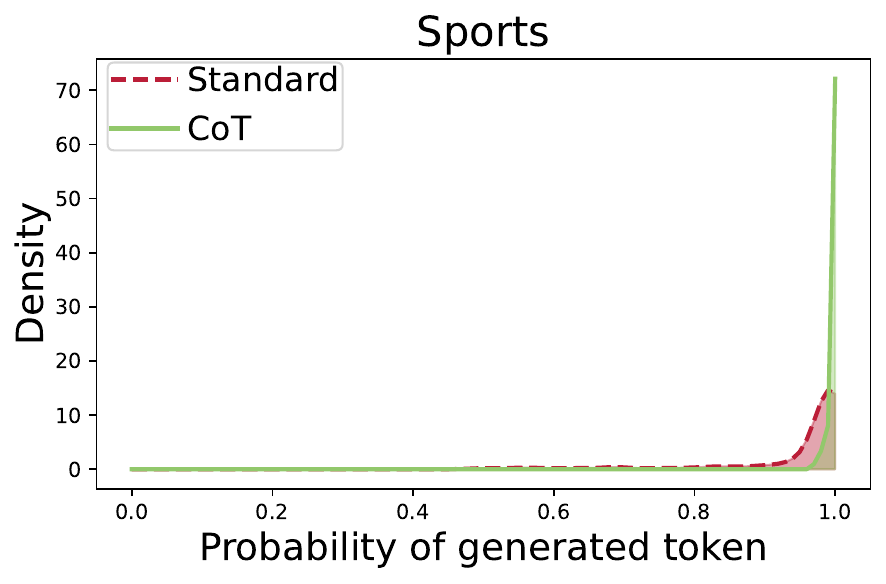}
    \includegraphics[width=0.24\linewidth]{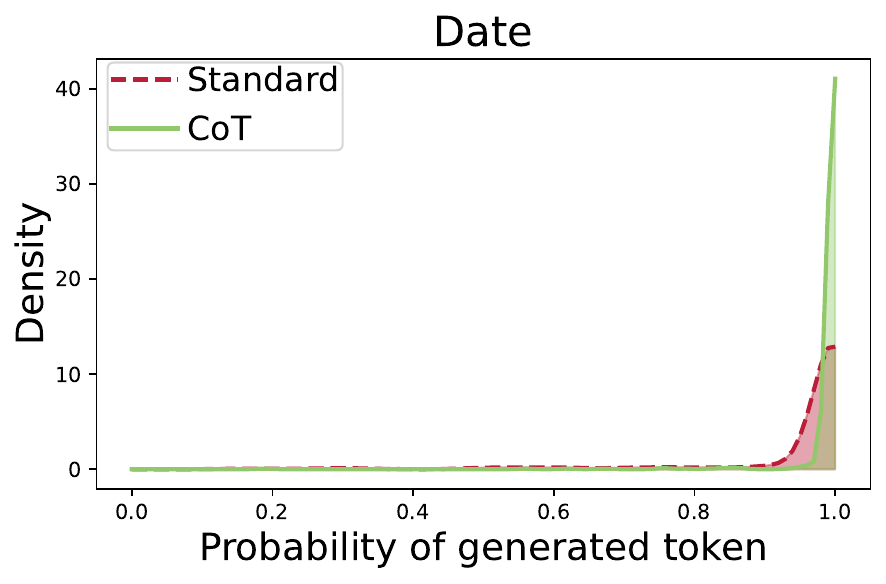}
     \includegraphics[width=0.24\linewidth]{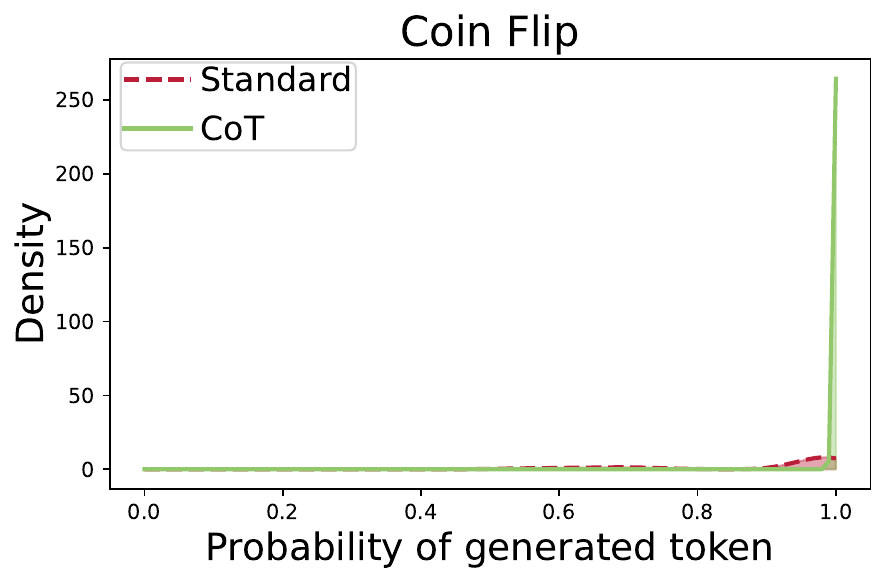}
    \includegraphics[width=0.24\linewidth]{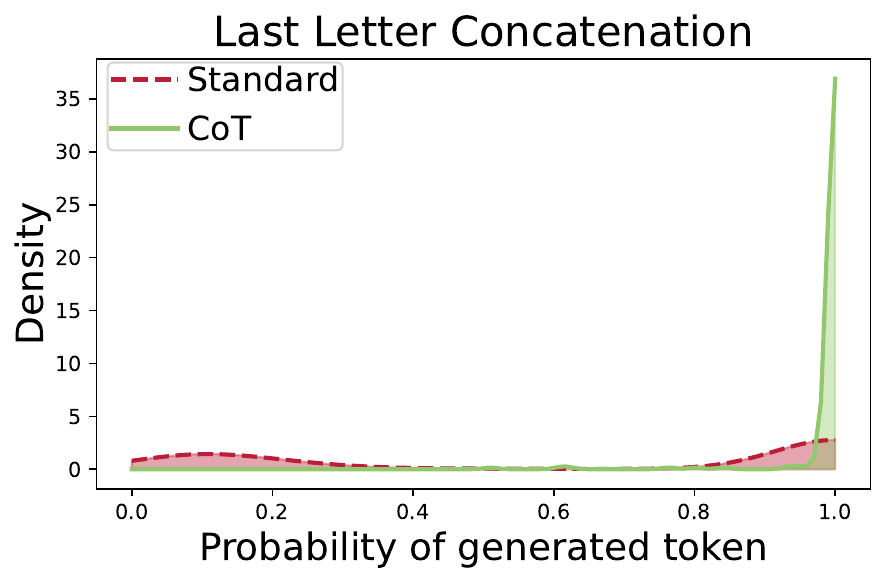}
    \caption{Kernal density for probabilities of sequence ``the answer is...'' (the results of Gemma2-27b is shown).}
    \label{fig:question2_logits_value_kernel_27b}
\end{figure*}

Kernel Density Estimation (KDE) is a non-parametric way to estimate the probability density function of a dataset. It smooths out the observed data using a kernel function to give a continuous estimation of the underlying distribution. The kernel function is essentially a smooth, localized function, like a Gaussian, which is summed over each data point to produce the density estimate. Equation \ref{eq:kde} is the mathematical Formulation of KDE.
\begin{equation}
\label{eq:kde}
\hat{f}(x) = \frac{1}{n h} \sum_{i=1}^{n} K\left( \frac{x - x_i}{h} \right)
\end{equation}
$\hat{f}(x)$ is the estimated density at point $x$, $n$  is the number of data points, $h$ is the bandwidth, a smoothing parameter, $x_i$ represents the observed data points, $K(\cdot)$ is a kernel function (e.g., Gaussian kernel). In this paper, we use the Gaussian KDE(Equation \ref{eq:gaussian}).
\begin{equation}
\label{eq:gaussian}
K(x) = \frac{1}{\sqrt{2\pi}} e^{-\frac{x^2}{2}}
\end{equation}
Since our data lies within a bounded range (0 to 1), we use the standard Gaussian kernel but adapting it for this boundary condition by restricting the domain during plotting. We conducted experiments on four models: Gemma2-2b, Gemma2-9b, LLaMA2-13b, and Gemma2-27b. The experimental results of Gemma2-2b are shown in Figure \ref{fig:question2_logits_value_kernel_2b}, the experimental results of Gemma2-9b are shown in Figure \ref{fig:question2_horizontal_2}. the experimental results of LLaMA2-13b are shown in Figure \ref{fig:question2_logits_value_kernel_13b}, and the experimental results of Gemma2-27b are shown in Figure \ref{fig:question2_logits_value_kernel_27b}.
\clearpage
\section{Entropy of the Answer Space Probability Distribution}\label{appendix:entropy}
\begin{figure*}[ht]
\centering
    \begin{subfigure}{0.32\linewidth}
        \includegraphics[width=\linewidth]{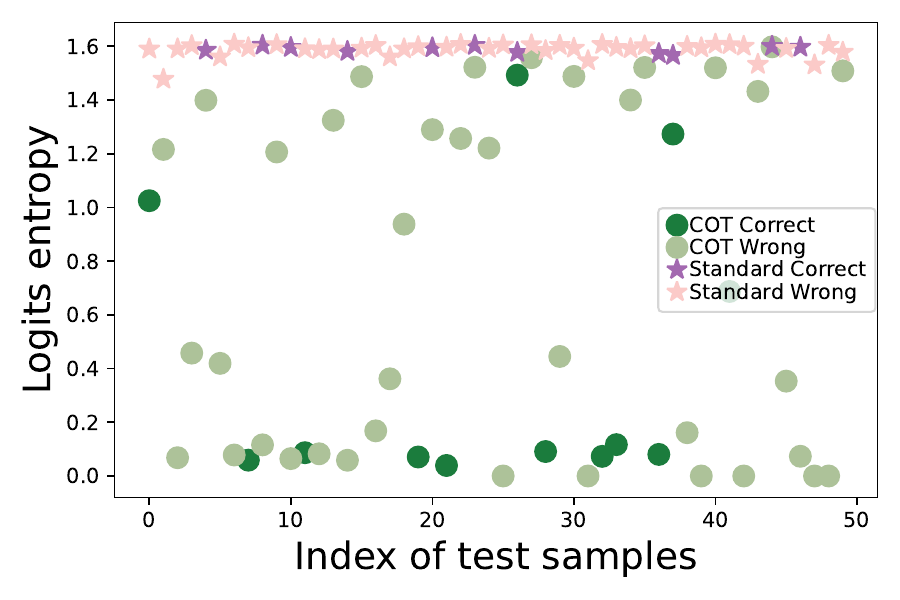}
        \caption{\textit{AQuA}}
    \end{subfigure}
    \begin{subfigure}{0.32\linewidth}
        \includegraphics[width=\linewidth]{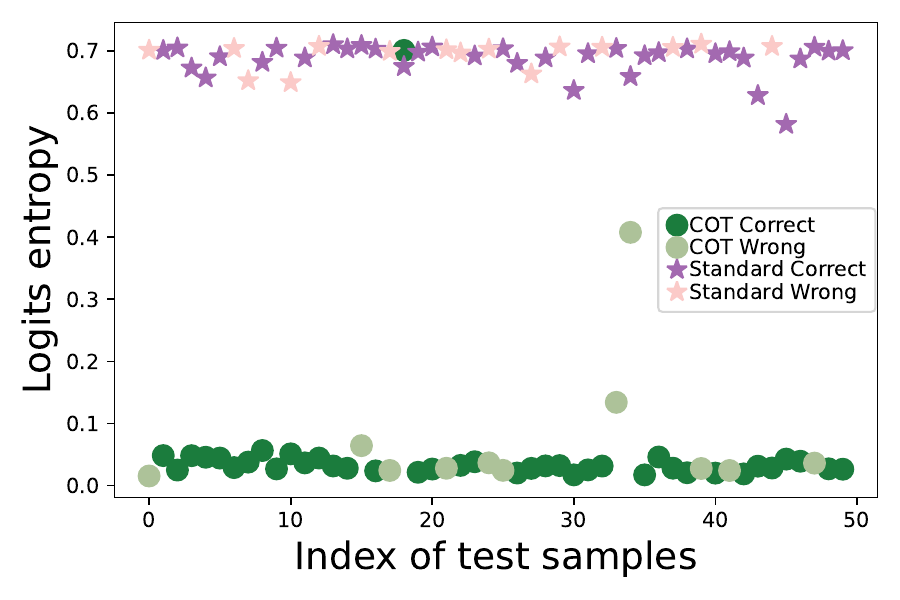}
        \caption{\textit{Sports}}
    \end{subfigure}
    \begin{subfigure}{0.32\linewidth}
        \includegraphics[width=\linewidth]{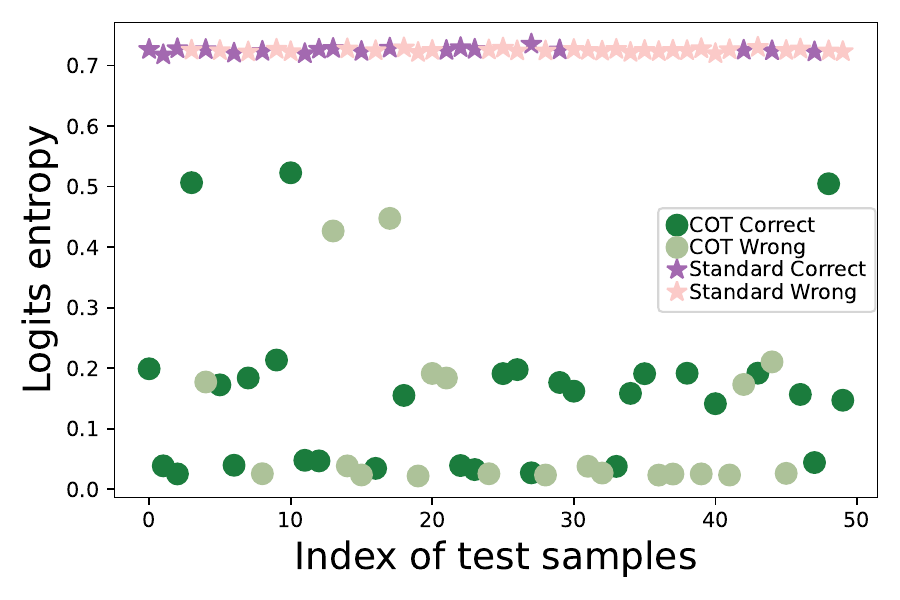}
        \caption{\textit{Coin Flip}}
    \end{subfigure}
\caption{Probability distribution entropy of answer generation step (results of Gemma2-2b is shown).}
\label{fig:question2_vertical_entropy_2b}
\end{figure*}

\begin{figure*}[ht]
\centering
    \begin{subfigure}{0.32\linewidth}
        \includegraphics[width=\linewidth]{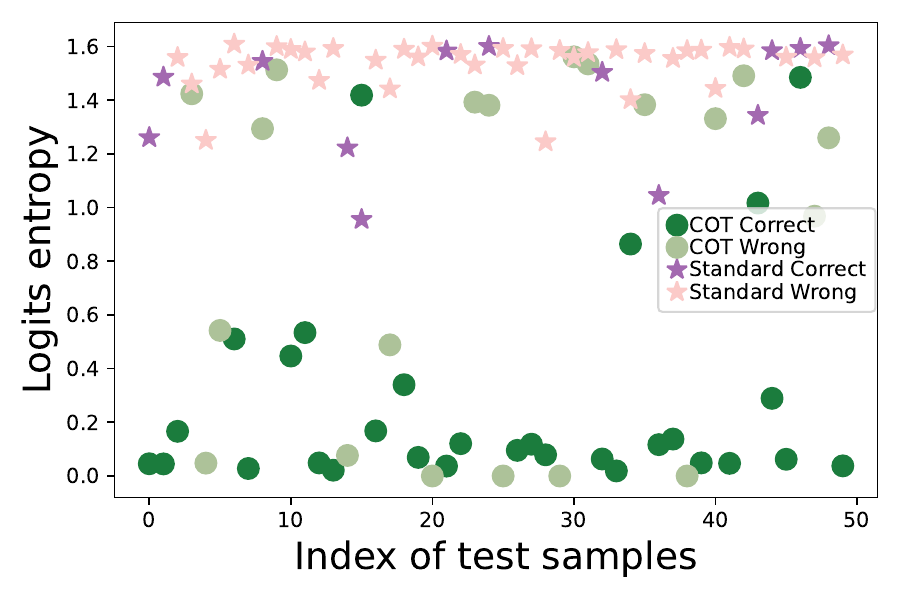}
        \caption{\textit{AQuA}}
    \end{subfigure}
    \begin{subfigure}{0.32\linewidth}
        \includegraphics[width=\linewidth]{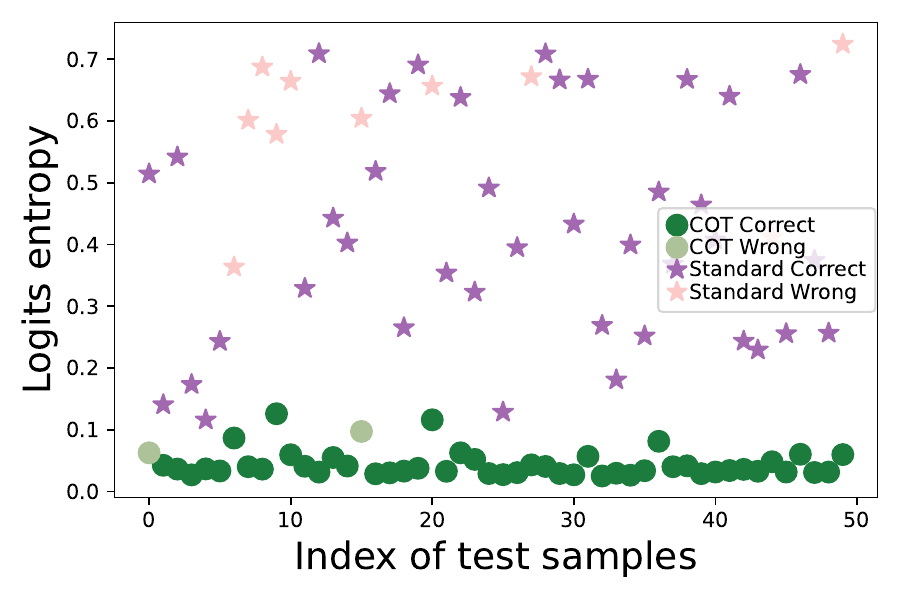}
        \caption{\textit{Sports}}
    \end{subfigure}
    \begin{subfigure}{0.32\linewidth}
        \includegraphics[width=\linewidth]{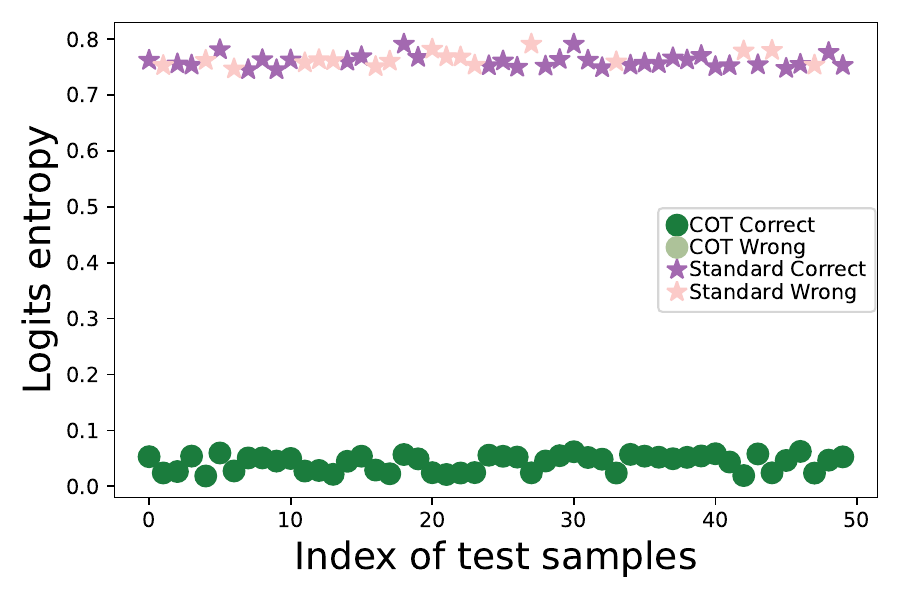}
        \caption{\textit{Coin Flip}}
    \end{subfigure}
\caption{Probability distribution entropy of answer generation step (the results of Gemma2-9b is shown).}
\label{question2_vertical_entropy_9b}
\end{figure*}

\begin{figure*}[ht]
\centering
    \begin{subfigure}{0.32\linewidth}
        \includegraphics[width=\linewidth]{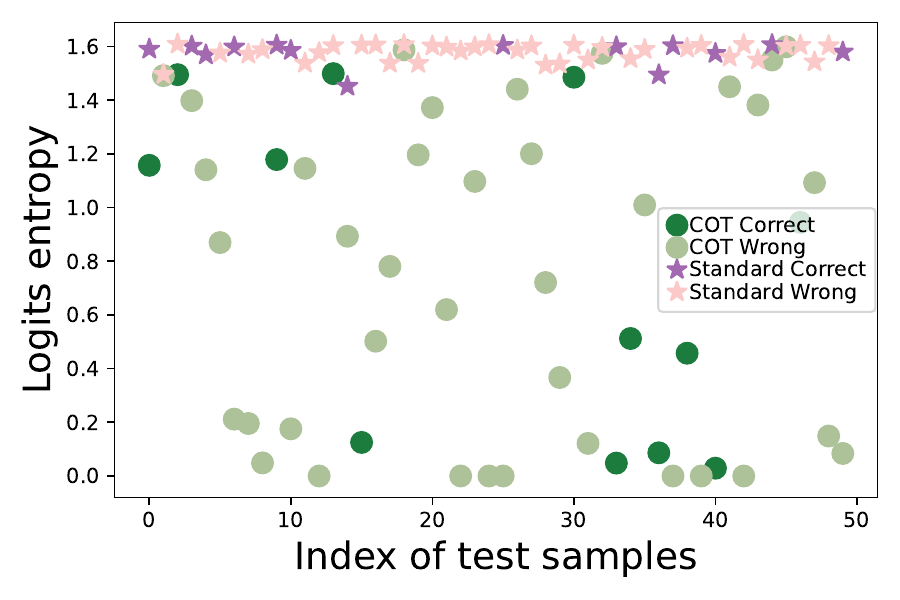}
        \caption{\textit{AQuA}}
    \end{subfigure}
    \begin{subfigure}{0.32\linewidth}
        \includegraphics[width=\linewidth]{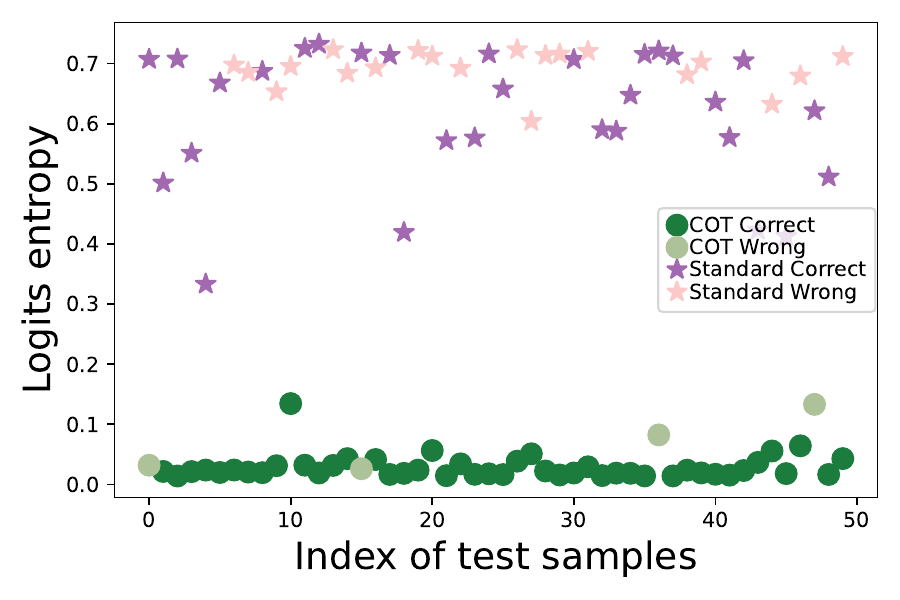}
        \caption{\textit{Sports}}
    \end{subfigure}
    \begin{subfigure}{0.32\linewidth}
        \includegraphics[width=\linewidth]{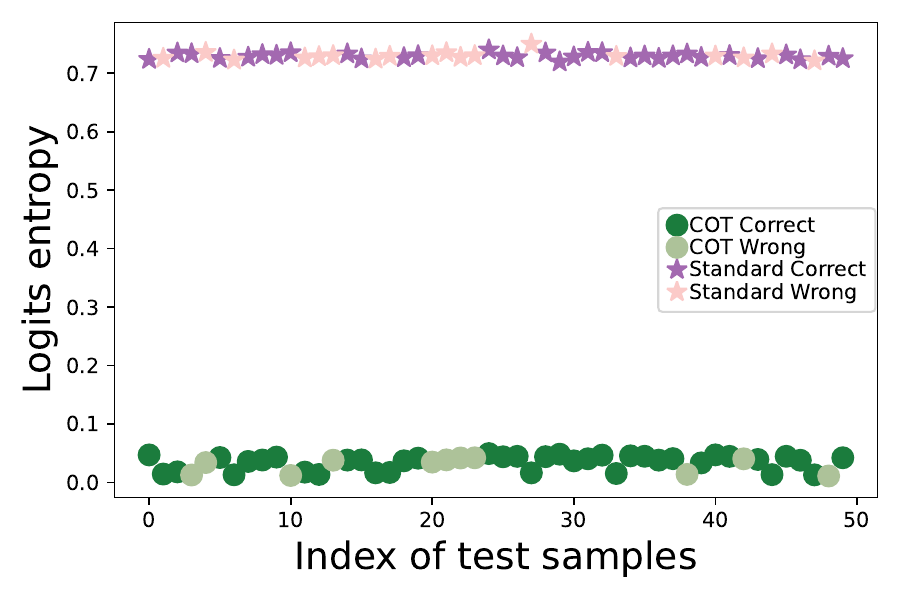}
        \caption{\textit{Coin Flip}}
    \end{subfigure}
\caption{Probability distribution entropy of answer generation step (the results of LLaMA2-13b is shown).}
\label{fig:question2_vertical_entropy_13b}
\end{figure*}

To investigate the model's confidence when predicting answers, we examined the probability distribution over candidate answers for the token generated at the answer prediction step. We focused on datasets AQuA, Sports, and Coin Flip, where the number of answer tokens is 1, and the answer space is finite. For datasets with multiple answer tokens and open-domain questions, it is not feasible to retrieve all potential answers and their corresponding probabilities.
Surprisingly, for the AQuA dataset, the top five probabilities at the answer prediction step corresponded exactly to the answer space options: ``a '', ``b '', ``c '', ``d '', and ``e''. Similarly, for the Sports and Coin Flip datasets, the top two probabilities corresponded to their answer spaces, ``yes'' and ``no''.
We selected the top k probabilities at the answer prediction step (where k is the size of the answer space), normalized them, calculated the entropy, and plotted a scatter plot to compare the entropy of answer probabilities generated using the Standard and CoT methods. Figure \ref{fig:question2_vertical_entropy_2b} shows the experimental results for Gemma2-2b, Figure \ref{fig:question2_vertical_entropy_13b} for LLaMA2-13b, and Figure \ref{question2_vertical_entropy_9b} for Gemma2-9b.
\clearpage
\section{Activation}
\begin{figure*}[ht]
\centering
 \includegraphics[width=0.18\linewidth]{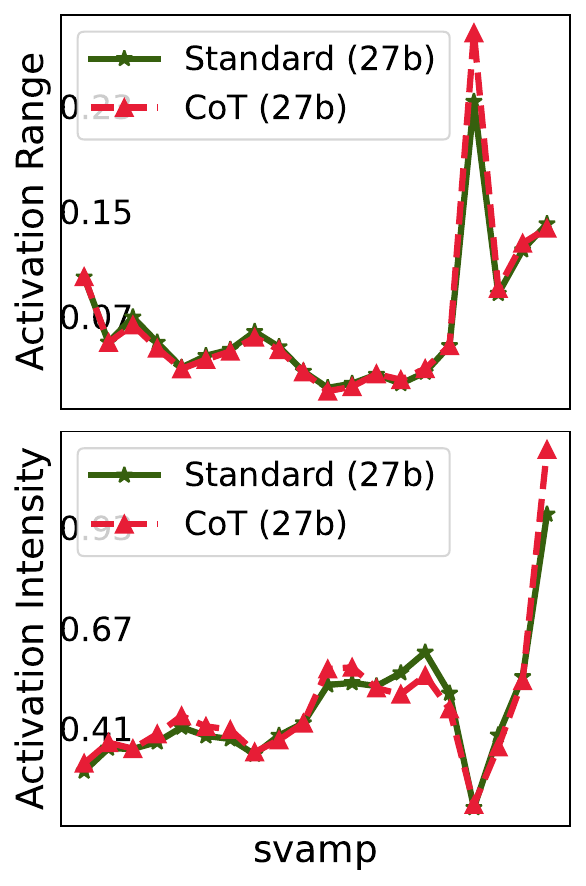}
    \includegraphics[width=0.18\linewidth]{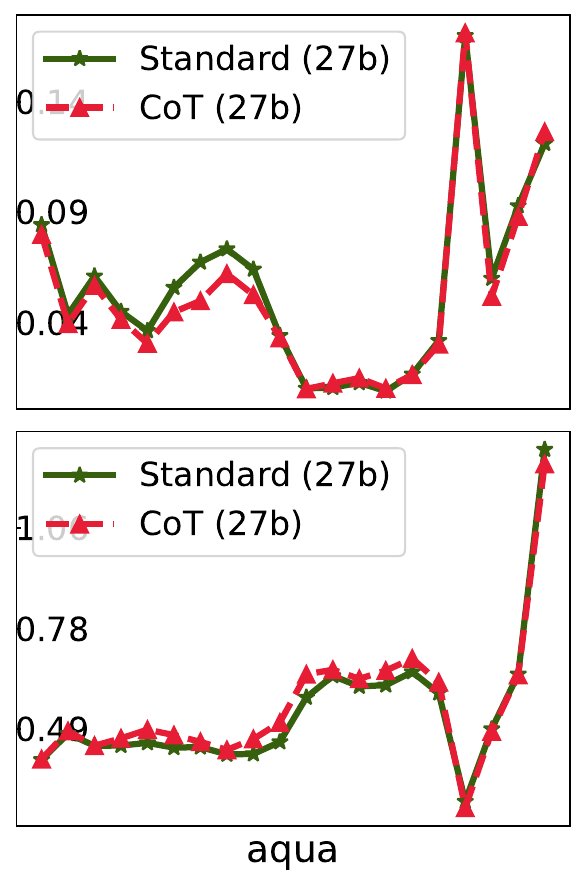}
    \includegraphics[width=0.18\linewidth]{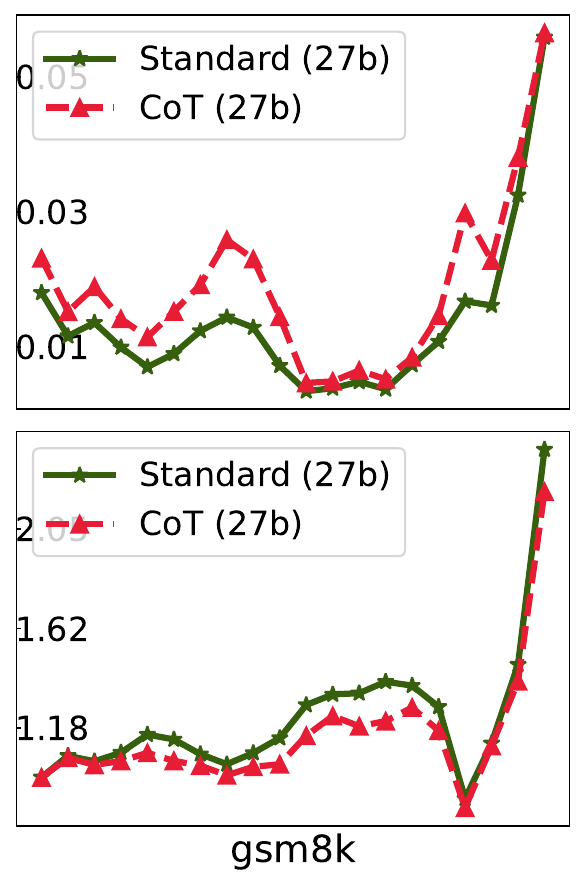}
    \includegraphics[width=0.18\linewidth]{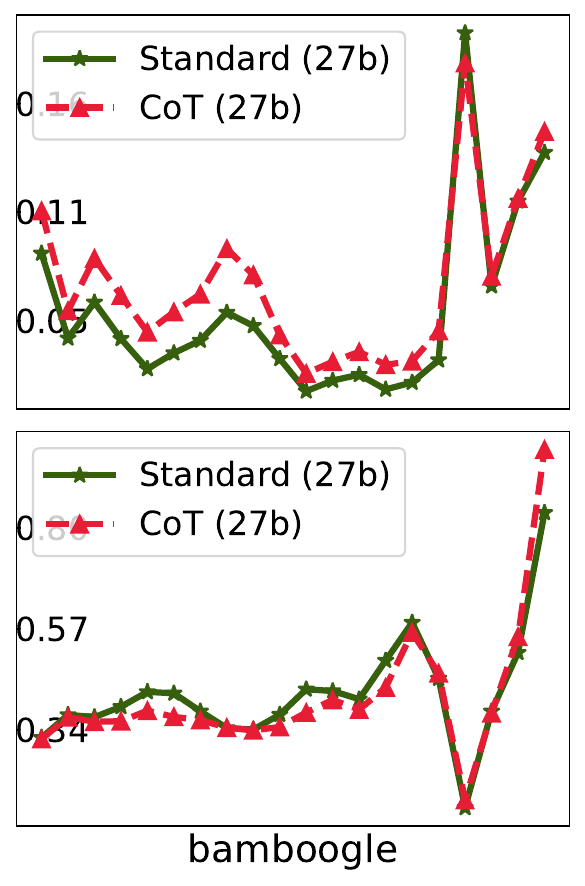}
    \includegraphics[width=0.18\linewidth]{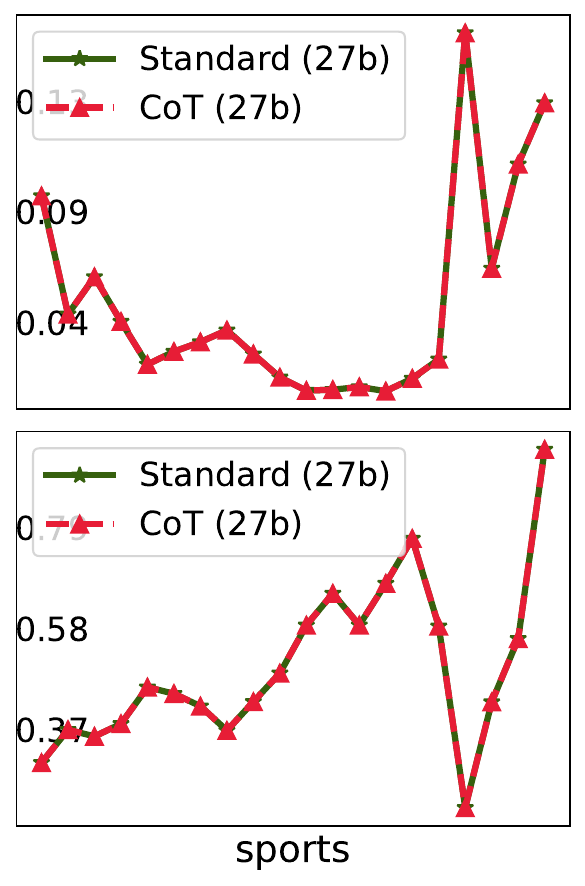}
    \includegraphics[width=0.18\linewidth]{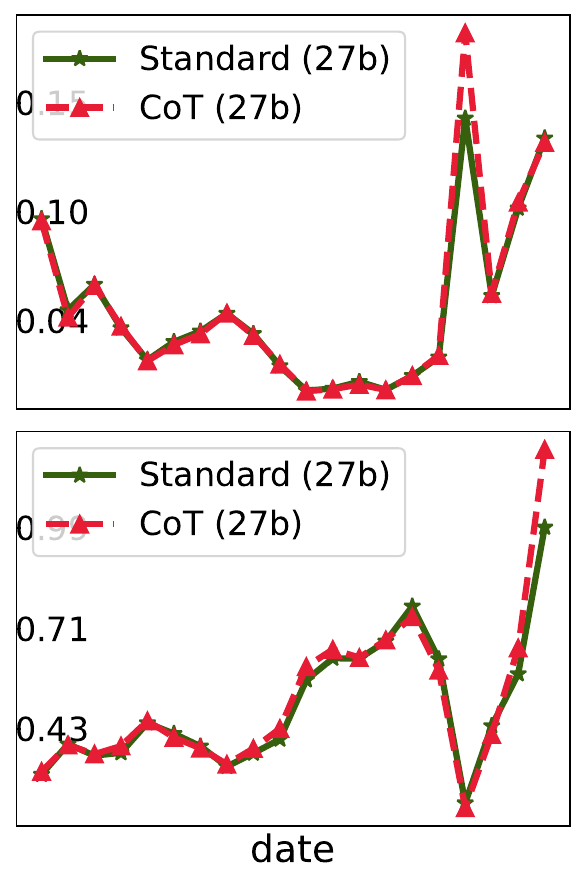}
    \includegraphics[width=0.18\linewidth]{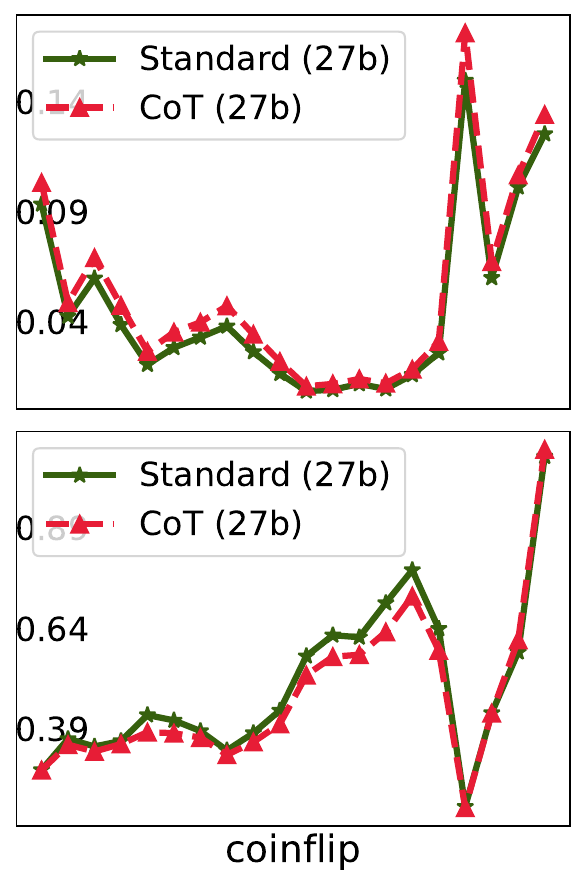}
    \includegraphics[width=0.18\linewidth]{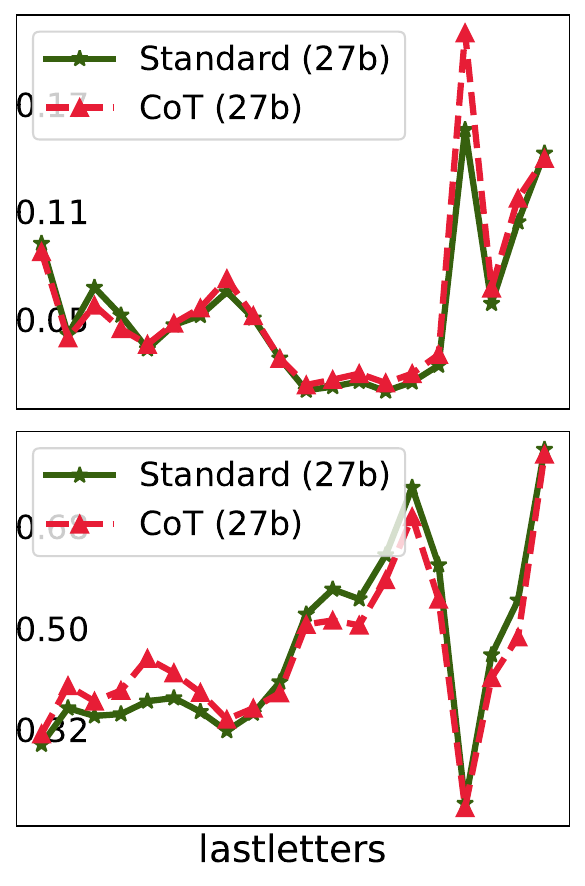}
    \\
    \includegraphics[width=0.18\linewidth]{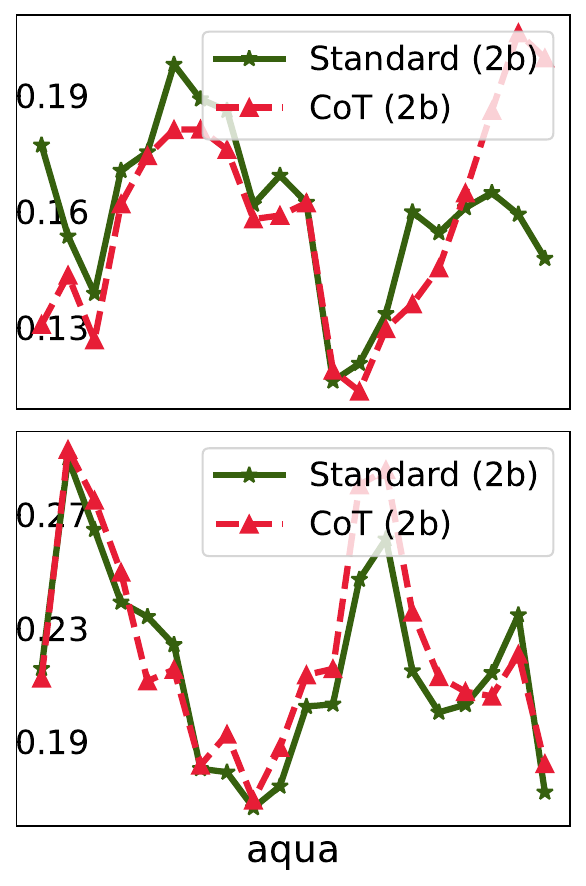}
    \includegraphics[width=0.18\linewidth]{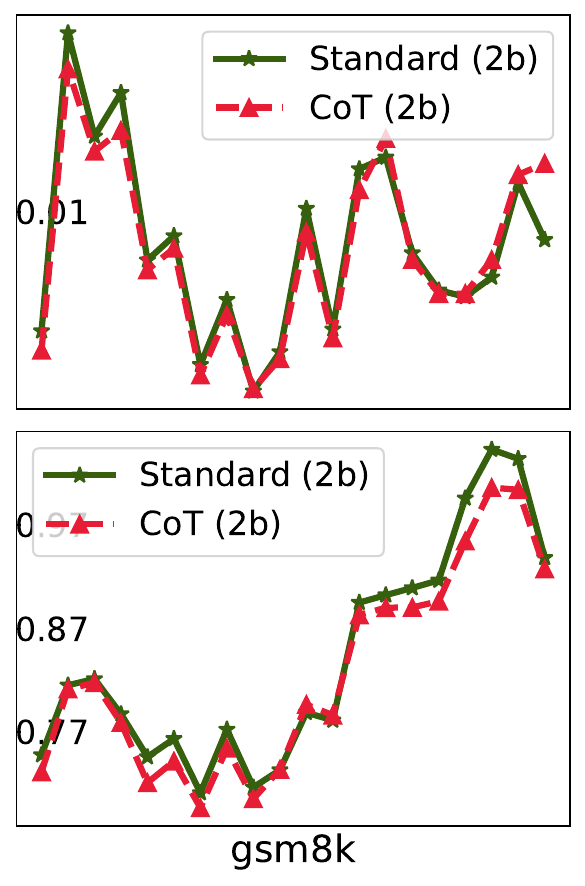}
    \includegraphics[width=0.18\linewidth]{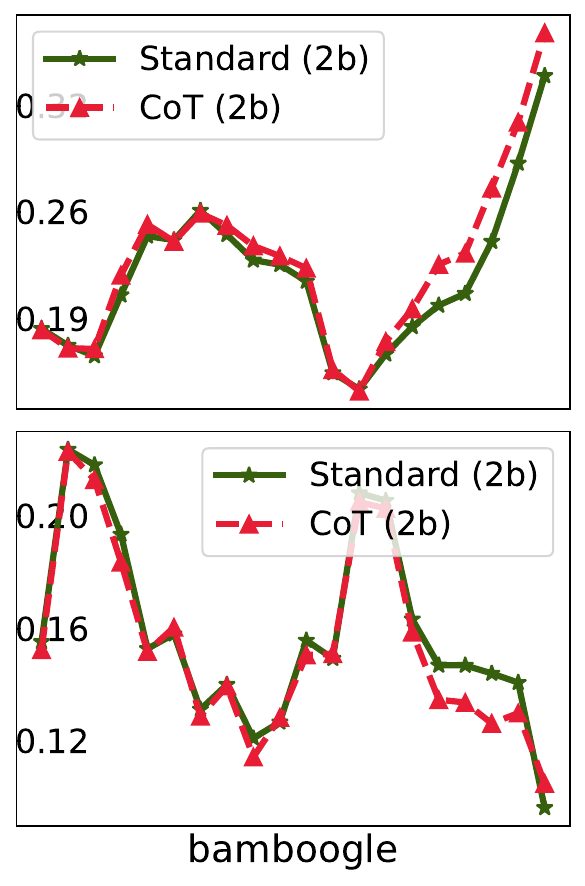}
    \includegraphics[width=0.18\linewidth]{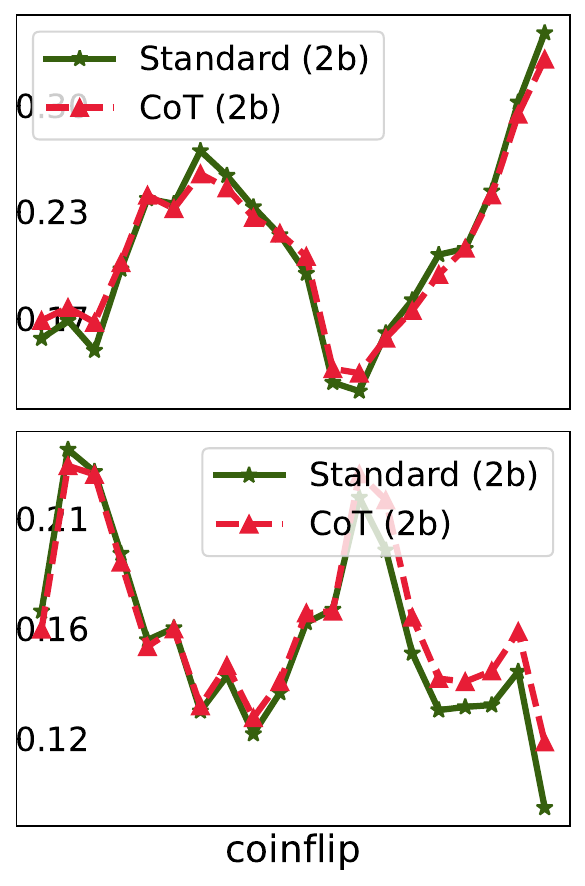}
    \includegraphics[width=0.18\linewidth]{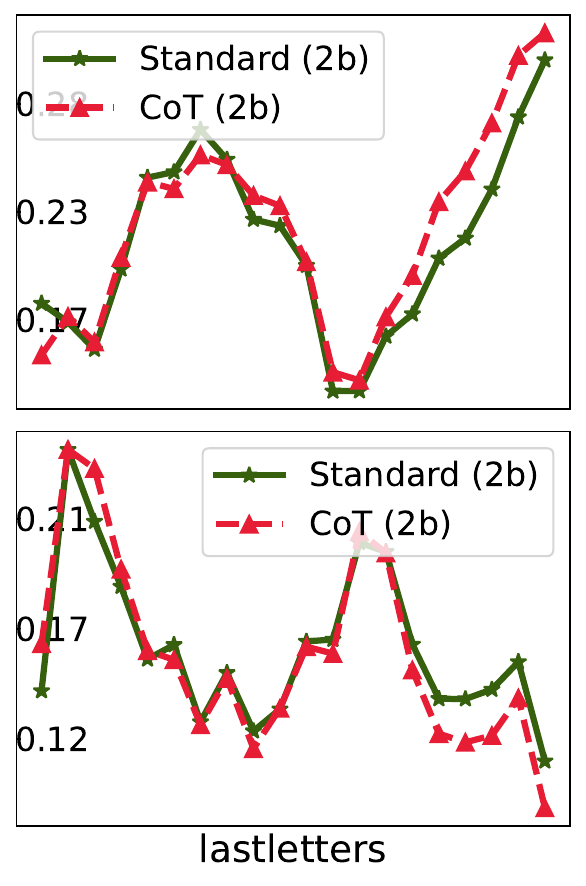}
    \caption{Neuron activation range and intensity across the final 20 layers of Gemma2-27b and Gemma2-2b. }
    \label{fig:question3_activation_2b_27b}
\end{figure*}
\end{document}